\documentclass[journal]{IEEEtran}
\ifCLASSINFOpdf
  \usepackage[pdftex]{graphicx}
  \graphicspath{{img/jpg/}{img/pdf/}{img/eps/}{img/png/}}
  \DeclareGraphicsExtensions{.jpg,.pdf,.eps,.png}
\else
\fi
\usepackage{amsmath}
\usepackage{amssymb}
\usepackage{amsthm}
\usepackage{algorithmic}
\usepackage[ruled,norelsize,linesnumbered,vlined]{algorithm2e}
\ifCLASSOPTIONcompsoc
 \usepackage[caption=false,font=normalsize,labelfont=sf,textfont=sf]{subfig}
\else
 \usepackage[caption=false,font=footnotesize]{subfig}
\fi

\usepackage{stfloats}

\usepackage{color}

\usepackage{siunitx}
\usepackage{hhline}
\usepackage{hyperref}

\usepackage{bm}
\usepackage{booktabs}

\usepackage{multicol}

\begin{document}
%
\title{Motion Planning for Variable Topology Trusses:
  Reconfiguration and Locomotion}
%
%
%

\author{Chao~Liu, Sencheng Yu, and Mark~Yim
  \thanks{Chao Liu and Mark Yim are with GRASP Lab and the Department of Mechanical
    Engineering and Applied Mechanics, University of Pennsylvania,
    Philadelphia, PA, 19104 USA e-mail: {\tt\footnotesize
      \{chaoliu, yim\}@seas.upenn.edu}. Sencheng Yu is with
    the Department of Mechanical Engineering, Texas A\&M University, TX 77843 USA e-mail:
    {\tt\footnotesize
      schyu@tamu.edu}}
}

\maketitle

\begin{abstract}
  Truss robots are highly redundant parallel robotic systems that can
  be applied in a variety of scenarios. The variable topology truss
  (VTT) is a class of modular truss robots. As self-reconfigurable
  modular robots, a VTT is composed of many edge modules that can be
  rearranged into various structures depending on the task. These
  robots change their shape by not only controlling joint positions as
  with fixed morphology robots, but also reconfiguring the
  connectivity between truss members in order to change their
  topology. The motion planning problem for VTT robots is difficult
  due to their varying morphology, high dimensionality, the high
  likelihood for self-collision, and complex motion constraints. In
  this paper, a new motion planning framework to dramatically alter
  the structure of a VTT is presented. It can also be used to solve
  locomotion tasks that are much more efficient compared with previous
  work. Several test scenarios are used to show its
  effectiveness. Supplementary materials are available
  at
  \href{https://www.modlabupenn.org/vtt-motion-planning/}{https://www.modlabupenn.org/vtt-motion-planning/}.
\end{abstract}


%
\IEEEpeerreviewmaketitle

\section{Introduction}
%
%
%
%
\IEEEPARstart{M}{odular} self-reconfigurable robots consist of
repeated building blocks (modules) with uniform docking interfaces
that allow the transfer of forces and moments, power, and
communication between modules~\cite{Yim-review-ram-2007}. These
systems are capable of reconfiguring themselves to handle failures and
adapt to changing environments. Many self-reconfigurable modular
robots have been developed, with the majority being lattice or chain
type systems \cite{Yim-review-ram-2007}.  In lattice modular robots
(\cite{Yim-telecube-icra-2002,Gilpin-miche-ijrr-2008}), modules are
regularly positioned on a three-dimensional grid. Chain modular
robots, such as~\cite{Yim-polybot-icra-2000,Yim-ckbot-2009}, consist
of modules that form tree-like structures with kinematic chains that
form articulated arms. Some systems are hybrid
(\cite{Salemi-superbot-iros-2006,Murata-mtran-trm-2002,Liu-smores-reconfig-ral-2019})
and move like chain systems for articulated tasks but reconfigure
using lattice-like actions.

\begin{figure}[t]
  \centering
  \includegraphics[width=0.25\textwidth]{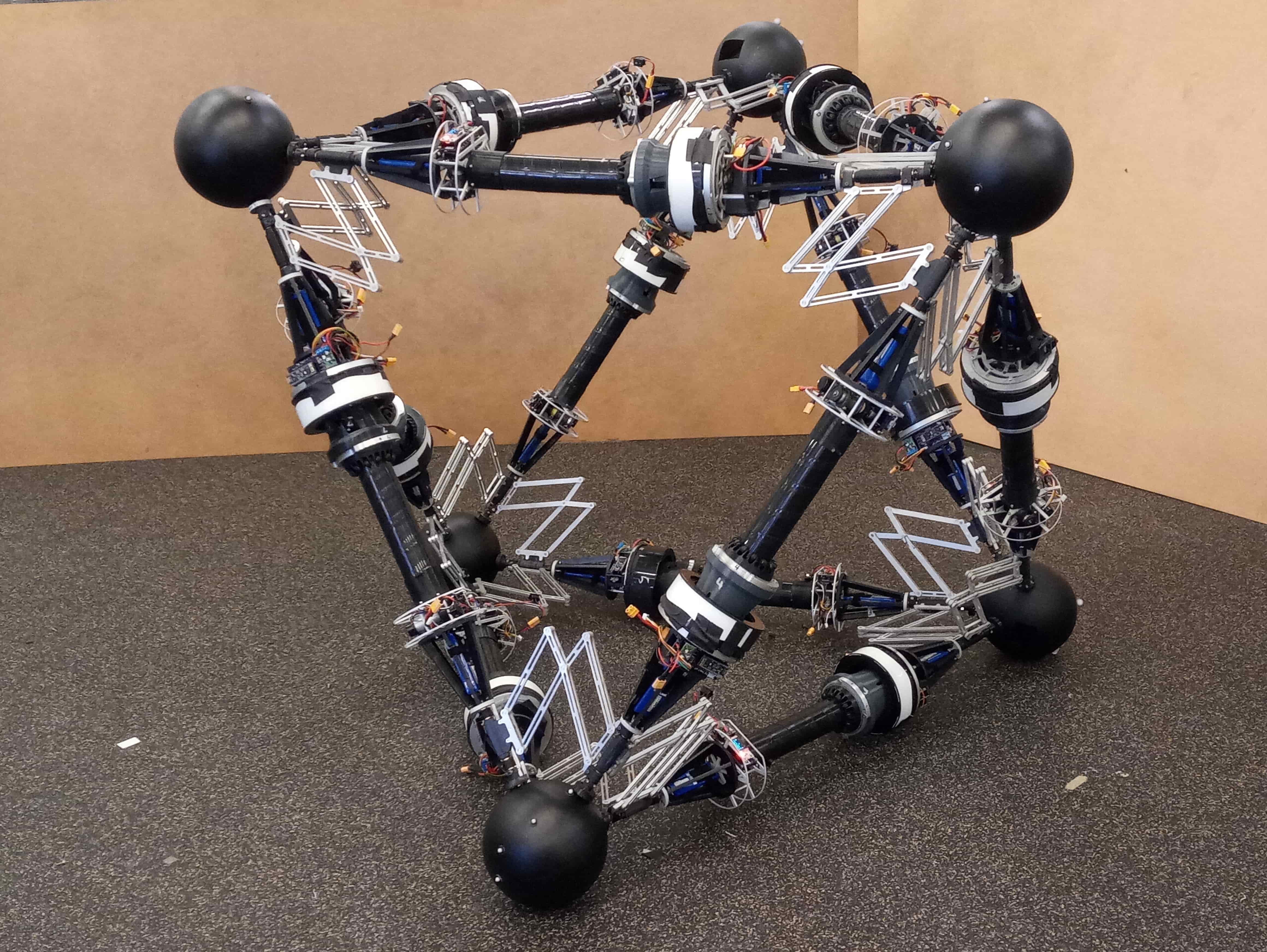}
  \caption{The hardware prototype of a VTT in octahedron configuration
    is composed of 12 members. Note that at least 18 members are
    required for topology reconfiguration
    \cite{Spinos-vtt-iros-2017}.}
  \label{fig:vtt-hardware}
\end{figure}

Modular truss robots are made of beams that form parallel
structures. The variable geometry truss
(VGT)~\cite{Miura-vgt-concept-1984} is a modular robotic truss system
with prismatic joints as truss
members~\cite{Hamlin-tetrobot-ram-1997,Lyder-odin-iros-2008,Schwager-lar-iros-2017,Komendera-truss-assembly-ijrr-2015}. These
truss members change their lengths to alter the shape of the truss.
The {\bf variable topology truss (VTT)} is similar to the variable
geometry truss robots with added ability to self-reconfigure the
connection between members changing the truss
topology~\cite{Spinos-vtt-iros-2017, vtt-review-ur-2018}. One current
hardware prototype is shown in Fig.~\ref{fig:vtt-hardware}.

A significant advantage for self-reconfigurable modular robots is
their ability to adapt their morphologies to suit different
requirements. For example, a VTT in which the members form a long
chain could be used as a radio tower, or it could reconfigure to form
a dome as a shelter in a disaster scenario. However, a fundamental
complication with VTT systems comes from the motion of the complex
parallel structure that has a high probability of
self-collision. Developing self-collision-free motion plans is
difficult due to the large number of degrees of freedom (DOFs) leading
to very large search spaces.

A truss is composed of truss members (beams) and nodes (the connection
points of multiple beams). A VTT is composed of {\em edge modules}
each of which is one active prismatic joint member and passive joint
ends that can actively attach or detach from other edge module
ends~\cite{Spinos-vtt-iros-2017}. The \textit{configuration} can be
fully defined by the set of member lengths and their node assignments
at which point the edge modules are joined. A node is constructed by
multiple edge module ends using a linkage system with a passive
rotational DOF. The node assignments define the topology or how truss
edge modules are connected, and the length of every member defines the
shape of the resulting system~\cite{Liu-vtt-planning-iros-2019}. Thus,
there are two types of reconfiguration motions: \textit{geometry
  reconfiguration} and \textit{topology reconfiguration}. Geometry
reconfiguration involves moving positions of nodes by changing lengths
of corresponding members and topology reconfiguration involves
changing the connectivity among members.

There are several physical constraints for a VTT to execute
reconfiguration activities. A VTT has to be a rigid structure in order
to maintain its shape and be statically determinant. A node in a VTT
must be of degree three, so has to be attached by at least three
members to ensure its controllability. In addition, A VTT requires at
least 18 members before topology reconfiguration is
possible~\cite{Spinos-vtt-iros-2017}. Thus, motion planning for VTT
systems has to deal with at least 18 dimensions and typically more
than 21 dimensions. These constraints complicate the motion planning
problem.

As VTTs are inherently parallel robots, it is easier to solve inverse
kinematics than forward kinematics. So, for geometry reconfiguration,
rather than planning for the active DOFs --- the member lengths --- we
plan the motion of the \textit{nodes} and then use inverse kinematics
to determine the member lengths. However, the edge module arrangement
in a VTT can result in a complicated configuration space even for a
single node (nominally a 3-DOF problem) while keeping the others
rigid.

Some desired paths for a node may be impossible to achieve without
self-collision.  Topology reconfiguration can often fix this. For
example, some nodes may be split into two nodes to go around blocking
members and then merge again on the other side. This process requires
motion planning for two nodes at the same time. In addition, multiple
nodes must move in coordination in order to achieve locomotion.

This paper builds on a sampling-based motion planning framework
presented in our previous conference paper~\cite{Liu-vtt-rss-2020}. In
this conference paper, we showed that the collision avoidance in
geometry reconfiguration planning for simplified VTT models (nodes are
considered as points and members are considered as line segments) can
be handled efficiently by explicitly computing the configuration space
for a group of nodes resulting in a smaller sampling space and a
simpler collision model. We also extended the approach
from~\cite{Liu-vtt-cspace-icra-2020} to compute all separated enclosed
free spaces for a single node so that a sequence of topology
reconfiguration actions can be computed by exploring these spaces
using a simple rule. This paper adds significantly beyond the
conference version in the following. We first solved the geometry
reconfiguration planning with realistic physical hardware constraints
being considered, including the sizes of the mechanical components,
the actuator limitations, robot stability, and motion manipulability
(to avoid singular configurations). For topology reconfiguration
planning, the simple rule from our conference paper that is only based
on the configuration space of simplified VTT models is limited when
considering all physical constraints. A better sampling strategy is
presented to handle this new difficulty so that enough samples can
span the free space as much as possible and also provide topology
reconfiguration actions as many as possible. These samples can then be
used to search a sequence of actions for a given reconfiguration
task. Finally, we present a novel locomotion framework based on this
motion planning approach which controls a VTT to locomote without
receiving external impact and achieves more robust and efficient
performance compared with existing locomotion planning algorithms for
truss robots.

The rest of the paper is organized as
follows. Section~\ref{sec:related-work} reviews relevant works and
some necessary concepts are introduced in
Section~\ref{sec:preliminary} as well as the problem statement. All
the motion constraints are introduced in
Section~\ref{sec:kinematics-constraints}. Section~\ref{sec:geometry}
presents the geometry reconfiguration algorithm with motion of
multiple nodes involved. Section~\ref{sec:topology} introduces the
approach to verify whether a topology reconfiguration action is needed
and the planning algorithm to do topology reconfiguration. The
locomotion planning approach is discussed in
Section~\ref{sec:locomotion}. Our framework is demonstrated in several
test scenarios and compared with other approaches in
Section~\ref{sec:experiment}. Finally, Section~\ref{sec:conclusion}
talks about the conclusion and some future work.

\section{Related Work}
\label{sec:related-work}

In order to enable modular robots to adapt themselves to different
activities and tasks, many reconfiguration planning algorithms have
been developed over several decades for a variety of modular robotic
systems~\cite{Yim-reconfiguration-chain-1999,
  Butler-decentralized-control-ijrr-2004,
  Hou-graph-reconfiguration-ras-2014, Liu-smores-reconfig-ral-2019}.
The planning framework includes topology reconfiguration where
undocking (disconnecting two attached modules) and docking (connecting
two modules) actions are involved. There are also some approaches for
shape morphing and manipulation tasks, including inverse kinematics
for highly redundant chains using
PolyBot~\cite{Yim-joint-solution-redundant-icra-2001}, constrained
optimization techniques with nonlinear
constraints~\cite{Fromherz-modular-robot-control-2001}, and a
real-time quadratic programming approach with linear
constraints~\cite{Liu-manipulation-ijrc-2021}. In these works, there
are no topology reconfiguration actions involved, but complicated
kinematic structures and planning in high dimensional spaces need to
be considered. However, these methods are not applicable to truss
systems which have a very different morphology and connection
architecture. Indeed the physical constraints and the collision models
are significantly different from all of the previous lattice and chain
type systems.

Some approaches have been developed for VGT systems that are similar
to VTT systems, but do not include topology reconfiguration. Kinematic
control is presented in~\cite{Hamlin-tetrobot-ram-1997} but is limited
to tetrahedrons or octahedrons. Linear actuator robots (LARs) with a
shape morphing algorithm are introduced
in~\cite{Schwager-lar-iros-2017}. These systems are in mesh graph
topology constructed by multiple convex hulls, and therefore
self-collision can be avoided easily. However, this does not apply to
VTT systems because edge modules span the workspace in a very
non-uniform manner. There has been some work on VTT motion
planning. The retraction-based RRT algorithm was developed
in~\cite{vtt-review-ur-2018} in order to handle this high dimensional
problem with narrow passage difficulty that is a well-known issue in
sampling-based planning approaches; nevertheless, this approach is not
efficient because it samples the whole workspace for every node and
the collision checking needs to be done for every pair of
members. Also, sometimes waypoints have to be assigned manually. A
reconfiguration motion planning framework inspired by the DNA
replication process --- the topology of DNA can be changed by cutting
and resealing strands as tanglements form --- is presented
in~\cite{Liu-vtt-planning-iros-2019}. This work is based on a new
method to discretize the workspace depending on the space density and
an efficient way to check self-collision. Both topology
reconfiguration actions and geometry reconfiguration actions are
involved if needed. However, only a single node is involved in each
step and the transition model is more complicated, which makes the
algorithm limited in efficiency.

A fast algorithm to compute the configuration space of a given node in
a VTT which is usually a non-convex space is presented
in~\cite{Liu-vtt-cspace-icra-2020} and this space can be then
decomposed into multiple convex polyhedrons so that a simple graph
search algorithm can be applied to plan a path for this node
efficiently. However, multiple nodes are usually involved in shape
morphing tasks. In this paper, we first extend this approach to
compute the obstacles for multiple nodes so that the search space can
be decreased significantly. In addition, only the collision among a
small number of edge modules needs to be considered when moving
multiple nodes at the same time. Hence sampling-based planners can be
applied efficiently. The idea has been discussed briefly
in~\cite{Liu-vtt-planning-idea-ur-2019}. For some motion tasks,
topology reconfiguration is required. An updated algorithm is
developed to compute the whole not fully connected free space and
required topology reconfiguration actions can then be generated using
a hybrid planning framework (sampling-based and search-based) which
can achieve behaviors that are similar to the DNA replication process.

The VTT locomotion process is similar to a locomotion mode of some
VGTs which is accomplished by tipping and contacting the
ground. TETROBOT systems and others have been shown with this mode in
simulation with generated paths for moving nodes
\cite{Lee-vgt-locomotion-tra-2002,Abrahantes-tetrahedron-gait-2010}. These
works divide the locomotion gait into several steps, but they have to
compute the motion of nodes beforehand which cannot be applied to
arbitrary configurations. Optimization approaches have been used for
locomotion planning. The locomotion process can be formulated as a
quadratic program by constraining the motion of the center of
mass~\cite{Schwager-lar-iros-2017}. The objective function is related
to the velocity of each node. A more complete quadratic programming
approach to locomotion is presented
in~\cite{Usevitch-lar-locomotion-tro-2020} to generate discrete
motions of nodes in order to follow a given trajectory or compute a
complete gait cycle. More hardware constraints were considered,
including length and collision avoidance. However, the approach has to
solve an optimization problem in high dimensional space, and it also
has to deal with non-convex and nonlinear constraints which may cause
numerical issues and is limited to a fully connected five-node graph
in order to avoid incorporating the manipulability constraints into
the quadratic program. This optimization-based approach is extended
in~\cite{Park-vtt-locomotion-ral-2019} by preventing a robot from
receiving impacts from the ground. Incorporated with a polygon-based
random tree search algorithm to output a sequence of supporting
polygons, a VTT can execute a locomotion task in an
environment~\cite{Park-vtt-locomotion-ral-2020}. However, in these
quadratic programs, numerical differentiation is required to relate
some physical constraints with these optimized parameters whenever
solving the problem. A locomotion step has to be divided into multiple
phases leading to more constraints. Also, these approaches are not
guaranteed to provide feasible solutions and they are also
time-consuming to solve. In this paper, we present a new locomotion
planning solution based on our efficient geometry reconfiguration
planning algorithm. Our solution can solve the problem much faster and
more reliably under several hardware constraints compared with
previous works. This approach can be applied to arbitrary truss
robots.

\section{Preliminaries and Problem Statement}
\label{sec:preliminary}

A VTT can be represented as an undirected graph $G=(V, E)$ where $V$
is the set of vertices of $G$ and $E$ is the set of edges of $G$: each
member can be regarded as an edge $e\in E$ of the graph and every
intersection among members can be treated as a vertex $v\in V$ of the
graph denoting a node. The Cartesian coordinates of a node $v\in V$ is
$q^v=[v_x, v_y, v_z]^\intercal\in \mathbb{R}^3$ and the
\textit{configuration space} of node $v$ denoted as $\mathcal{C}^v$ is
simply $\mathbb{R}^3$. In this way, the state of a member
$e = (v_1, v_2)\in E$ where $v_1$ and $v_2$ are two vertices of edge
$e$ can be fully defined by $q^{v_1}$ and $q^{v_2}$. The position of a
given node $v\in V$ is controlled by changing the lengths of all
attached members denoted as $E^v\subseteq E$.

\begin{figure}[b]
    \centering
    \begin{subfloat}[]{\includegraphics[width=0.16\textwidth]{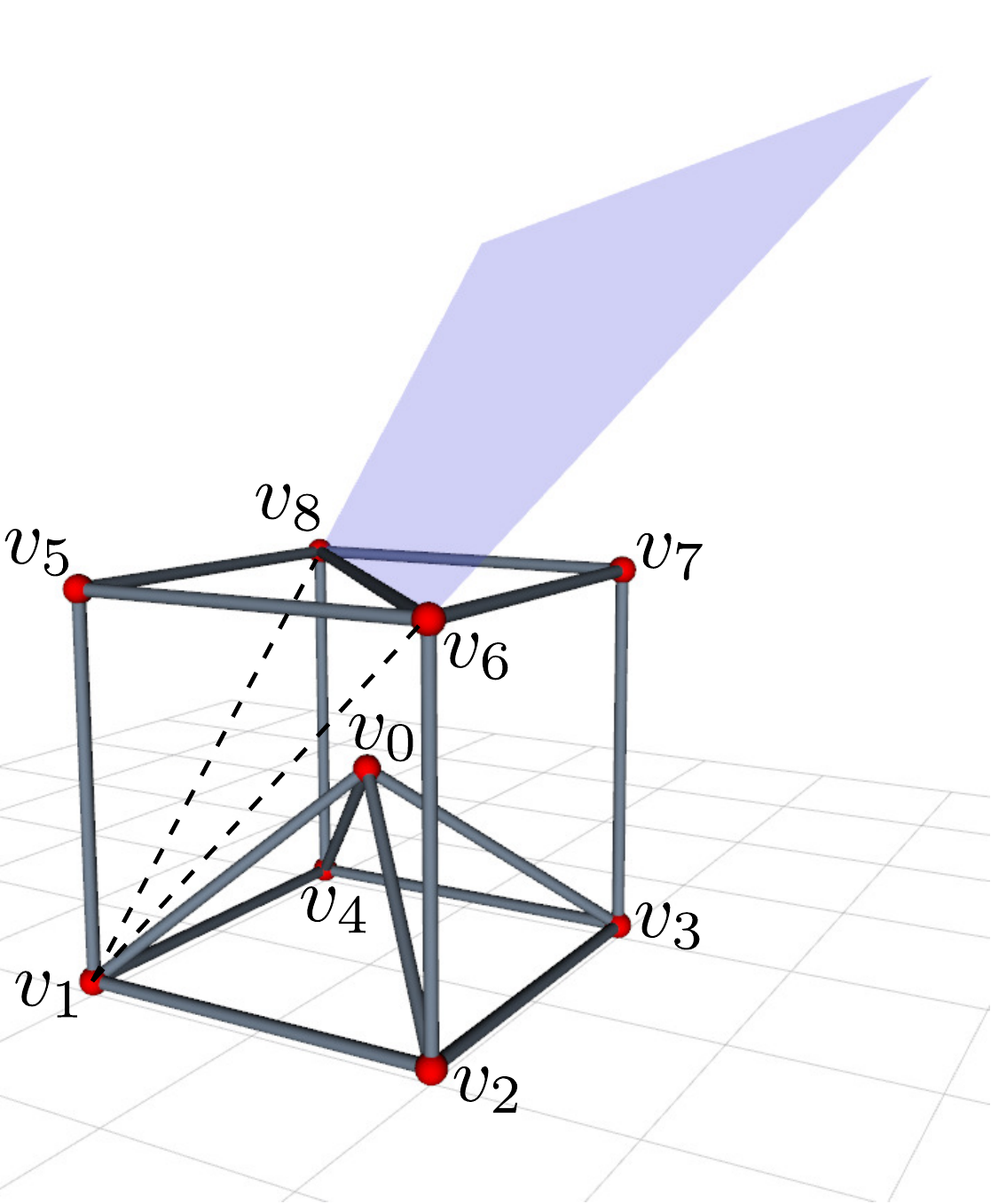}\label{fig:single-polygon}}
    \end{subfloat}
    \begin{subfloat}[]{\includegraphics[width=0.21\textwidth]{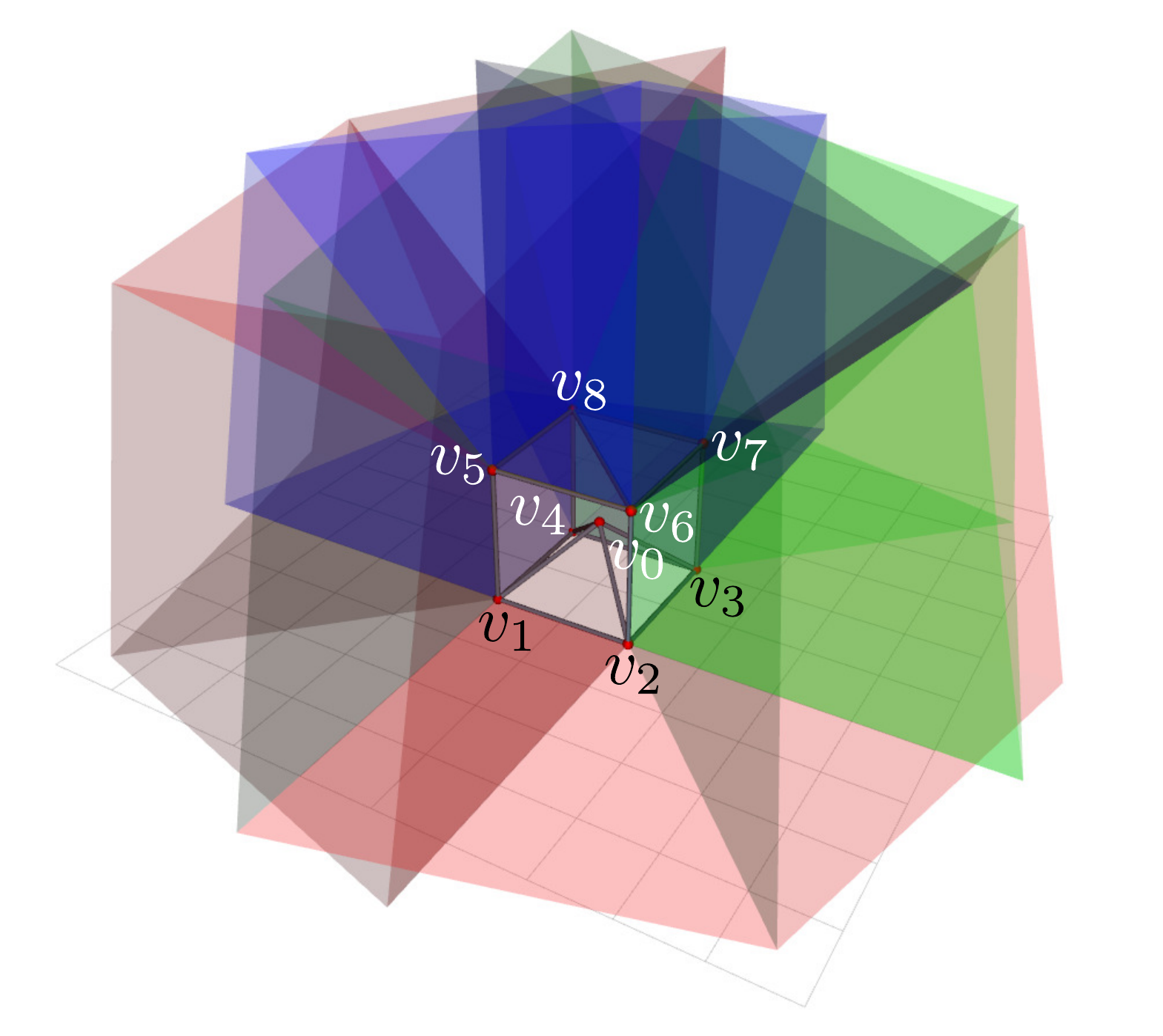}\label{fig:all-polygons-single-node}}
    \end{subfloat}
    \caption{(a) Given node $v_{0}$, one of its neighbors $v_{1}$ and
      a member $(v_{6}, v_{8})$ can define the blue polygon which is
      part of $\mathcal{C}_{\mathrm{obs}}^{v_0}$. (b) The obstacle region
      $\mathcal{C}_{\mathrm{obs}}^{v_0}$ is composed of polygons and the
      leftover space of $\mathbb{R}^3$ is $\mathcal{C}_{\mathrm{free}}^v$.}
\end{figure}

\begin{figure}[b!]
  \centering
  \includegraphics[width=0.18\textwidth]{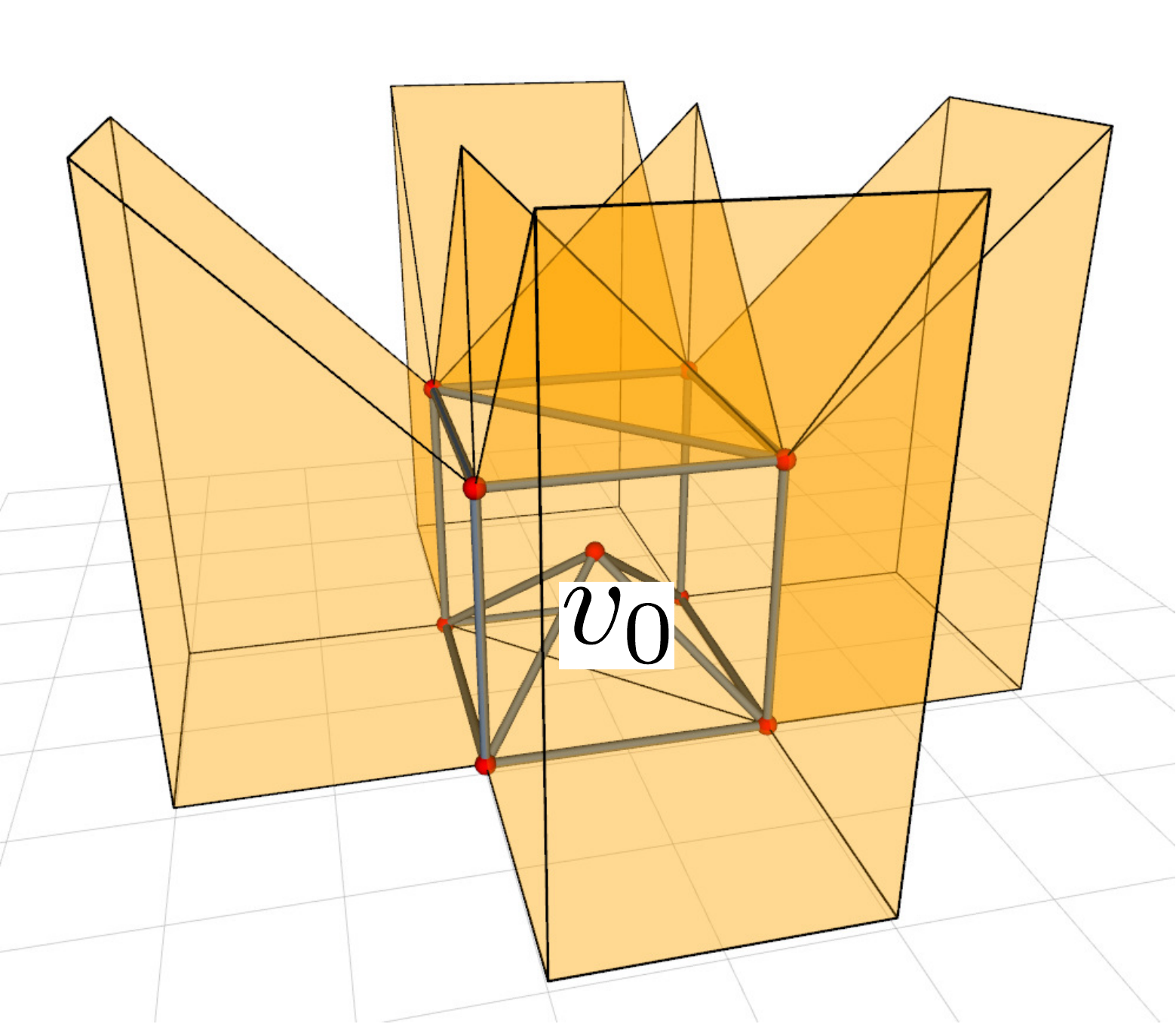}
  \caption{$\mathcal{C}_{\mathrm{free}}^{v_0}(q^{v_0})$ is bounded by polygons
    and workspace boundaries.}
  \label{fig:bounded-c-free}
\end{figure}

The geometry reconfiguration motion planning of a VTT is achieved by
planning the motion of the involved nodes then determining the
required member length trajectories. Given $v\in V$ in $G = (V, E)$,
the state of every member $e\in E^v$, denoted as $\mathcal{A}^v(q^v)$,
can be altered by changing $q^v$. The \textit{obstacle region} of this
node that includes self-collision with other members as obstacles
$\mathcal{C}^v_{\mathrm{obs}}\subseteq \mathcal{C}^v = \mathbb{R}^3$
is defined as
\begin{equation}
  \label{eq:obstacle}
  \mathcal{C}^v_{\mathrm{obs}} = \{q^v\in \mathbb{R}^3\vert \mathcal{A}^v(q^v) \cap
   \mathcal{O}^v \neq \emptyset\}
\end{equation}
where $\mathcal{O}^v$ is the obstacle for $E^v$. It is proved that
this obstacle region is fully defined by the states of
$\forall e\in E\setminus E^v$ and composed of multiple
polygons~\cite{Liu-vtt-cspace-icra-2020}. For a simple VTT shown in
Fig.~\ref{fig:single-polygon}, the obstacle region
$\mathcal{C}_{\mathrm{obs}}^{v_0}$ is shown in
Fig.~\ref{fig:all-polygons-single-node}. The \textit{free space} of
$v$ is just the leftover configurations denoted as
\begin{equation}
  \label{eq:free-space}
  \mathcal{C}_{\mathrm{free}}^v = \mathbb{R}^3 \setminus \mathcal{C}_{\mathrm{obs}}^v
\end{equation}
However, $\mathcal{C}_{\mathrm{free}}^v$ may not be fully connected
and is usually partitioned by $\mathcal{C}_{\mathrm{obs}}^v$. Only the
enclosed subspace containing $q^v$ which is denoted as
$\mathcal{C}_{\mathrm{free}}^v(q^v)$ is free for node $v$ to move. A
fast algorithm to compute the boundary of this subspace is presented
in~\cite{Liu-vtt-cspace-icra-2020}. For example, given the VTT in
Fig.~\ref{fig:single-polygon}, $\mathcal{C}_{\mathrm{free}}^{v_0}$ ---
the free space of $v_0$ --- is partitioned by
$\mathcal{C}_{\mathrm{obs}}^{v_0}$ in
Fig.~\ref{fig:all-polygons-single-node}, and the subspace
$\mathcal{C}_{\mathrm{free}}^{v_0}(q^{v_0})$ is shown in
Fig.~\ref{fig:bounded-c-free}.

\begin{figure}[t]
  \centering
  \includegraphics[width=0.3\textwidth]{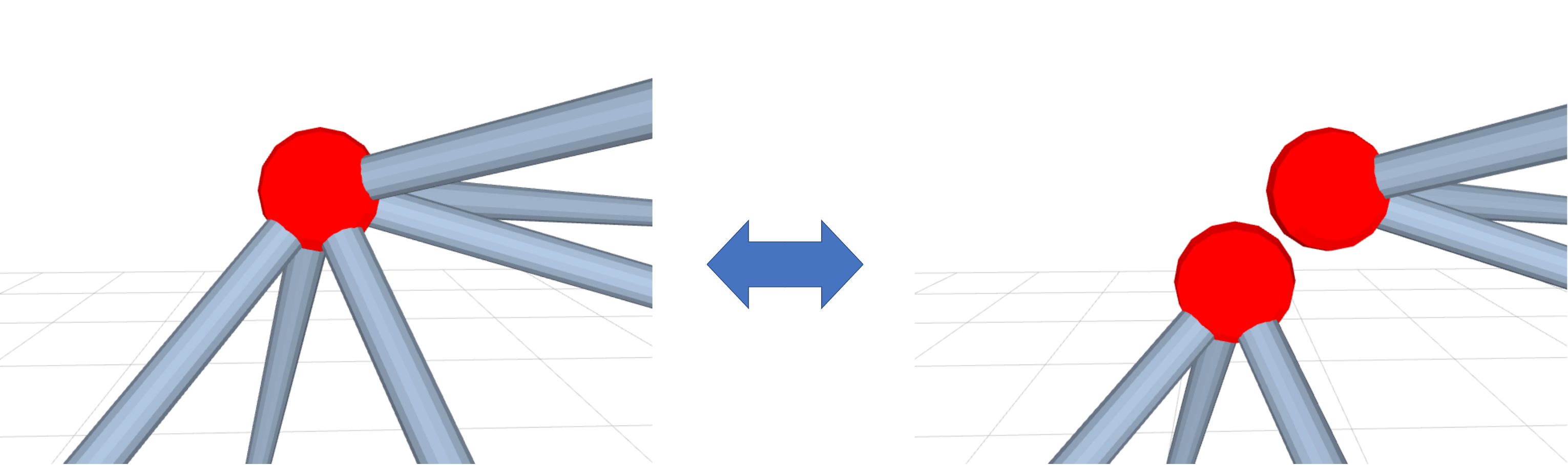}
  \caption{A single node with six members can be split into a pair of nodes and two separate nodes can also merge into an individual node.}
  \label{fig:merge-split}
\end{figure}

$\mathcal{C}_{\mathrm{free}}^v$ is usually partitioned by
$\mathcal{C}_{\mathrm{obs}}^v$ into multiple enclosed subspaces, and
it is impossible to move $v$ from one enclosed subspace to another one
without topology reconfiguration. The physical system constraints
shown in~\cite{Spinos-vtt-iros-2017} allow two atomic actions on nodes
that enable topology reconfiguration: \texttt{Split} and
\texttt{Merge}. Since the physical system must be statically
determinate with all nodes of degree three, a node $v$ must be
composed of six or more edge modules to undock and split into two new
nodes $v^\prime$ and $v''$, and both nodes should still have three or
more members. This process is called \texttt{Split}. Two separate
nodes are able to merge into an individual one in a \texttt{Merge}
action. The simulation of these two actions is shown in
Fig.~\ref{fig:merge-split}. We are not considering splitting a node
recursively, namely after splitting a node $v$ into $v^{\prime}$ and
$v''$, further splitting $v^{\prime}$ or $v''$, because it is
difficult to attach more than nine members to a single node due to
physical hardware constraints. In a topology reconfiguration process,
the number of nodes can change, but the number of members that are
physical elements remains constant.

In this work, given a VTT $G = (V,E)$, the reconfiguration planning
problem can be stated as the following:
\begin{itemize}
\item \textbf{Geometry Reconfiguration}. For a set of $n$ nodes
  $\{v_t\in V\vert t = 1,2,\cdots,n\}$, compute paths
  $\tau_t:\left[0, 1\right]\to \mathcal{C}_{\mathrm{free}}^{v_t}$ such
  that $\tau_t(0) = q_i^{v_t}$ and $\tau_t(1) = q_g^{v_t}$ in which
  $t=1,2,\cdots, n$, $q_i^{v_t}$ is the initial position of $v_t$ and
  $q_g^{v_t}$ is the goal position of $v_t$.
\item \textbf{Topology Reconfiguration}. Compute the topology
  reconfiguration actions, \texttt{Merge} and \texttt{Split}, and find
  collision-free path(s) to move a node $v$ from its initial position
  $q_i^v$ to its goal position $q_g^v$.
\end{itemize}

The shape of a VTT $G=(V,E)$ can be regarded as a polyhedron with flat
polygonal facets formed by members. This polyhedron consists of
vertices (a subset of $V$), edges $E^G$, facets $F^G$, and an
incidence relation on them (e.g., every edge is incident to two
vertices and every edge is incident to two
facets~\cite{Kettner-polyhedron-data-structure-1999}). One of the
facets $f\in F^G$ is the current support polygon. In a rolling
locomotion step, the support polygon is changed from $f$ to an
adjacent facet $f^{\prime}$. The non-impact locomotion problem for a
given VTT $G=(V,E)$ can be stated as the following:
\begin{itemize}
\item \textbf{Non-impact Rolling Locomotion}. Compute the motions of a
  set of nodes in $V$ such that a sequence of support polygons
  $f\in F^G$ can be generated in which adjacent facet
  $f^{\prime}\in F^G$ are used for each support polygon and the
  transition between two adjacent support polygons occur with all
  nodes from both facets are co-planar in contact with the ground.
\end{itemize}

\section{Motion Constraints}
\label{sec:kinematics-constraints}

\begin{figure}[t]
  \centering
  \includegraphics[width=0.15\textwidth]{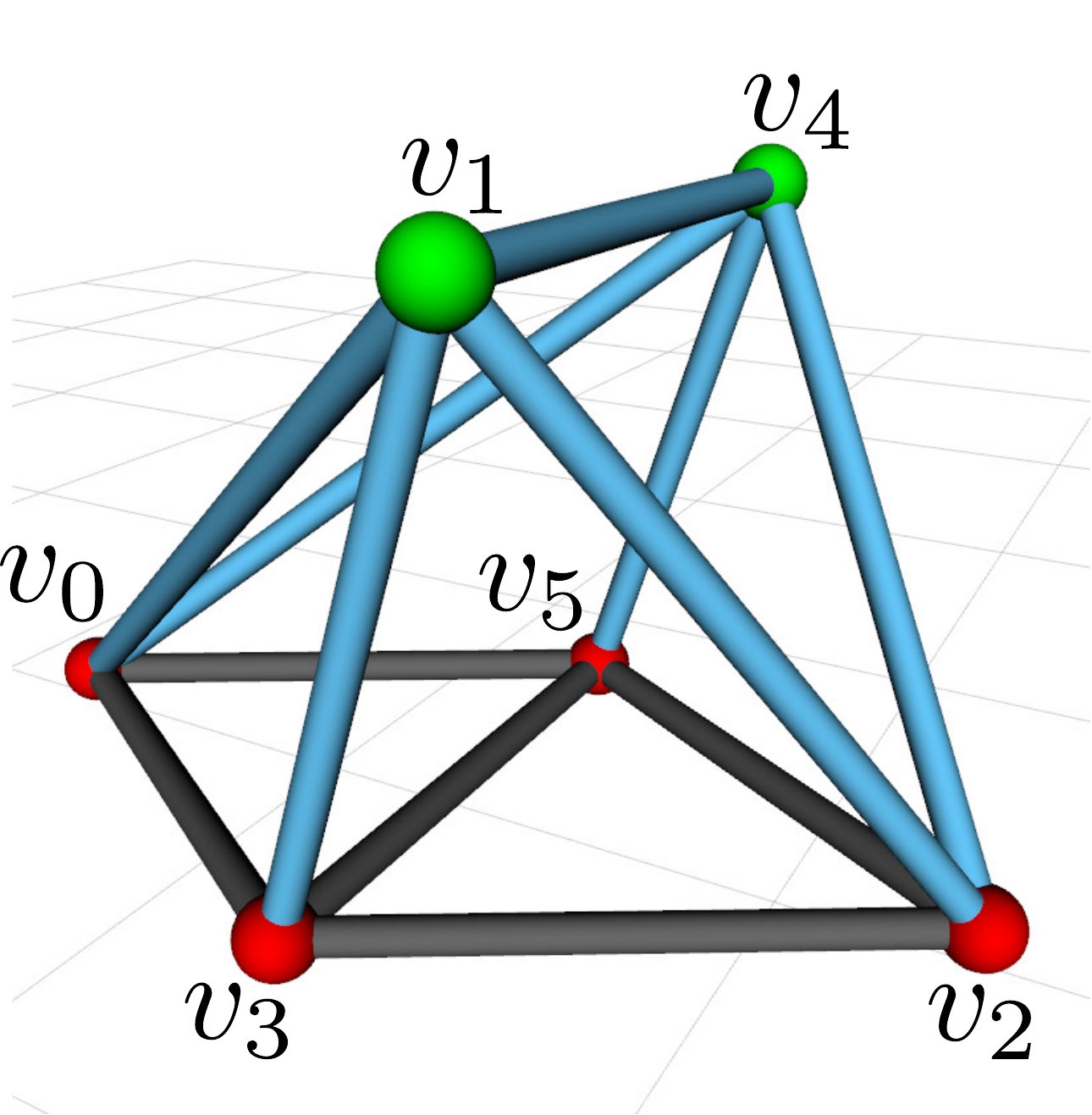}
  \caption{This VTT is composed of twelve members. Currently, the
    motion of node $v_1$ and node $v_4$ are under control by seven
    blue members.}
  \label{fig:vtt-octahedron}
\end{figure}

Given a VTT $G=(V, E)$, the motions of nodes are controlled by all
attached actuated members and the system is often an overconstrained
parallel robot. We follow the modeling way
in~\cite{Hamlin-tetrobot-ram-1997} to model a VTT and additionally
provide a general kinematics model for truss robots. The set of all
nodes $V$ is separated into two groups: $V_F$ and $V_C$ where $V_F$
contains all the fixed or stationary nodes and $V_C$ contains all the
controlled nodes. For example, given the VTT shown in
Fig.~\ref{fig:vtt-octahedron}, when controlling the motion of node
$v_1$ and node $v_4$, $V_C=\{v_1, v_4\}$ and
$V_F=\{v_0, v_2, v_3, v_5\}$. This is a 6-DOF system since there are
two controlled nodes. In addition, this system is overconstrained and
the motion of two nodes are controlled by seven members. Note that
$V_F$ and $V_C$ are not constant and they can be changed during the
motion of a VTT (reconfiguration and locomotion). All motions of a VTT
have to satisfy physical constraints in order to be feasible. In
addition to these physical constraints, a VTT has to maintain
infinitesimally rigidity or avoid singular configurations.

\subsection{Hardware and Environmental Constraints}
\label{sec:constraints}

\subsubsection{Length Constraints}
\label{sec:length}

In a VTT, each edge module has an active prismatic joint called Spiral
Zipper~\cite{Collins-spiral-zipper-icra-2016} that is able to achieve
a high extension ratio and form a high strength-to-weight-ratio
column. The mechanical components determine the minimum member length
and the total material determines the maximum member length. Hence, in
a VTT $G=(V, E)$, $\forall e=(v_i, v_j)\in E$, we have the following
constraint
\begin{equation}
  \label{eq:length-constraint}
  \overline{L}_{\mathrm{min}} \le \|q^{v_i} - q^{v_j}\| \le \overline{L}_{\mathrm{max}}
\end{equation}

\subsubsection{Collision Avoidance}
\label{sec:collision}

During the motion task, a VTT has to avoid self-collision. The
distance between every two member axes must be greater than
$\bar{d}_{\mathrm{min}}$, the diameter of an edge module which is a
cylinder. The minimum distance between member $(v_i, v_j)$ and
$(v_m, v_n)$ can be expressed as
\begin{equation}
  \label{eq:min-distance-edges}
  \min \|(q^{v_i} + \alpha(q^{v_j}-q^{v_i})) -
  (q^{v_m} + \gamma(q^{v_n}-q^{v_m}))\|
\end{equation}
in which $\alpha, \gamma \in (0, 1)$. This is not easy to compute and
can be more complicated when both members are moving. In
Section~\ref{sec:geometry}, we will show a way to avoid this
problem. We also need to ensure that the angle between connected
members remains larger than a minimum physical joint limit. The angle
constraint between member $(v_i, v_j)$ and $(v_i, v_k)$ can be
expressed as
\begin{equation}
  \label{eq:angle-constraint}
  \arccos\left(\frac{(q^{v_j}-q^{v_i})\bullet(q^{v_k}-q^{v_i})}{\|q^{v_j}-q^{v_i}\|\|q^{v_k}-q^{v_i}\|}\right)
  \ge \bar{\theta}_{\mathrm{min}}
\end{equation}

\subsubsection{Stability}
\label{sec:com}

The truss structure of a VTT is meant to be statically stable under
gravity when interacting with the environment and sufficient
constraints must be satisfied to ensure the location of the structure
is fully defined. At least three still nodes should be on the ground
in order to form a valid support polygon. For the VTT shown in
Fig.~\ref{fig:vtt-octahedron}, $v_0$, $v_2$, $v_3$, and $v_5$ are
stationary on the ground to form the current support polygon. In
addition, the vertical projection of the center of mass of a VTT on
the ground has to be inside this support polygon. Furthermore, no
collision is allowed between a VTT and the environment, and the
simplest condition is that all nodes have to be above the ground.

\subsection{Manipulability Maintenance Constraint}
\label{sec:manipulability}

In order to control the motions of nodes, a VTT has to maintain the
manipulability for these moving nodes to make sure the system is not
close to singularity. For a VTT $G=(V,E)$, given $V_C$ with $\Lambda$
nodes under control, the position vector of this system is simply the
stack of $q^v$ where $v\in V_C$. This system is controlled by all
members that are attached with these nodes, namely
$\bigcup_{v\in V_C} E^v$. Let $l_{ij}$ be the \textit{link vector}
from a controlled node $v_i$ pointing to any node $v_j$. There are two
types of link vectors: 1. if $v_j\in V_F$, then $l_{ij}$ is an
\textit{attachment link vector}; 2. if $v_j\in V_C$, then $l_{ij}$ is
a \textit{connection link vector}. The link vector satisfies the
following equation:
\begin{equation}
  \label{eq:inverse_kinematics}
  l_{ij} = q^{v_j} - q^{v_i}
\end{equation}
in which $v_i\in V_C$.  Taking the time derivative of both sides of
Eq.~(\ref{eq:inverse_kinematics}) for an attachment link vector and a
connection link vector to relate node velocities to joint velocities
yields
\begin{subequations}
  \begin{equation}
    \label{eq:derivative-eq-1}
    l_{ij}^\intercal\dot{l}_{ij} = (q^{v_i} - q^{v_j})^{\intercal}\dot{q}^{v_i}\quad \forall
    v_j\in V_F
  \end{equation}
  \begin{equation}
    \label{eq:derivative-eq-2}
    \dot{l}_{ij} = \dot{q}^{v_j} - \dot{q}^{v_i}\quad \forall v_j\in V_C
  \end{equation}
\end{subequations}

Assume $V_C = \{\bar{v}_{\alpha}\vert \alpha = 1,2,\cdots,\Lambda\}$,
and for a controlled node $\bar{v}_{\alpha}$, all the fixed nodes in
its neighborhood $\mathcal{N}_G(\bar{v}_{\alpha})$ are denoted as
$\hat{v}_1^{\alpha}, \hat{v}_2^\alpha, \cdots, \hat{v}_N^\alpha$, and
the corresponding attachment link vectors are denoted as
${^{\alpha}}\hat{l}_1, {^{\alpha}}\hat{l}_2, \cdots,
{^{\alpha}}\hat{l}_N$. Eq.~\eqref{eq:derivative-eq-1} is true for any
${^\alpha}\hat{l}_t$ where $t \in \left[1, N\right]$, so we can
rewrite it for this controlled vertex $\bar{v}_{\alpha}$
\begin{equation}
  \label{eq:vertex_kinematics}
  B_\alpha\dot{L}_\alpha = A_\alpha\dot{q}^{\bar{v}_{\alpha}}
\end{equation}
in which
\begin{gather*}
  \dot{L}_\alpha = \left[
    \begin{array}{cccc}
      {^\alpha}\dot{\hat{l}}_1^\intercal&{^\alpha}\dot{\hat{l}}_2^\intercal&\cdots&{^\alpha}\dot{\hat{l}}_N^\intercal
    \end{array}
  \right]^\intercal_{3N\times1}\\
  B_\alpha = \left[
    \begin{array}{cccc}
      {^\alpha}\hat{l}_1^{\intercal}&0&\cdots&0\\
      0&{^\alpha}\hat{l}_2^{\intercal}&\cdots&0\\
      \vdots&\vdots&\ddots&\vdots\\
      0&0&\cdots&{^\alpha}\hat{l}_N^{\intercal}
    \end{array}
  \right]_{N\times3N}\\
  A_\alpha = \left[
    \begin{array}{cccc}
      q^{\bar{v}_{\alpha}} - q^{\hat{v}_1^{\alpha}} &
      q^{\bar{v}_{\alpha}} - q^{\hat{v}_2^{\alpha}} &
      \cdots&
      q^{\bar{v}_{\alpha}} - q^{\hat{v}_N^{\alpha}}
    \end{array}
  \right]^\intercal_{N\times3}
\end{gather*}
If another controlled node
$\bar{v}_{\beta}\in \mathcal{N}_G(\bar{v}_{\alpha})$, namely
$\bar{v}_{\beta}$ is adjacent to $\bar{v}_{\alpha}$, then we have
\begin{equation}
  \label{eq:connect_link_vector_change_new}
  \dot{\bar{l}}_{\alpha\beta} = \dot{q}^{\bar{v}_{\beta}} - \dot{q}^{\bar{v}_{\alpha}}
\end{equation}
Combining Eq.~(\ref{eq:vertex_kinematics}) and
Eq.~(\ref{eq:connect_link_vector_change_new}), we get
\begin{equation}
  \label{eq:kinematics_eq}
  \mathcal{B}\dot{\mathcal{L}} = \mathcal{A}\dot{p}
\end{equation}
in which
\begin{gather*}
  p = \left[
    \begin{array}{cccc}
      (q^{\bar{v}_1})^{\intercal}& (q^{\bar{v}_2})^{\intercal} & \cdots& (q^{\bar{v}_\Lambda})^{\intercal}
    \end{array}
  \right]^\intercal\\
  \dot{\mathcal{L}} = \left[
    \begin{array}{ccccccc}
      \dot{L}_1^\intercal&\dot{L}_2^\intercal&\cdots&\dot{L}_\Lambda^\intercal&\cdots
      &\dot{\bar{l}}_{\alpha\beta}& \cdots
    \end{array}\right]^\intercal\\
  \mathcal{B} =\mathrm{diag} (B_1, B_2, \cdots,
  B_\Lambda, \cdots, I, \cdots)\quad
  \mathcal{A} = \left[\mathcal{A}_1, \mathcal{A}_2\right]^\intercal\\
  \mathcal{A}_1 = \mathrm{diag} (A_1, A_2, \cdots,
  A_\alpha, \cdots, A_\beta, \cdots, A_\Lambda)\\
  \mathcal{A}_2 = \left[
    \begin{array}{ccccc}
      \vdots&\vdots&\vdots&\vdots&\vdots\\
      0_{3\times3\alpha}&I_{3\times3}&0_{3\times(\beta-\alpha-1)}&-I_{3\times3}&0_{3\times3(\Lambda-\beta)}\\
      \vdots&\vdots&\vdots&\vdots&\vdots
    \end{array}
  \right]
\end{gather*}

The size of $\mathcal{A}_2$ is determined by the number of controlled
vertices and connection link vectors (if there are $\omega$ connection
link vectors, $\mathcal{A}_2$ is a $3\omega\times 3\Lambda$
matrix). For the system shown in Fig.~\ref{fig:vtt-octahedron},
$\mathcal{A}_2 = \left[I, -I\right]_{3\times 6}$ since there are only
two controlled nodes and one connection link vector. It can be shown
that $\mathcal{B}$ has full rank as long as there are no zero-length
members whereas $\mathcal{A}$ may not have full rank. There must exist
a set of link velocities, but some link velocities may result in
invalid motions of nodes since the system can be overconstrained.

We can rearrange Eq.~\eqref{eq:kinematics_eq} in two ways:
\begin{subequations}
  \begin{equation}
    \label{eq:jacobian_1}
    \dot{\mathcal{L}} =
    \mathcal{B}^+\mathcal{A}\dot{p} = J_{\mathrm{BA}}\dot{p}
  \end{equation}
  \begin{equation}
    \label{eq:jacobian_2}
    \dot{p} =
    \mathcal{A}^+\mathcal{B}\dot{\mathcal{L}} =
    J_{\mathrm{AB}}\dot{\mathcal{L}}
  \end{equation}
\end{subequations}
where $\mathcal{B}^+$ and $\mathcal{A}^+$ are the pseudo-inverse of
$\mathcal{B}$ and $\mathcal{A}$ respectively, and both
$J_{\mathrm{BA}}$ and $J_{\mathrm{AB}}$ matrices are the
\textit{Jacobian}. We use these two equations to describe the
relationship between the link velocities and the controlled node
velocities. $J_{BA}$ is always defined (as long as there is no
zero-length member), but $J_{AB}$ may not be defined. Given $\dot{p}$
is known, Eq.~(\ref{eq:jacobian_1}) gives the minimum norm solution to
Eq.~(\ref{eq:kinematics_eq}), namely minimizing
$\|\dot{\mathcal{L}}\|$. On the other hand, Eq.~(\ref{eq:jacobian_2})
results in the unique least square solution to
Eq.~(\ref{eq:kinematics_eq}) if $\dot{\mathcal{L}}$ is known, namely
minimizing $\|\mathcal{B}\dot{\mathcal{L}}-\mathcal{A}\dot{p}\|$.

By applying singular value decomposition on $J_{\mathrm{AB}}$, its
maximum singular value $\sigma_{\max}(J_{\mathrm{AB}})$ and minimum
singular value $\sigma_{\min}(J_{\mathrm{AB}})$ can be derived, and
the manipulability of the current moving nodes can be constrained as
\begin{equation}
  \label{eq:manipulability-constraint}
  \mu = \frac{\sigma_{\min}(J_{\mathrm{AB}})}{\sigma_{\max}(J_{\mathrm{AB}})} \ge \bar{\mu}_{\min}
\end{equation}

\section{Geometry Reconfiguration}
\label{sec:geometry}

The overall shape of a VTT is altered by moving nodes around in the
workspace. For an individual node, its configuration space is
$\mathbb{R}^3$, and the configuration space for $n$ nodes is
$\mathbb{R}^{3n}$. Our strategy to avoid this high dimensionality is
to divide the moving nodes into multiple groups where each group
contains one or a pair of nodes. The motion planning space for each
group is either in $\mathbb{R}^3$ or $\mathbb{R}^6$ and we just need
to solve a sequence of $\mathbb{R}^3$ or $\mathbb{R}^6$ path planning
problems. Even with this lower-dimensional space, it is still a
challenge to search for a valid solution. In addition, the resolution
of a discretized space must be fine enough to ensure the motion of a
member does not skip over a possible self-collision which leads to
slow planning.  We overcome these issues by computing the free space
of the group in advance so that the sampling space is significantly
decreased.

\subsection{Obstacle Region and Free Space}
\label{sec:obstacle-free}

A group can contain either one node or a pair of nodes. Given a VTT
$G = (V, E)$, for an individual node $v\in V$, an efficient algorithm
to compute $\mathcal{C}_{\mathrm{obs}}^v$ and
$\mathcal{C}_{\mathrm{free}}^v(q^v)$ with its boundary is introduced
in~\cite{Liu-vtt-cspace-icra-2020}. If there are two nodes $v_i\in V$
and $v_j\in V$ in a group, then any collision among members in
$E^{v_i}$ and $E^{v_j}$ is treated as self-collision inside the group,
and all the members in $E\setminus(E^{v_i}\cup E^{v_j})$ define the
obstacle region of this group denoted as
$\widehat{\mathcal{C}}_{\mathrm{obs}}^{v_i}$ (the obstacle region of
$v_i$ in the group) and $\widehat{\mathcal{C}}_{\mathrm{obs}}^{v_j}$
(the obstacle region of $v_j$ in the group) respectively, namely
\begin{gather*}
  \widehat{\mathcal{C}}_{\mathrm{obs}}^{v_i} = \left\{q^{v_i}\in \mathbb{R}^3\vert
    \mathcal{A}^{v_i}(q^{v_i})\cap \mathcal{O}^{v_i,v_j}\neq \emptyset\right\}\\
  \widehat{\mathcal{C}}_{\mathrm{obs}}^{v_j} = \left\{q^{v_j}\in \mathbb{R}^3\vert
    \mathcal{A}^{v_j}(q^{v_j})\cap \mathcal{O}^{v_i,v_j}\neq \emptyset\right\}
\end{gather*}
in which $\mathcal{O}^{v_i, v_j}$ is formed by
$\forall e\in E\setminus(E^{v_i}\cup E^{v_j})$.

\begin{figure}[b]
  \centering
  \includegraphics[width=0.13\textwidth]{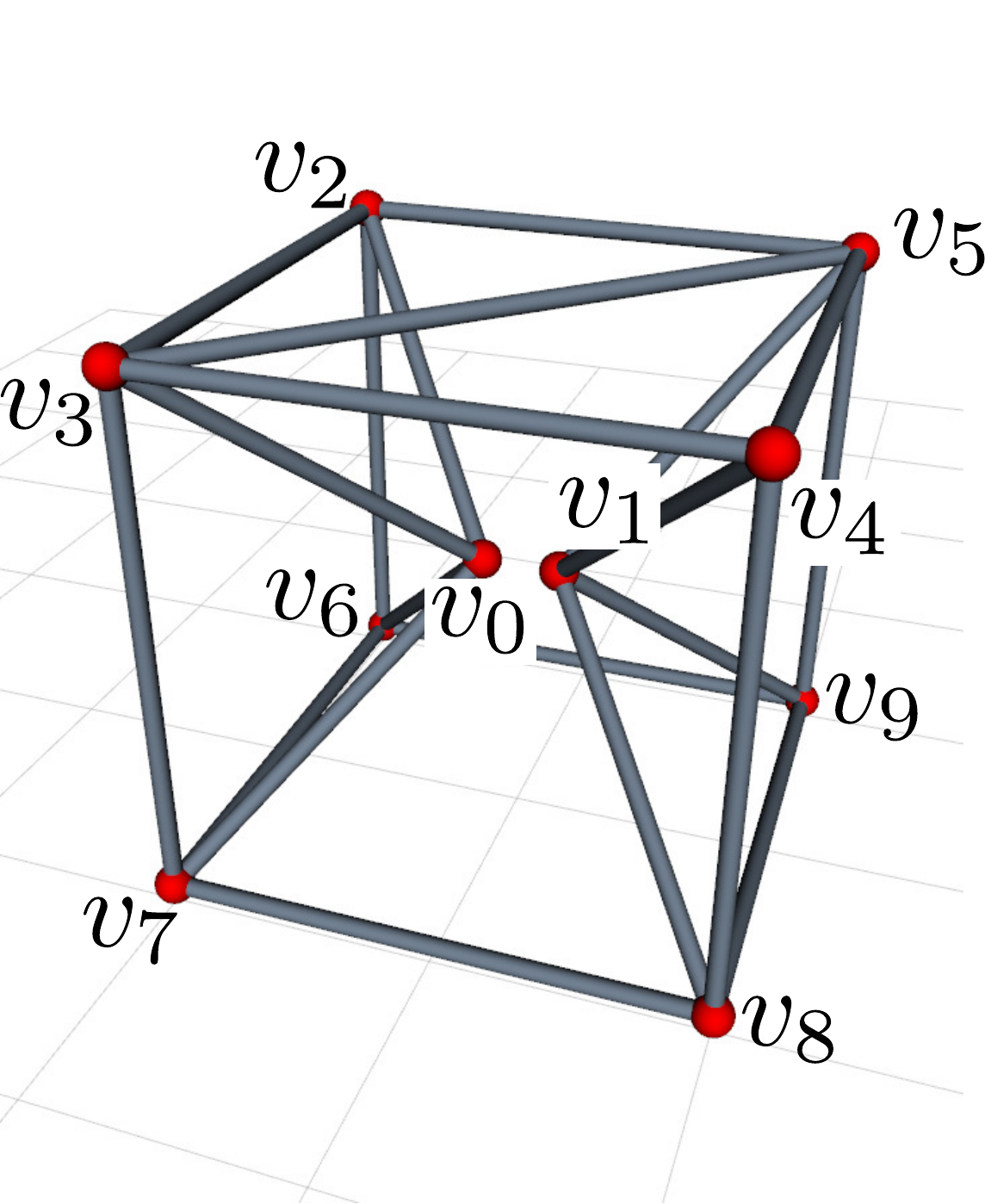}
  \caption{This VTT is composed of 21 edge modules with 10 nodes among
    which $v_0$ and $v_1$ form a group.}
  \label{fig:geometry-truss}
\end{figure}

\begin{figure}[b!]
  \centering
  \subfloat[]{\includegraphics[width=0.2\textwidth]{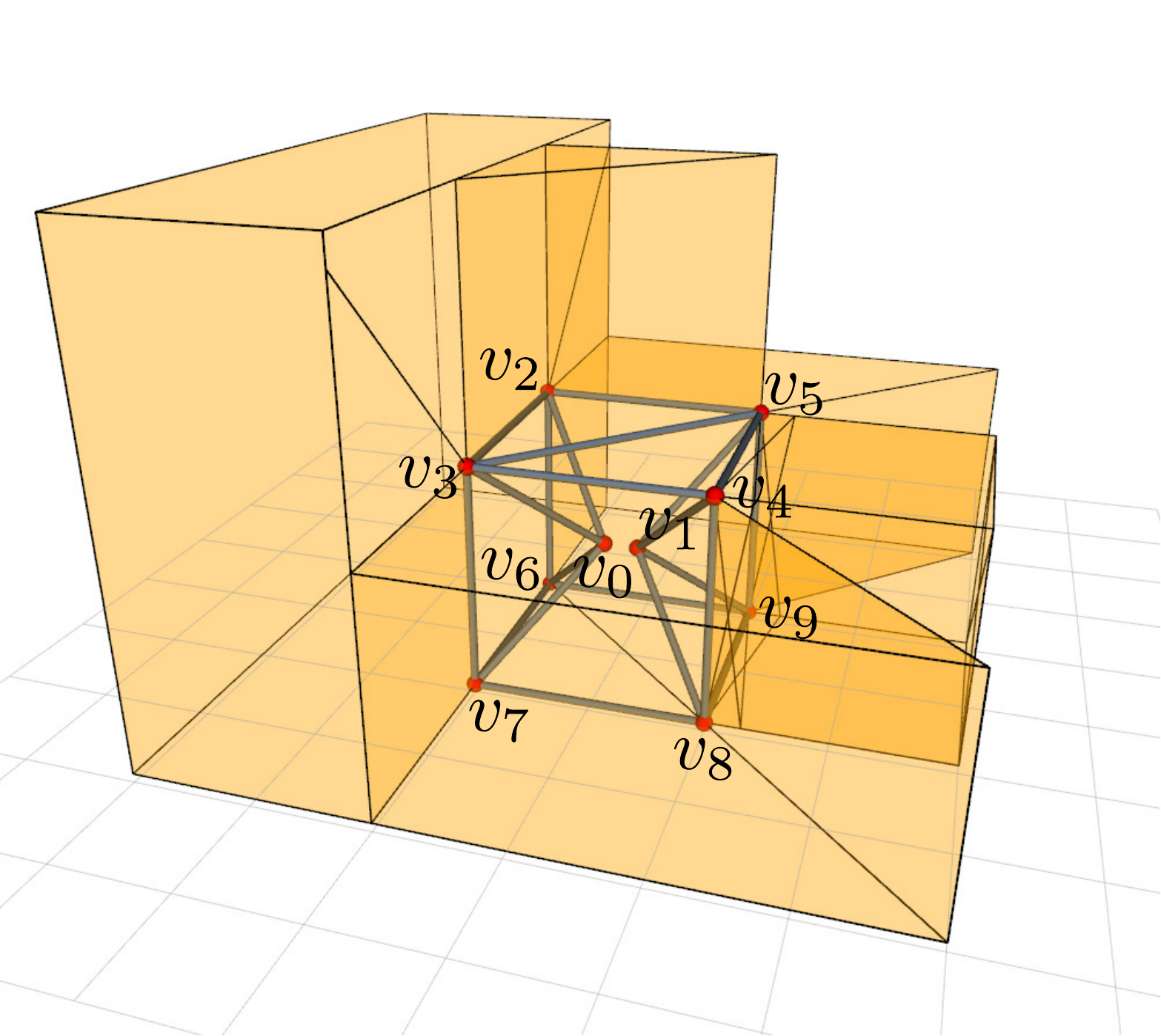}\label{fig:c-free-v0}}
  \hfil
  \subfloat[]{\includegraphics[width=0.2\textwidth]{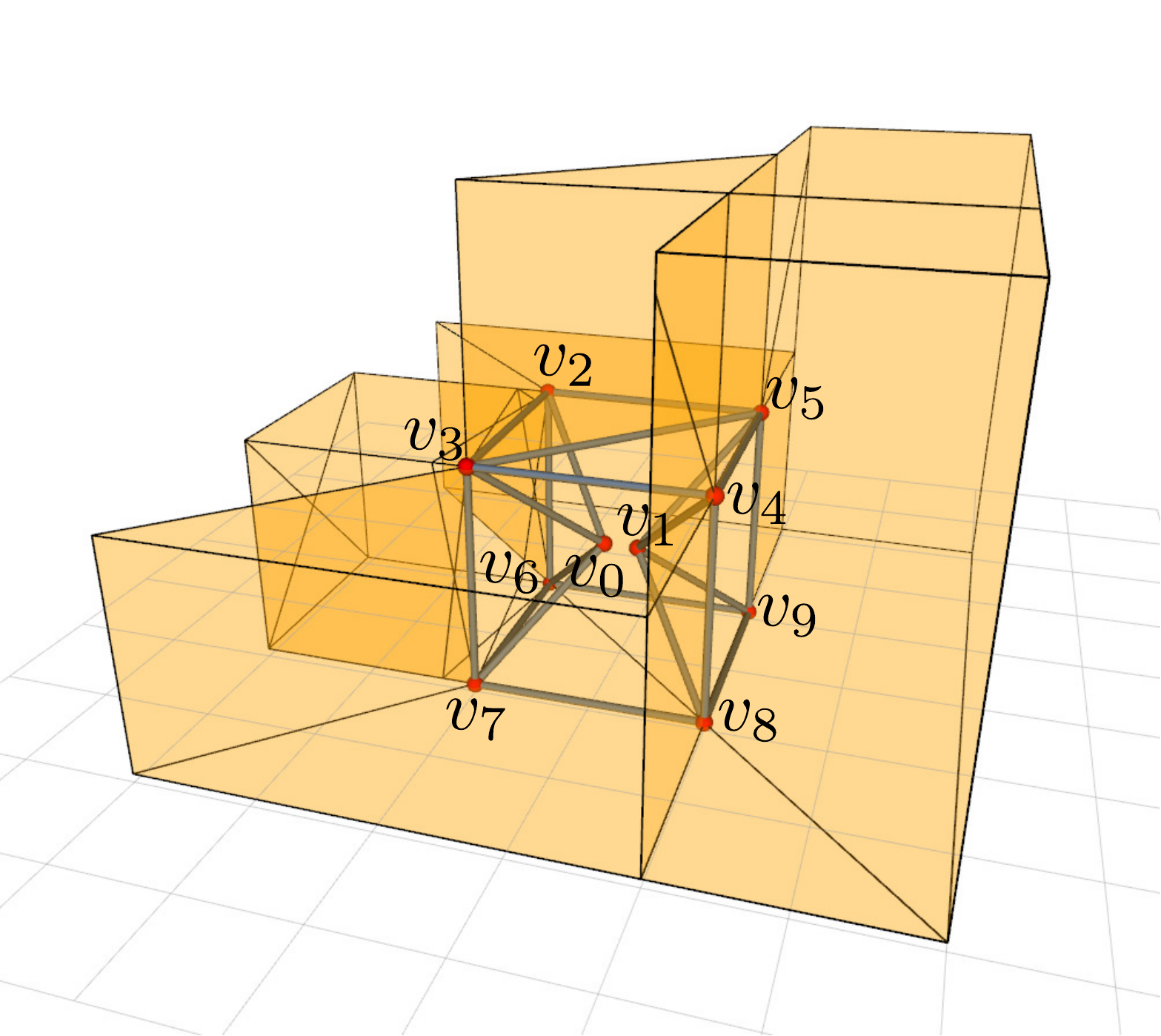}\label{fig:c-free-v1}}
  \caption{(a) $\widehat{\mathcal{C}}_{\mathrm{free}}^{v_0}(q^{v_0})$ is
    computed with all members controlling $v_1$ ignored. (b)
    $\widehat{\mathcal{C}}_{\mathrm{free}}^{v_1}(q^{v_1})$ is computed with all
    members controlling $v_0$ ignored.}
  \label{fig:c-free-group-node}
\end{figure}

Then the free space of this group can be derived as
\begin{gather}
  \label{eq:free-space-group}
  \widehat{\mathcal{C}}_{\mathrm{free}}^{v_i} = \mathbb{R}^3\setminus
  \widehat{\mathcal{C}}_{\mathrm{obs}}^{v_i}\\
  \widehat{\mathcal{C}}_{\mathrm{free}}^{v_j} = \mathbb{R}^3\setminus
  \widehat{\mathcal{C}}_{\mathrm{obs}}^{v_j}
\end{gather}
Using the same boundary search approach
in~\cite{Liu-vtt-cspace-icra-2020}, the boundary of
$\widehat{\mathcal{C}}_{\mathrm{free}}^{v_i}(q^{v_i})$ --- the
enclosed subspace containing the current position of node $v_i$ ---
can be obtained efficiently. Similarly, the boundary of
$\widehat{\mathcal{C}}_{\mathrm{free}}^{v_j}(q^{v_j})$ can be
obtained. For example, given the VTT shown in
Fig.~\ref{fig:geometry-truss}, if node $v_0$ and $v_1$ form a group,
then $\widehat{\mathcal{C}}_{\mathrm{free}}^{v_0}(q^{v_0})$ and
$\widehat{\mathcal{C}}_{\mathrm{free}}^{v_1}(q^{v_1})$ can be computed
and shown in Fig.~\ref{fig:c-free-group-node}.
$\mathcal{C}_{\mathrm{free}}^{v_0}(q^{v_0})$ contains some space on
the right side of $v_1$ because $E^{v_1}$ is ignored. This space shows
that it is possible to move $v_0$ to locations that are currently
blocked by $v_1$ since $v_1$ can be moved away. It is guaranteed that
as long as $v_0$ is moving inside
$\widehat{\mathcal{C}}_{\mathrm{free}}^{v_0}(q^{v_0})$ (the space
shown in Fig.~\ref{fig:c-free-v0}), there must be no collision between
any member in $E^{v_0}$ and any member in
$E\setminus (E^{v_0}\cup E^{v_1})$. Similarly, no collision between
any member in $E^{v_1}$ and any member in
$E\setminus (E^{v_0}\cup E^{v_1})$ can happen if $v_1$ is moving
inside $\widehat{\mathcal{C}}_{\mathrm{free}}^{v_1}(q^{v_1})$ (the
space shown in Fig.~\ref{fig:c-free-v1}). In this way, when planning
the motion of node $v_0$ and $v_1$ using RRT, a sample will only be
generated inside
$\widehat{\mathcal{C}}_{\mathrm{free}}^{v_0}(q^{v_0})$ and
$\widehat{\mathcal{C}}_{\mathrm{free}}^{v_1}(q^{v_1})$, and we only
need to consider self-collision in the group, namely the collision can
only happen among members in $E^{v_0}\cup E^{v_1}$. There is a special
case when these two nodes in the group are connected by a member. Both
ends of the member are moving which is not considered by our obstacle
model. This is an extra case that we need to check when doing the
planning.

\begin{figure}[b]
  \centering
  \includegraphics[width=0.2\textwidth]{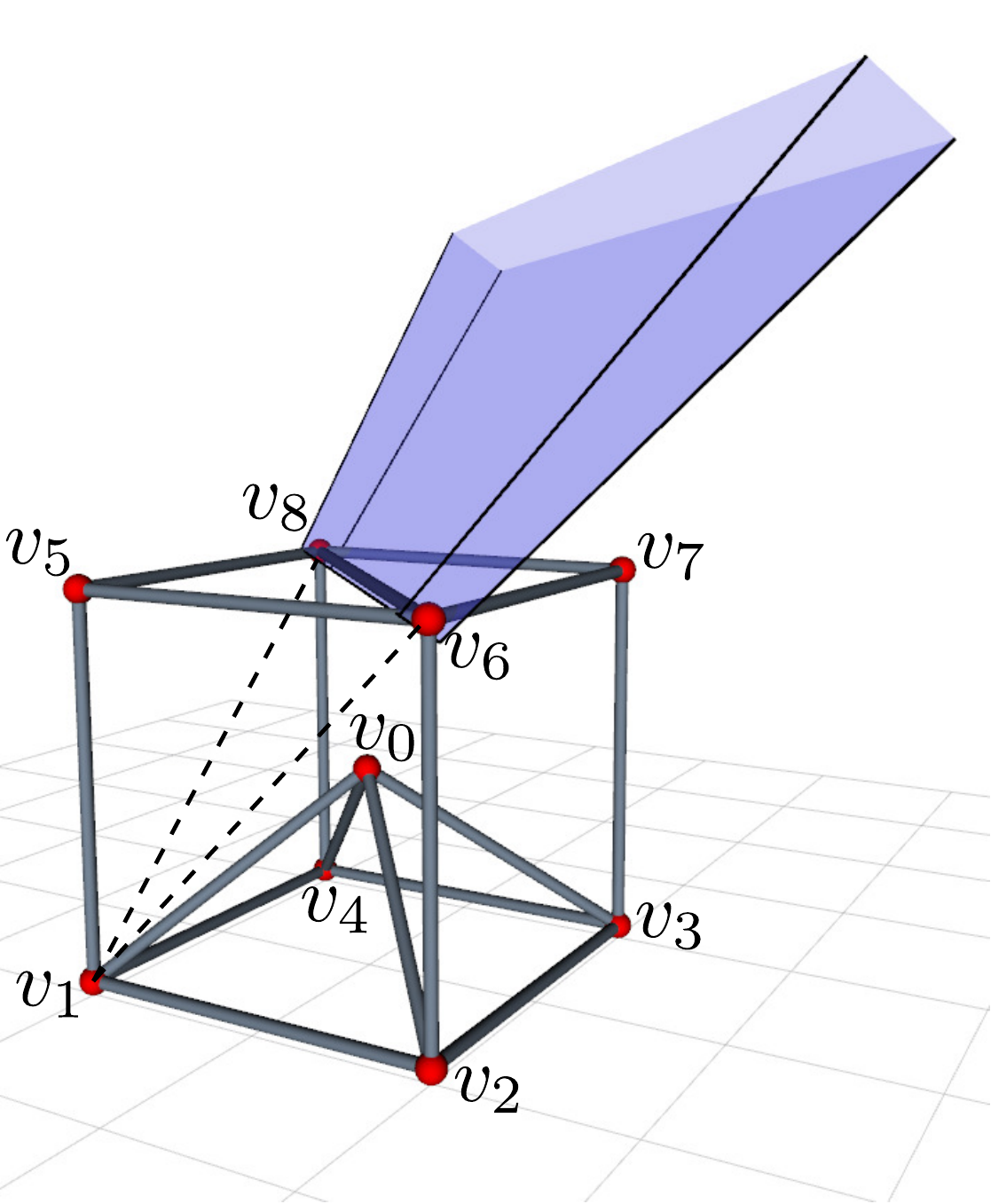}
  \caption{One obstacle polygon of $\mathcal{C}_{\mathrm{obs}}^{v_0}$
    shown in Fig.~\ref{fig:single-polygon} becomes a polyhedron
    bounded by five polygons if the sizes of VTT components are
    considered.}
  \label{fig:exp-obs}
\end{figure}

\begin{figure}[b!]
  \centering
  \subfloat[]{\includegraphics[width=0.2\textwidth]{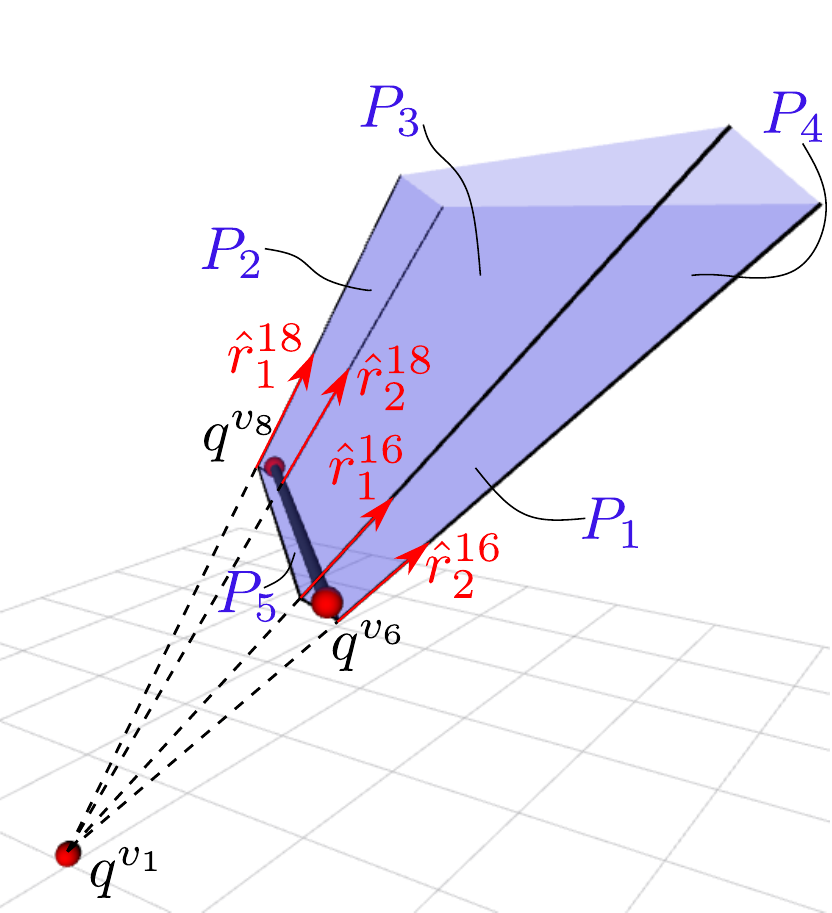}\label{fig:detail_exp_obs}}
  \hfil
  \subfloat[]{\includegraphics[width=0.2\textwidth]{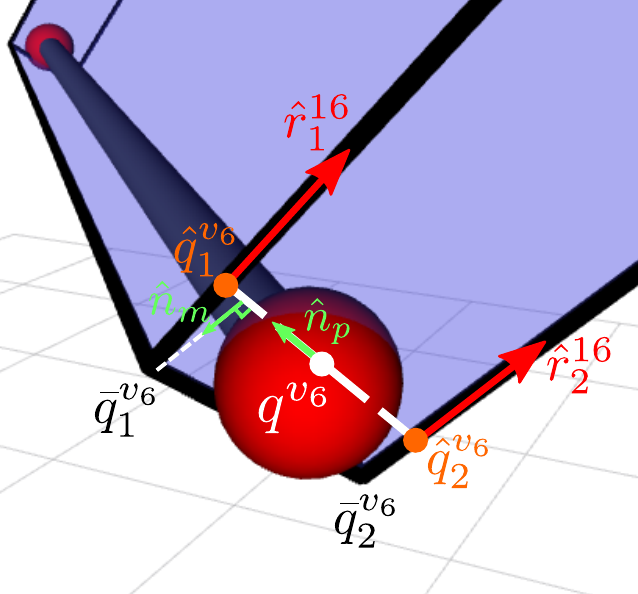}\label{fig:enlarge_exp_obs}}
  \caption{(a) Detailed illustration of the formation of the obstacle
    polyhedron. (b) Close view of the obstacle polyhedron.}
\end{figure}

The previous free space and the obstacle region are derived by
ignoring the physical sizes of members and nodes
(Fig.~\ref{fig:single-polygon}). In practice, a node is a sphere and a
member is a cylinder with non-zero radius. To make the system more
realistic we can increase the derived obstacle region by considering
the physical radii of VTT components. In the obstacle region, every
polygon is converted into a polyhedron. For example, for the VTT shown
in Fig.~\ref{fig:single-polygon}, the purple polygon defined by node
$v_1$ and edge module $(v_6, v_8)$ is one obstacle polygon for node
$v_0$, and this polygon is converted into a polyhedron shown in
Fig.~\ref{fig:exp-obs}. The boundary of this obstacle polyhedron is
composed of five polygons that can be obtained as follows:

The current locations of nodes $v_1$, $v_6$, and $v_8$ are $q^{v_1}$,
$q^{v_6}$, and $q^{v_8}$ respectively, and the unit vector normal to
the plane formed by these three nodes is $\hat{n}_p$ ($-\hat{n}_p$ is
also a normal unit vector but in the opposite direction). As shown in
Fig.~\ref{fig:enlarge_exp_obs}, we first adjust $q^{v_6}$ and
$q^{v_8}$ by moving them along $\hat{n}_p$ and $-\hat{n}_p$
respectively:
\begin{gather*}
    \hat{q}^{v_6}_{1}=q^{v_6}+\lambda\hat{n}_p,\ \hat{q}^{v_6}_{2}=q^{v_6}-\lambda\hat{n}_p\\
    \hat{q}^{v_8}_{1}=q^{v_8}+\lambda\hat{n}_p,\ \hat{q}^{v_8}_{2}=q^{v_8}-\lambda\hat{n}_p
\end{gather*}
where $\lambda$ is the growing size that needs to consider the
physical sizes of the mechanical components. $\lambda$ is usually set
to be slightly larger than the sum of the radius of the node and the
radius of the edge. Then, the four rays shown in
Fig.~\ref{fig:detail_exp_obs} can be derived as
\begin{gather*}
    \hat{r}^{16}_{1}=\frac{\hat{q}^{v_6}_{1}-q^{v_1}}{\left\|\hat{q}^{v_6}_{1}-q^{v_1}\right\|},\ \hat{r}^{16}_{2}=\frac{\hat{q}^{v_6}_{2}-q^{v_1}}{\left\|\hat{q}^{v_6}_{2}-q^{v_1}\right\|}\\
    \hat{r}^{18}_{1}=\frac{\hat{q}^{v_8}_{1}-q^{v_1}}{\left\|\hat{q}^{v_8}_{1}-q^{v_1}\right\|},\ \hat{r}^{18}_{2}=\frac{\hat{q}^{v_8}_{2}-q^{v_1}}{\left\|\hat{q}^{v_8}_{2}-q^{v_1}\right\|}
\end{gather*}
$\hat{n}_m$ is the unit vector perpendicular to the edge
$\left(v_6,\ v_8\right)$, pointing to $v_1$ and lying on the plane
formed by $v_1$, $v_6$, and $v_8$. Then $\hat{q}^{v_6}_{1}$,
$\hat{q}^{v_6}_{2}$, $\hat{q}^{v_8}_{1}$, and $\hat{q}^{v_8}_{2}$ are
moved by $\lambda/2$ projected on $\hat{n}_m$ along
$-\hat{r}^{16}_{1}$, $-\hat{r}^{16}_{2}$, $-\hat{r}^{18}_{1}$, and
$-\hat{r}^{18}_{2}$, respectively:
\begin{align*}
    \bar{q}^{v_6}_{1}&=\hat{q}^{v_6}_{1}-\frac{\lambda}{2\left|\hat{n}_m\cdot\hat{r}^{16}_{1}\right|}\hat{r}^{16}_{1}\\
    \bar{q}^{v_6}_{2}&=\hat{q}^{v_6}_{2}-\frac{\lambda}{2\left|\hat{n}_m\cdot\hat{r}^{16}_{2}\right|}\hat{r}^{16}_{2}\\
    \bar{q}^{v_8}_{1}&=\hat{q}^{v_8}_{1}-\frac{\lambda}{2\left|\hat{n}_m\cdot\hat{r}^{18}_{1}\right|}\hat{r}^{18}_{1}\\
    \bar{q}^{v_8}_{2}&=\hat{q}^{v_8}_{2}-\frac{\lambda}{2\left|\hat{n}_m\cdot\hat{r}^{18}_{2}\right|}\hat{r}^{18}_{2}
\end{align*}
Thus, we can encode the polyhedron boundary that is formed by the five
polygons shown in Fig.~\ref{fig:detail_exp_obs} as:
\begin{align*}
    P_1&=\left\{V=\left(\bar{q}^{v_6}_{1},\ \bar{q}^{v_6}_{2}\right),\ R=\left(\hat{r}^{16}_{1},\ \hat{r}^{16}_{2}\right)\right\}\\
    P_2&=\left\{V=\left(\bar{q}^{v_8}_{1},\ \bar{q}^{v_8}_{2}\right),\ R=\left(\hat{r}^{18}_{1},\ \hat{r}^{18}_{2}\right)\right\}\\
    P_3&=\left\{V=\left(\bar{q}^{v_6}_{1},\ \bar{q}^{v_8}_{1}\right),\ R=\left(\hat{r}^{16}_{1},\ \hat{r}^{18}_{1}\right)\right\}\\
    P_4&=\left\{V=\left(\bar{q}^{v_6}_{2},\ \bar{q}^{v_8}_{2}\right),\ R=\left(\hat{r}^{16}_{2},\ \hat{r}^{18}_{2}\right)\right\}\\
    P_5&=\left\{V=\left(\bar{q}^{v_6}_{1},\ \bar{q}^{v_6}_{2},\ \bar{q}^{v_8}_{1},\ \bar{q}^{v_8}_{2}\right)\right\}
\end{align*}
Note that $P_5$ consists of no rays.

\begin{figure}[b!]
  \centering
  \includegraphics[width=0.20\textwidth]{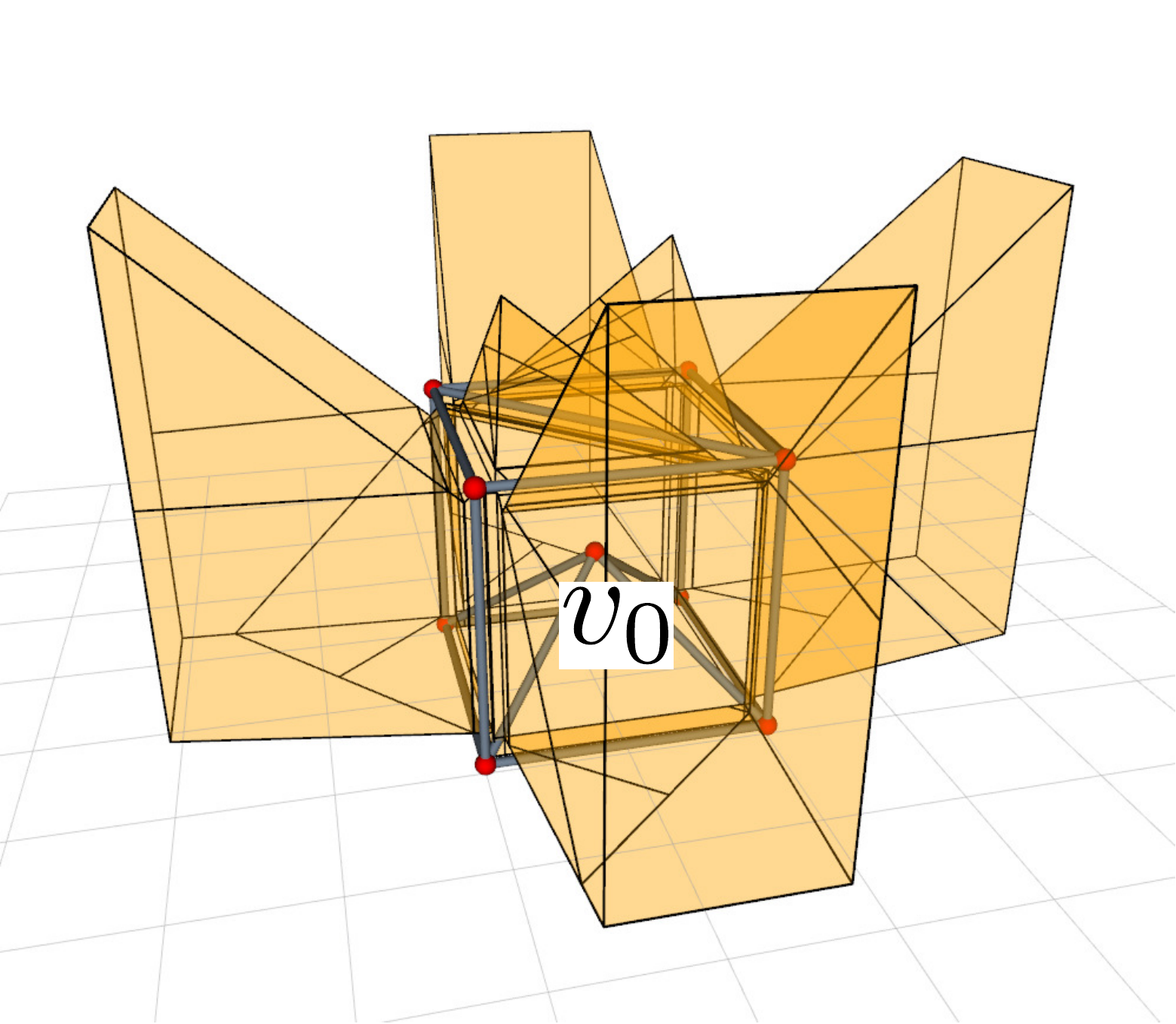}
  \caption{$\mathcal{C}_{\mathrm{free}}^{v_0}(q^{v_0})$ with physical
    sizes of components being considered.}
  \label{fig:bounded-c-free-shrink}
\end{figure}

\begin{algorithm}[t]
  \caption{Boundary Search Algorithm}\label{alg:bsa}
  \KwIn{One obstacle polygon $P_{\mathrm{s}}$,
    the set of obstacle polygons $\mathcal{P}_{\mathrm{obs}}$}
  \KwOut{A set of boundary polygons $\mathcal{P}_{\mathrm{b}}$}
  $\mathcal{P}_{\mathrm{b}}\leftarrow\emptyset$\;
  $\mathcal{Q}_{P}\leftarrow\emptyset$\;
  $\mathcal{Q}_{P}\text{.enqueue(}P_{\mathrm{s}}\text{)}$\;
  \While{$\mathcal{Q}_{P}\neq \emptyset$}{
    $P_{i}\leftarrow \mathcal{Q}_{P}\text{.dequeue()}$\;
    $\mathcal{P}_{\mathrm{b}}\leftarrow\mathcal{P}_{\mathrm{b}}\cup\left\{P_{i}\right\}$\;
    \ForEach{$s_{ij}\in S_{i}$}{ $\overline{P}_{ij} \leftarrow$ the innermost
      polygon in $\mathcal{N}_{ij}$\;
      \If{$\overline{P}_{ij}\notin \mathcal{P}_{\mathrm{b}}\wedge \overline{P}_{ij}\notin
        \mathcal{Q}_{P}$}{
        $\mathcal{Q}_{P}\text{.enqueue(}\overline{P}_{ij}\text{)}$\; }
    } } \Return $\mathcal{P}_{\mathrm{b}}$
\end{algorithm}

We can convert every obstacle polygon in
Fig.~\ref{fig:all-polygons-single-node} into an obstacle polyhedron
(bounded by five polygons) and then process these polygons by
\textit{Polygon Intersection} and \textit{Boundary Search}
from~\cite{Liu-vtt-cspace-icra-2020} to derive
$\mathcal{C}_{\mathrm{free}}^{v_0}(q^{v_0})$ shown in
Fig.~\ref{fig:bounded-c-free-shrink}. The \textit{Boundary Search}
step can be simplified and the modified boundary search algorithm is
shown in Algorithm~\ref{alg:bsa}. Recall that the free space of a node
is usually partitioned by its obstacle region into multiple enclosed
subspaces, and given an obstacle polygon $P_{\mathrm{s}}$ and the set
of all obstacle polygons $\mathcal{P}_{\mathrm{obs}}$ generated by
\textit{Polygon Intersection} step~\cite{Liu-vtt-cspace-icra-2020},
this algorithm can find the enclosed subspace with $P_{\mathrm{s}}$
being part of the boundary. Note that after \textit{Polygon
  Intersection}, each polygon in $\mathcal{P}_{\mathrm{obs}}$ can only
bound one enclosed subspace because every obstacle polygon is the
boundary between the free space and the obstacle region. If we want to
find the boundary of $\mathcal{C}_{\mathrm{free}}^{v_0}(q^{v_0})$, we
can set $P_{\mathrm{s}}$ to be the obstacle polygon that is closest to
$q^{v_0}$. In the algorithm, $S_i$ is the set of all edges of polygon
$P_i$, $s_{ij}$ is the $j$th edge of $P_i$, and $\mathcal{N}_{ij}$ is
the set of all polygons that share $s_{ij}$ with $P_i$. For each
$P_i$, its normal vector pointing outwards the obstacle region is
computed, and $\overline{P}_{ij}$ in \textit{Line 8} is the innermost
polygon in $\mathcal{N}_{ij}$ along the normal vector of $P_i$.

\begin{figure}[t]
  \centering
  \subfloat[]{\includegraphics[width=0.20\textwidth]{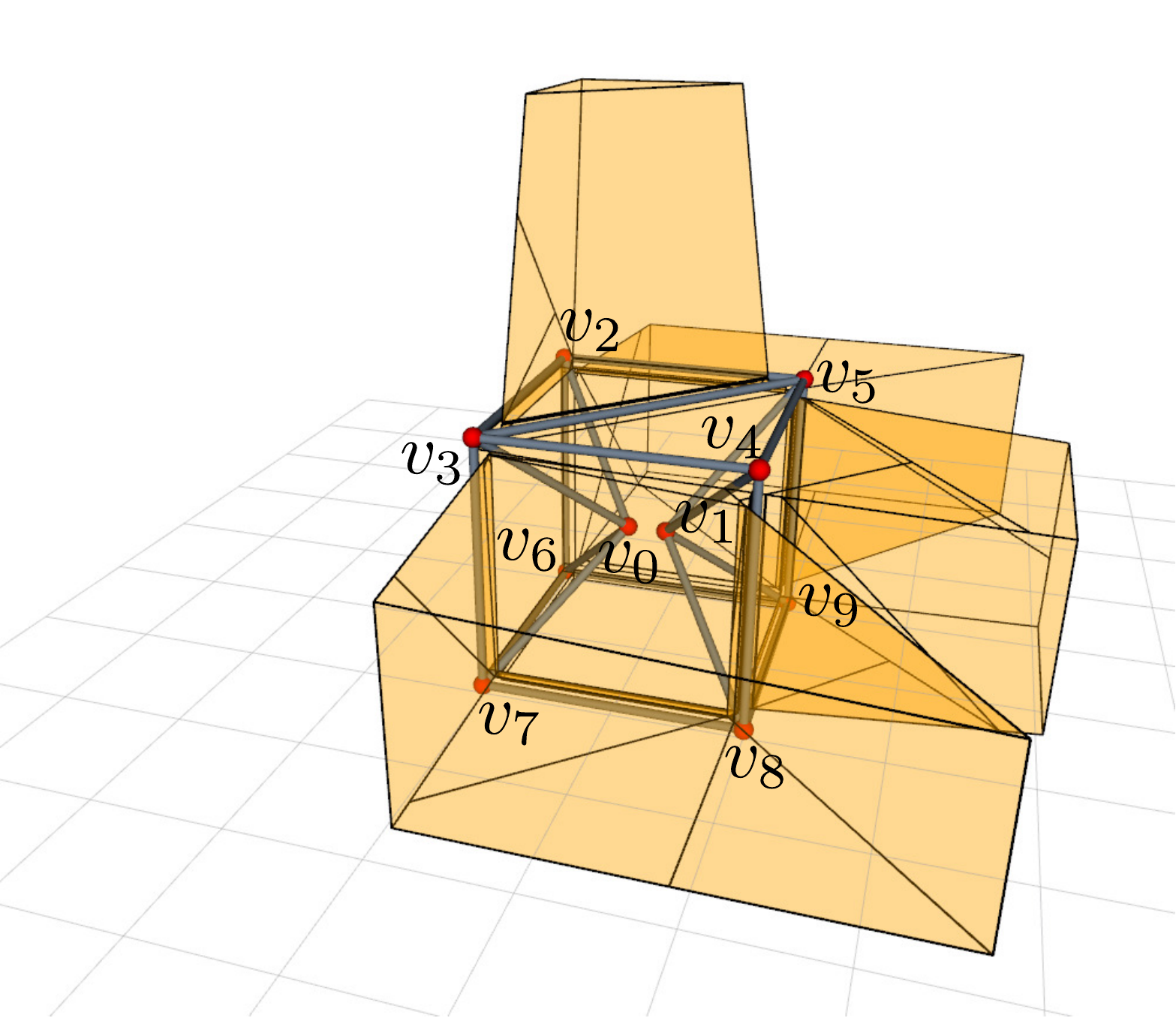}\label{fig:c-free-v0-shrink}}
  \hfil
  \subfloat[]{\includegraphics[width=0.20\textwidth]{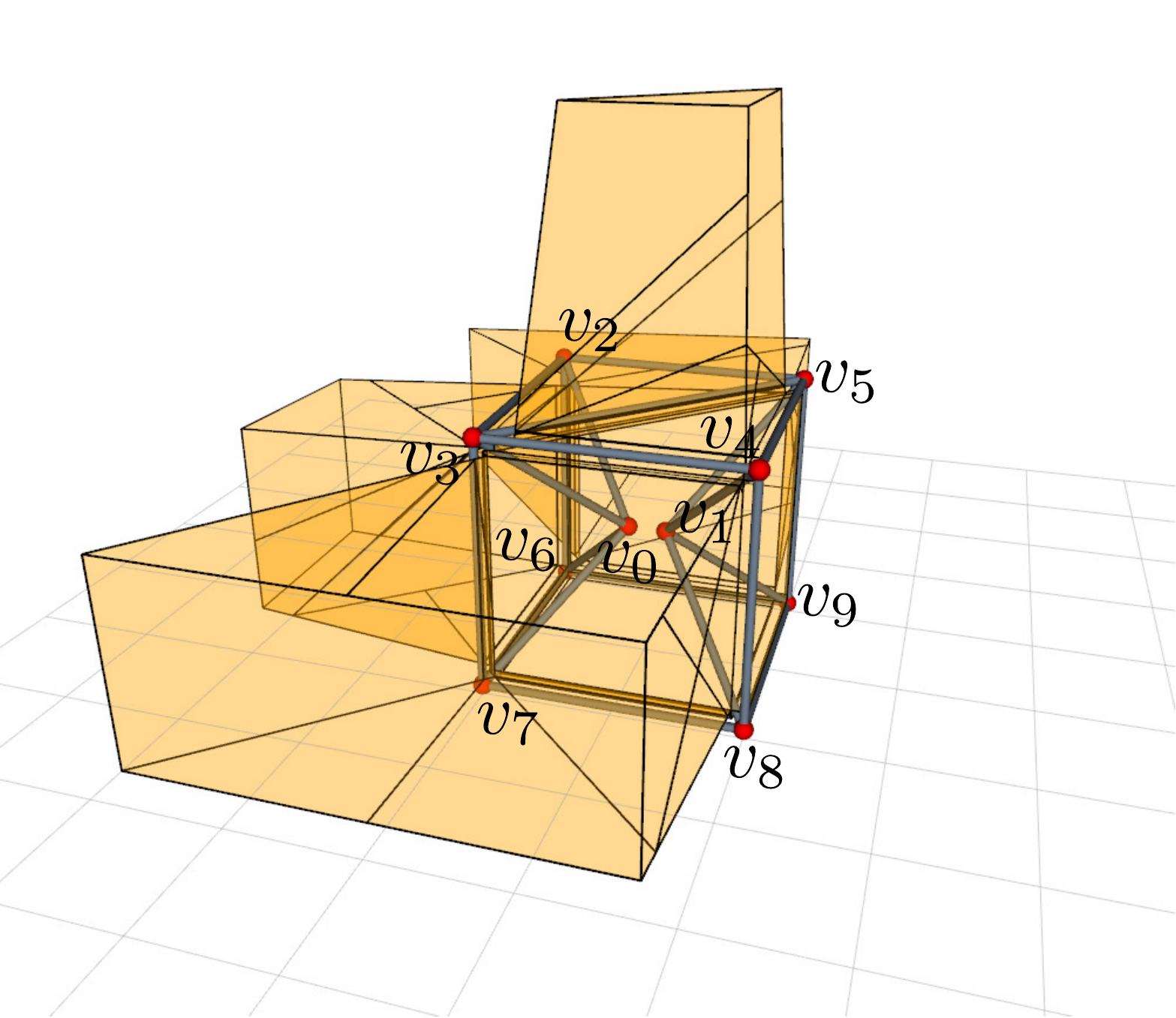}\label{fig:c-free-v1-shrink}}
  \caption{(a) $\widehat{\mathcal{C}}_{\mathrm{free}}^{v_0}(q^{v_0})$ is
    computed with all members controlling $v_1$ ignored. (b)
    $\widehat{\mathcal{C}}_{\mathrm{free}}^{v_1}(q^{v_1})$ is computed with all
    members controlling $v_0$ ignored.}
  \label{fig:c-free-group-node-shrink}
\end{figure}

The new $\mathcal{C}_{\mathrm{free}}^{v_0}(q^{v_0})$ is smaller and
bounded by more polygons. For a single node, if it is on the same
plane with all of its neighbor nodes, then it is in a singular
configuration and we will lose controllability of that node. To avoid
this, we also add this plane to the obstacle polygon set when running
the boundary search algorithm. By taking these constraints into
consideration, we can derive
$\widehat{\mathcal{C}}_{\mathrm{free}}^{v_0}(q^{v_0})$ and
$\widehat{\mathcal{C}}_{\mathrm{free}}^{v_1}(q^{v_1})$ for the group
containing $v_0$ and $v_1$ in the VTT shown in
Fig.~\ref{fig:geometry-truss} and the result is shown in
Fig.~\ref{fig:c-free-group-node-shrink}. In the rest of the paper, we
will use the simpler free space ignoring these physical constraints
for visualization purposes.

\subsection{Path Planning for a Group of Nodes}
\label{sec:path-planning}
If there is only one node $v$ in the group and the motion task is to
move the node from its initial position $q_i^v$ to its goal position
$q_g^v$ where $q_i^v\in \mathcal{C}_{\mathrm{free}}^v(q^v_i)$ and
$q_g^v\in \mathcal{C}_{\mathrm{free}}^v(q^v_i)$ ($q_i^v$ and $q_g^v$
are in the same enclosed subspace), then it is straightforward to
apply RRT approach in $\mathcal{C}_{\mathrm{free}}^v(q^v_i)$ and no
collision can happen as long as the motion of each step is inside
$\mathcal{C}_{\mathrm{free}}^v(q^v_i)$ since this space is usually not
convex.

When moving two nodes $v_i$ and $v_j$ in a group, sampling will only
happen inside $\widehat{\mathcal{C}}_{\mathrm{free}}^{v_i}(q^{v_i})$
and $\widehat{\mathcal{C}}_{\mathrm{free}}^{v_j}(q^{v_j})$ for $v_i$
and $v_j$ respectively. If there is no edge module connecting $v_i$
and $v_j$, then when applying RRT approach, the collision between
moving members and fixed members can be ignored as long as the motion
of both nodes in each step are inside
$\widehat{\mathcal{C}}_{\mathrm{free}}^{v_i}(q^{v_i})$ and
$\widehat{\mathcal{C}}_{\mathrm{free}}^{v_j}(q^{v_j})$
respectively. Only self-collision inside the group --- the collision
among members in $E^{v_i}\cup E^{v_j}$ --- needs to be considered. If
there is an edge module $e=(v_i, v_j)$ which connects $v_i$ and $v_j$,
since this case is not included in our obstacle model when computing
the obstacle region for the group, it is also necessary to check the
collision between $e = (v_i, v_j)$ and every edge module in
$E\setminus (E^{v_i}\cup E^{v_j})$.

In summary, when planning node $v_i$ and $v_j$ in a VTT $G = (V, E)$,
for each step, in order to avoid collision, it is required to ensure
the following:
\begin{itemize}
\item The motion of both node $v_i$ and $v_j$ are inside
  $\widehat{\mathcal{C}}_{\mathrm{free}}^{v_i}(q^{v_i})$ and
  $\widehat{\mathcal{C}}_{\mathrm{free}}^{v_j}(q^{v_j})$ respectively;
\item No collision happens among edge modules in
  $E^{v_i}\cup E^{v_j}$;
\item No collision happens between edge module $e=(v_i, v_j)$ and
  every edge module in $E\setminus(E^{v_i}\cup E^{v_j})$ if
  $e=(v_i, v_j)$ exists.
\end{itemize}

\begin{figure}[b]
  \centering
  \begin{subfloat}[]{\includegraphics[width=0.20\textwidth]{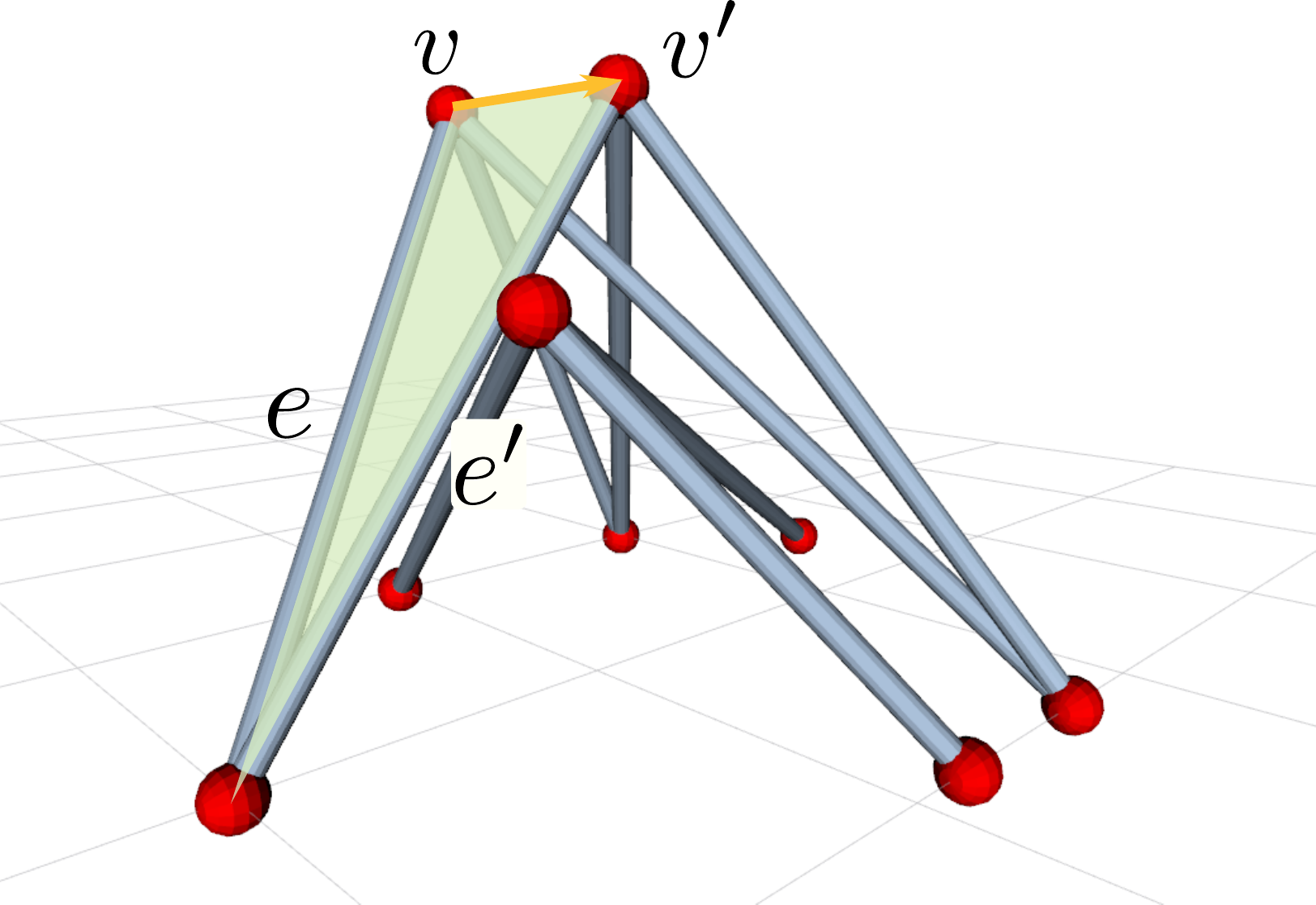}\label{fig:sweep_tri}}
  \end{subfloat}
  \begin{subfloat}[]{\includegraphics[width=0.20\textwidth]{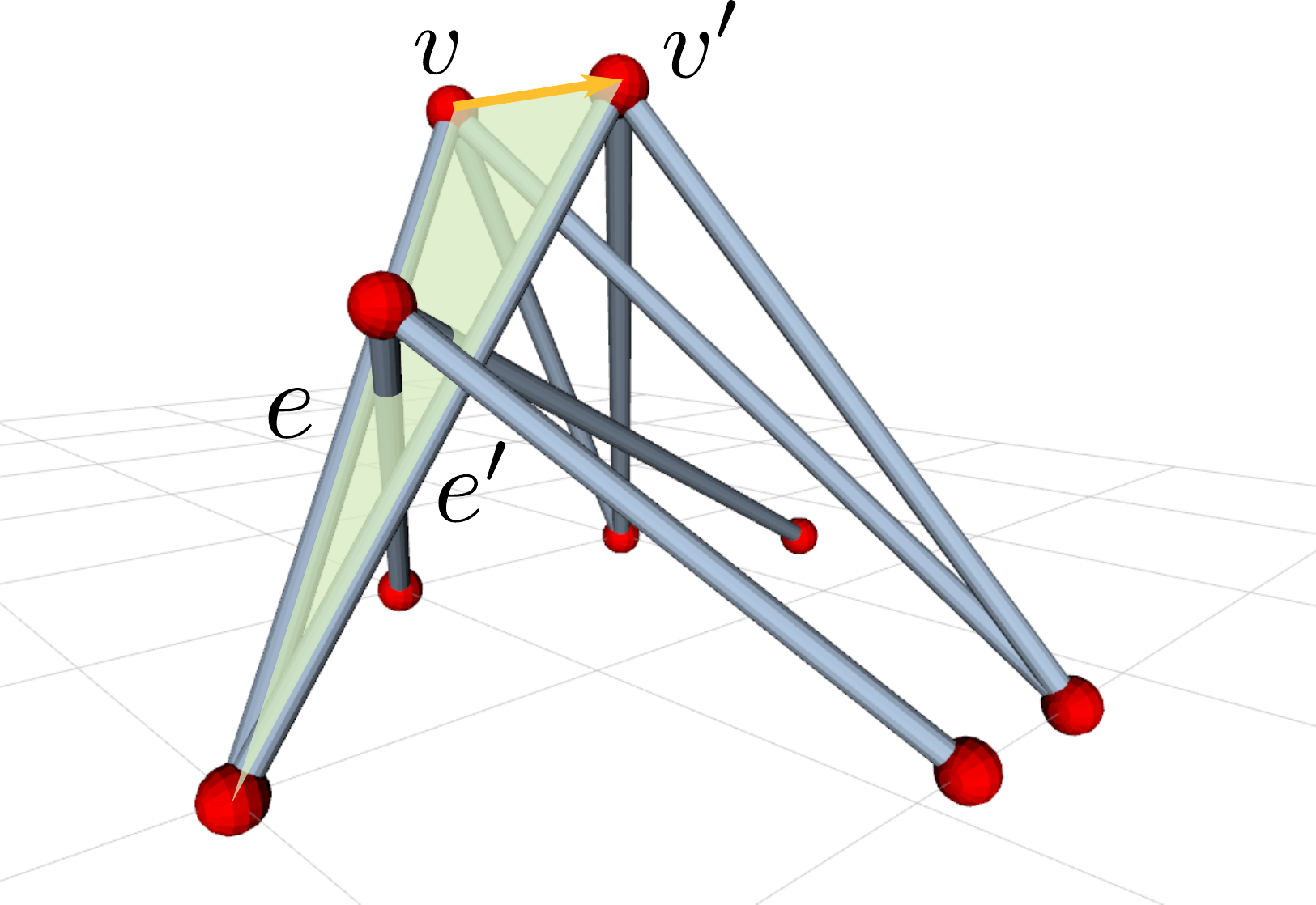}\label{fig:sweep_tri_collision}}
  \end{subfloat}
  \caption{Light green triangle
    ({\color[rgb]{0.847,0.941,0.788}{$\blacktriangle$}}) is sweeped by
    member $e$ when moving node $v$ to new location $v^\prime$ along
    the yellow ({\color[rgb]{0.9922, 0.7490,
        0.1765}{$\bm{\rightarrow}$}}) trajectory and the new position
    of member $e$ is $e^\prime$. $e$ doesn't collide with any other
    members in (a) but does collide with two members in (b) during the
    motion~\cite{Liu-vtt-planning-iros-2019}.}
  \label{fig:collision-model}
\end{figure}

It is difficult to check the second and third collision case during
the motion if both nodes are moving simultaneously. But since the step
size for each node is limited and both of them are moving in straight
lines, we can first check the collision during the motion of $v_i$
while keeping $v_j$ fixed, and then check the motion of
$v_j$~\cite{Liu-vtt-planning-iros-2019}. Every edge module can be
modeled as a line segment in space, thus, when moving node $v$, every
$e\in E^{v}$ sweeps a triangle area, and if this member collides with
another member $\bar{e}\in E$, then $\bar{e}$ must intersect with the
triangle generated by $e$ (Fig.~\ref{fig:collision-model}). Just as in
Section~\ref{sec:obstacle-free}, we can buffer this triangle area to
consider the component radii. Another way is to compute the local
obstacle region of $v_i$ by only taking $E^{v_j}$ into account when
moving $v_i$, and similarly compute the local obstacle region of $v_j$
by only taking $E^{v_i}$ into account when moving $v_j$.  For $v_i$,
given $\mathcal{N}_G(v_i)$ that is the neighbors of $v_i$ and
$E^{v_j}$, the local obstacle region of $v_i$ is the union of the
obstacle polyhedrons
$\forall (v, e)\in \mathcal{N}_G(v_i)\times E^{v_j}$. Collisions occur
if the line segment trajectory of $v_i$ intersects with its local
obstacle region. Similarly for $v_j$. If edge module $(v_i, v_j)$
exists, when moving $v_i$, we also consider the obstacle polyhedron
defined by $v_j$ and $e\in E\setminus (E^{v_i}\cup E^{v_j})$, and
similarly the obstacle polyhedron defined by $v_i$ and
$e\in E\setminus (E^{v_i}\cup E^{v_j})$ when moving $v_j$. In this
way, we avoid solving Eq.~\eqref{eq:min-distance-edges}.

In addition to this constraint we have the length constraint
(Eq.~\ref{eq:length-constraint}), the angle constraint
(Eq.~\ref{eq:angle-constraint}), the manipulability constraint
(Eq.~\ref{eq:manipulability-constraint}), and the stability constraint
when interacting with the environment. For these constraints, it is
straightforward to check discretized states for validity. When
checking the static stability constraint, we first find all supporting
nodes by checking a nodes distance from the ground. The projected
center of mass onto the ground must be within the convex hull formed
by the supporting nodes on the ground.

We can efficiently check the state validity and motion validity, and
RRT-type approaches can be applied. Open Motion Planning Library
(OMPL)~\cite{Sucan-ompl-ram-2012} is used to implement RRT for this
path planning problem.

\subsection{Geometry Reconfiguration Planning}
\label{sec:shape}

Assuming there are $n$ nodes $\{v_t\in V\vert t=1,2\cdots,n\}$ that
should be moved from their initial positions
$q_i^{v_1}, q_i^{v_2}, \cdots, q_i^{v_n}$ to their goal positions
$q_g^{v_1}, q_g^{v_2}, \cdots, q_g^{v_n}$ respectively, we first
divide these nodes into $\lceil n /2\rceil$ groups. Each group
contains at most two nodes. The motion task is achieved by moving
nodes one group at a time. Then this geometry reconfiguration problem
results in a search for a sequence of group motions that can achieve
the task. This approach solves the geometry reconfiguration planning
problem faster than that in~\cite{Liu-vtt-cspace-icra-2020}.

\section{Topology Reconfiguration}
\label{sec:topology}

Topology reconfiguration involves changing the connectivity among edge
modules by docking and undocking. This reconfiguration process is
often difficult for modular robotic systems so the necessary
requirement for topology reconfiguration should be verified. Recall
that the free space of a node is usually not a single connected
component and if a motion task spans multiple connected components, if
a solution exists it must include topology reconfiguration.

\subsection{Enclosed Subspace in Free Space}
\label{sec:cells}

As mentioned before, each polygon after \textit{Polygon Intersection}
in $\mathcal{C}_{\mathrm{obs}}^v$ bounds only one enclosed
subspace. Therefore we can compute all the enclosed subspaces by
repeatedly applying Algorithm~\ref{alg:bsa} as shown in
Algorithm~\ref{alg:fssa}.

\begin{figure}[t]
  \centering
  \subfloat[]{\includegraphics[width=0.2\textwidth]{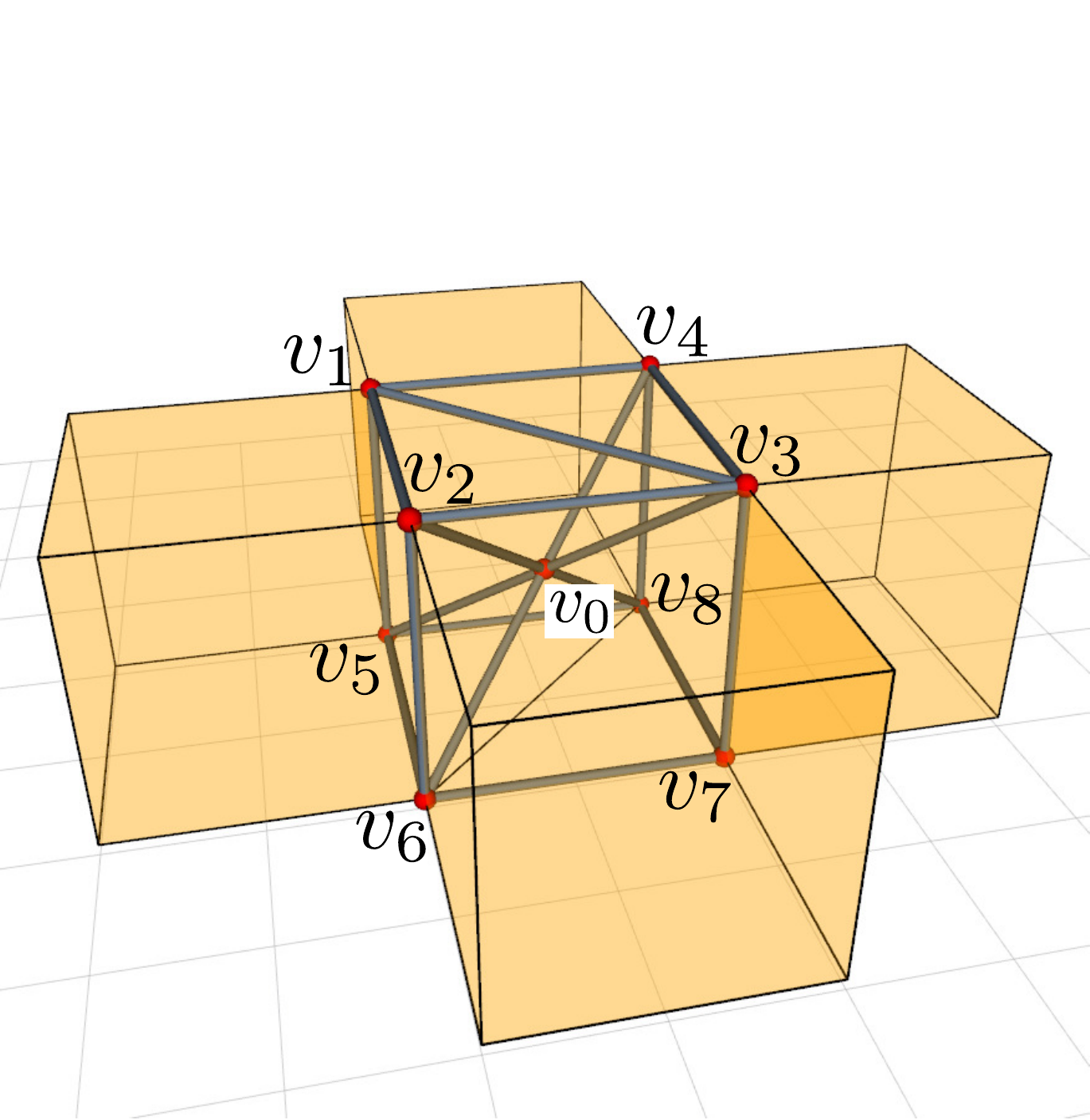}\label{fig:c-free-topo1}}
  \hfil
  \subfloat[]{\includegraphics[width=0.2\textwidth]{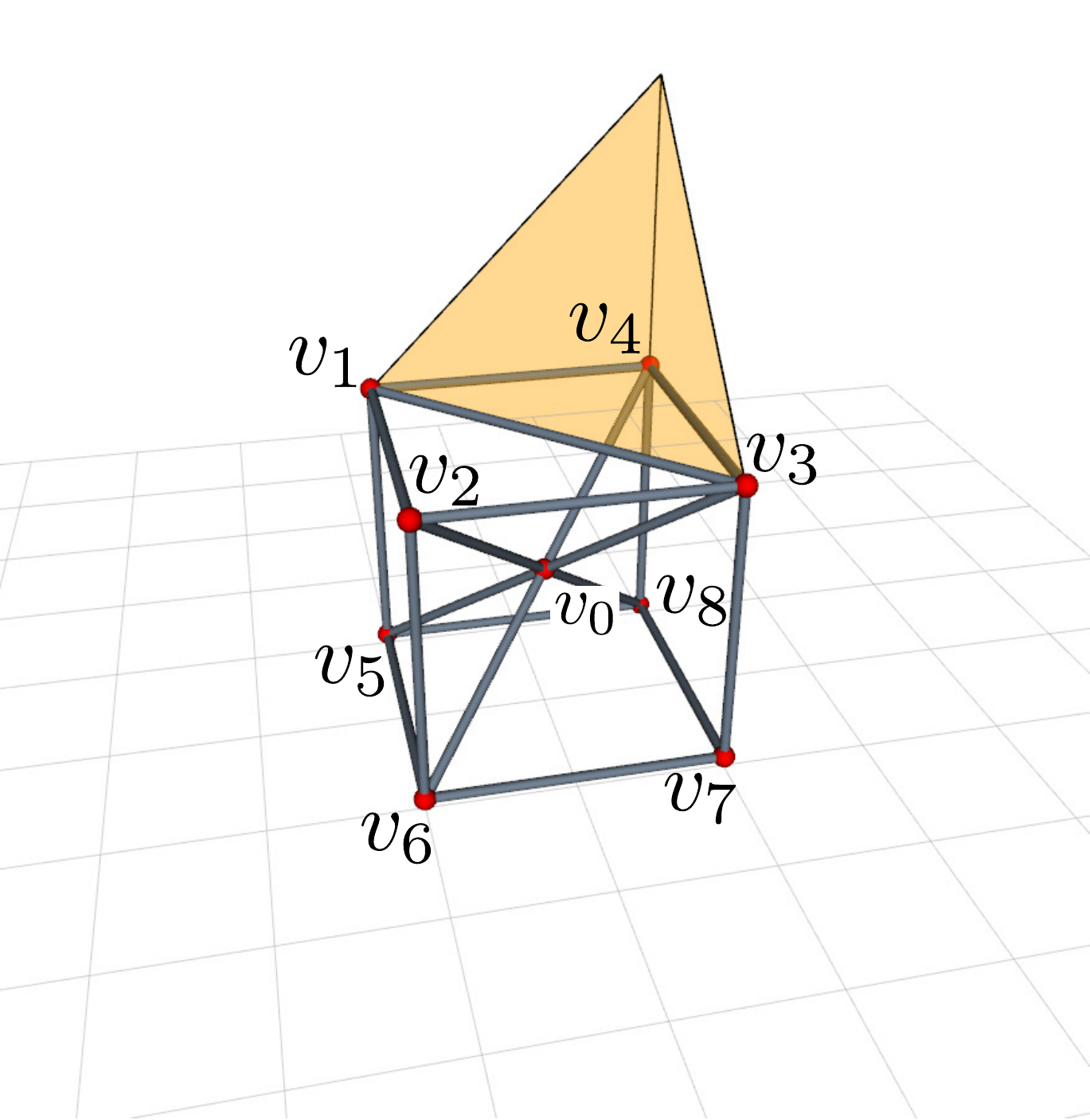}\label{fig:c-free-topo2}}
  \caption{(a) Enclosed subspace
    $\mathcal{C}_{\mathrm{free}}^{v_0}(q^{v_0})$ contains the current
    position of $v_0$. (b) Another enclosed subspace is separated from
    $\mathcal{C}_{\mathrm{free}}^{v_0}(q^{v_0})$ by obstacles.}
  \label{fig:enclosed-subspace-example}
\end{figure}

In this algorithm, we first obtain the set of all obstacle polygons
$\mathcal{P}_{\mathrm{obs}}^v$ by \textit{Polygon Intersection}. Then,
search for the enclosed subspace containing the current node
configuration --- $\mathcal{C}_{\mathrm{free}}^v(q^v)$ --- from a
starting polygon $P_{\mathrm{s}}$ that is the nearest one to the
node~\cite{Liu-vtt-cspace-icra-2020}. Afterward, we compute all other
enclosed subspaces in $\mathcal{C}_{\mathrm{free}}^v$ by searching the
boundary starting from any polygon that has not been
used. Fig.~\ref{fig:enclosed-subspace-example} shows two enclosed
subspaces of node $v_0$ in a simple cubic truss. In total, there are
33 enclosed subspaces in $\mathcal{C}_{\mathrm{free}}^{v_0}$ above the
ground.

\begin{algorithm}[t ]
  \caption{Enclosed Subspace Search}\label{alg:fssa}
  \SetKwFunction{boundarySearch}{BoundarySearch}
  \KwIn{VTT $G=(V,E)$, node $v\in V$}
  \KwOut{The set of all enclosed subspaces
    $\mathcal{C}_{\mathrm{free}}^v$}
  Compute $\mathcal{P}_{\mathrm{obs}}$\;
  $P_{\mathrm{s}}\leftarrow$polygon closest to node $v$\;
  $\mathcal{C}_{\mathrm{free}}^v(q^v)\leftarrow\boundarySearch{$P_{\mathrm{s}},\ \mathcal{P}_{\mathrm{obs}}$}$\;
  $\mathcal{C}_{\mathrm{free}}^v\leftarrow\left\{\mathcal{C}_{\mathrm{free}}^v(q^v)\right\}$\;
  \ForEach{$P_i\in \mathcal{P}_{\mathrm{obs}}^v$}{
    \If{$P_i\notin\mathcal{C}\ \forall\mathcal{C}\in\mathcal{C}_{\mathrm{free}}^v$}{
      $\mathcal{C}_{\mathrm{new}}\leftarrow\boundarySearch{$P_i,\ \mathcal{P}_{\mathrm{obs}}$}$\;
      $\mathcal{C}_{\mathrm{free}}^v\leftarrow\mathcal{C}_{\mathrm{free}}^v\cup\{\mathcal{C}_{\mathrm{new}}\}$\;
    }
  }
\end{algorithm}

\subsection{Topology Reconfiguration Actions}
\label{sec:topology-action}

\begin{figure}[t]
  \centering
  \subfloat[]{\includegraphics[width=0.2\textwidth]{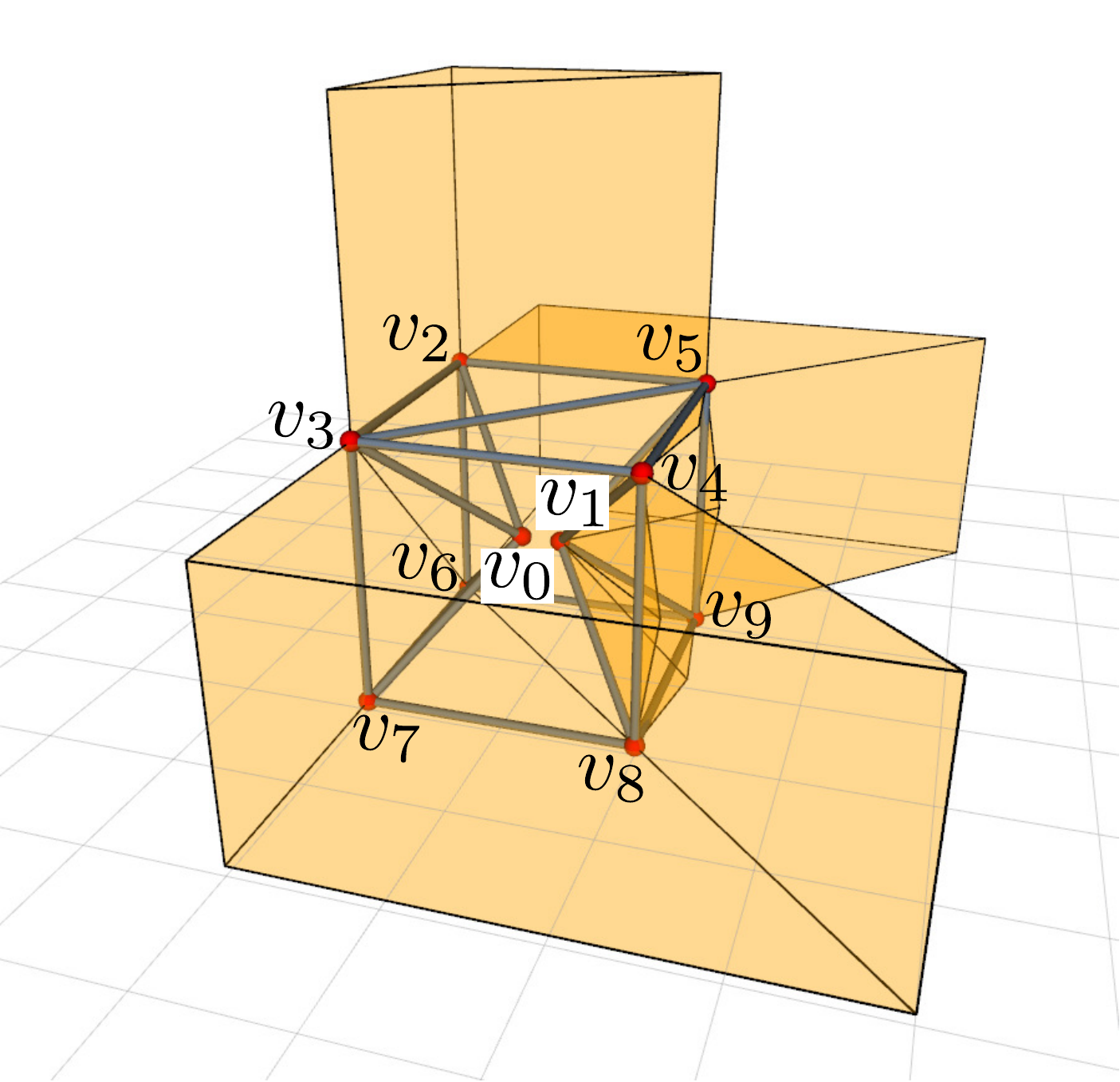}\label{fig:split-cspace}}
  \hfil
  \subfloat[]{\includegraphics[width=0.2\textwidth]{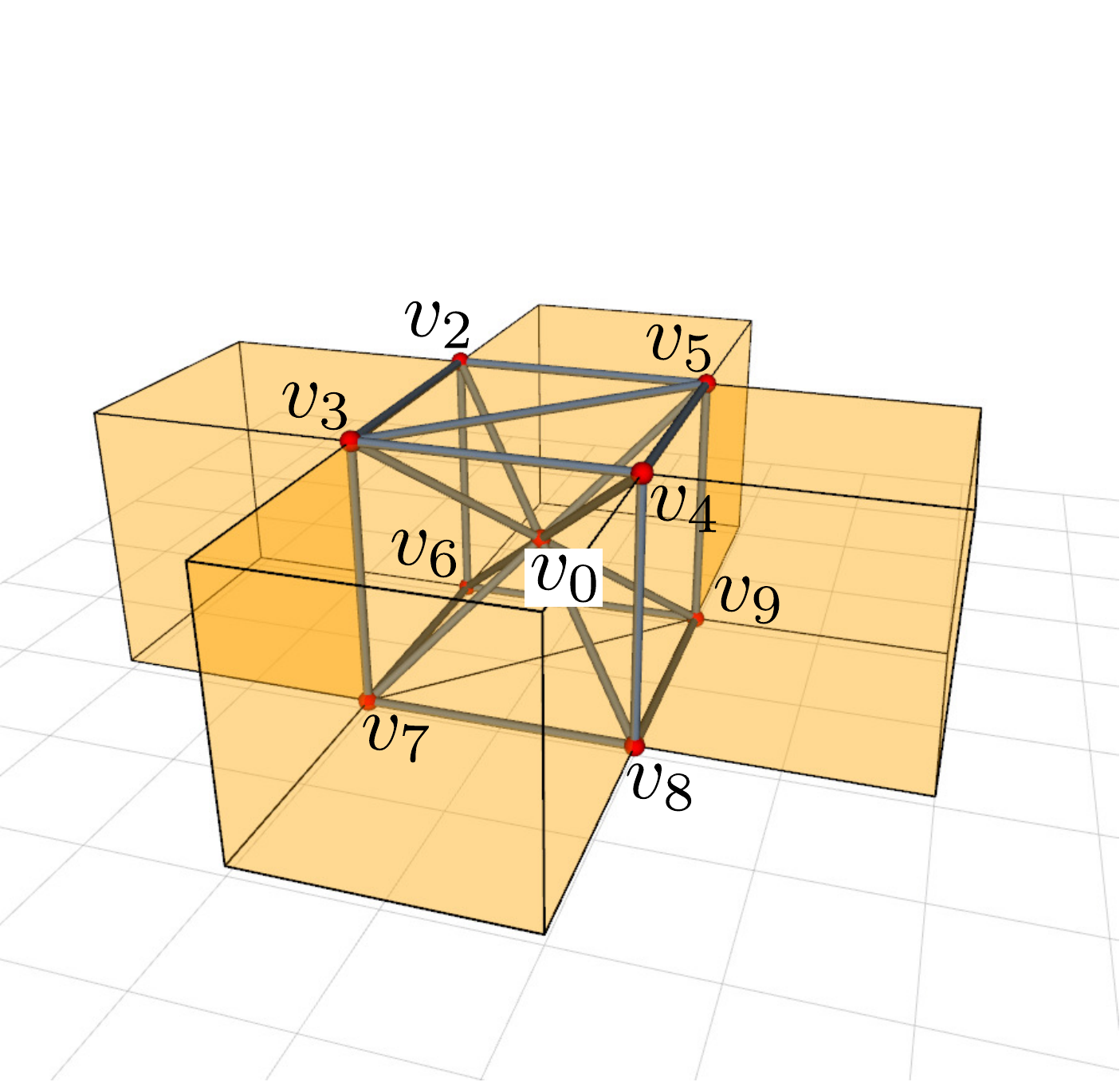}\label{fig:merge-cspace}}
  \caption{(a) Enclosed subspace
    $\mathcal{C}_{\mathrm{free}}^{v_0}(q^{v_0})$ when $v_0$ and $v_1$
    are separated; (b) Enclosed subspace
    $\mathcal{C}_{\mathrm{free}}^{v_0}(q^{v_0})$ after merging $v_1$
    with $v_0$.}
\end{figure}

There are two topology reconfiguration actions for VTT: \texttt{Split}
and \texttt{Merge}~\cite{Spinos-vtt-iros-2017}. A node constructed by
six or more edge modules can be split into two separate nodes (every
node must have at least three edges). Two separate nodes can merge
into one node. These actions can significantly affect the motions of
the involved nodes with a variety of constraints. For the VTT shown in
Fig.~\ref{fig:geometry-truss},
$\mathcal{C}_{\mathrm{free}}^{v_0}(q^{v_0})$ that is the enclosed
subspace containing the current location of node $v_0$ is shown in
Fig.~\ref{fig:split-cspace}. This node can move to some locations
outside the truss but its motion is also blocked in some
directions. The members attached with node $v_1$ blocked the motion of
node $v_0$ on this side, and the plane formed by node $v_2$, $v_3$,
$v_6$, and $v_7$ also blocks the motion of $v_0$ due to: 1. collision
avoidance with member $(v_2,v_3)$, $(v_2,v_6)$, $(v_3,v_7)$, and
$(v_6,v_7)$; 2. singularity avoidance that $v_0$ cannot move through
the square formed by $v_2$, $v_3$, $v_6$, and $v_7$. If merging $v_1$
and $v_0$ as $v_0$ shown in Fig.~\ref{fig:merge-cspace},
$\mathcal{C}_{\mathrm{free}}^{v_0}(q^{v_0})$ will be changed. The
boundary originally blocked by members attached with node $v_1$
disappears because all members controlling $v_0$ and $v_1$ are
combined into a single set. In addition, this \texttt{Merge} action
increases the motion manipulability of node $v_0$ so it can move
through the square formed by $v_2$, $v_3$, $v_6$, and $v_7$. However,
the reachable space is reduced such as the space above the truss and
around the member $(v_4, v_8)$.

\begin{figure}[t]
  \centering
  \subfloat[]{\includegraphics[width=0.13\textwidth]{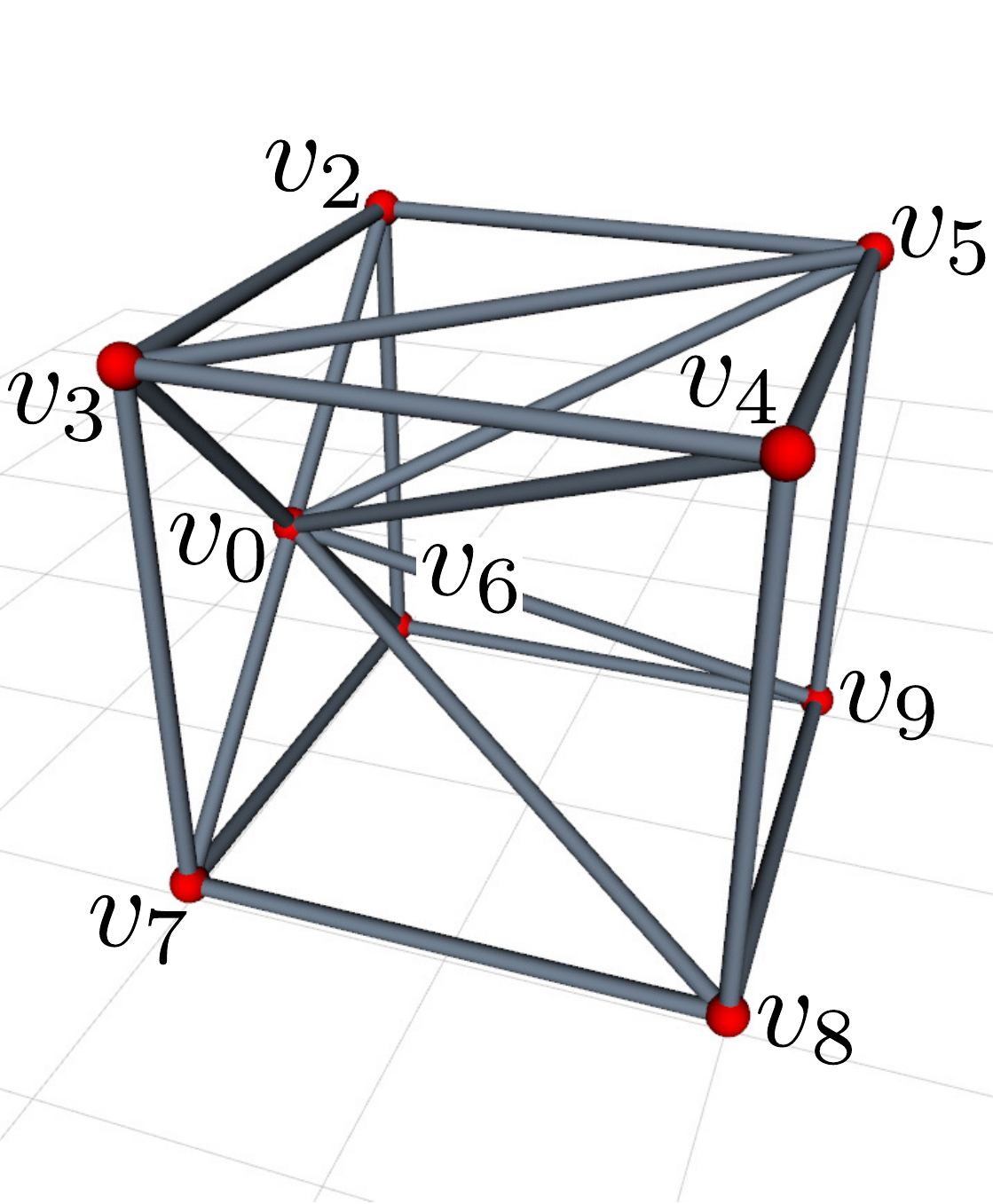}\label{fig:merge-singular}}
  \hfil
  \subfloat[]{\includegraphics[width=0.13\textwidth]{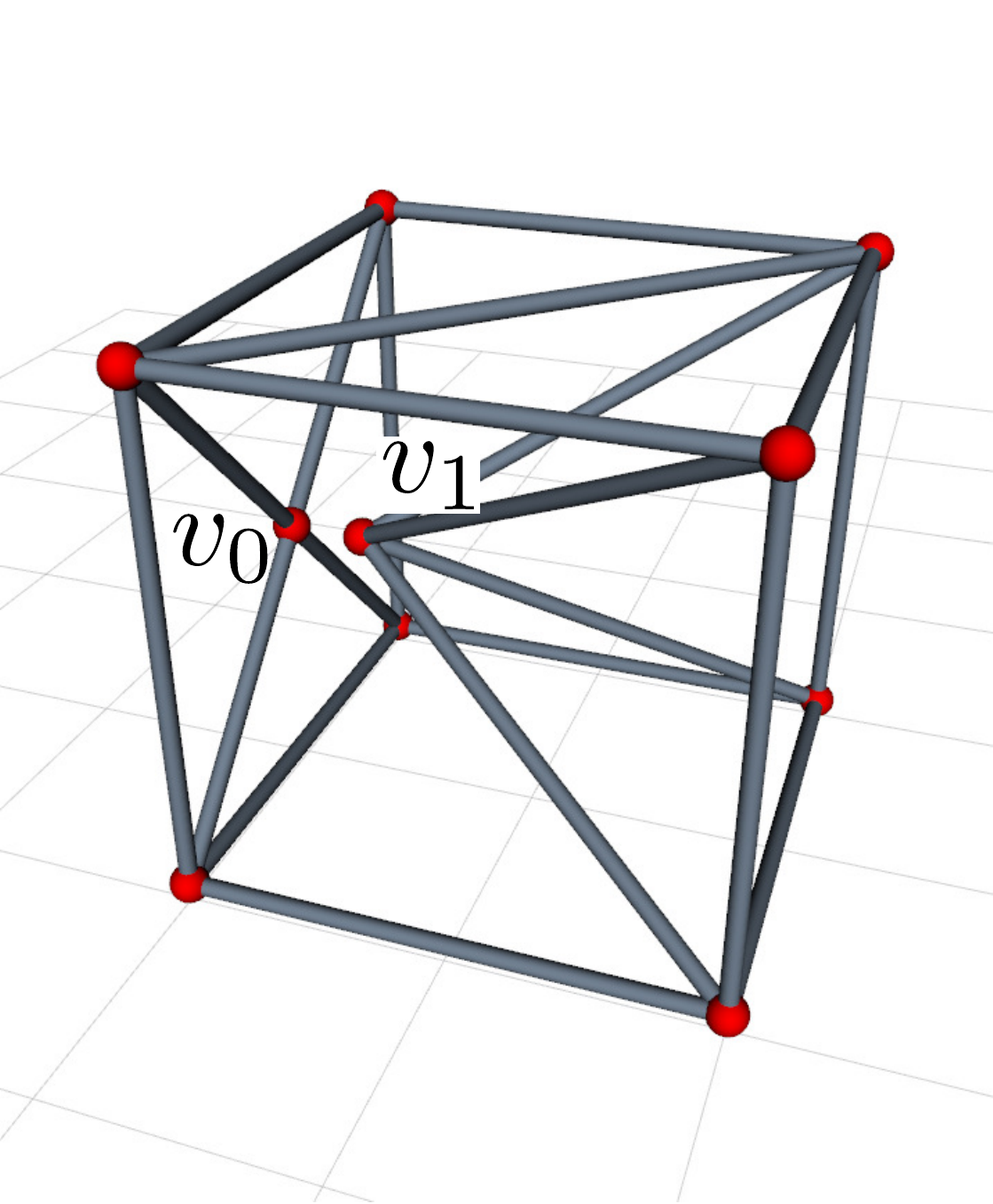}\label{fig:split-singular}}
  \caption{(a) Node $v_0$, $v_2$, $v_3$, $v_6$, and $v_7$ are on the
    same plane. (b) Splitting node $v_0$ in this way to separate
    $E^{v_0}$ into two sets is not valid, because node $v_0$ will be
    in singular configuration. Similarly, it is not valid to merge
    $v_1$ with $v_0$.}
  \label{fig:topology-singular}
\end{figure}

\begin{figure}[t!]
  \centering
  \subfloat[]{\includegraphics[width=0.13\textwidth]{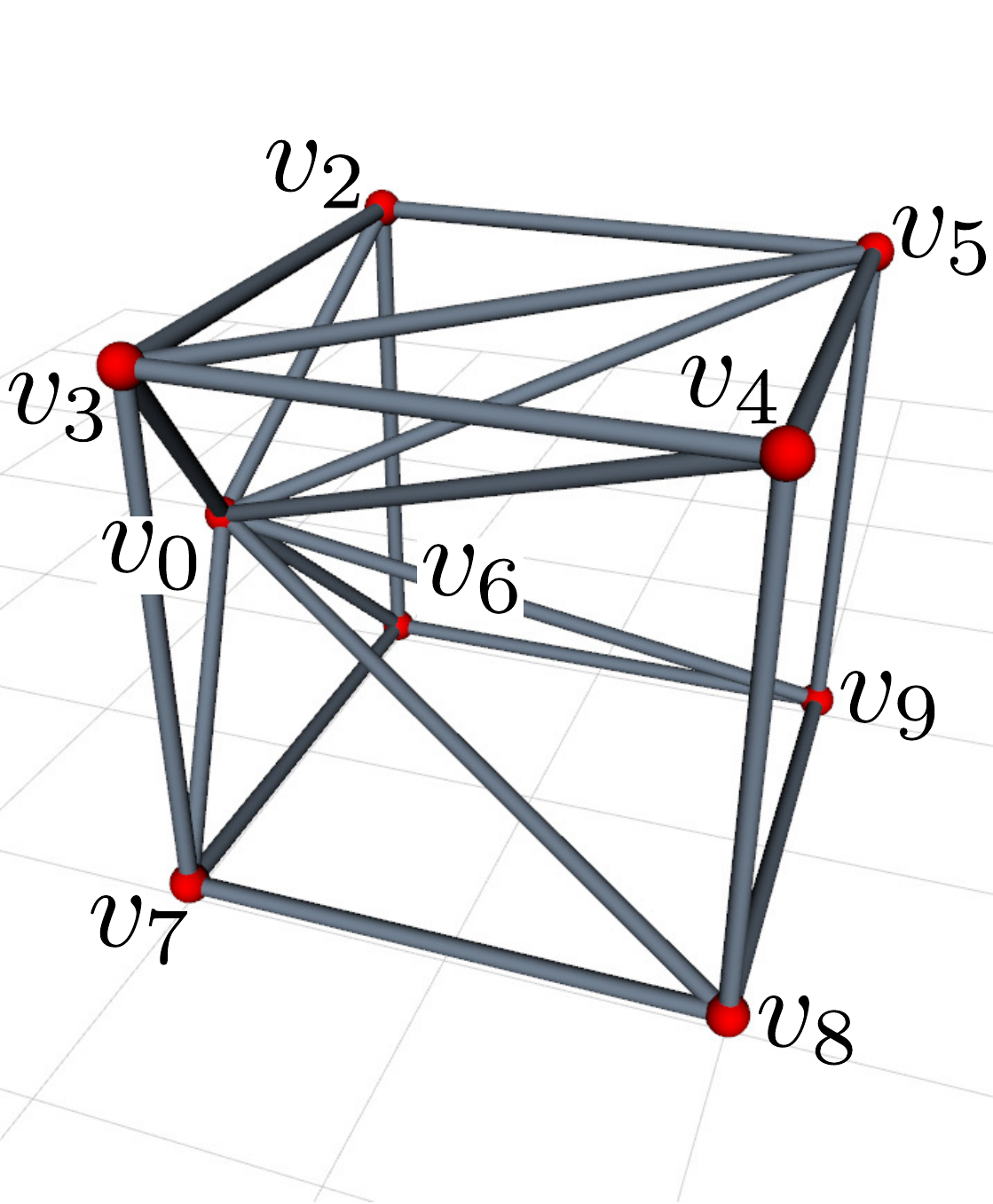}\label{fig:merge-nosingular}}
  \hfil
  \subfloat[]{\includegraphics[width=0.13\textwidth]{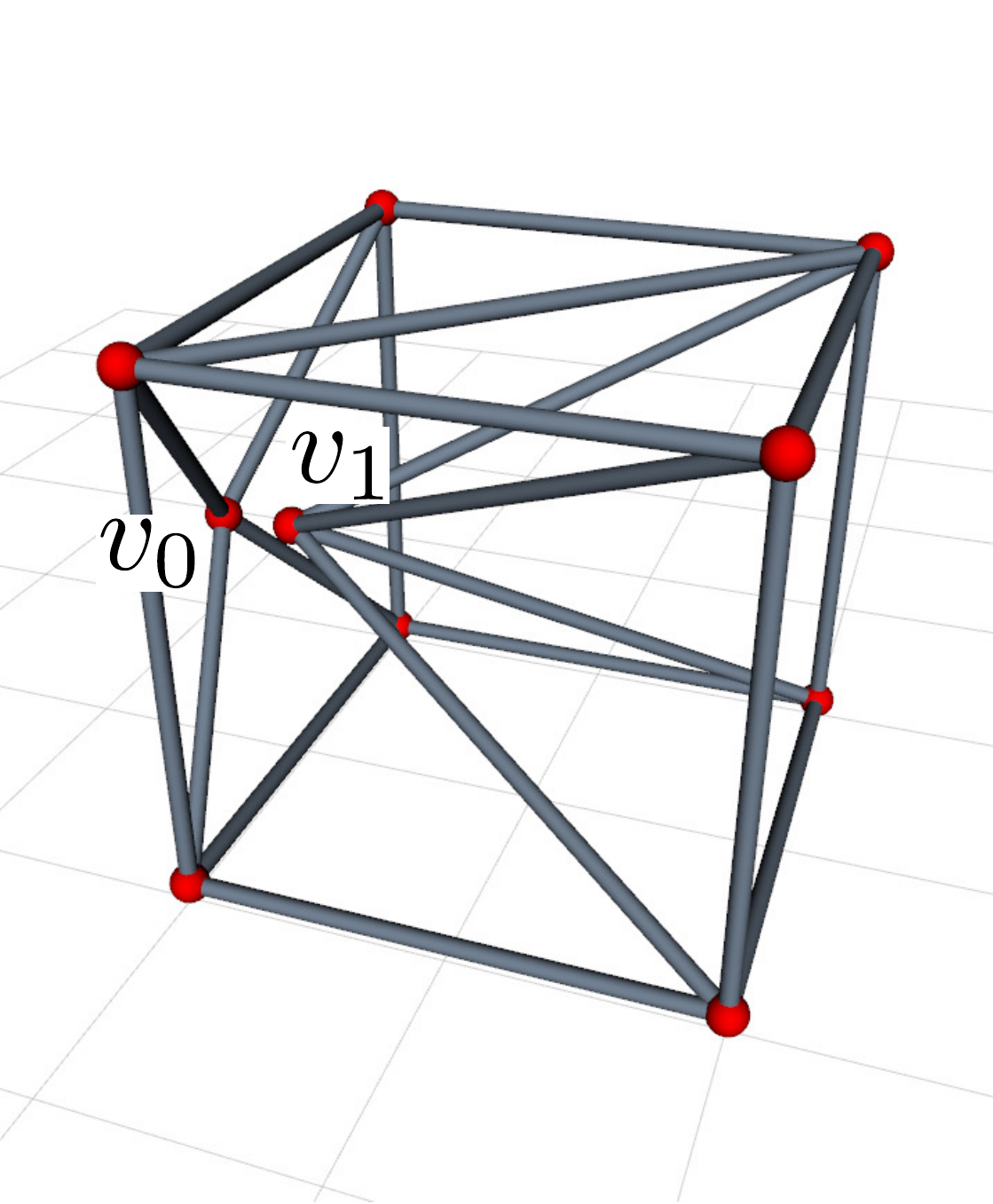}\label{fig:split-nosingular}}
  \caption{(a) Node $v_0$ is outside the truss. (b) It is possible to
    split node $v_0$ in this way to generate $v_1$. Reversely, $v_0$
    and $v_1$ can be merged.}
\end{figure}

When executing a \texttt{Split} action on a node, one requirement is
that this node must have at least six edge modules. Here we physically
split one node into two and move them apart. In order to guarantee
motion controllability, singularities should be avoided. For example,
when node $v_0$ is on the same plane with node $v_2$, $v_3$, $v_6$,
and $v_7$ shown in Fig.~\ref{fig:merge-singular}, we cannot split node
$v_0$ as in Fig.~\ref{fig:split-singular}, because one of the newly
generated nodes ($v_0$ in this case) will be in a singular
configuration. In comparison, if $v_0$ is outside the truss shown in
Fig.~\ref{fig:merge-nosingular}, then it is feasible to apply the same
\texttt{Split} action to generate two new nodes shown in
Fig.~\ref{fig:split-nosingular}. This constraint also applies for the
\texttt{Merge} action. In Fig.~\ref{fig:split-singular}, since node
$v_0$ is in a singular configuration, we cannot merge $v_1$ with $v_0$
to become a new truss shown in Fig.~\ref{fig:merge-singular}. However,
it is feasible to merge $v_0$ and $v_1$ in
Fig.~\ref{fig:split-nosingular} to become $v_0$ in
Fig.~\ref{fig:merge-nosingular}.

\begin{figure}[t]
  \centering
  \subfloat[]{\includegraphics[width=0.13\textwidth]{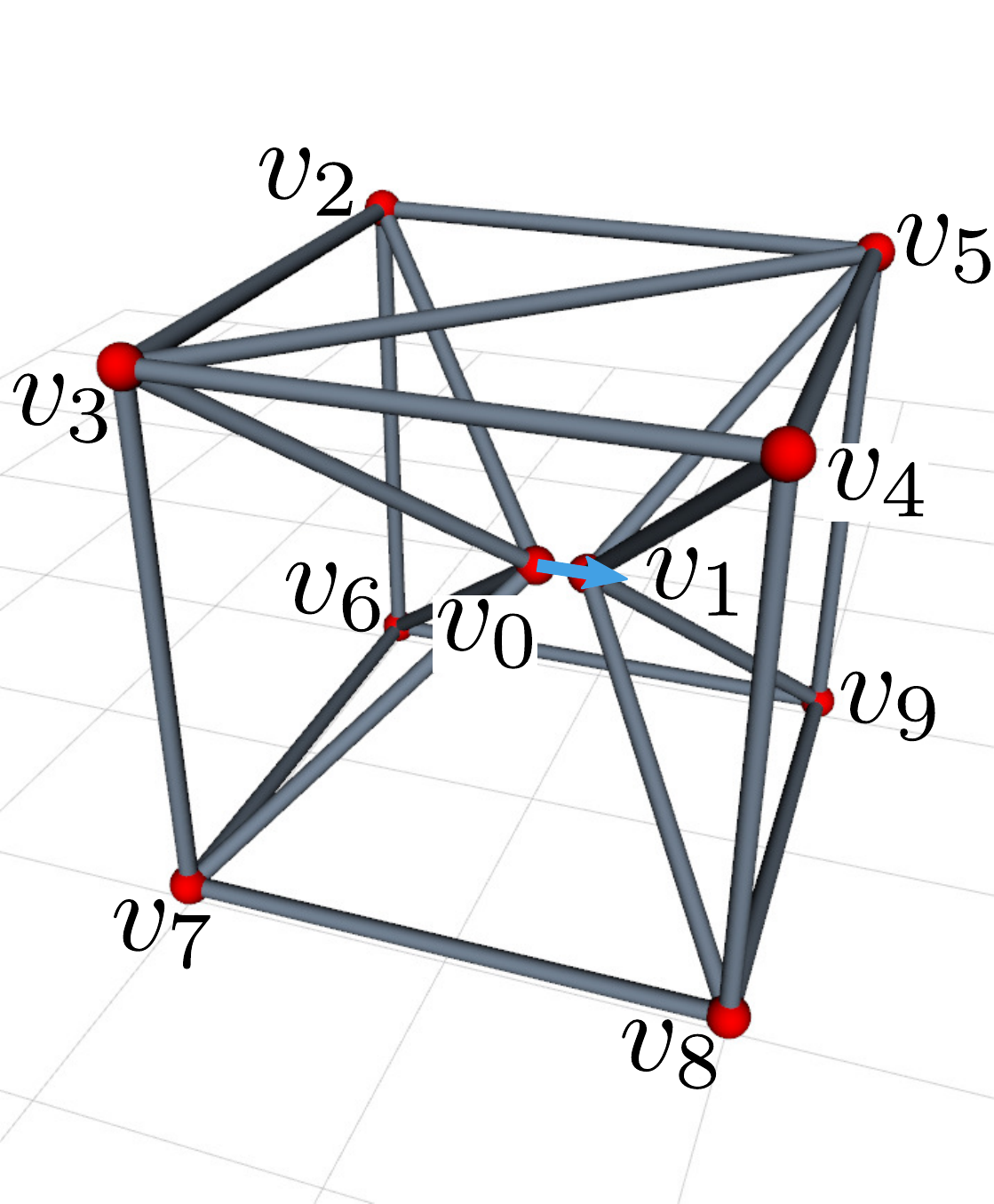}\label{fig:split-rightmove}}
  \hfil
  \subfloat[]{\includegraphics[width=0.13\textwidth]{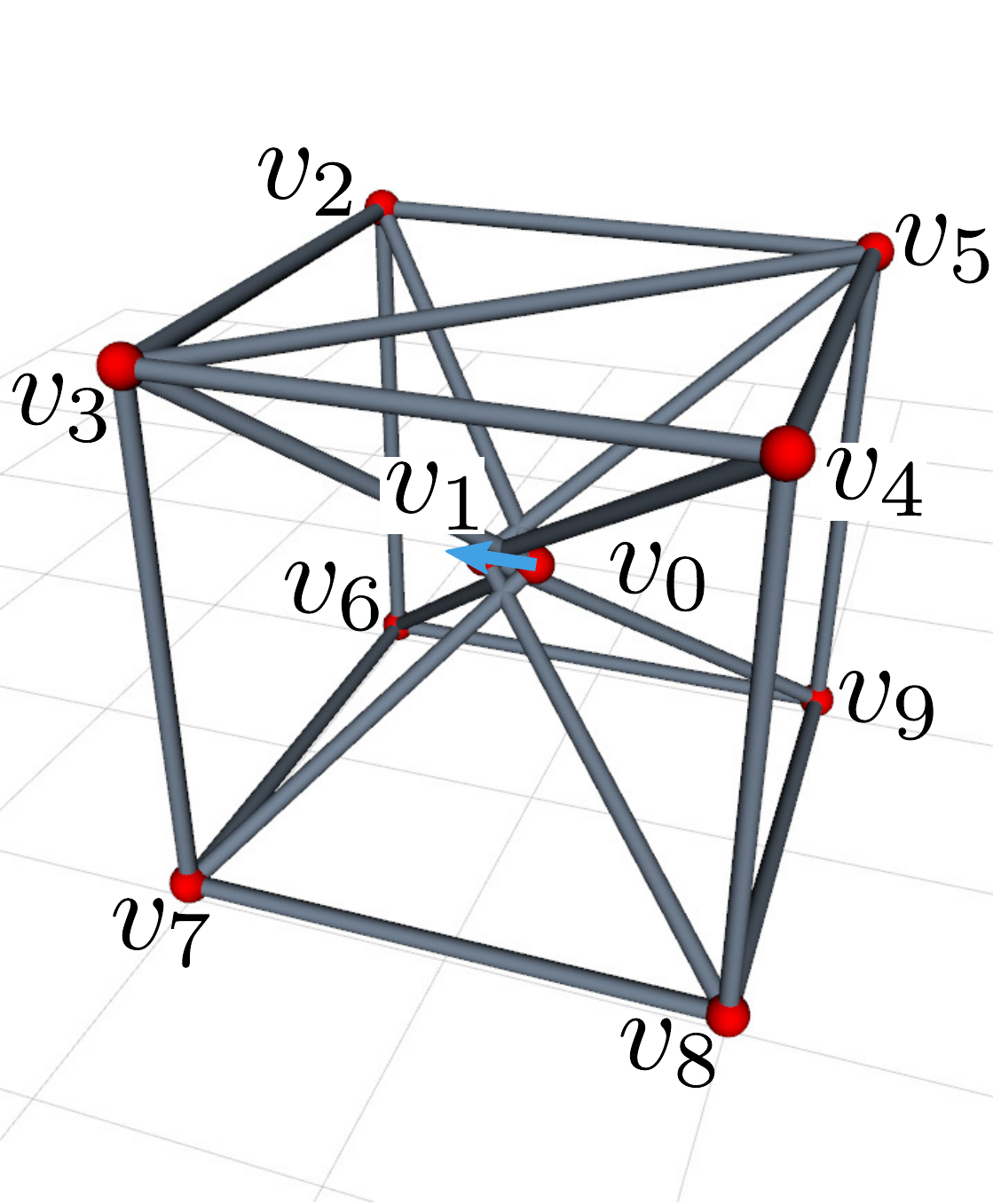}\label{fig:split-wrongmove}}
  \caption{Split node $v_0$ into $v_0$ and $v_1$: (a) a right way to
    move node $v_1$ away from node $v_0$; (b) a wrong way to move node
    $v_1$ away from node $v_0$.}
\end{figure}

After splitting a node, the members attached to this node are
separated into two groups that can be controlled
independently. However, we need to consider how to move the two newly
generate nodes away from each other. In
Fig.~\ref{fig:split-rightmove}, after splitting $v_0$ into $v_0$ and
$v_1$, $v_1$ is moved towards the right side of $v_0$ and this motion
can separate the members without collision. However, in
Fig.~\ref{fig:split-wrongmove}, for the same \texttt{Split} action,
$v_1$ is moved to the left side of $v_0$ which is not feasible because
members will collide.  When merging two nodes, we first need to move
them to some locations that are close to each other. If the new
locations are not selected correctly, it may be impossible to merge
them.

\subsection{Topology Reconfiguration Planning}
\label{sec:topology-planning}

Given a VTT $G = (V, E)$ and a motion task that is to move node $v$
from $q_i^v$ to $q_g^v$, if $q_i^v$ and $q_g^v$ belong to the same
enclosed subspace, then geometry reconfiguration planning is able to
handle this problem by either the approach
in~\cite{Liu-vtt-cspace-icra-2020} or the approach introduced in
Section~\ref{sec:path-planning}. Otherwise, topology reconfiguration
is needed. The node has to execute a sequence of \texttt{Split} and
\texttt{Merge} to avoid collision with other members.

There are multiple ways for a node $v$ to split into nodes $v^\prime$
and $v''$ as there are multiple ways to take the members into two
groups.  It is straightforward to compute all possible ways to split
$E^v$ into two groups in which both sets contain at least three edge
modules. Let $\mathcal{A}$ be the set of all possible ways to separate
$E^v$ into two groups. If $v$ is split into $v'$ and $v''$, then $E^v$
is separated into $E^{v'}$ and $E^{v''}$ accordingly. This split
process is denoted as $(E^{v'}, E^{v''})$. Not all possible ways in
$\mathcal{A}$ can be applied on node $v$. Given a valid VTT $G=(V,E)$
and a node $v$, a \texttt{Split} action can be encoded as a tuple
$(E^{v'}, E^{v''}, q_{\mathrm{s}}^{v'}, q_{\mathrm{s}}^{v''})$ where
$v'$ and $v''$ are the newly generated node after splitting, and
$q_{\mathrm{s}}^{v'}$ and $q_{\mathrm{s}}^{v''}$ are the locations of
these two new nodes. For simplicity, $q_{\mathrm{s}}^{v'} = q^v$ and
$v''$ is moved away from $v'$ Function~\ref{fun:split-valid} is used
to search $q_{\mathrm{s}}^{v''}$ and compute a valid \texttt{Split}
action.

\begin{function}[t!]
  \caption{ComputeSplitAction($G$, $q^v$, $a \in \mathcal{A}$)}\label{fun:split-valid}
  \KwData{$D$, $\Delta D$, $D_{\max}$}
  \SetKwFunction{VTTValid}{VTTValidation}
  \SetKwFunction{VTTCollision}{VTTCollisionCheck}
  Move node $v$ to $q^v$\;
  \If{\VTTValid{$G$} = \textsc{False}}{
    \KwRet \textsc{Null}\;
  }
  Split node $v$ into $v'$ and $v''$ according to $a\in \mathcal{A}$\;
  $q_{\mathrm{s}}^{v'}\leftarrow q^v$, $q_{\mathrm{s}}^{v''}\leftarrow q^v$\;
  Compute $d_v$\;
  \Repeat{$D > D_{\max}$}{
    $q_{\mathrm{s}}^{v''} = q^v + Dd_v$\;
    Move node $v''$ to $q^{v''}$\;
    \eIf{\VTTCollision($G$) = \textsc{True}}{
      $D\leftarrow (D+\Delta D)d_v$\;
    }{
      \eIf{\VTTValid{$G$} = \textsc{True}}{
         \KwRet $(E^{v'}, E^{v''}, q_{\mathrm{s}}^{v'}, q_{\mathrm{s}}^{v''})$\;
       }{
         \KwRet \textsc{Null}\;
      }
    }
  }
  \KwRet \textsc{Null}\;
\end{function}

In Function~\ref{fun:split-valid}, $G$ is the given VTT,
$a\in \mathcal{A}$ is one possible split, and $D$, $\Delta D$, and
$D_{\max}$ are three parameters. The function \texttt{VTTValidation}
checks the length constraints (Eq.~\eqref{eq:length-constraint}), the
angle constraints (Eq.~\eqref{eq:angle-constraint}), the stability
constraint (Sec.~\ref{sec:com}), and the manipulability constraint
(Eq.~\eqref{eq:manipulability-constraint}), and returns true if a
given VTT satisfies all these hardware constraints, otherwise returns
false. The function \texttt{VTTCollisionCheck} returns false if there
is no collision among all members, otherwise it returns true. The
basic idea is to move $v''$ gradually away from $v'$ along a unit
vector $d_v$ until a valid VTT is found. The direction of $d_v$ is
determined by $q_{\mathrm{s}}^{v'}$ (the current location of $v'$) and
$\mathcal{N}_G(v')$ (the neighbors of $v'$), and can be derived by
normalizing the following vector:
\begin{equation}
  \label{eq:split-dir-normalize}
  \sum_{\hat{v}\in\mathcal{N}_G(v')}\frac{q_{\mathrm{s}}^{v'}-q^{\hat{v}}}{\|q_{\mathrm{s}}^{v'}-q^{\hat{v}}\|}
\end{equation}
and the distance moved along vector $d_v$ can be tried iteratively.
The maximum distance between $v'$ and $v''$ for searching is
$D_{\max}$ that should be a small value. Within this small range,
self-collision is a more significant constraint, such as in
Fig.~\ref{fig:split-wrongmove}. Hence collision checking occurs first
then the other hardware constraints are checked. Conversely, if
merging $v'$ and $v''$ as $v$ at a location $q^v$, with this computed
\texttt{Split} action
$(E^{v'}, E^{v''}, q_{\mathrm{s}}^{v'}, q_{\mathrm{s}}^{v''})$, first
check if
$q_{\mathrm{s}}^{v'} = q^v\in
\mathcal{C}_{\mathrm{free}}^{v'}(q^{v'})$ and
$q_{\mathrm{s}}^{v''}\in \mathcal{C}_{\mathrm{free}}^{v''}(q^{v''})$,
and if true, move $v'$ to $q_{\mathrm{s}}^{v'}$ and move $v''$ to
$q_{\mathrm{s}}^{v''}$, and then merge them at $q^v$.

Given $q^v$, the current position of node $v$, and
$\mathcal{C}_{\mathrm{free}}^v =
\{{^t}\mathcal{C}_{\mathrm{free}}^v\vert t = 1,2,\cdots, T\}$ that
contains $T$ enclosed subspaces in the free space of node $v$, apply a
valid \texttt{Split} action on this node to separate $E^v$ into
$E^{v'}$ and $E^{v''}$ and generate two new nodes $v'$ and
$v''$. $q^{v'}$ and $q^{v''}$ are the locations of $v'$ and $v''$, and
$\mathcal{C}_{\mathrm{free}}^{v'}(q^{v'})$ and
$\mathcal{C}_{\mathrm{free}}^{v''}(q^{v''})$ can be computed
accordingly. Assuming there is a position
$q\in {^t}\mathcal{C}_{\mathrm{free}}^v$ where node $v$ can also be
split in the same way ($E^v$ can be separated into $E^{v'}$ and
$E^{v''}$ with \texttt{Split} action
$(E^{v'}, E^{v''}, q_{\mathrm{s}}^{v'}, q_{\mathrm{s}}^{v''})$), if
$q_{\mathrm{s}}^{v'}\in \mathcal{C}_{\mathrm{free}}^{v'}(q^{v'})$ and
$q_{\mathrm{s}}^{v''}\in \mathcal{C}_{\mathrm{free}}^{v''}(q^{v''})$,
namely $v'$ and $v''$ can navigate to $q_{\mathrm{s}}^{v'}$ and
$q_{\mathrm{s}}^{v''}$ respectively and merge at $q$, then this node
$v$ can navigate from $\mathcal{C}_{\mathrm{free}}^v(q^v)$ to
${^t}\mathcal{C}_{\mathrm{free}}^v$ by a pair of \texttt{Split} and
\texttt{Merge} actions, and these two enclosed subspaces can be
connected under this action pair. This is the transition model when
applying topology reconfiguration actions.

From Section~\ref{sec:topology-action}, it is shown that the possible
ways to split a node are highly dependent on the location of the node,
and the resulting enclosed subspaces of the newly generated nodes can
be different when splitting the node at different locations, so
multiple samples are needed for every enclosed subspace. When applying
the transition model described above, it is necessary to compute
$\mathcal{C}_{\mathrm{free}}^{v'}(q^{v'})$ and
$\mathcal{C}_{\mathrm{free}}^{v''}(q^{v''})$ many times (for every
sample and every valid split action) and can be time-consuming. An
alternative is to regard $v^\prime$ and $v''$ as a group and compute
their group free space
$\widehat{\mathcal{C}}_{\mathrm{free}}^{v'}(q^{v'})$ and
$\widehat{\mathcal{C}}_{\mathrm{free}}^{v''}(q^{v''})$ respectively
for every possible way to split $v$ in advance. For $v^\prime$, we can
apply Algorithm~\ref{alg:fssa} to search all enclosed subspaces
ignoring all edge modules in $E^{v''}$ that is
$\{{^t}\widehat{\mathcal{C}}_{\mathrm{free}}^{v'}\vert t = 1,2,\cdots,
T^{\prime}\}$. Similarly, we can also derive
$\{{^t}\widehat{\mathcal{C}}_{\mathrm{free}}^{v''}\vert t =
1,2,\cdots, T''\}$ for $v''$ ignoring all edge modules in
$E^{v'}$. These two sets are independent of the location of node $v$.
If
$q_{\mathrm{s}}^{v'}\in
\widehat{\mathcal{C}}_{\mathrm{free}}^{v'}(q^{v'})$ and
$q_{\mathrm{s}}^{v''}\in
\widehat{\mathcal{C}}_{\mathrm{free}}^{v''}(q^{v''})$, then these two
enclosed subspaces can be connected. This transition model based on
the group free space can be much more efficient but may cause failures
when applying the path planning for a group of nodes since it is less
strict than the previous one.

\begin{algorithm}[t!]
  \caption{Sample Generation}\label{alg:sample-generation}
  \SetKw{del}{del}
  \SetKwData{updateFlag}{UpdateFlag}
  \SetKwFunction{splitValid}{ComputeSplitAction}
  \SetKwFunction{sampleNum}{SampleNumber}
  \SetKwFunction{size}{SpaceSize}
  \SetKwFunction{randomPos}{RandomPosition}
  \SetKwFunction{validAction}{ValidSplitActions}
  \KwIn{${^t}\mathcal{C}_{\mathrm{free}}^v$, $G$, $\mathcal{A}$}
  \KwOut{$\mathcal{S}$, $\mathcal{A}^{\mathcal{S}}$}
  $\mathcal{S}\leftarrow \emptyset$\;
  Initialize an empty map $\mathcal{A}^{\mathcal{S}}$\;
  Initialize $d_{\min}$\;
  $N_{\max}\leftarrow\sampleNum{${^t}\mathcal{C}_{\mathrm{free}}^v$}$\;
  \For{$k=1$ \KwTo $K$}{
    $\mathcal{A}_{\mathrm{valid}}\leftarrow \emptyset$;\;
    $q_{\mathrm{rand}}\leftarrow\randomPos{${^t}\mathcal{C}_{\mathrm{free}}^v$}$\;
    \ForEach{$a\in \mathcal{A}$}{
      $\hat{a}=\splitValid{$G$, $q$, $a$}$\;
      \If{$\hat{a}$ is not \textsc{Null}}{
         $\mathcal{A}_{\mathrm{valid}}=\mathcal{A}_{\mathrm{valid}}\cup\{\hat{a}\}$\;
      }
    }
    $\mathcal{S}_{\mathrm{close}}\leftarrow\left\{q\vert q\in \mathcal{S} \wedge \|q - q_{\mathrm{rand}}\|\le
      d_{\min}\right\}$\;
    \uIf{$\mathcal{S}_{\mathrm{close}}=\emptyset\wedge \vert\mathcal{S}\vert<N_{\max}$}{
      $\mathcal{S}\leftarrow \mathcal{S}\cup \{q_{\mathrm{rand}}\}$\;
      $\mathcal{A}^{\mathcal{S}}\left[q\right] = \mathcal{A}_{\mathrm{valid}}$\;
    }
    \ElseIf{$\mathcal{S}_{\mathrm{close}}\neq\emptyset\wedge
      \vert\mathcal{S}\vert<N_{\max}$}{
      \eIf{$\vert\mathcal{A}_{\mathrm{valid}}\vert = \vert\mathcal{A}\vert$}{
        \ForEach{$q\in \mathcal{S}_{\mathrm{close}}$}{
          \del $\mathcal{A}^{\mathcal{S}}\left[q\right]$\;
        }
        $\mathcal{S} \leftarrow
        \mathcal{S}\setminus\mathcal{S}_{\mathrm{close}}+\{q_{\mathrm{rand}}\}$\;
        $\mathcal{A}^{\mathcal{S}}\left[q_{\mathrm{rand}}\right] = \mathcal{A}_{\mathrm{valid}}$\;
      }{
        $\updateFlag\leftarrow \textsc{False}$\;
        \ForEach{$q\in \mathcal{S}_{\mathrm{close}}$}{
          \uIf{$\mathcal{A}^{\mathcal{S}}\left[q\right]\subset
            \mathcal{A}_{\mathrm{valid}}$}{
            \del $\mathcal{A}^{\mathcal{S}}\left[q\right]$\;
            $\mathcal{S}\leftarrow \mathcal{S}\setminus \{q\}$\;
            $N_{\max}\leftarrow N_{\max} - 1$\;
            $\updateFlag \leftarrow \textsc{True}$\;
          }
          \ElseIf{$\mathcal{A}_{\mathrm{valid}}\setminus\mathcal{A}^{\mathcal{S}}\left[q\right]\neq\emptyset$}{
            $\updateFlag \leftarrow \textsc{True}$\;
          }
        }
        \If{$\updateFlag = \textsc{True}$}{
          $\mathcal{S} \leftarrow\mathcal{S}+\{q\}$\;
          $\mathcal{A}^{\mathcal{S}}\left[q_{\mathrm{rand}}\right] =
          \mathcal{A}_{\mathrm{valid}}$\;
          $N_{\max}\leftarrow N_{\max}+1$\;
        }
      }
    }
  }
\end{algorithm}

There are two phases in the topology reconfiguration planning:
\textit{sample generation} and \textit{graph search}. In
\textit{sample generation}, samples are generated providing as many
valid \texttt{Split} actions as possible spanning the enclosed
subspaces. We introduce Algorithm~\ref{alg:sample-generation} to
sample a given enclosed subspace. $\mathcal{S}$ is the set containing
all generated samples and $\mathcal{A}^{\mathcal{S}}$ stores the
\texttt{Split} actions for every sample. $N_{\max}$ is the maximum
number of samples for a given enclosed subspace
${^t}\mathcal{C}^v_{\mathrm{free}}$ determined by the size of the
space. In our setup, we find the range of a given space along
$x$-axis, $y$-axis, and $z$-axis denoted as $x_{\mathrm{range}}$,
$y_{\mathrm{range}}$, and $z_{\mathrm{range}}$ respectively, then
$N_{\max}$ is
$\left\lceil\sqrt{x_{\mathrm{range}}^2+y_{\mathrm{range}}^2+z_{\mathrm{range}}^2}\Big/d_{\min}\right\rceil$
in which $d_{\min}$ is the minimum distance between every pair of
samples. $K$ is the maximum number of iterations set by users and can
be related to $N_{\max}$. For each iteration, we first randomly
generate a sample $q_{\mathrm{rand}}$ that is inside the given
enclosed subspace, then find all valid \texttt{Split} actions stored
in $\mathcal{A}_{\mathrm{valid}}$ (\textit{Line 6 --- 11}). Then find
all previous valid samples that are close to $q_{\mathrm{rand}}$
(\textit{Line 12}). If $q_{\mathrm{rand}}$ is far from all previous
valid samples and the size of $\mathcal{S}$ is less than $N_{\max}$,
add this new sample and store its valid \texttt{Split} actions
(\textit{Line 13---15}). If $q_{\mathrm{rand}}$ is close to some
previous valid samples, there are two cases for consideration. If this
new sample provides all possible \texttt{Split} actions, then remove
all valid nearby samples and only keep this new sample for this area
since this is the best option (\textit{Line 17---21}). Otherwise, we
check two conditions: 1. whether there exists any valid nearby sample
that has fewer valid \texttt{Split} actions than $q_{\mathrm{rand}}$;
2. whether the new sample $q_{\mathrm{rand}}$ introduces any new
\texttt{Split} actions compared with all valid nearby samples. If the
first condition is true, remove the old sample and add the new sample
to $\mathcal{S}$. If the second condition is true, add the new sample
to $\mathcal{S}$ (\textit{Line: 23---35}). We run
Algorithm~\ref{alg:sample-generation} for every enclosed subspace to
generate enough samples, and then add $q_i^v$ (the initial location of
node $v$) to the sample set of $\mathcal{C}^v_{\mathrm{free}}(q_i^v)$
and add $q_g^v$ (the goal location of node $v$) to the sample set of
$\mathcal{C}^v_{\mathrm{free}}(q_g^v)$. After the \textit{sample
  generation} phase, $^{\mathcal{C}}\mathcal{S}$ and the corresponding
\texttt{Split} actions for all samples
$^{\mathcal{C}}\mathcal{A}^{\mathcal{S}}$ are obtained
$\forall \mathcal{C}\in \mathcal{C}_{\mathrm{free}}^v$.

\begin{figure}[t]
  \centering
  \includegraphics[width=0.25\textwidth]{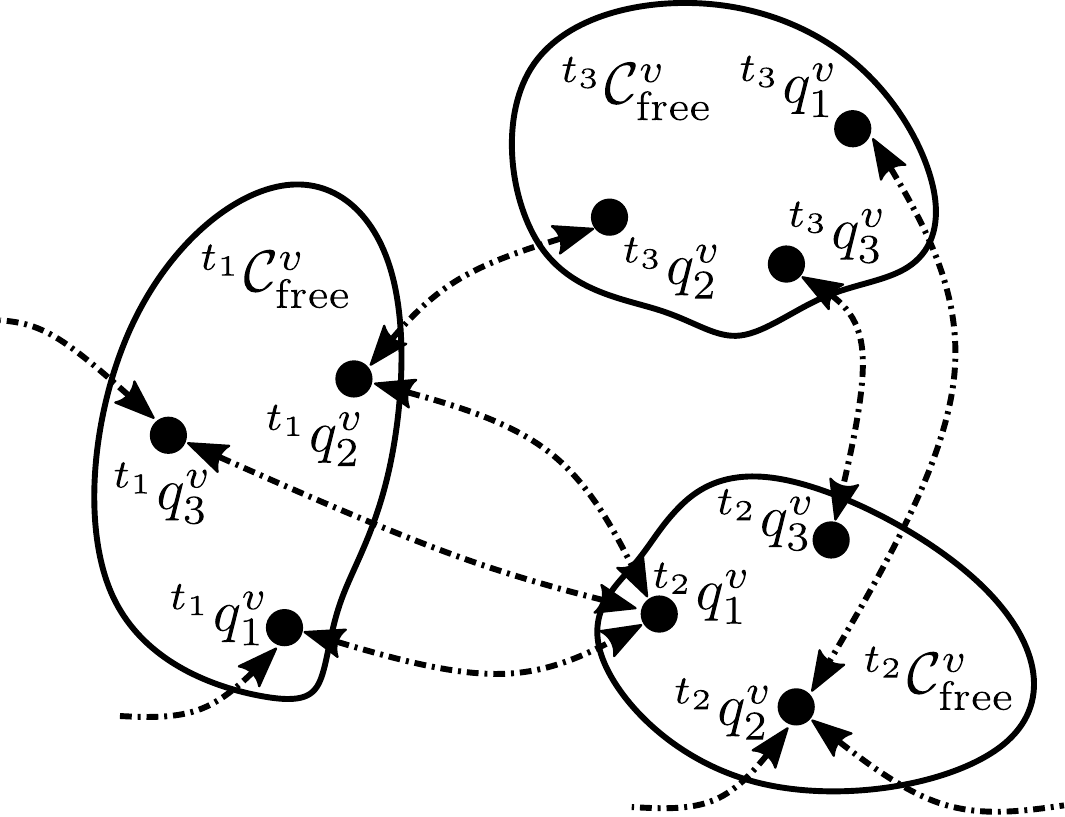}
  \caption{Topology connections among three enclosed subspaces of node
  $v$.}
  \label{fig:topology-graph}
\end{figure}

The \textit{graph search} phase follows. It generates a sequence of
topology reconfiguration actions using the transition model discussed
earlier. A graph search algorithm can be applied to compute a sequence
of enclosed subspaces starting from
$\mathcal{C}_{\mathrm{free}}^v(q_i^v)$ to
$\mathcal{C}_{\mathrm{free}}^v(q_g^v)$ while exploring the topology
connections among these enclosed subspaces.
Fig.~\ref{fig:topology-graph} is an example where the graph has
enclosed subspaces as vertices. An edge in this graph connecting two
enclosed subspaces denotes that node $v$ can move from a sample in one
enclosed subspace to a sample in the other. The graph is built from
$\mathcal{C}_{\mathrm{free}}^v(q^v)$, grows as valid transitions among
enclosed subspaces are found, and stops when the enclosed subspace
containing $q_g^v$ is visited. A graph search algorithm based on
Dijkstra's framework is shown in Algorithm~\ref{alg:topology}.

\textit{Line 1 --- 4}: If $q_i^v$ and $q_g^v$ are in the same enclosed
subspace, then no topology reconfiguration is needed. Otherwise, we
make two sets $\mathcal{Q}$ and $\overline{\mathcal{Q}}$ where
$\mathcal{Q}$ contains all newly checked or non-visited enclosed
subspaces and $\overline{\mathcal{Q}}$ contains all visited enclosed
subspaces. The sizes of these two sets will change as the algorithm
explores $\mathcal{C}_{\mathrm{free}}^v$. Initially, only the enclosed
subspace containing $q_i^v$ that is
$\mathcal{C}_{\mathrm{free}}^v(q_i^v)$ and the enclosed subspace
containing $q_g^v$ that is $\mathcal{C}_{\mathrm{free}}^v(q_g^v)$ are
in $\mathcal{Q}$, and the algorithm starts with
$\mathcal{C}_{\mathrm{free}}^v(q_i^v)$. The value $g(\mathcal{C})$ is
the cost of the path from $q_i^v$ to the enclosed subspace
$\mathcal{C}$, so $g(\mathcal{C}_{\mathrm{free}}^v(q_i^v)) = 0$ and
$g(\mathcal{C}_{\mathrm{free}}^v(q_g^v)) = \infty$ at the beginning.

\begin{algorithm}[t]
  \caption{Topology Reconfiguration Planning}\label{alg:topology}
  \SetKwFunction{motionValid}{ValidMotion}
  \SetKwData{validFlag}{ValidFlag}
  \SetKw{Break}{break}
  \KwIn{$G$, $q_i^v$, $q_g^v$, $\mathcal{C}_{\mathrm{free}}^v$,
    $\{({^{\mathcal{C}}}\mathcal{S},
    {^{\mathcal{C}}}\mathcal{A}^{\mathcal{S}})\vert \mathcal{C}\in \mathcal{C}_{\mathrm{free}}^v\}$}
  \KwOut{Tree of enclosed subspaces $p$}
  \If{$\mathcal{C}_{\mathrm{free}}^v(q_i^v) = \mathcal{C}_{\mathrm{free}}^v(q_g^v)$}{
    \Return \textsc{Null}\;
  }
  $\mathcal{Q} \leftarrow \left\{\mathcal{C}_{\mathrm{free}}^v(q_i^v),
  \mathcal{C}_{\mathrm{free}}^v(q_g^v)\right\}$, $\overline{\mathcal{Q}}\leftarrow \emptyset$\;
  $g(\mathcal{C}_{\mathrm{free}}^v(q_i^v))\leftarrow 0$,
  $g(\mathcal{C}_{\mathrm{free}}^v(q_g^v))\leftarrow \infty$\;
  \While{$\mathcal{C}_{\mathrm{free}}^v(q_g^v)\in \mathcal{Q}$}{
    $\overline{\mathcal{C}}\leftarrow \underset{\mathcal{C}\in
      \mathcal{Q}}{\arg\min}\ g(\mathcal{C})$\;
    $\mathcal{Q}\leftarrow \mathcal{Q}\setminus \{\overline{\mathcal{C}}\}$\;
    $\overline{\mathcal{Q}}\leftarrow \overline{\mathcal{Q}} \cup
    \{\overline{\mathcal{C}}\}$\;
    \ForEach{$\mathcal{C}\in
      \mathcal{C}_{\mathrm{free}}^v\setminus\{\overline{\mathcal{C}}\} \wedge \mathcal{C}\notin
      \overline{\mathcal{Q}}$}{
      \ForEach{$({^{\overline{\mathcal{C}}}}q, {^{\mathcal{C}}}q)\in
        {^{\overline{\mathcal{C}}}}\mathcal{S}\times {^{\mathcal{C}}}\mathcal{S}$}{
        \If{\motionValid{$^{\overline{\mathcal{C}}}q$, $^{\mathcal{C}}q$} =
          \textsc{True}}{
          \eIf{$\mathcal{C}\in \mathcal{Q}$}{
            \If{$g(\overline{\mathcal{C}}) +
              c({^{\overline{\mathcal{C}}}}q, {^{\mathcal{C}}}q) < g(\mathcal{C})$}{
              $g(\mathcal{C})\leftarrow g(\overline{\mathcal{C}})
              + c({^{\overline{\mathcal{C}}}}q, {^{\mathcal{C}}}q)$\;
              $p(\mathcal{C})\leftarrow \overline{\mathcal{C}}$\;
            }
          }{
            $\mathcal{Q}\leftarrow \mathcal{Q} + \{\mathcal{C}\}$\;
            $g(\mathcal{C})\leftarrow g(\overline{\mathcal{C}}) +
            c({^{\overline{\mathcal{C}}}}q, {^{\mathcal{C}}}q)$\;
            $p(\mathcal{C})\leftarrow \overline{\mathcal{C}}$\;
          }
          \Break\;
        }
      }
    }
  }
\end{algorithm}

\textit{Line 6 --- 8}: Every iteration starts with the enclosed
subspace that has the lowest cost $g(\mathcal{C})$ in
$\mathcal{Q}$. At the beginning,
$\mathcal{C}_{\mathrm{free}}^v(q_i^v)$ has the lowest cost. After
selecting an enclosed subspace, update $\mathcal{Q}$ and
$\overline{\mathcal{Q}}$.

\textit{Line 9 --- 20}: Iterate every enclosed subspace $\mathcal{C}$
except $\overline{\mathcal{C}}$ in $\mathcal{C}_{\mathrm{free}}^v$ and
check if it is already visited. If so, then this potential transition
is not a new transition. Otherwise, check if there exists a valid
motion to move the node from any sample in $\overline{\mathcal{C}}$
that is the enclosed subspace with the lowest cost to any sample in
$\mathcal{C}$ handled by Function \texttt{ValidMotion}. Both
transition models are implemented and compared in the test
scenarios. If this is true, then there are two cases: this subspace is
not checked for the first time namely that there is already a
connection between this enclosed subspace and another enclosed
subspace, or this subspace has never been checked which means it has
no connection before. For the first case, we need to check whether its
cost needs to be updated.
$c({^{\overline{\mathcal{C}}}}q, {^{\mathcal{C}}}q)$ is the cost of
the motion from ${^{\overline{\mathcal{C}}}}q$ to
${^{\mathcal{C}}}q$. This cost can be related to the estimated
distance or other factors. In our setup, all valid motions from one
enclosed subspace to another one have the same cost. If its cost is
updated, then its parent $p(\mathcal{C})$ should also be updated
accordingly. For the second case, initialize the cost and the parent
of this newly checked enclosed subspace, and update set
$\mathcal{Q}$. There can be multiple ways to transit between two
enclosed subspaces. For example, there are two edges between
$^{t_2}\mathcal{C}_{\mathrm{free}}^v$ and
$^{t_3}\mathcal{C}_{\mathrm{free}}^v$ shown in
Fig.~\ref{fig:topology-graph}. In our setup, we just keep the first
one being found. Since ${^{\overline{{\mathcal{C}}}}}\mathcal{S}$ and
$^{\mathcal{C}}\mathcal{S}$ are randomly generated, it is possible
that the condition in \textit{Line 11} is failed although there does
exist $^{\overline{\mathcal{C}}}q$ and $^{\mathcal{C}}q$ that can pass
this condition.

Once $\mathcal{C}_{\mathrm{free}}^v(q_g^v)$ is visited, the algorithm
ends. With $p$, a tree with visited enclosed subspaces as vertices, it
is straightforward to find the optimal path connecting
$\mathcal{C}_{\mathrm{free}}^v(q_i^v)$ and
$\mathcal{C}_{\mathrm{free}}^v(q_g^v)$ as well as all the samples that
the node needs to traverse. For example, in
Fig.~\ref{fig:topology-graph}, when node $v$ traverses from
$^{t_1}q_1^v$ to $^{t_3}q_1^v$, it first moves to $^{t_1}q_2^v$, then
\texttt{Split} and \texttt{Merge} at $^{t_3}q_2^v$, and finally move
to $^{t_3}q_1^v$. Moving a node inside one of its enclosed subspaces
can be solved easily by geometry reconfiguration planning. After
splitting the node, we can apply geometry reconfiguration planning to
move them to the computed positions in the next enclosed subspace for
merging.

\section{Locomotion}
\label{sec:locomotion}

\begin{figure}[t]
  \centering
  \subfloat[]{\includegraphics[width=0.16\textwidth]{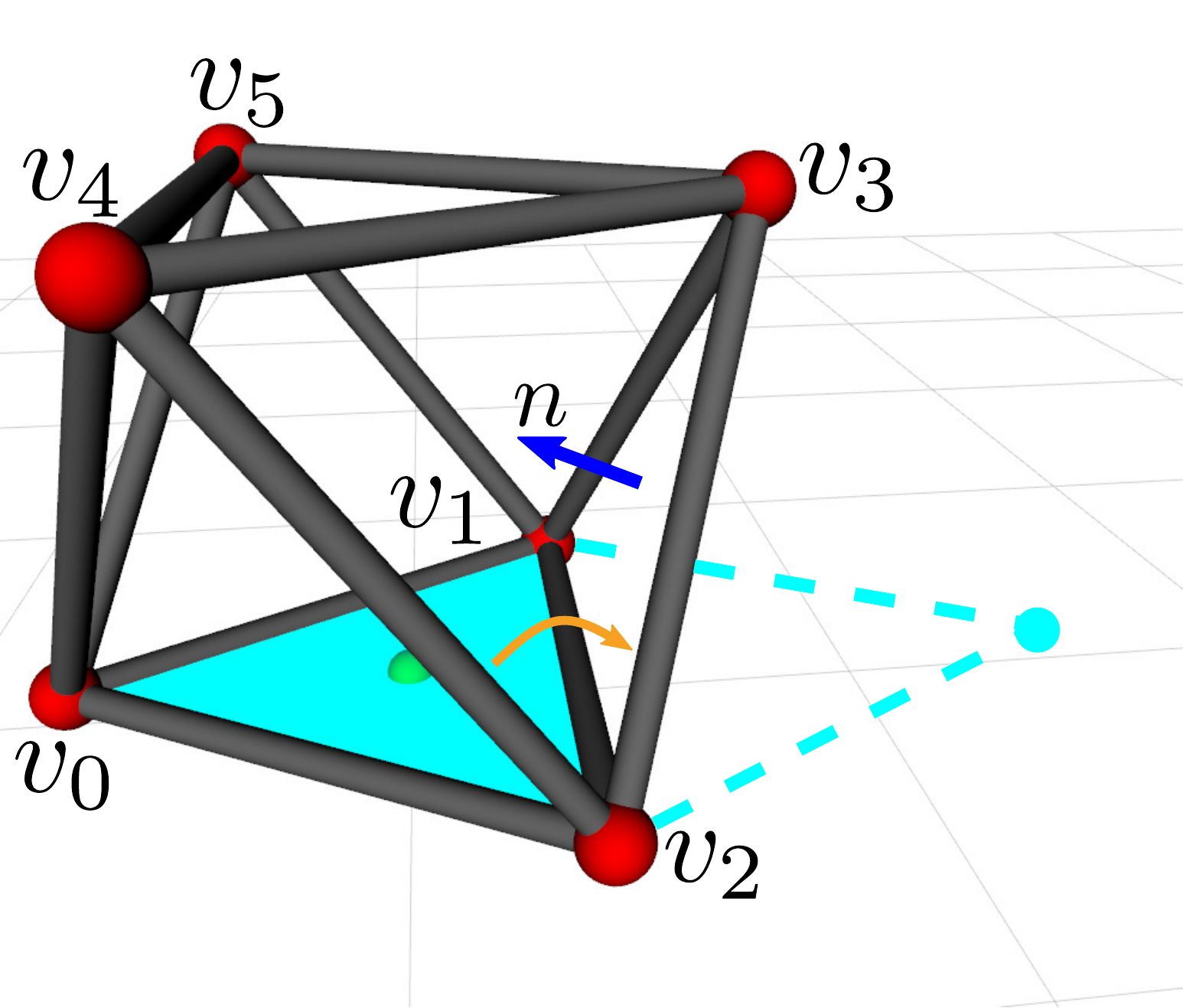}\label{fig:locomotion-step-1}}
  \hfil
  \subfloat[]{\includegraphics[width=0.16\textwidth]{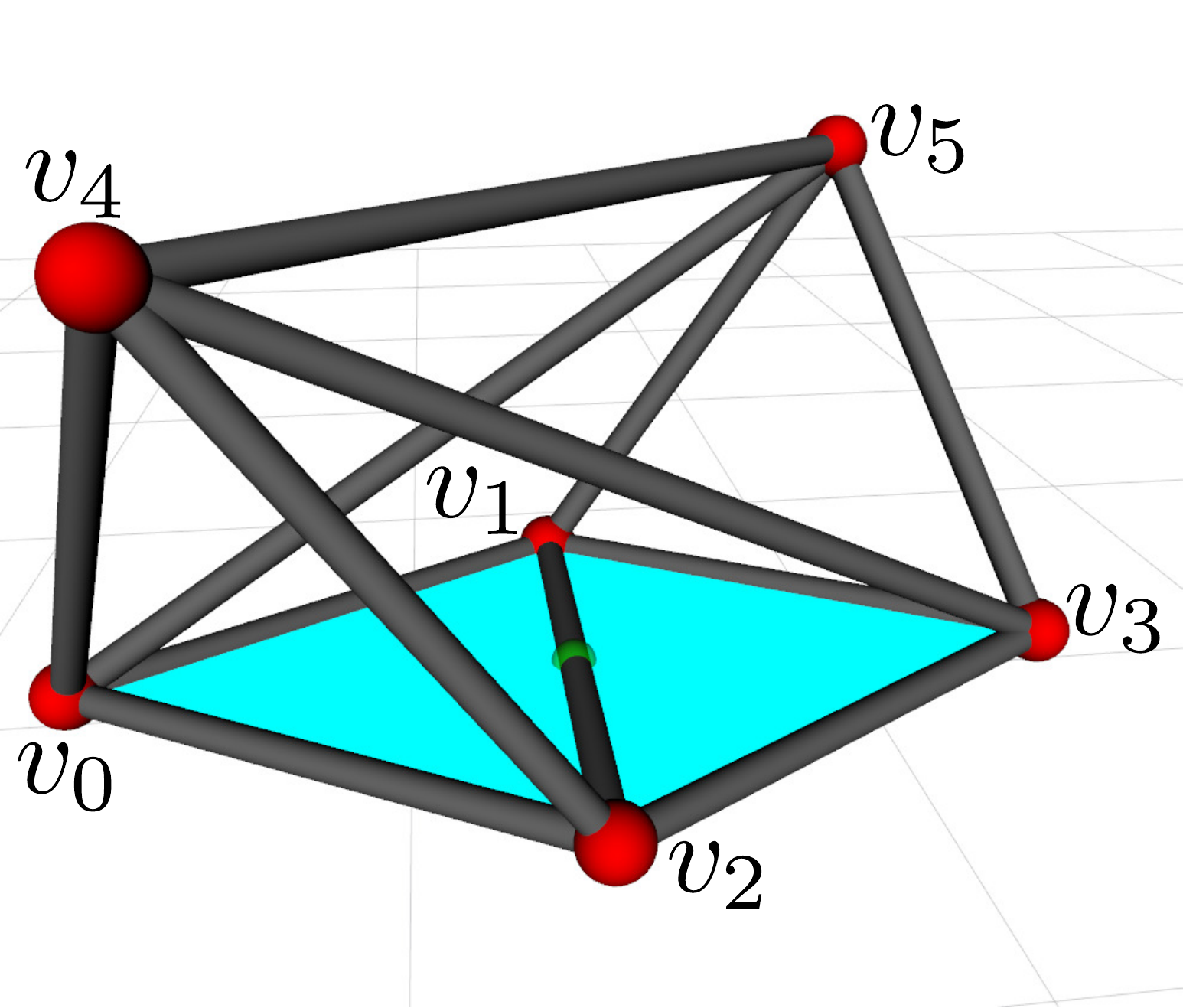}\label{fig:locomotion-step-2}}
  \hfil
  \subfloat[]{\includegraphics[width=0.16\textwidth]{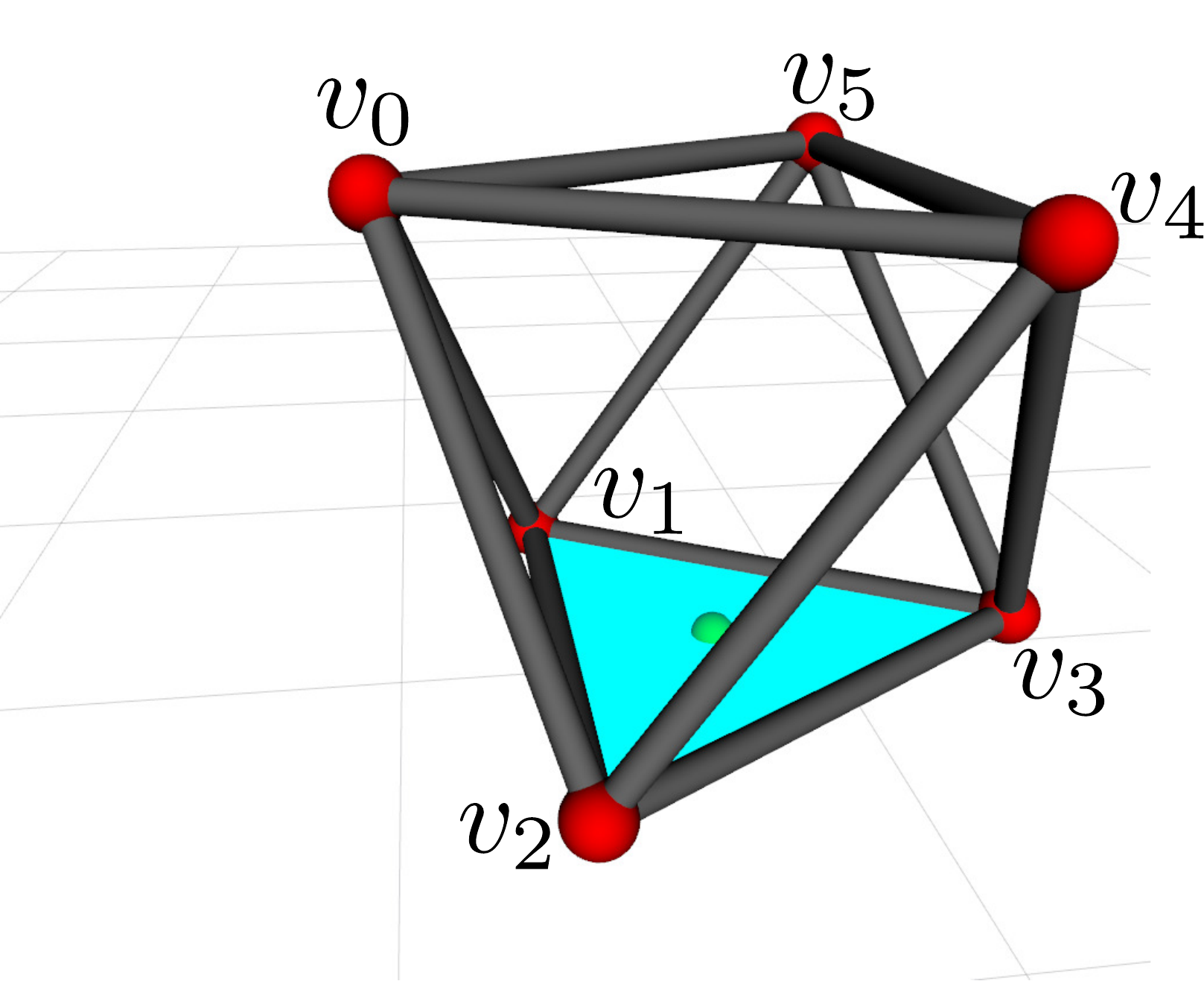}\label{fig:locomotion-step-3}}
  \caption{A VTT in octahedron configuration executes a single rolling
    locomotion step. (a) Initially node $v_0$, $v_1$, and $v_2$ forms
    the support polygon shown as the aqua region
    ({\color[rgb]{0,1,1}$\blacktriangle$}), and the center of mass
    projected onto the ground ({\color{green}{$\bullet$}}) is within
    this support polygon. The truss wants to roll from its current
    support polygon to an adjacent support polygon formed by node
    $v_1$, $v_2$, and the new tipping location. (b) $v_3$ and $v_5$
    are moved so that the support polygon is expanded and the center
    of mass projected onto the ground is on member $(v_1, v_2)$. (c)
    $v_0$ and $v_4$ are moved to their destinations to finish this
    locomotion step, and the center of mass projected onto the ground
    is within the new support polygon formed by node $v_1$, $v_2$, and
    $v_3$.}
  \label{fig:single-step-locomotion}
\end{figure}

Unlike the previous reconfiguration planning, the center of mass
during locomotion moves over a large range. In a rolling locomotion
step, a VTT rolls from one support polygon to an adjacent support
polygon (Fig.~\ref{fig:locomotion-step-1}). In this process, it is
useful to maintain statically stable locomotion to prevent the system
from receiving impacts from the ground as repeated impacts may damage
the reconfiguration nodes. Due to this requirement, previous
implementations of a rolling locomotion step has manually divided into
several
phases~\cite{Park-vtt-locomotion-ral-2020,Usevitch-lar-locomotion-tro-2020}. Our
approach can deal with this issue automatically. A high-level path
planner generates a sequence of support polygons for a given
environment and the locomotion planner ensures that the truss can
follow this support polygon trajectory without violating constraints
and receiving external impacts.

\subsection{Truss Polyhedron}
\label{sec:polyhedron}

The boundary representation of a VTT is modeled as a convex polyhedron
which can be either pre-defined or computed from the set of node
positions by the convex hull~\cite{Barber-quickhull-1996}. Given a VTT
$G=(V,E)$, its polyhedron representation can be fully defined as
$P^G=(E^G, F^G)$ in which $F^G$ is the set of facets forming the
boundary of this VTT and $E^G$ is the set of edges forming all
facets. For example, the boundary representation of the VTT in
Fig.~\ref{fig:locomotion-step-1} is an octahedron in which edge
$(v_1, v_2)$ is incident to facet $(v_1, v_2, v_3)$ and facet
$(v_0, v_1, v_2)$.

Suppose the set of edges forming a facet $f\in F^G$ is
$E_f^G\subset E^G$, and $\forall e\in E_f^G$, the other incident facet
of edge $e$ is $f^e\in F^G$. Assume $f\in F^G$ is the current support
polygon and a single rolling locomotion step is equivalent to rotating
the truss with respect to an edge of this facet $e\in E_f^G$ until the
other incident facet of this edge $f^e$ becomes the new support
polygon. An example is shown in
Fig.~\ref{fig:single-step-locomotion}. The motion of the robot from
Fig.~\ref{fig:locomotion-step-1} to Fig.~\ref{fig:locomotion-step-3}
is equivalent to rotating the truss with respect to the fixed edge
$e=(v_1,v_2)$ until the support polygon becomes
$f^e=(v_1,v_2,v_3)$. This means the input command for each locomotion
step can be simply an edge of the current support polygon, e.g.,
$(v_1,v_2)$ for the scenario in Fig.~\ref{fig:single-step-locomotion}.

\subsection{Locomotion Planning}
\label{sec:locomotion-step}

Given a VTT $G=(V,E)$, its truss polyhedron model $P^G=(E^G, F^G)$ and
its current support polygon $f\in F^G$ can be derived. Given a
locomotion command $e\in f$, we can find $f^e$ and compute its normal
vector $n_{f^e}$ pointing inward $P^G$ (e.g., vector $n$ for facet
$(v_1,v_2,v_3)$ in Fig.~\ref{fig:locomotion-step-1}), then the
rotation angle $\alpha$ between $n_{f^e}$ and the normal vector of the
ground $\left[0,0,1\right]^{\intercal}$ is simply
\begin{equation}
  \label{eq:rotating-angle}
  \alpha=\arccos\left(\frac{n_{f^e}\bullet\left[0,0,1\right]^{\intercal}}{\|n_{f^e}\|}\right)
\end{equation}
The rotation axis is along edge $e=(v_1^e, v_2^e)$ and its direction
can be determined by the right-hand rule. For the example shown in
Fig.~\ref{fig:single-step-locomotion}, the rotation axis is simply
$\frac{q^{v_1}-q^{v_2}}{\|q^{v_1}-q^{v_2}\|}$. Assume the rotation
axis is pointing from $v_1^e$ to $v_2^e$, then the rotation angle and
the rotation axis determine a rotation matrix $R^e$, and for every
node $v\in V\setminus \{v^e_1, v^e_2\}$ during this locomotion
process, its initial position $q_i^v$ is known as $q^v$ and its goal
position $q_g^v$ can be derived as
\begin{equation}
  \label{eq:des-pos-locomotion}
  q_g^v = R^e(q^v-q^{v_1^e}) + q^{v_1^e}
\end{equation}

Then we can make use of the approach presented in
Section~\ref{sec:shape} to plan the motion for all involved nodes. The
workspace can either be an known space or be set such that it can
include the initial VTT and the goal VTT. Due to the stability
constraint where the projected center of mass has to be within its
support polygon, the planner will expand the support polygon by
placing an additional node on the ground, so the center of mass can be
moved in a larger range. An example of this non-impact rolling
locomotion planning result is shown in
Fig.~\ref{fig:single-step-locomotion}. In this task, the initial
support polygon is facet $(v_0, v_1, v_2)$ and we want to roll the
truss so that facet $(v_1, v_2, v_3)$ becomes the new support
polygon. The locomotion command is simply $(v_1, v_2)$, the rotation
axis and the rotation angle can be computed as mentioned, and four
nodes ($v_0$, $v_3$, $v_4$, $v_5$) have to move to new locations which
can be derived from Eq.~\eqref{eq:des-pos-locomotion}. Because of the
stability constraint, node $v_0$ cannot be moved at the beginning, or
the robot won't be stable since there will be no support polygon. The
planner chooses to move $v_3$ and $v_5$ first to expand the support
polygon which is formed by node $v_0$, $v_1$, $v_2$, and $v_3$, and
the center of mass is moved toward the target support polygon
(Fig.~\ref{fig:locomotion-step-2}). The center of mass projected onto
the ground is always within the initial support polygon. After the
support polygon is expanded, $v_0$ and $v_4$ are moved, and, in the
meantime, the projected center of mass enters the target support
polygon (Fig.~\ref{fig:locomotion-step-3}).

\begin{figure}[b]
  \centering
  \begin{subfloat}[]{\includegraphics[width=0.13\textwidth]{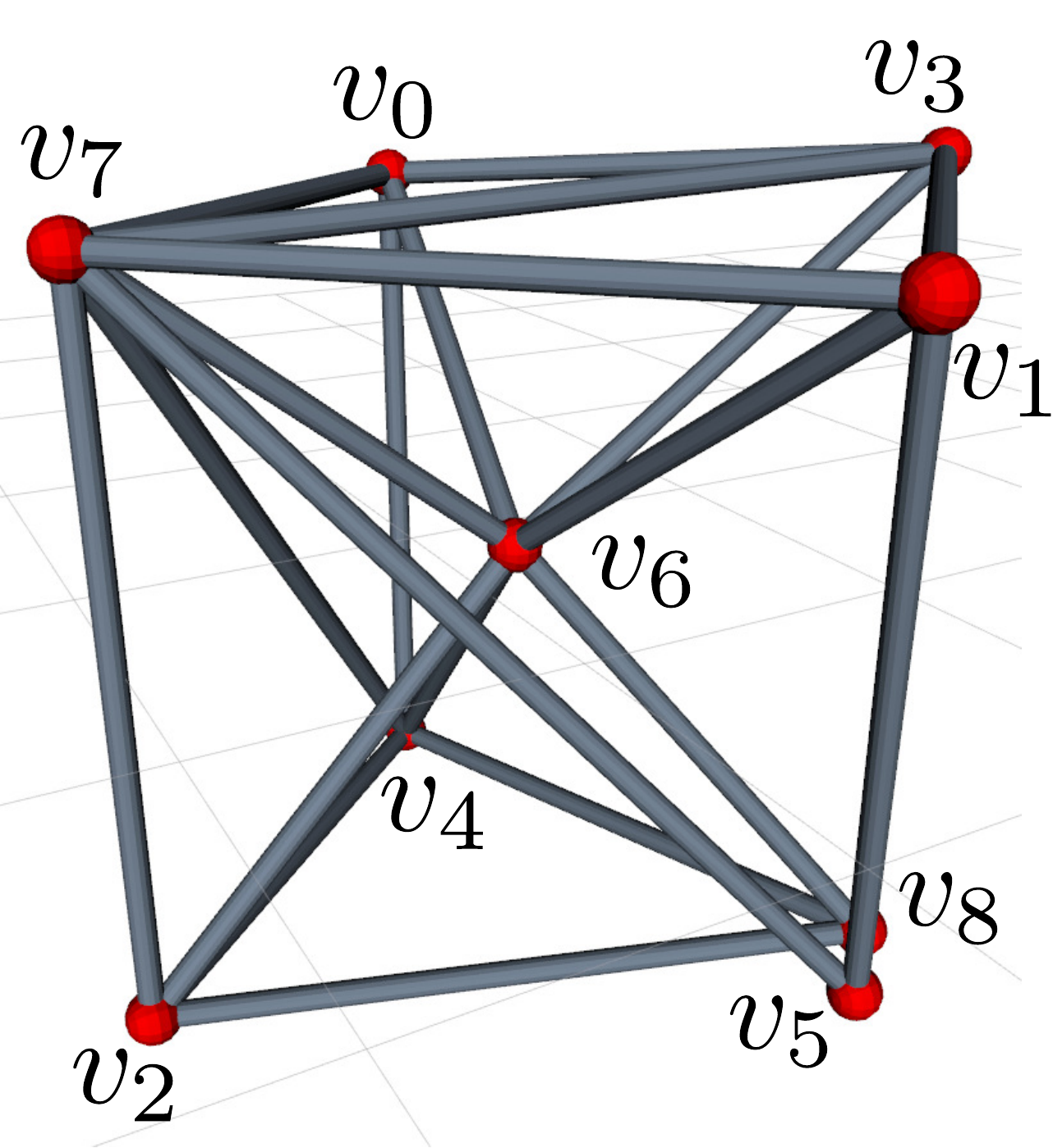}\label{fig:multi-node-init-truss}}
  \end{subfloat}
  \hfil
  \begin{subfloat}[]{\includegraphics[width=0.09\textwidth]{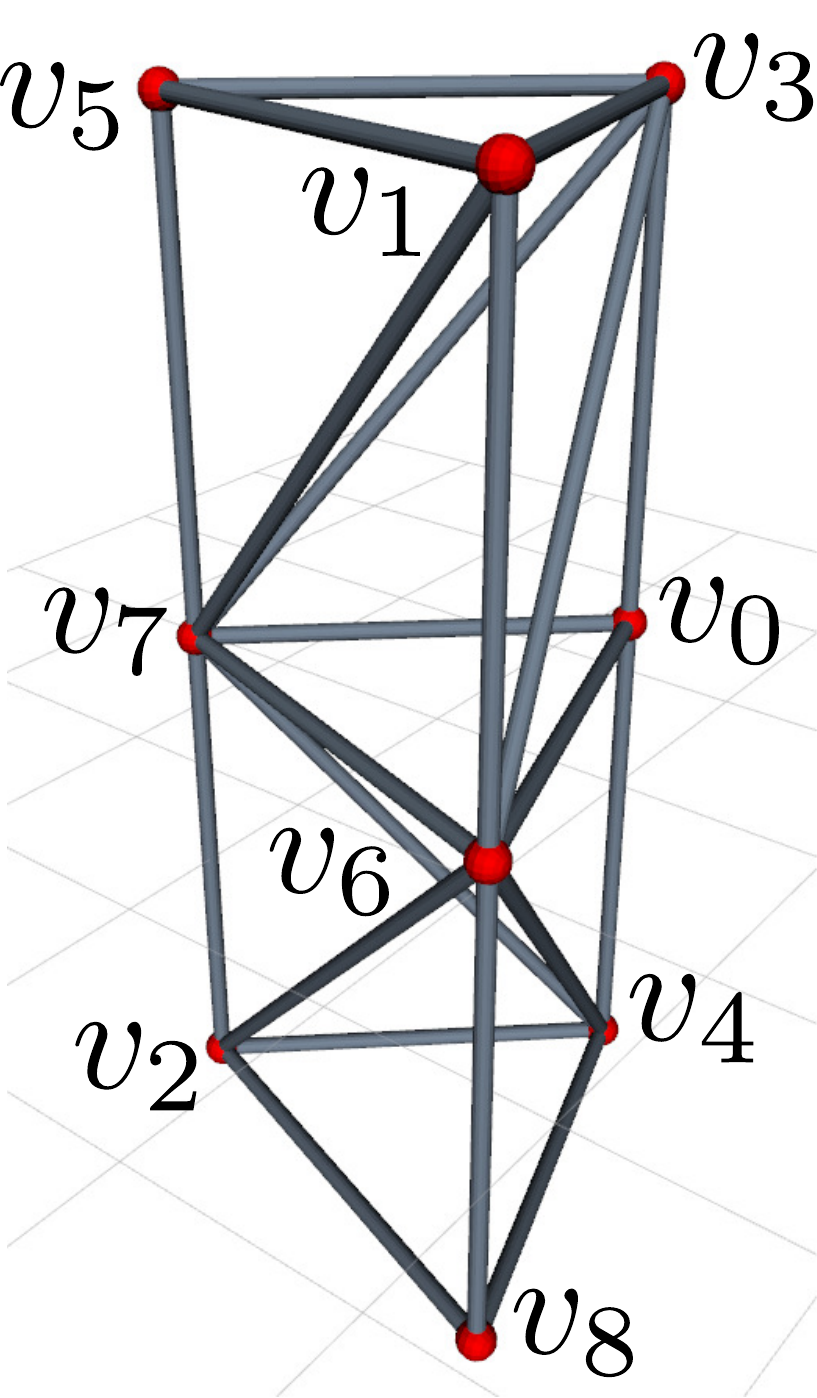}\label{fig:multi-node-goal-truss}}
  \end{subfloat}
  \caption{The motion task is to change the shape of a VTT from (a) a
    cubic truss for rolling locomotion to (b) a tower truss for
    shoring.}
  \label{fig:geometry-reconfig-test}
\end{figure}

\begin{figure*}[t]
  \centering
  \subfloat[]{\includegraphics[height=0.18\textwidth]{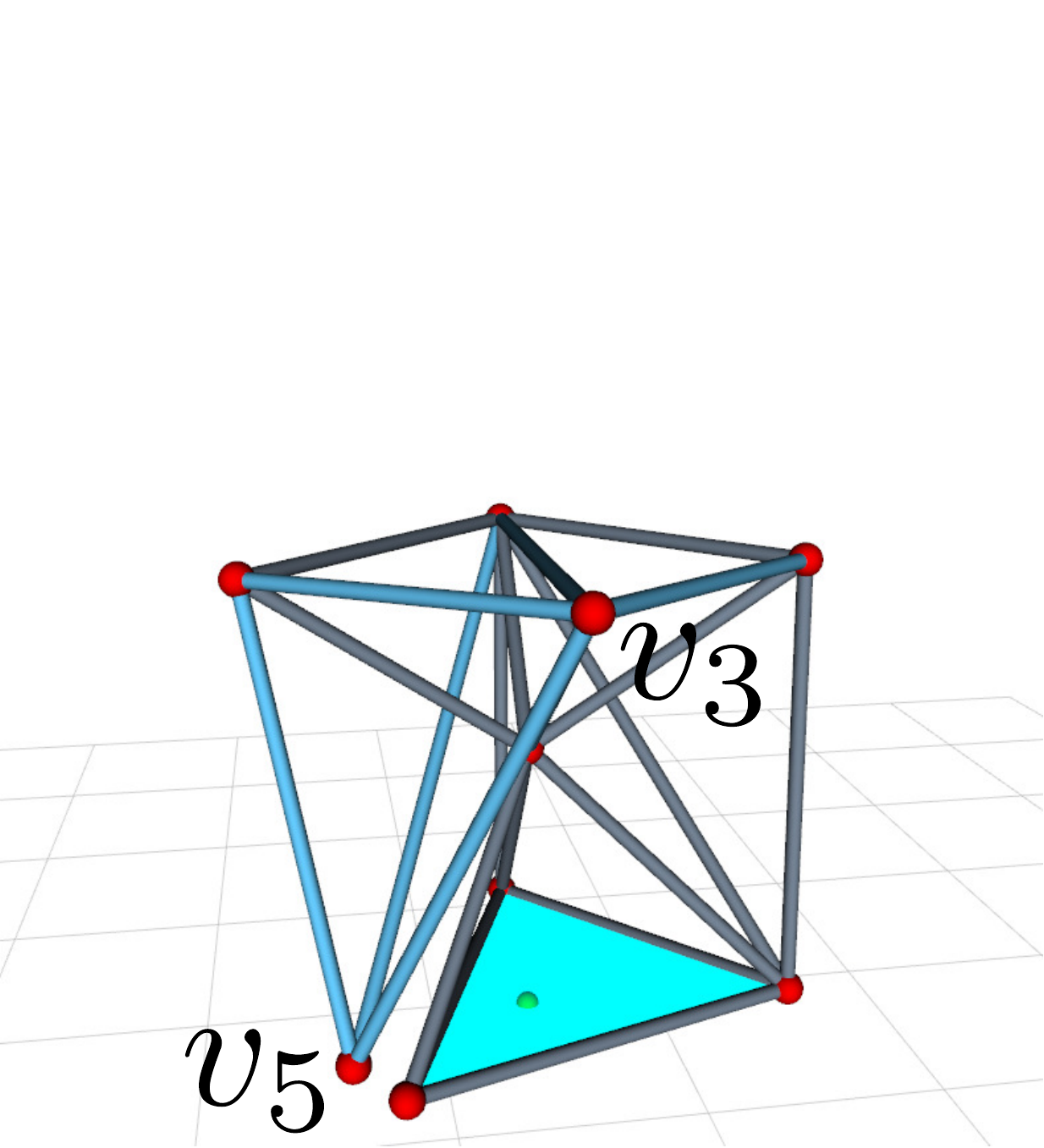}}
  \hfil
  \subfloat[]{\includegraphics[height=0.18\textwidth]{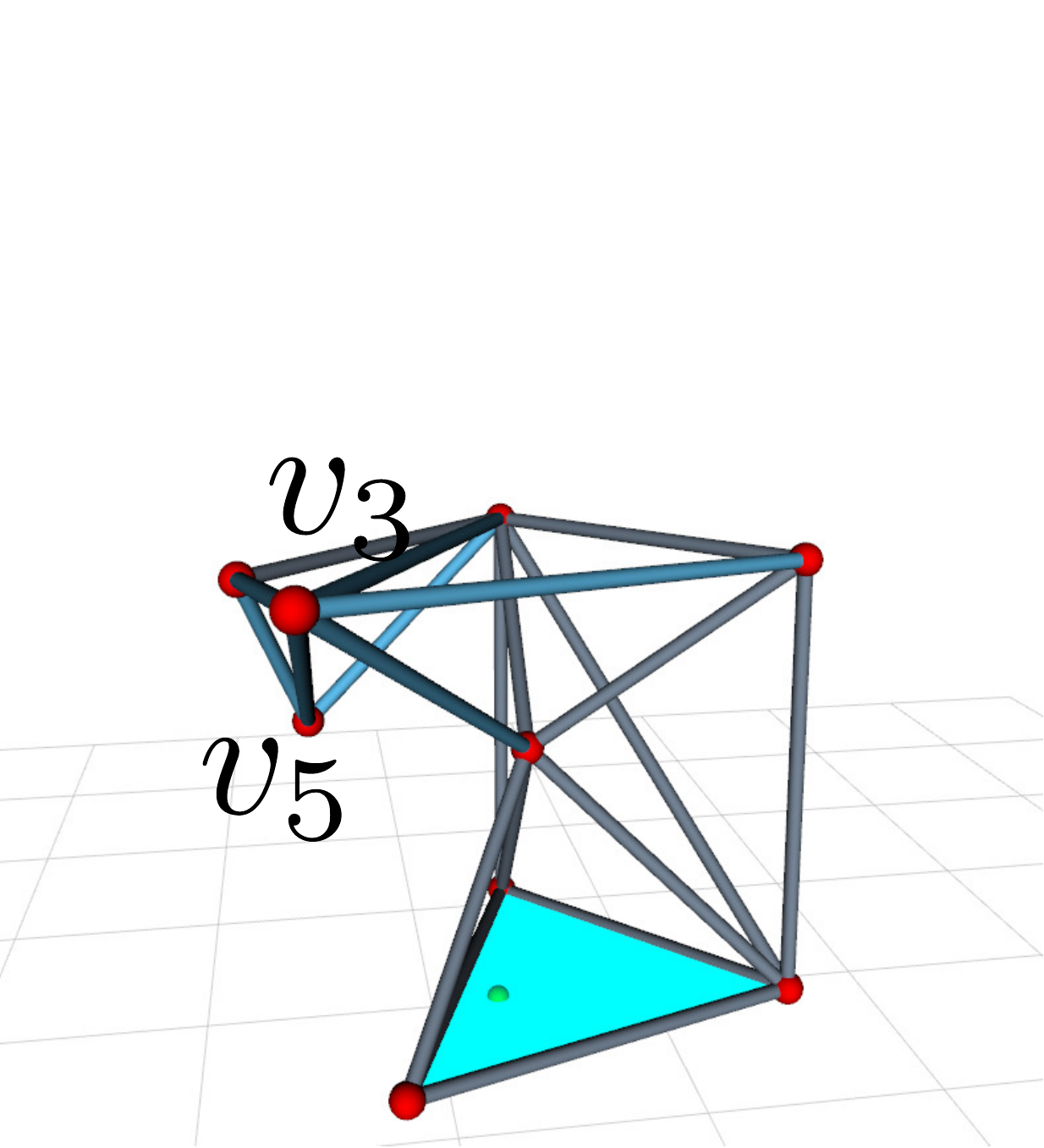}}
  \hfil
  \subfloat[]{\includegraphics[height=0.18\textwidth]{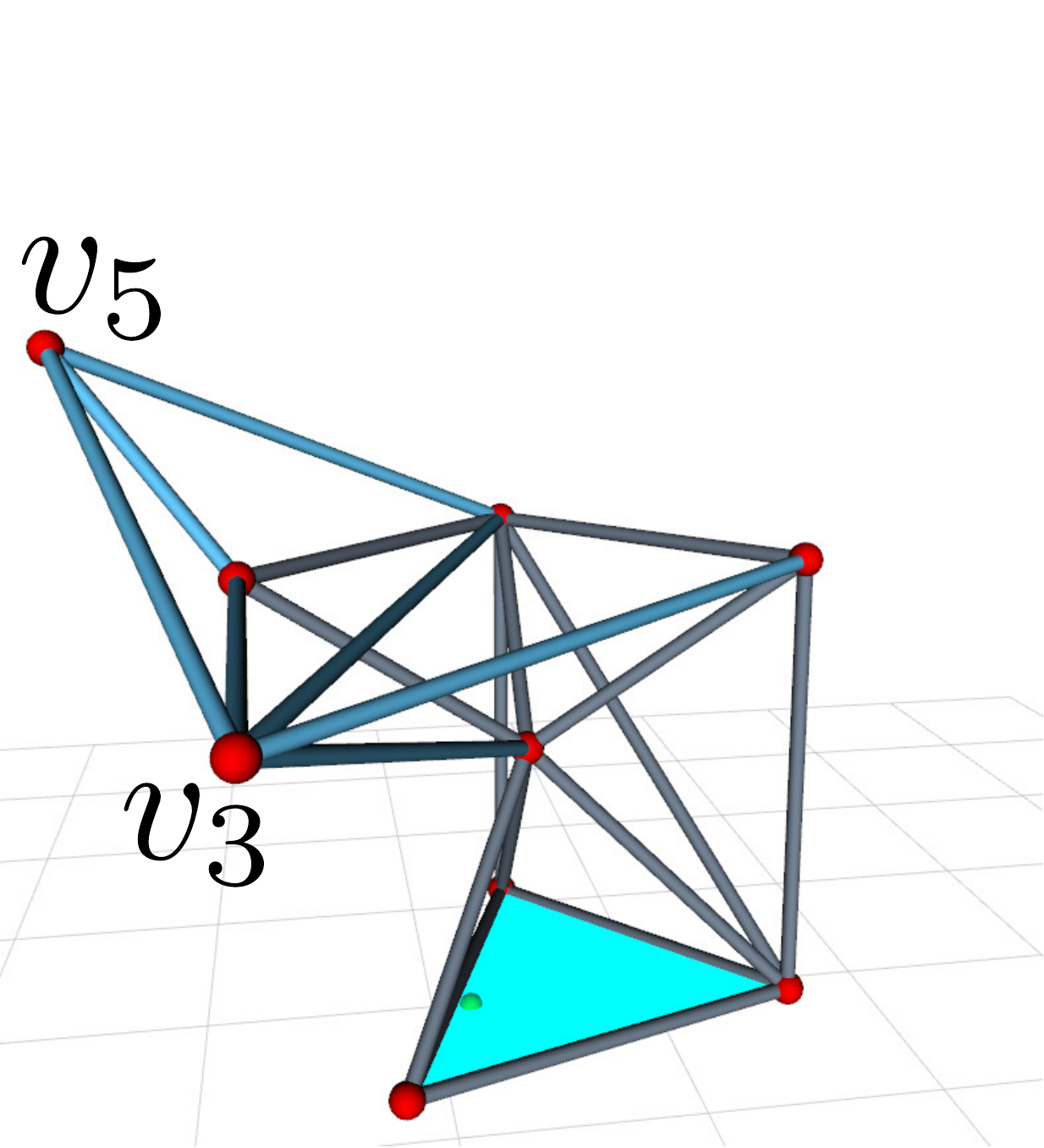}}
  \hfil
  \subfloat[]{\includegraphics[height=0.18\textwidth]{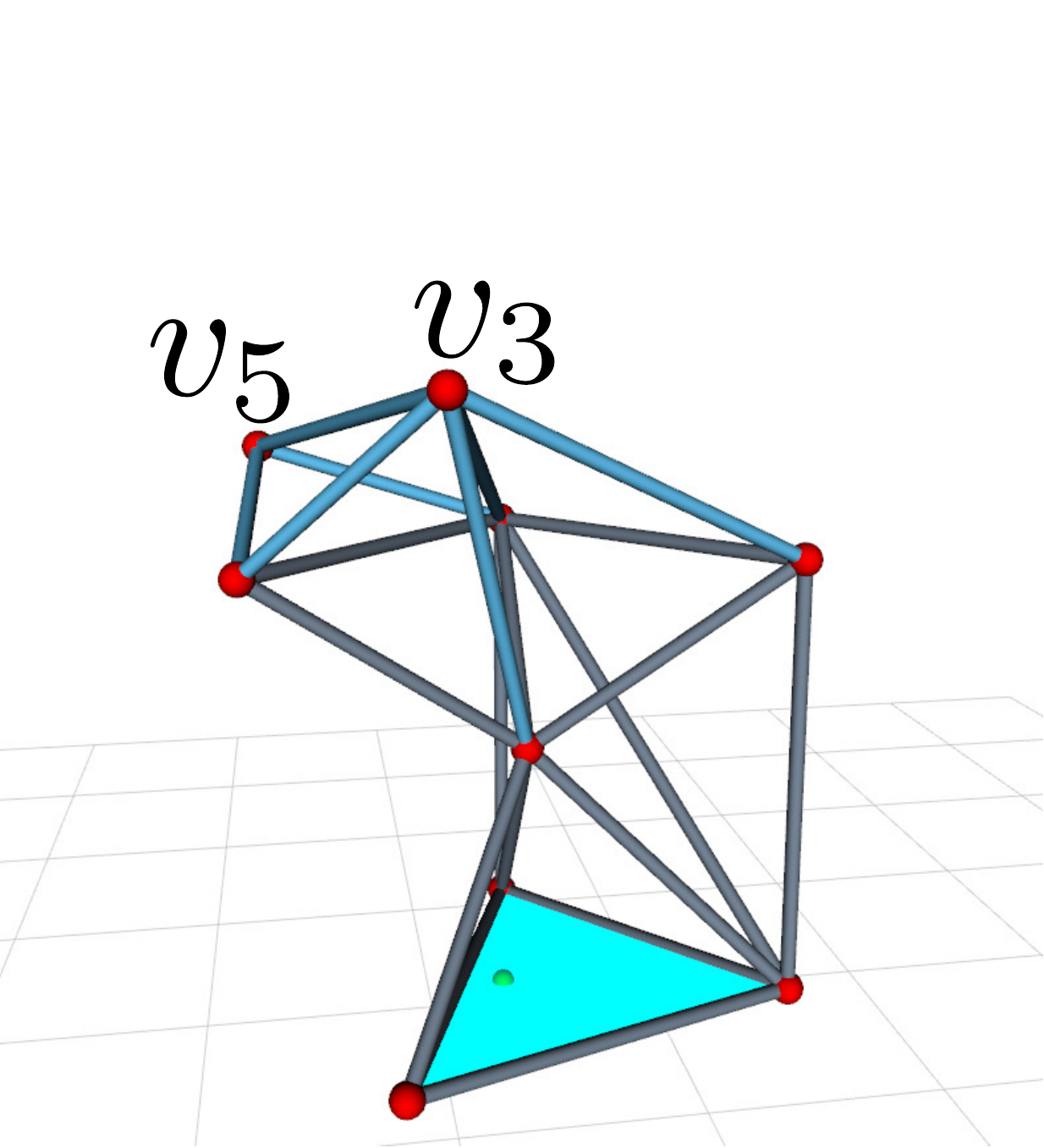}}
  \hfil
  \subfloat[]{\includegraphics[height=0.18\textwidth]{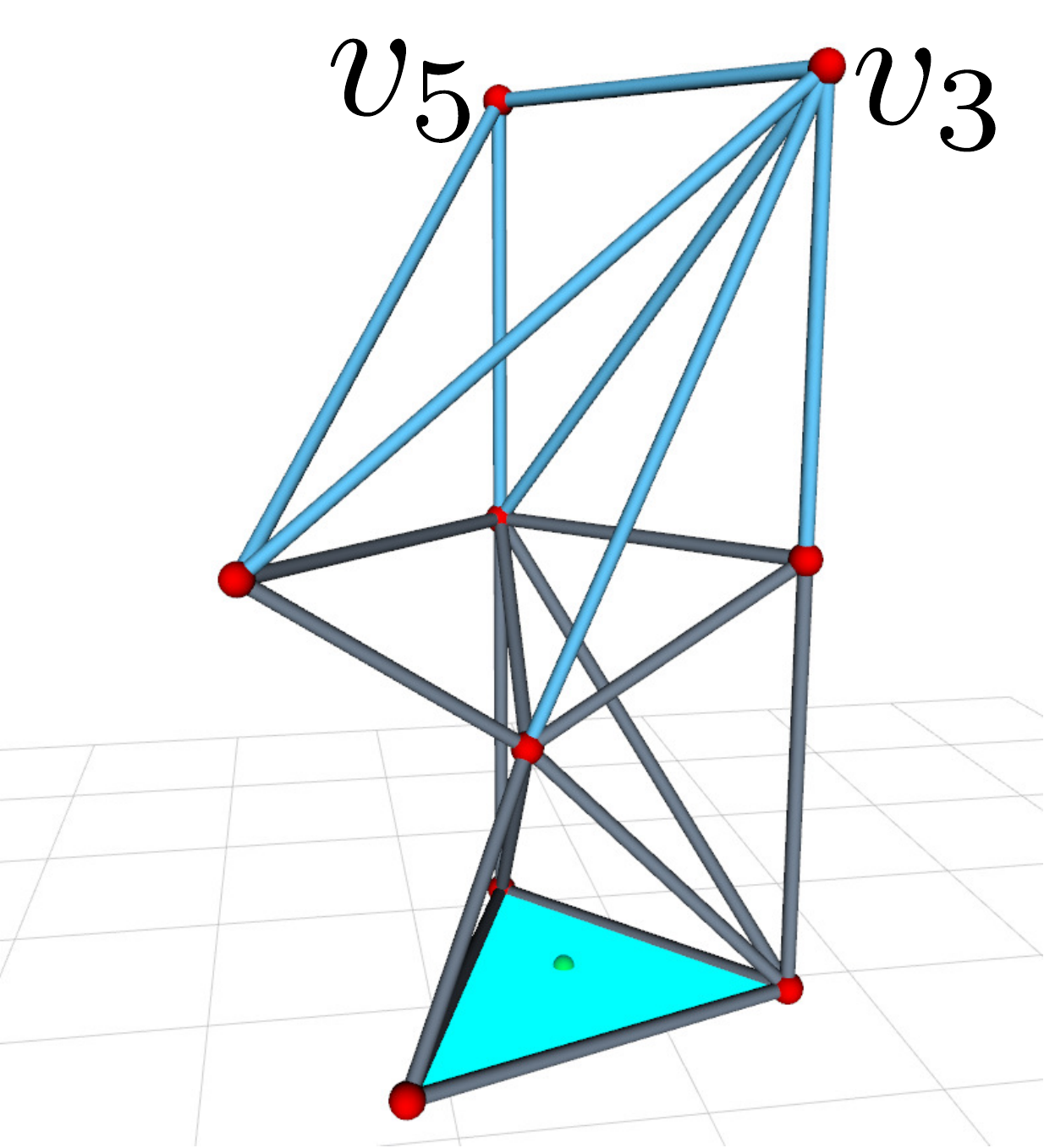}}
  \caption{$v_3$ and $v_5$ firstly extend outward, and then move
    upward to their goal positions. The support polygon is formed by
    three nodes ($v_2$, $v_4$, $v_5$) on the ground shown as the aqua
    region ({\color[rgb]{0,1,1}$\blacktriangle$}) and the green dot
    ({\color{green}{$\bullet$}}) is the center of mass represented on
    the ground.}
  \label{fig:geometry-reconfig-node-3-5}

  \centering
  \subfloat[]{\includegraphics[height=0.18\textwidth]{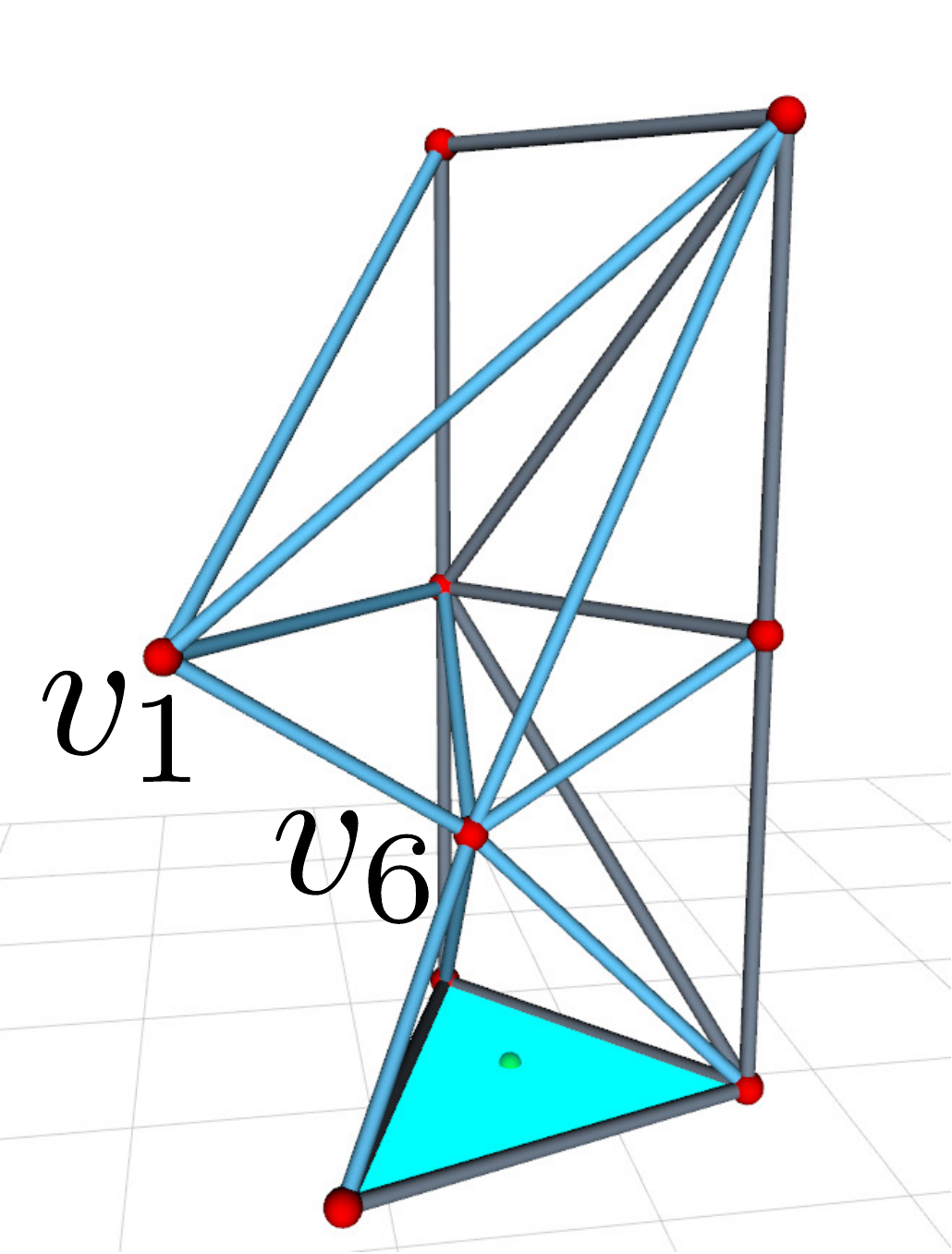}}
  \hfil
  \subfloat[]{\includegraphics[height=0.18\textwidth]{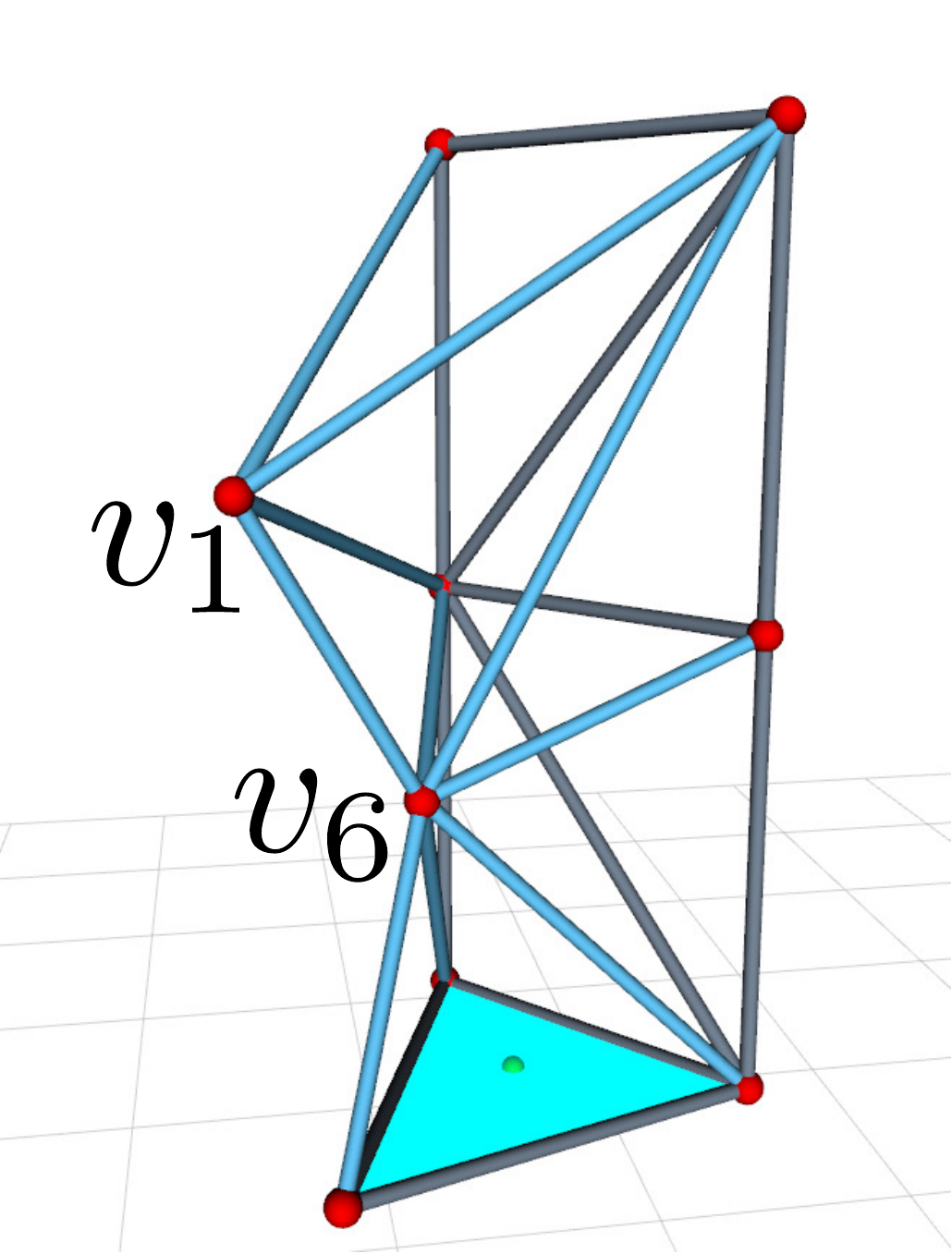}}
  \hfil
  \subfloat[]{\includegraphics[height=0.18\textwidth]{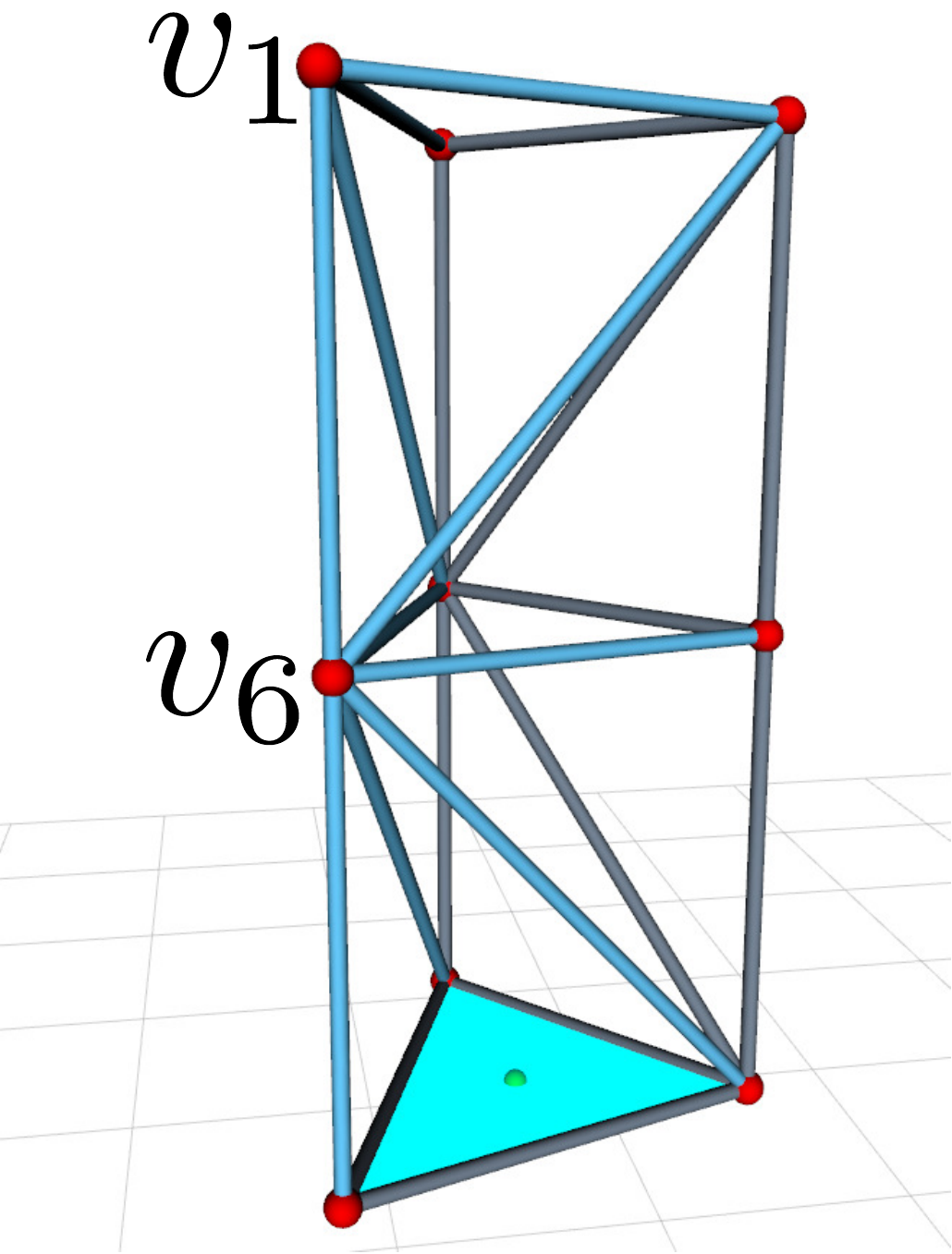}}
  \caption{$v_1$ and $v_6$ can navigate to their goal positions easily
    since $\widehat{\mathcal{C}}_{\mathrm{free}}^{v_1}(q_i^{v_1})$ and
    $\widehat{\mathcal{C}}_{\mathrm{free}}^{v_6}(q_i^{v_6})$ almost cover the
    whole workspace.}
  \label{fig:geometry-reconfig-node-1-6}
\end{figure*}

\section{Test Scenarios}
\label{sec:experiment}

The motion planning framework is implemented in C++. Four example
scenarios were conducted to measure the effectiveness of our
approach. The performance of the reconfiguration framework is compared
with the geometry reconfiguration approach
in~\cite{vtt-review-ur-2018} and the topology reconfiguration approach
in~\cite{Liu-vtt-planning-iros-2019}. The performance of the
locomotion framework is compared with the approach
in~\cite{Park-vtt-locomotion-ral-2020}
and~\cite{Usevitch-lar-locomotion-tro-2020}. All the comparisons are
based on the test scenarios from these previous works. These
experiments are also tested under different constraints and with
different parameters to show the universality of our framework. All
tests run on a laptop computer (Intel Core i7-8750H CPU, 16GB RAM) and
the workspace is a cuboid.

\subsection{Geometry Reconfiguration}

\subsubsection{Cube to Tower}

The geometry reconfiguration planning test changes the cube shape of a
VTT, Fig.~\ref{fig:multi-node-init-truss}, to a tower shape,
Fig.~\ref{fig:multi-node-goal-truss}.  The constraints for this task
are $\overline{L}_{\min}=\SI{1.0}{m}$,
$\overline{L}_{\max}=\SI{3.5}{m}$,
$\bar{\theta}_{\min}=\SI{0.3}{\radian}$, and $\bar{\mu}_{\min}=0.1$.

\begin{figure}[t]
  \centering
  \includegraphics[width=0.25\textwidth]{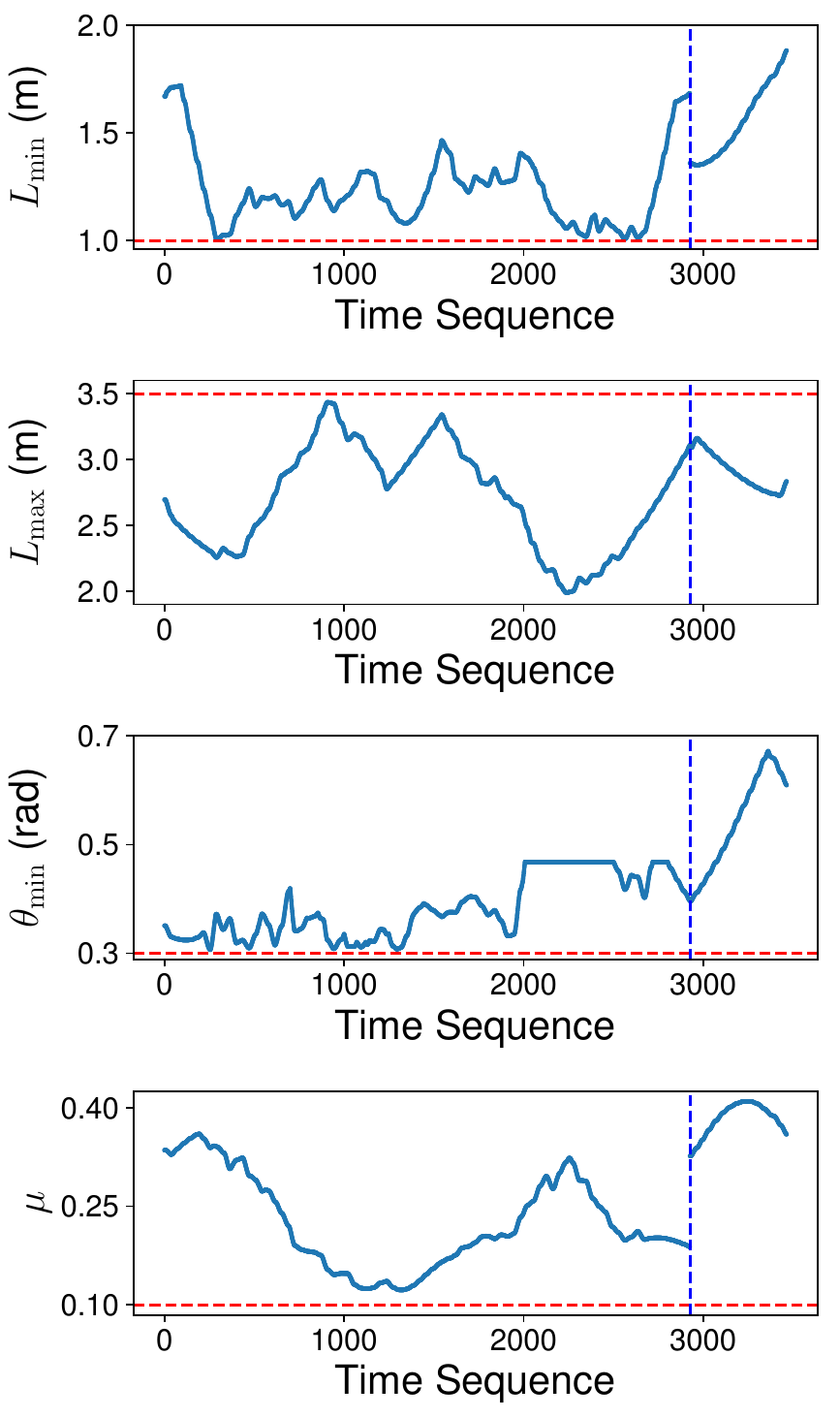}
  \caption{The minimum length ($L_{\min}$) and the maximum length
    ($L_{\max}$) of all moving edge modules, the minimum angle between
    every pair of edge modules ($\theta_{\min}$), and the motion
    manipulability ($\mu$) are measured throughout the geometry
    reconfiguration process in
    Fig.~\ref{fig:geometry-reconfig-node-3-5} and
    Fig.~\ref{fig:geometry-reconfig-node-1-6}.}
  \label{fig:geometry-test-data}
\end{figure}

Four nodes $v_1$, $v_3$, $v_5$, and $v_6$ are involved in this motion
task. These four nodes are separated into two groups $\{v_3, v_5\}$
and $\{v_1, v_6\}$ which are randomly selected. We first compute
$\widehat{\mathcal{C}}_{\mathrm{free}}^{v_3}(q_i^{v_3})$ and
$\widehat{\mathcal{C}}_{\mathrm{free}}^{v_5}(q_i^{v_5})$, and then do
planning for these two nodes. The motions of node $v_3$ and $v_5$ are
shown in Fig.~\ref{fig:geometry-reconfig-node-3-5}. Most of the
obstacle regions in this step are surrounded by the subspace
$\widehat{\mathcal{C}}_{\mathrm{free}}^{v_3}(q_i^{v_3})$ and
$\widehat{\mathcal{C}}_{\mathrm{free}}^{v_5}(q_i^{v_5})$, hence it is
easier for them to extend outward first in order to navigate to the
goal positions. This motion process moves the projected center of mass
toward one edge of the support polygon but the planner can constrain
the projected center of mass within the support polygon. After
planning for $v_3$ and $v_5$, the truss is updated, and
$\widehat{\mathcal{C}}_{\mathrm{free}}^{v_1}(q_i^{v_1})$ and
$\widehat{\mathcal{C}}_{\mathrm{free}}^{v_6}(q_i^{v_6})$ are computed
accordingly. Finally the planning for $v_1$ and $v_6$ finishes this
motion task with the result shown in
Fig.~\ref{fig:geometry-reconfig-node-1-6}. For $v_1$ and $v_6$ in this
updated truss, the enclosed subspace
$\widehat{\mathcal{C}}_{\mathrm{free}}^{v_1}(q_i^{v_1})$ and
$\widehat{\mathcal{C}}_{\mathrm{free}}^{v_6}(q_i^{v_6})$ almost covers
the whole workspace so it is also easy for them to navigate to the
goal positions. The minimum length ($L_{\min}$) and the maximum length
($L_{\max}$) of all moving edge modules, the minimum angle between
every pair of edge modules ($\theta_{\min}$), and the motion
manipulability ($\mu$) are shown in
Fig.~\ref{fig:geometry-test-data}. Note that $L_{\min}$, $L_{\max}$,
and $\mu$ are not necessarily to be continuous because nodes that are
under control are changing. When moving $v_3$ and $v_5$, only compute
the lengths of all blue members in
Fig.~\ref{fig:geometry-reconfig-node-3-5} and compute the
manipulability by setting $V_C=\{v_3, v_5\}$ (before the blue dashed
line in Fig.~\ref{fig:geometry-test-data}). Similarly, when moving
$v_1$ and $v_6$, only compute the lengths of all blue members in
Fig.~\ref{fig:geometry-reconfig-node-1-6} and compute the
manipulability by setting $V_C=\{v_1, v_6\}$ (after the blue dashed
line in Fig.~\ref{fig:geometry-test-data}). This motion task is also
demonstrated in~\cite{vtt-review-ur-2018} using the retraction-based
RRT algorithm. This algorithm cannot solve this motion planning task
in 100 trials unless an intermediate waypoint is manually specified to
mitigate the narrow passage issue. The success rate for the planning
from the initial to the waypoint is 99\% and 98\% from the waypoint to
the goal. In comparison, with \texttt{RRTConnect}, our algorithm
doesn't need any additional waypoints, and we did 1000 trials and the
mean running time is \SI{4.29}{s} with a standard deviation of
\SI{1.99}{s} and the success rate is 100\%. In these trials, the
maximum running time is \SI{7.56}{s} and the minimum is
\SI{1.01}{s}. The mean computing time for configuration space is
\SI{1.04}{s} with a standard deviation of \SI{0.38}{s}, and the
maximum computing time is \SI{1.61}{s} and the minimum is
\SI{0.52}{s}. The planing time is longer than that in our previous
conference paper because more constraints are considered leading to
more complicated configuration space computation and state validation.

\begin{figure}[b]
  \centering
  \subfloat[]{\includegraphics[height=0.18\textwidth]{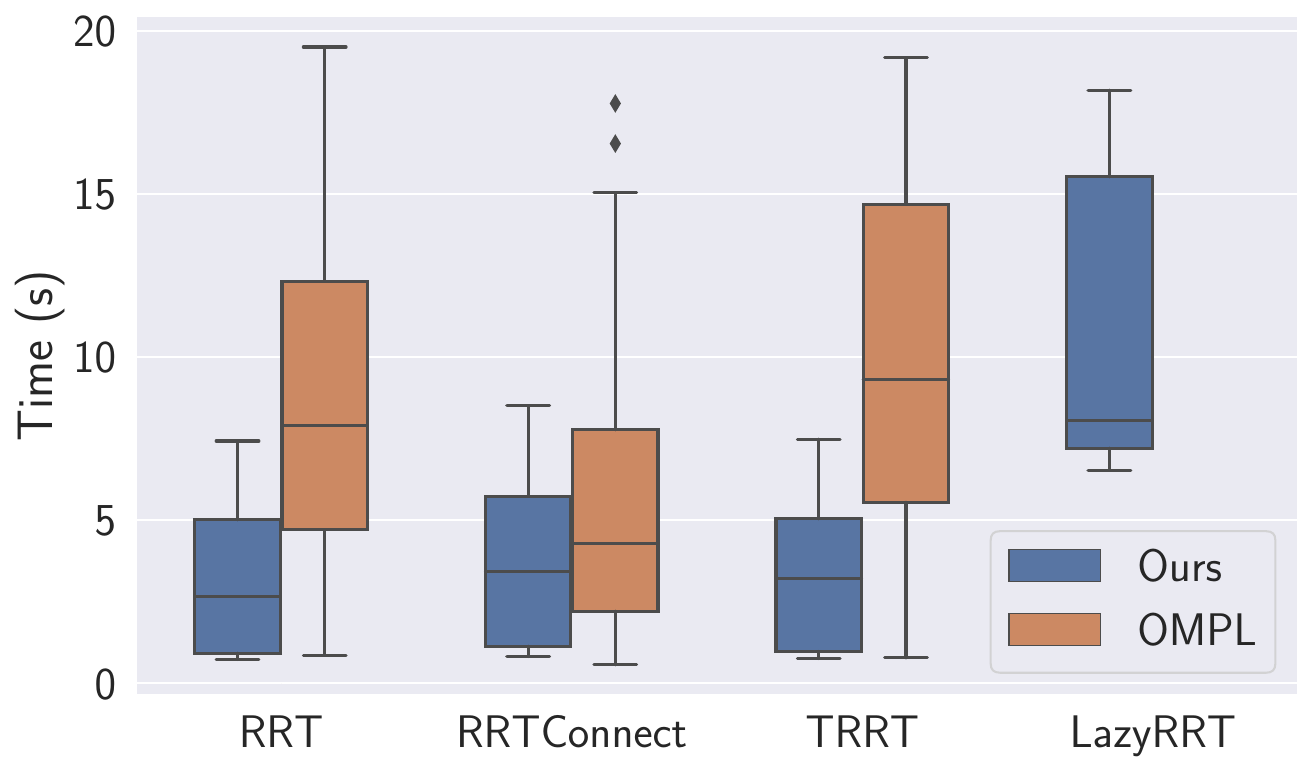}\label{fig:geometry-reconfig-benchmark}}
  \hfil
  \subfloat[]{\includegraphics[height=0.18\textwidth]{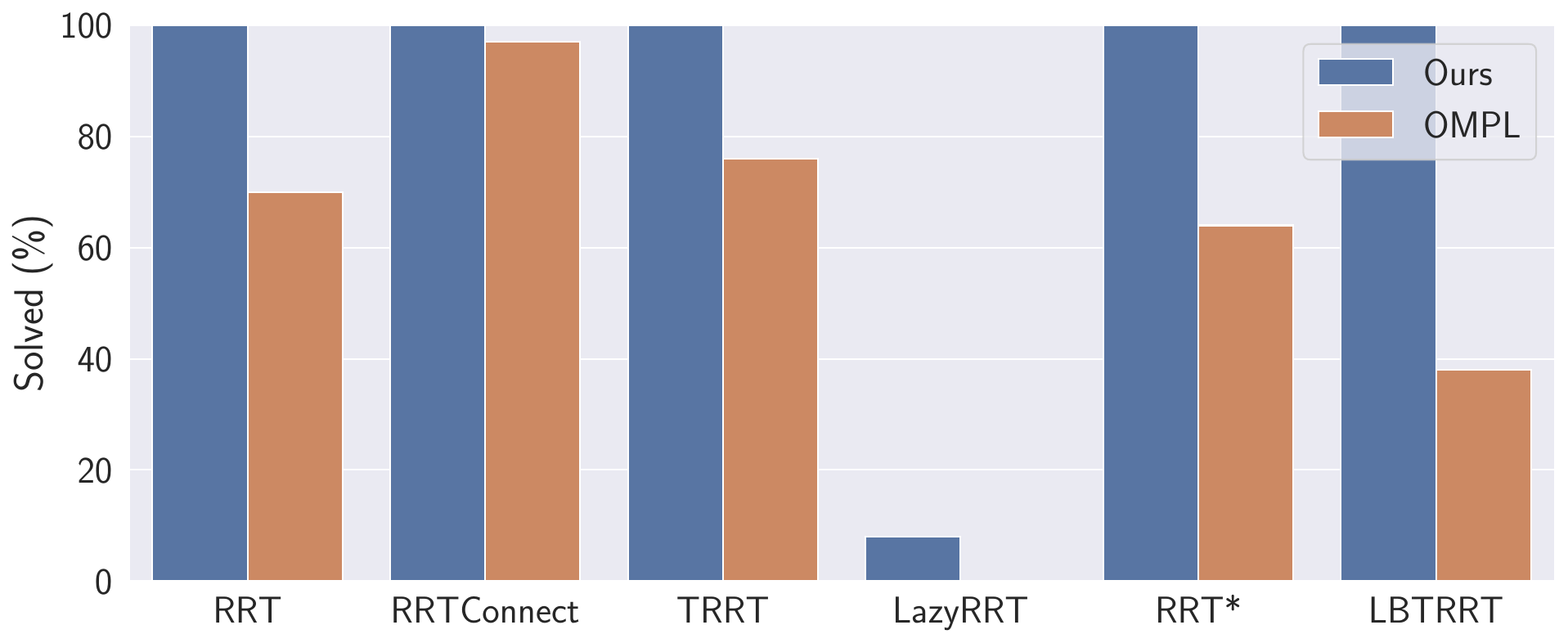}\label{fig:geometry-reconfig-success}}
  \caption{(a) The running time benchmark and (b) the solved
    percentage benchmark against standard OMPL planners.}
  \label{fig:geometry-benchmark}
\end{figure}

\begin{figure*}[b]
  \centering
  \subfloat[]{\includegraphics[height=0.18\textwidth]{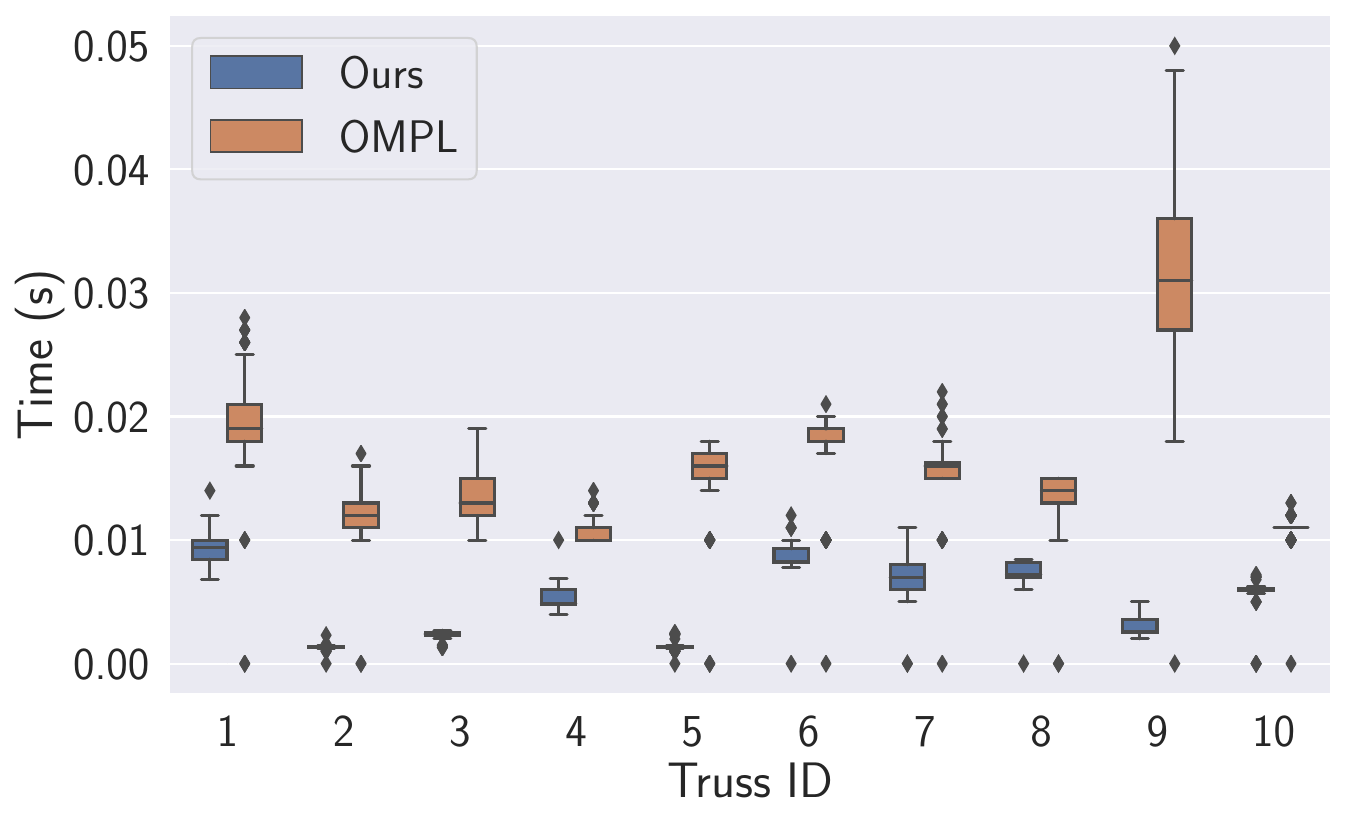}\label{fig:random-benchmark-1}}
  \hfil
  \subfloat[]{\includegraphics[height=0.18\textwidth]{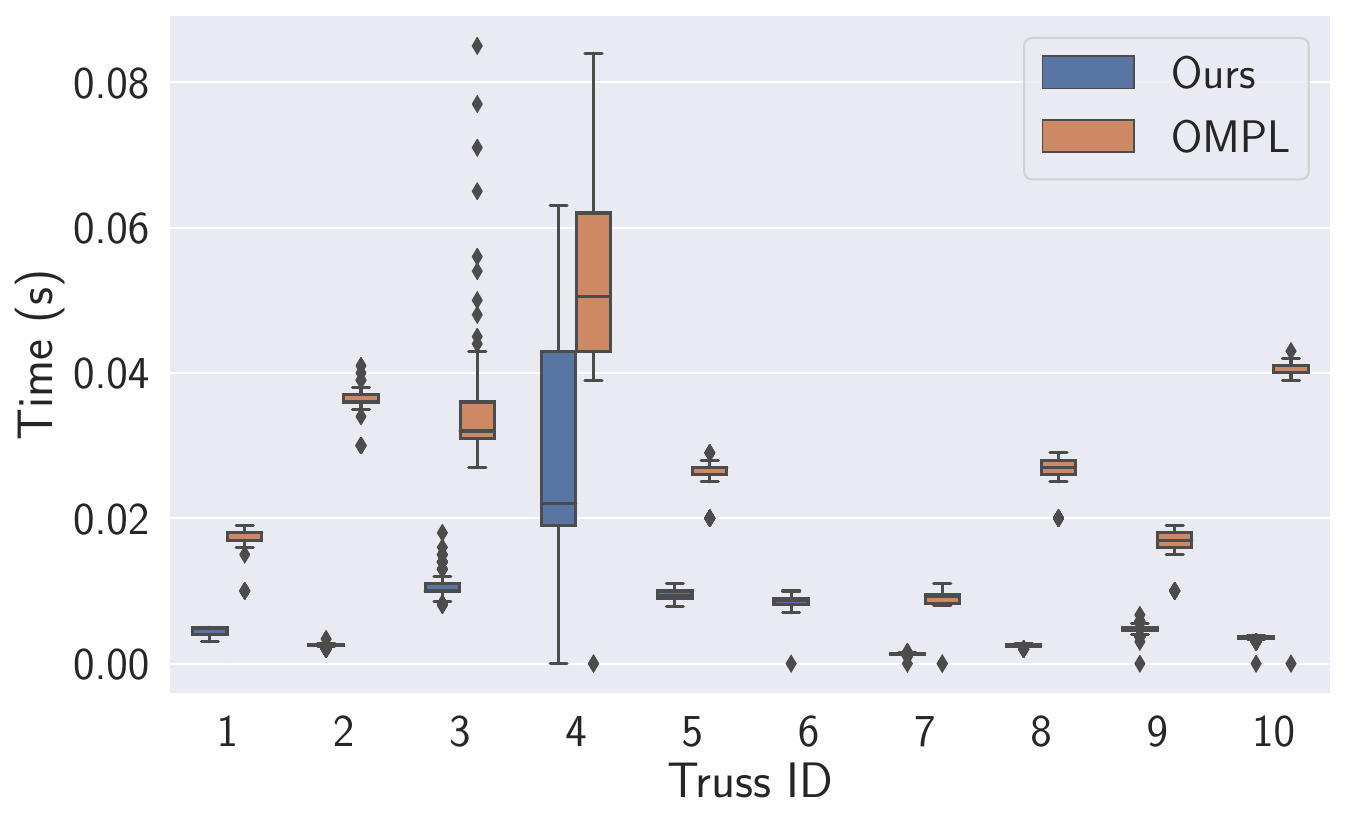}\label{fig:random-benchmark-2}}
  \hfil
  \subfloat[]{\includegraphics[height=0.18\textwidth]{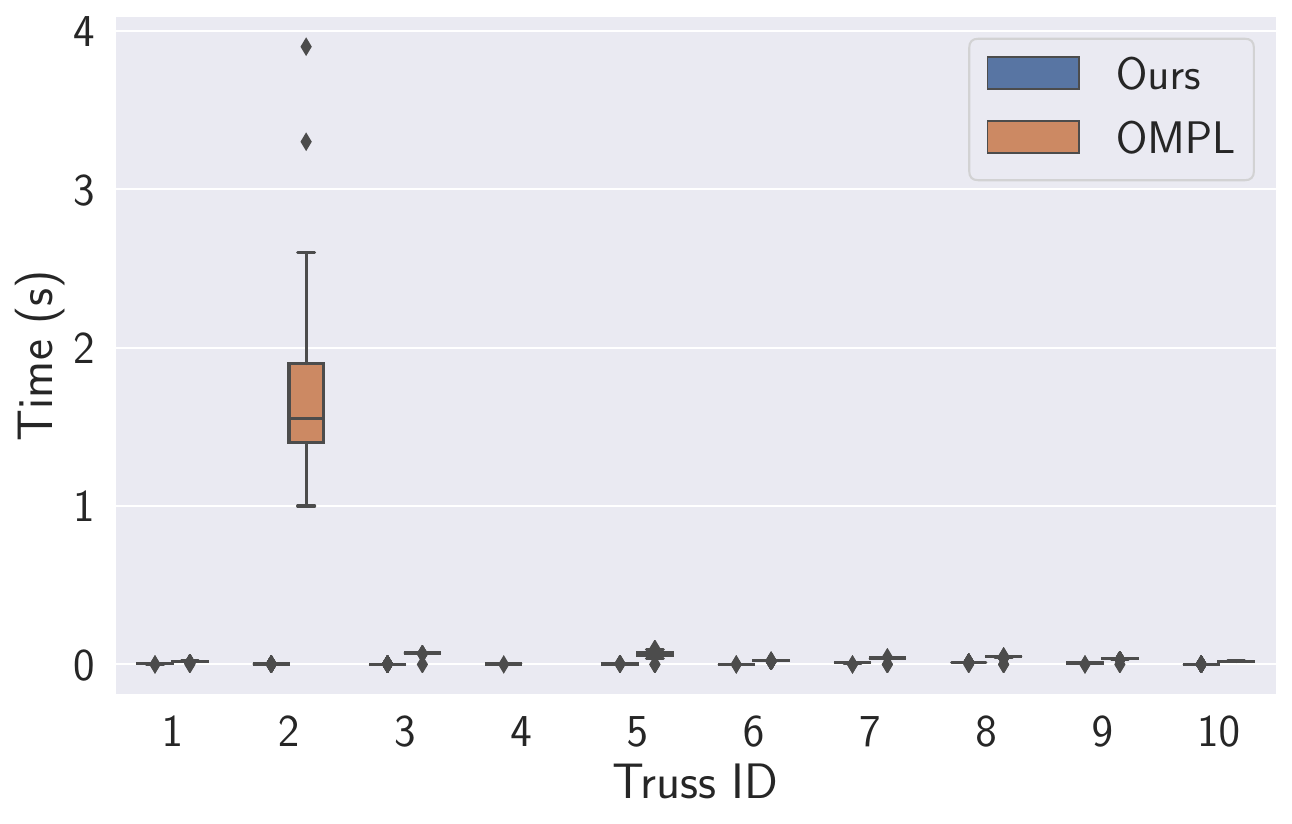}\label{fig:random-benchmark-3}}
  \caption{Benchmarks of randomly generated VTTs: (a) Group 1, (b)
    Group 2, (c) Group 3.}
  \label{fig:random-benchmark}
\end{figure*}

\subsubsection{Benchmark Test}

\begin{figure}[t]
  \centering
  \subfloat[]{\includegraphics[width=0.16\textwidth]{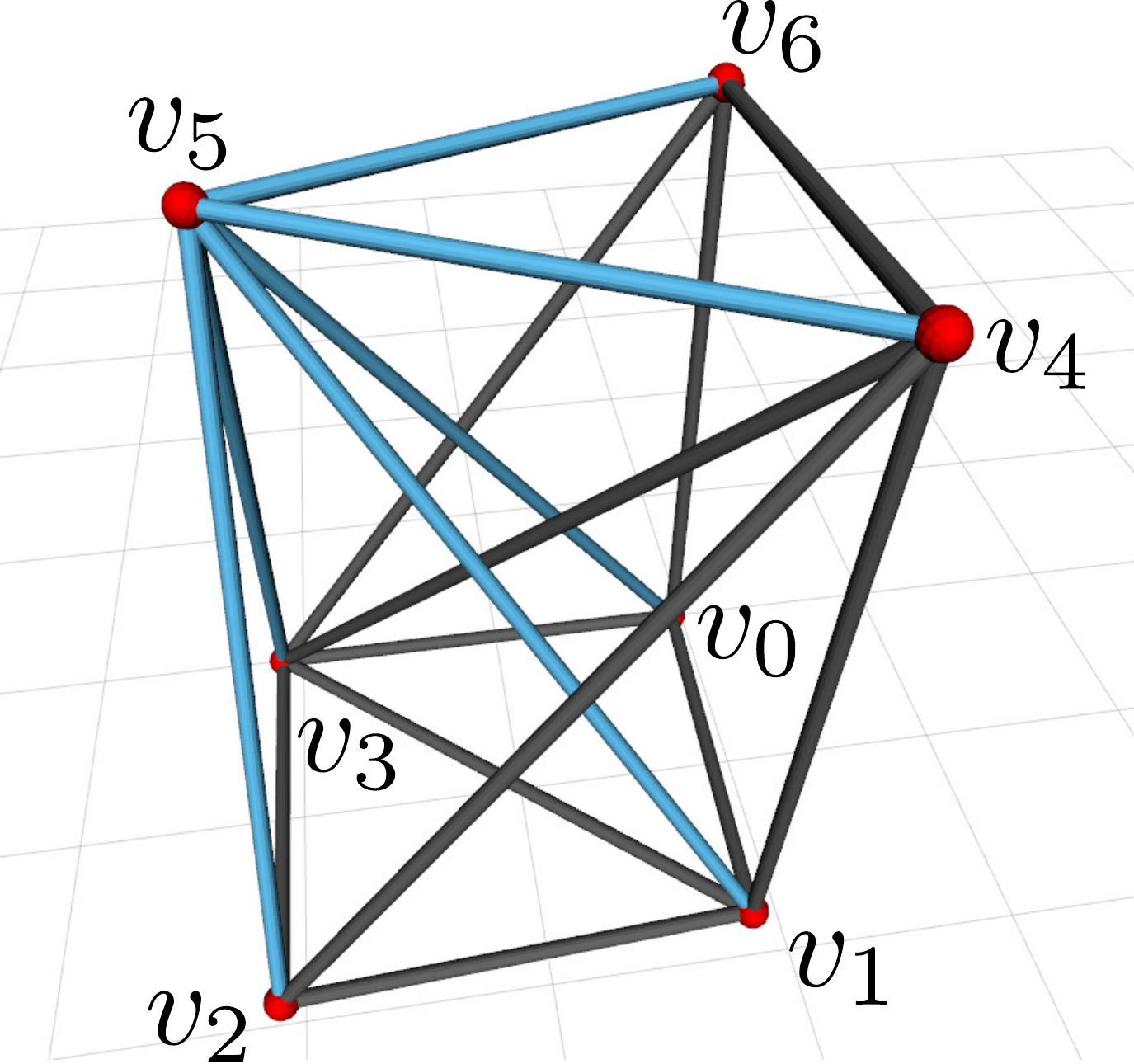}\label{fig:topology-reconfig-vtt}}
  \hfil
  \subfloat[]{\includegraphics[width=0.16\textwidth]{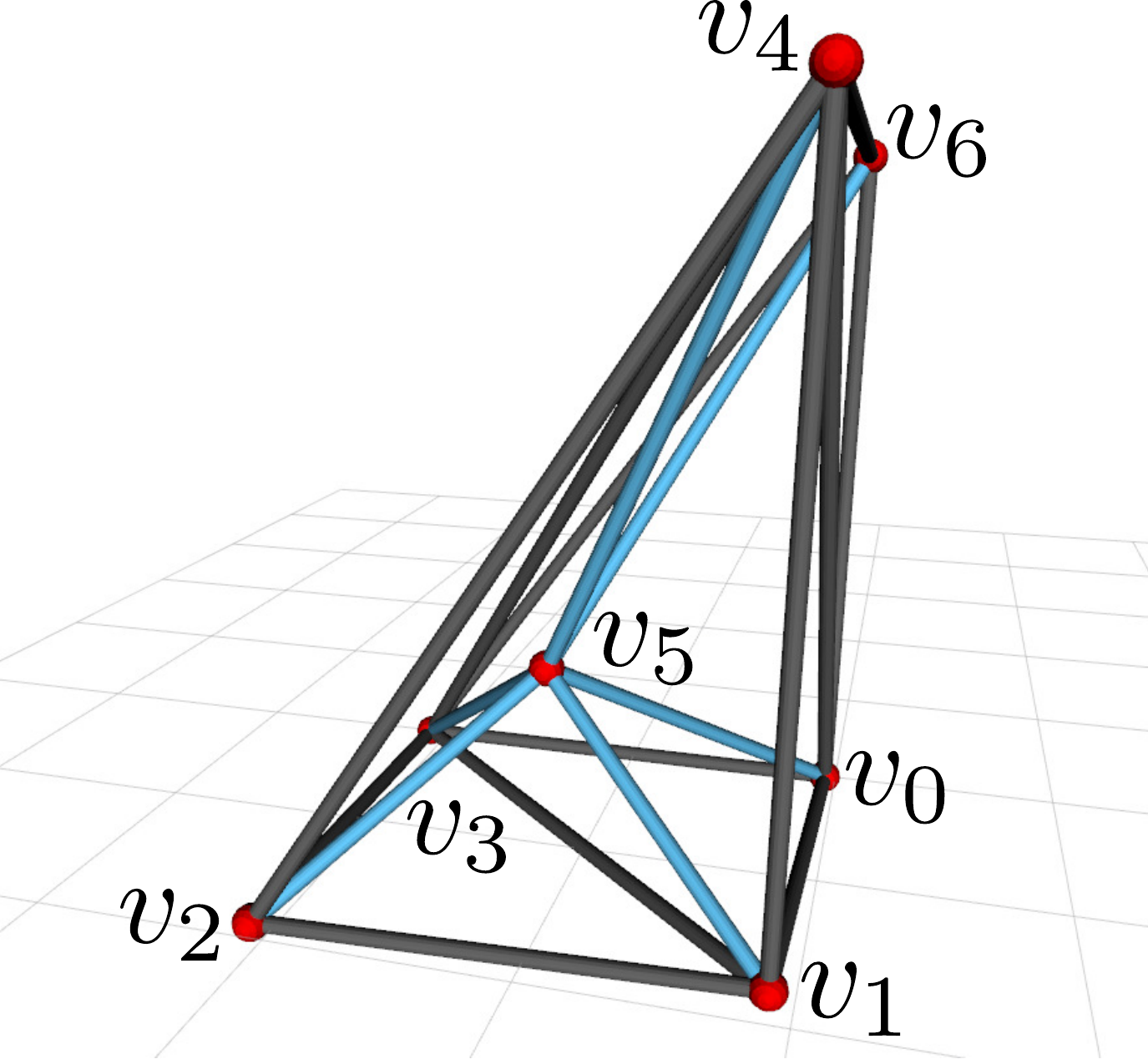}\label{fig:topology-reconfig-goal}}
  \caption{(a) A VTT is constructed from 18 edge modules with 6
    nodes. (b) The goal is to move $v_5$ from its initial position to
    a position inside the truss.}
\end{figure}

\begin{figure}[t]
  \centering
  \includegraphics[width=0.18\textwidth]{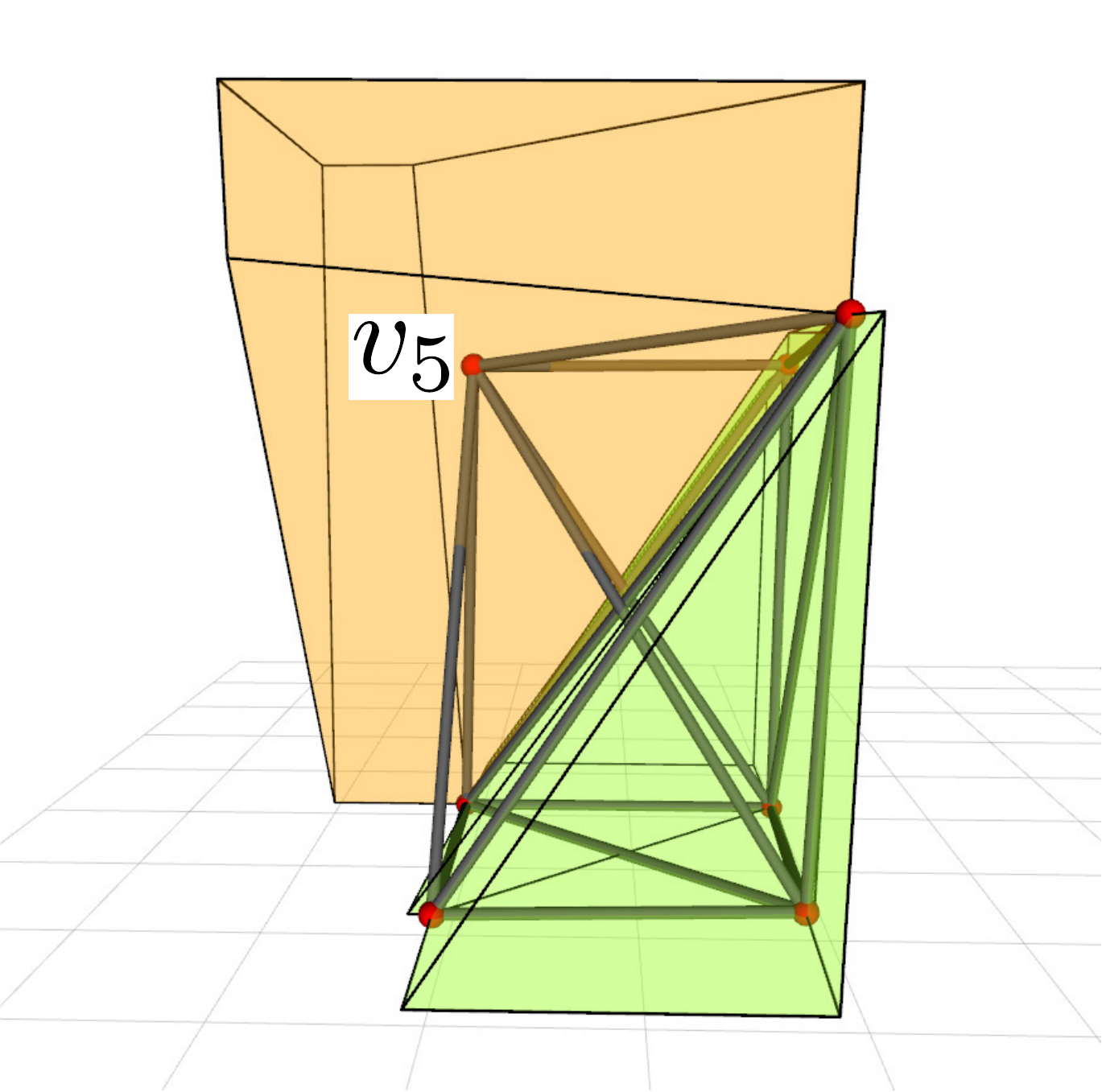}
  \caption{$\mathcal{C}_{\mathrm{free}}^{v_5}(q_i^{v_5})$ is the
    yellow space on the upper left and
    $\mathcal{C}_{\mathrm{free}}^{v_5}(q_g^{v_5})$ is the green space
    on the lower right. They are not connected and are separated by
    the obstacle region generated from edge $(v_3, v_4)$.}
  \label{fig:topology-free-space}
\end{figure}

The performance of our framework is tested rigorously with several
benchmarks. We first test the task shown in
Fig.~\ref{fig:geometry-reconfig-test} and compare the performance with
standard OMPL planners. We use the Flexible Collision Library
(FCL)~\cite{Jia-fcl-icra-2012} for collision detection. In a VTT,
every node is modeled as a sphere and every member is modeled as a
cylinder. For standard OMPL planners, we have to set much smaller
resolution at which state validity needs to be verified in order to
ensure the motion does not skip over self-collision. Here we compare
our planner with standard OMPL planners using several commonly used
sampling-based planning approaches. The benchmark is shown in
Fig.~\ref{fig:geometry-benchmark}. Our planner can solve the task
faster and also achieve higher success rate (100\% except for using
\texttt{Lazy RRT} and standard OMPL using \texttt{Lazy RRT} cannot
find a solution). For optimal (\texttt{RRT*}) or near-optimal
(\texttt{LBTRRT}) version of RRT, our planner also shows better
performance in terms of success rate.

This benchmark shows our planner outperforms standard OMPL
planners. The configuration space calculation significantly benefits
the sampling-based planning phase in the following. First, both the
maximum length of a motion to be added when searching and the
resolution to validate state can be large because we have better way
to validate motions. Second, our planner doesn't heavily rely on the
sampling-based planners. In our test, we set the maximum running time
of the sampling-based planners to be only \SI{2}{s} for our planner,
but we need to set this parameter to be \SI{20}{s} for standard OMPL
planners in order to achieve higher success rate. Also some specific
sampling-based planners, such as \texttt{Transition-based RRT}
(\texttt{TRRT}), can have much better performance. \texttt{Lazy RRT}
performs badly for truss robots. OMPL cannot solve this task and our
planner can only achieve 8\% success rate. This is because
\texttt{Lazy RRT} is optimistic and attempts to find a solution as
soon as possible without checking collision. Once a solution is found,
validity checking is applied, and if collisions are found, the invalid
path segments are removed and the search process is
continued. However, VTT robots have a high probability of
self-collision.

\begin{table}[b]
  \centering
  \caption{Random Truss Test}
  \begin{tabular}{cccc}
    \toprule
    Group&\# of Nodes&\# of Members&\# of Moving Nodes\\
    \midrule
    1&6&12&$\le 2$\\
    2&8&15&$\le 3$\\
    3&9&20&$\le 4$\\
    \bottomrule
  \end{tabular}
  \label{tab:random-truss}
\end{table}

\begin{figure}[t!]
  \centering
  \subfloat[]{\includegraphics[width=0.11\textwidth]{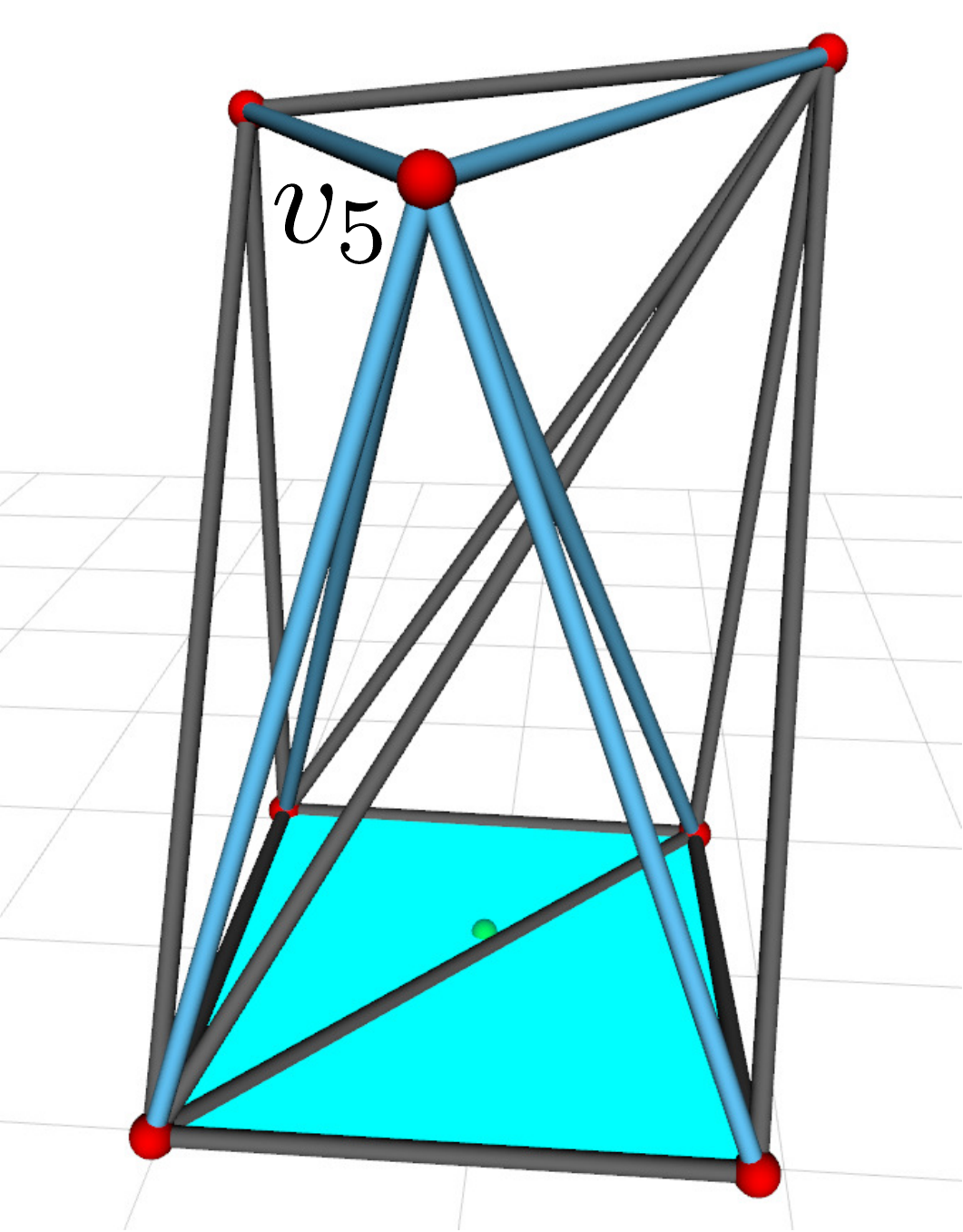}}
  \hfil
  \subfloat[]{\includegraphics[width=0.11\textwidth]{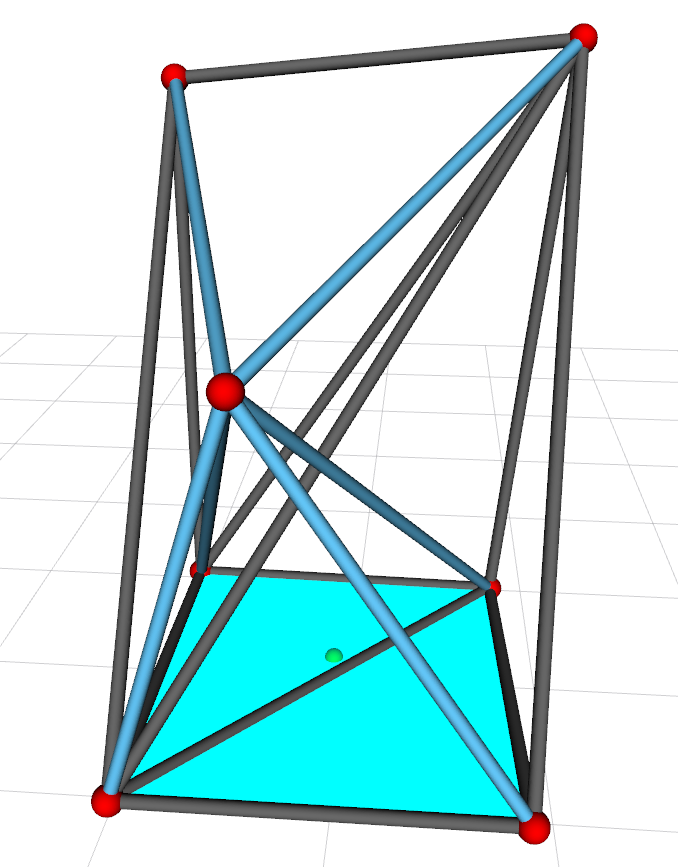}}
  \hfil
  \subfloat[]{\includegraphics[width=0.11\textwidth]{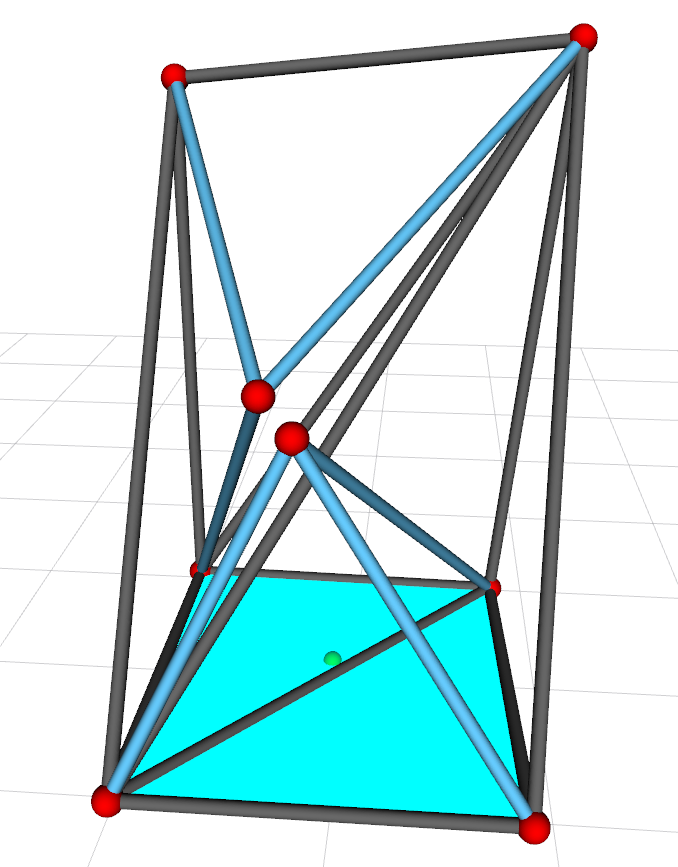}}
  \hfil
  \subfloat[]{\includegraphics[width=0.11\textwidth]{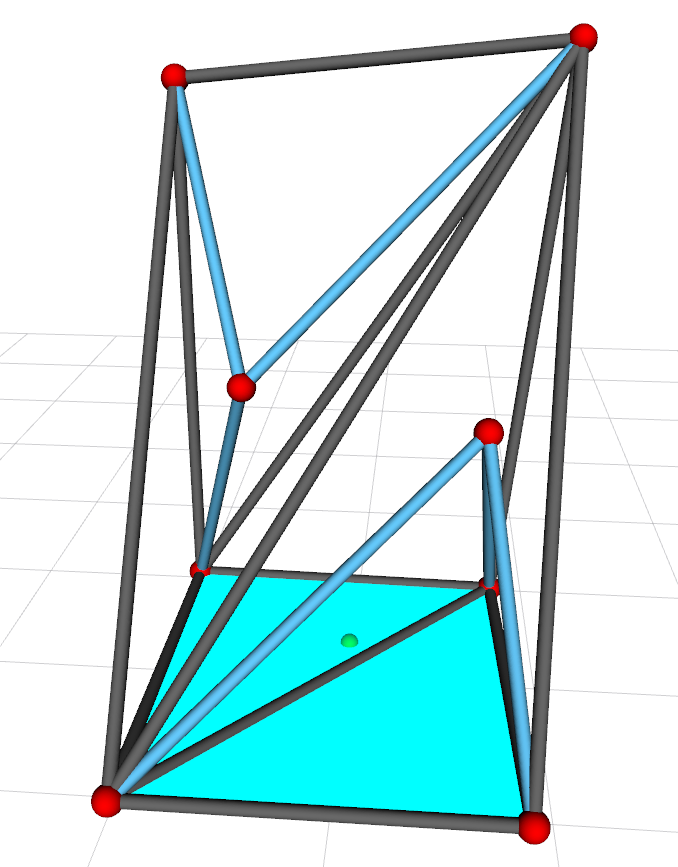}}
  \hfil
  \subfloat[]{\includegraphics[width=0.11\textwidth]{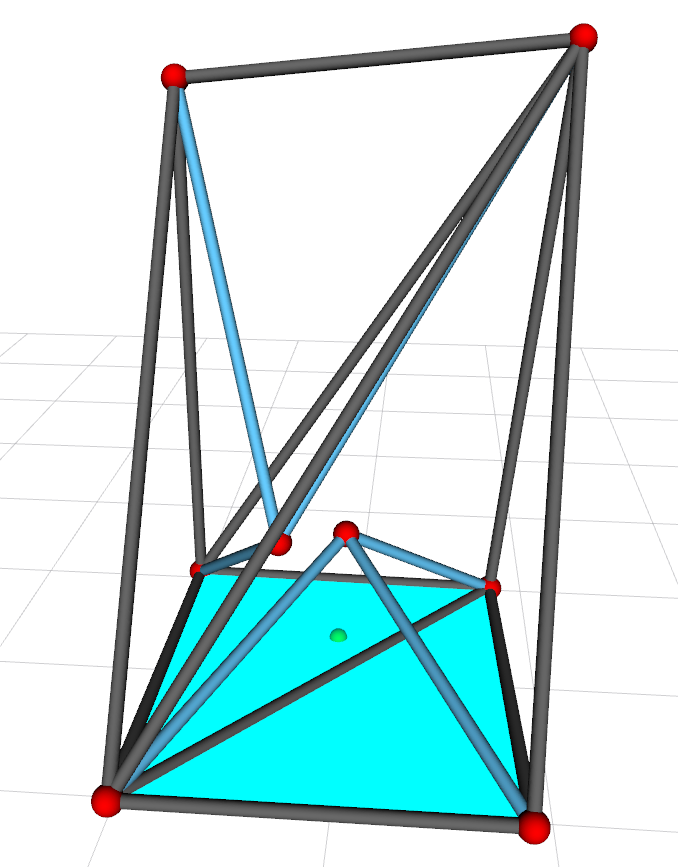}}
  \hfil
  \subfloat[]{\includegraphics[width=0.11\textwidth]{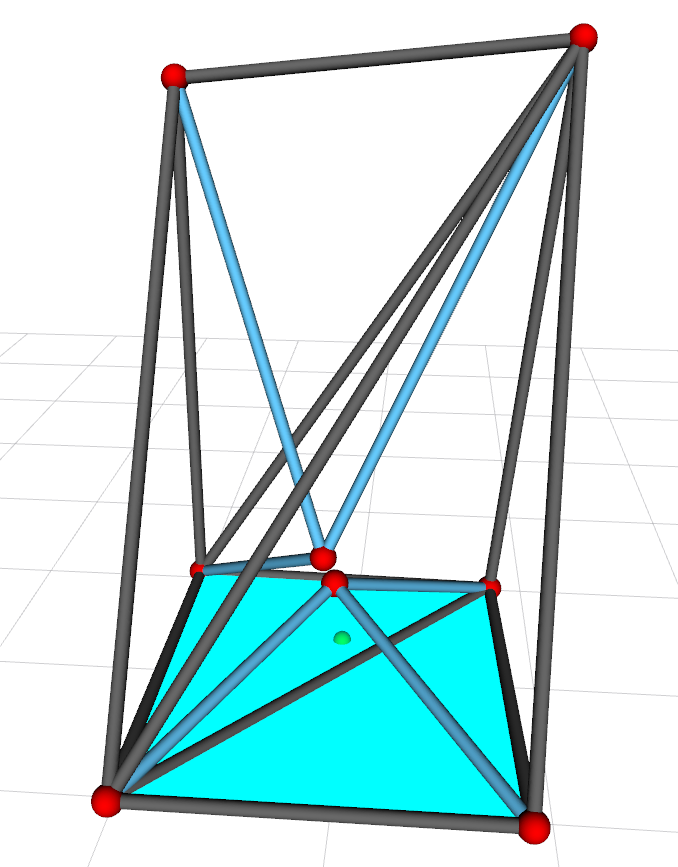}}
  \hfil
 \subfloat[]{\includegraphics[width=0.11\textwidth]{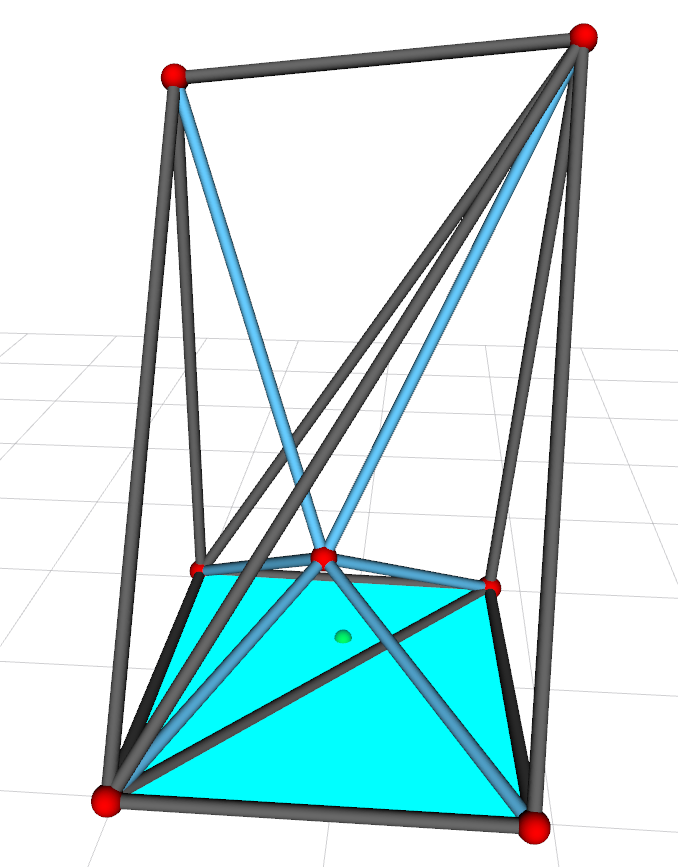}}
 \hfil
  \subfloat[]{\includegraphics[width=0.11\textwidth]{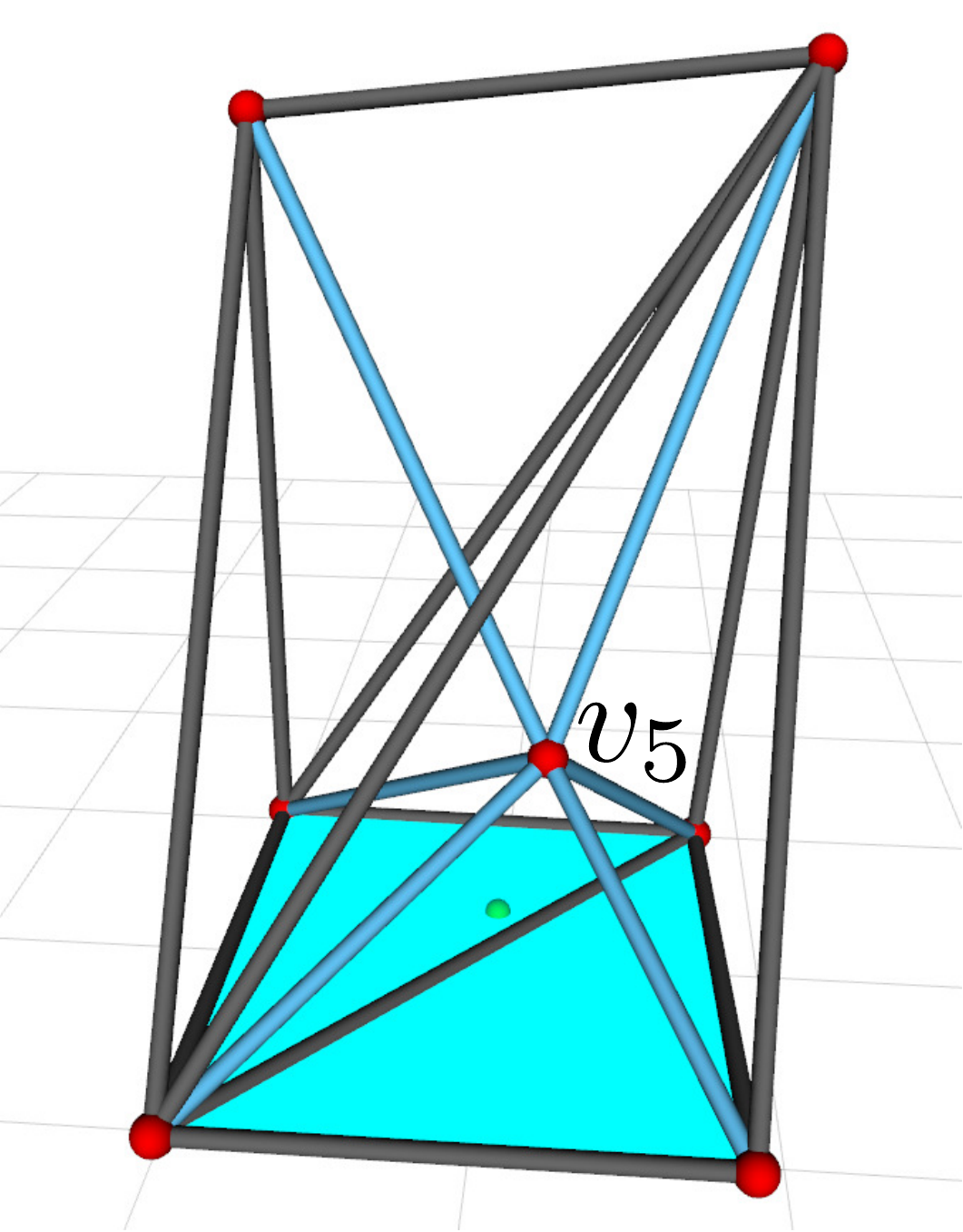}}
  \caption{The sequence to move $v_5$ from $q_i^{v_5}$ to $q_g^{v_5}$
    is shown. The support polygon is the aqua region
    ({\color[rgb]{0,1,1}$\blacksquare$}) and the green dot
    ({\color{green}{$\bullet$}}) is the center of mass represented on
    the ground. (a) --- (b) First move $v_5$ to a new location. (c)
    Split $v_5$ into a pair. (d) --- (e) Move these two newly
    generated nodes in different directions to go around the edge
    module $(v_3, v_4)$. (f) --- (h) Merge them into an individual
    node and move this node to $q_g^{v_5}$.}
  \label{fig:topology-reconfig-process}
\end{figure}

We also randomly generate three groups of trusses in different sizes
and randomly generate motion goals shown in
Table~\ref{tab:random-truss}. For every truss, we randomly select the
nodes to move with randomly generated goals. In order to ensure valid
goals (without topology reconfiguration), we compute the free space of
every moving node and sample a goal inside this free space, which may
lead to easy tasks because the free space for a group of nodes is
larger than the union of the free space of every individual node in
the group. Here we mainly compare the running time of the
sampling-based planning phase using \texttt{RRTConnect} and the result
(Fig.~\ref{fig:random-benchmark}) shows that the configuration space
computation can greatly improve the efficiency because our approach
generates samples in a much smaller space. The success rates of both
our planner and OMPL planner are 100\% for most of the tests except
for VTT 6 in Group 2 and VTT 4 in Group 3 in which OMPL cannot solve
the task.

\subsection{Topology Reconfiguration}

\subsubsection{Scenario 1}

The VTT configuration used for this topology reconfiguration example
is shown in Fig.~\ref{fig:topology-reconfig-vtt}. The constraints for
this task are $\overline{L}_{\min}=\SI{1.0}{m}$,
$\overline{L}_{\max}=\SI{5.0}{m}$,
$\bar{\theta}_{\min}=\SI{0.2}{\radian}$, and $\bar{\mu}_{\min}=0.1$.

\begin{figure}[t!]
  \centering
  \includegraphics[width=0.25\textwidth]{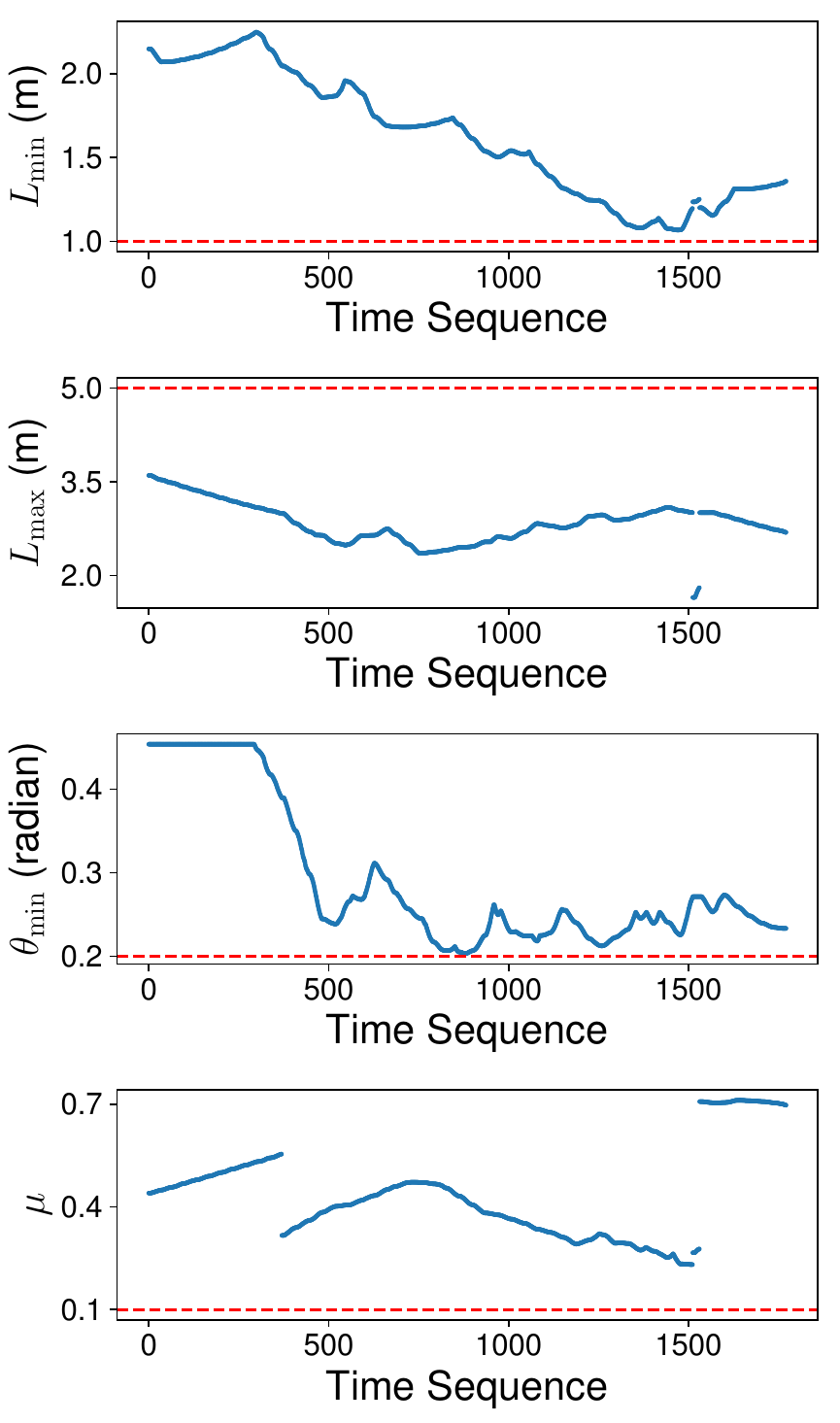}
  \caption{The minimum length ($L_{\min}$) and the maximum length
    ($L_{\max}$) of all moving edge modules, the minimum angle between
    every pair of edge modules ($\theta_{\min}$), and the motion
    manipulability ($\mu$) are measured throughout the topology
    reconfiguration process in
    Fig.~\ref{fig:topology-reconfig-process}.}
  \label{fig:topology-test1-data}
\end{figure}

The motion task is to move $v_5$ from its initial position $q_i^{v_5}$
to a goal position $q_g^{v_5}$
(Fig.~\ref{fig:topology-reconfig-goal}). This motion cannot be
executed with only geometry reconfiguration because
$\mathcal{C}_{\mathrm{free}}^{v_5}(q_i^{v_5})$ and
$\mathcal{C}_{\mathrm{free}}^{v_5}(q_g^{v_5})$ shown in
Fig.~\ref{fig:topology-free-space} are separated by the obstacle
region generated from edge module $(v_3, v_4)$. For this task, the
minimum distance between two nodes is $d_{\min} = \SI{1.0}{m}$ and
$N_{\max}$ is constrained to be greater than or equal to $3$, namely
it is expected to have at least 3 samples for every enclosed subspace,
and the maximum number of iterations $K = 5N_{\max}$.

With our topology reconfiguration planning algorithm
(Algorithm~\ref{alg:topology}), one pair of \texttt{Split} and
\texttt{Merge} actions is sufficient. $v_5$ is moved to a new
location, then split into a pair of nodes ($v^\prime_5$ and $v''_5$)
so that both of these two newly generated nodes can navigate to
$\mathcal{C}_{\mathrm{free}}^{v_5}(q_g^{v_5})$ and merge into a single
node. Then the geometry motion planning is used to plan the motions of
$v^\prime_5$ and $v''_5$ and control them to the target positions for
merging. Finally, merge them back to an individual node and then move
the node to $q_g^{v_5}$. The detailed process is shown in
Fig.~\ref{fig:topology-reconfig-process}. The minimum length
($L_{\min}$) and the maximum length ($L_{\max}$) of all moving edge
modules, the minimum angle between every pair of edge modules
($\theta_{\min}$), and the motion manipulability ($\mu$) are shown in
Fig.~\ref{fig:topology-test1-data}.

This motion task has been solved in~\cite{Liu-vtt-planning-iros-2019}
with the graph search algorithm exploring 8146 VTT configurations with
a more complex action model in order to find a valid sequence of
motion actions. With the proposed framework, the free space of $v_5$
is partitioned into 53 enclosed subspaces and it takes on average
\SI{56.61}{s} to solve this motion task when using the first
transition model with a standard deviation of \SI{5.65}{s} in 1000
trials, including the enclosed subspace computation, topology
reconfiguration planning, and two-node geometry reconfiguration
planning, and the success rate is 100\%. In these trials, the maximum
planning time is \SI{77.88}{s} and the minimum is \SI{38.97}{s}. With
the transition model based on the group free space, the average
planning time for 1000 trials can be as fast as \SI{21.99}{s} with a
standard deviation of \SI{0.67}{s}, and the success rate is 100\%. In
these trials, the maximum planning time is \SI{24.01}{s} and the
minimum is \SI{24.45}{s}.

\begin{figure}[t]
  \centering
  \subfloat[]{\includegraphics[width=0.19\textwidth]{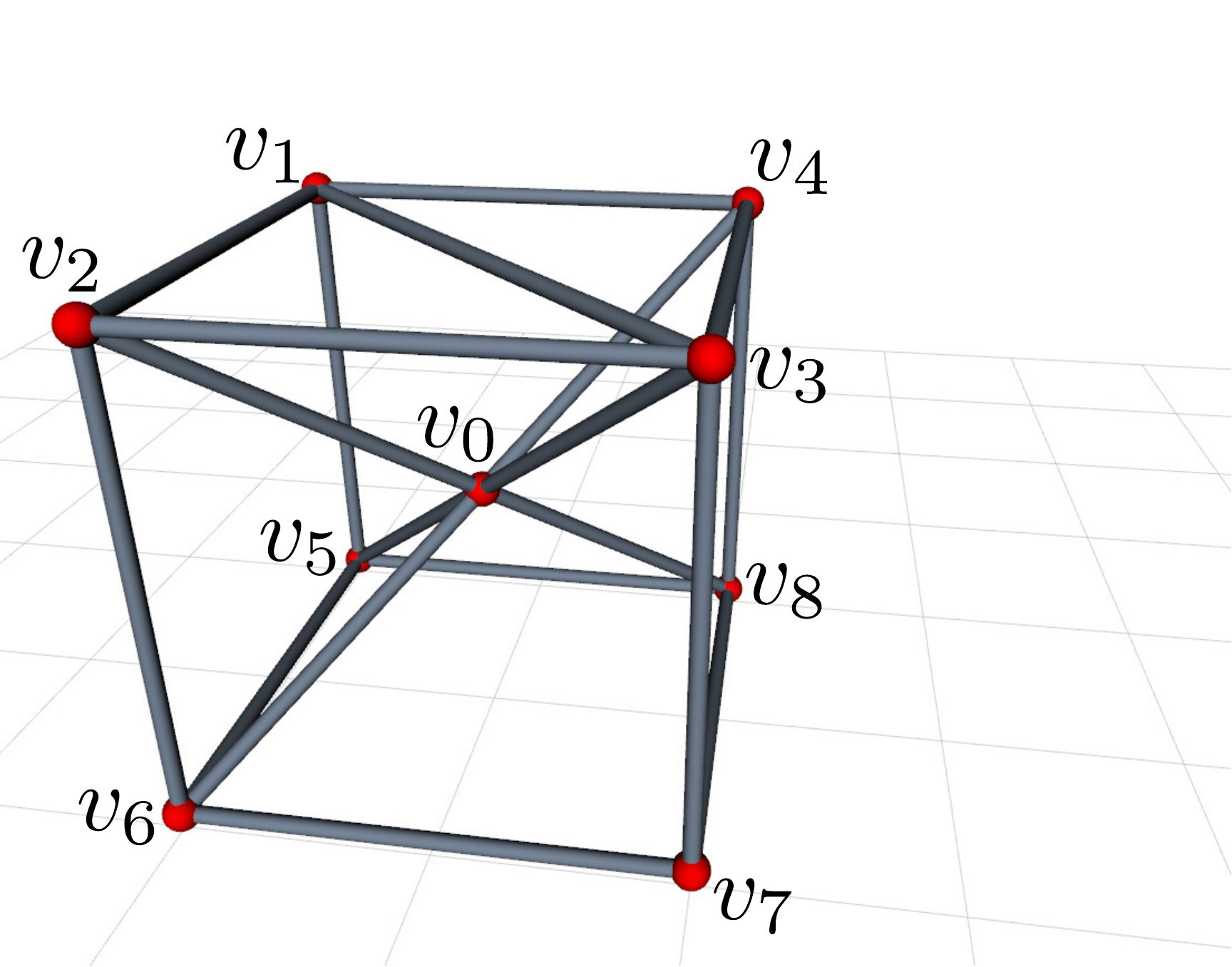}\label{fig:topology-reconfig-init-2}}
  \hfil
  \subfloat[]{\includegraphics[width=0.19\textwidth]{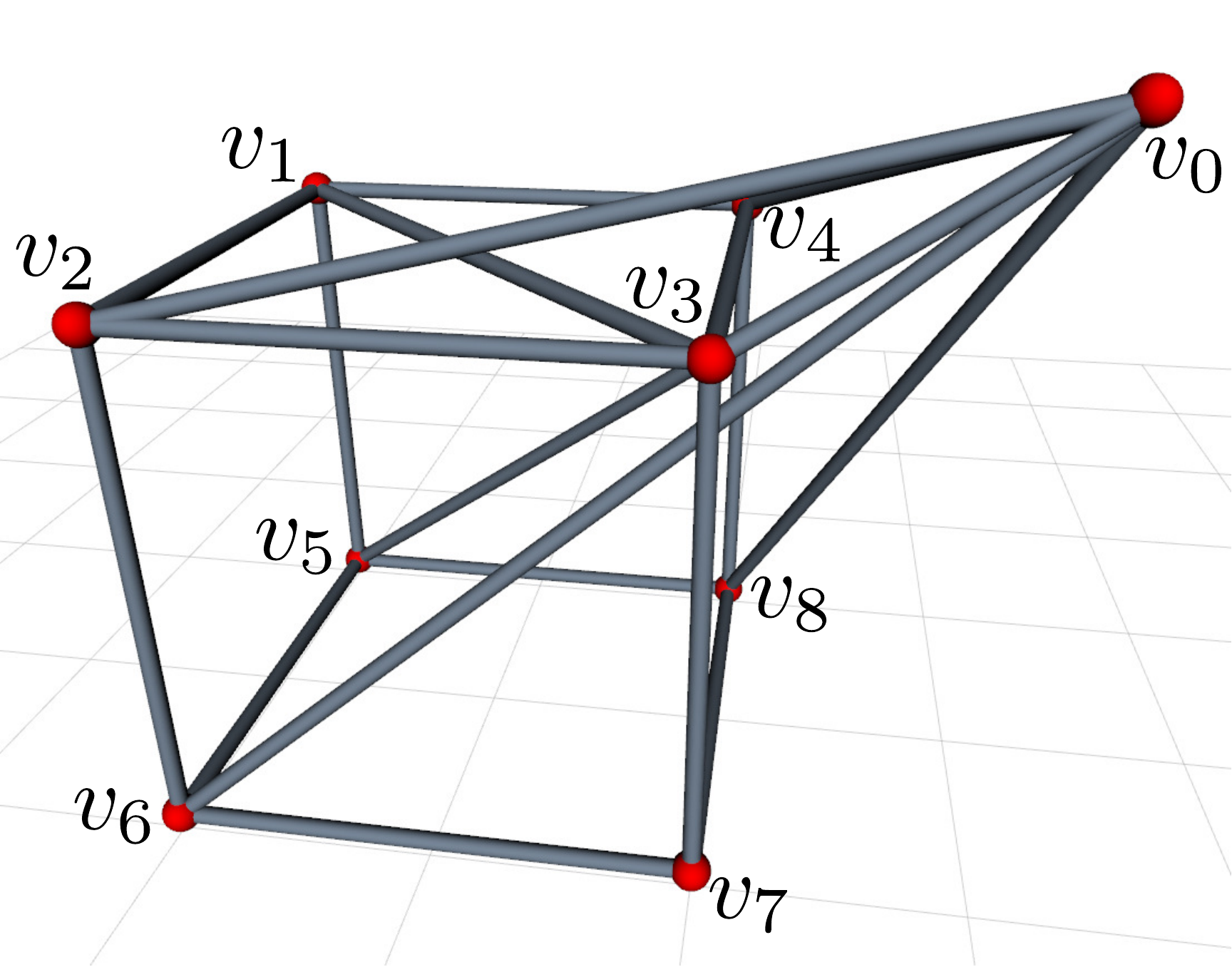}\label{fig:topology-reconfig-goal-2}}
  \caption{(a) A VTT is constructed from 19 members with 9 nodes. (b)
    The task is to move $v_0$ from its initial position to a position
    outside the cubic truss.}
  \label{fig:topology-reconfig-task-2}
\end{figure}

\begin{figure}[t!]
  \centering
  \includegraphics[width=0.22\textwidth]{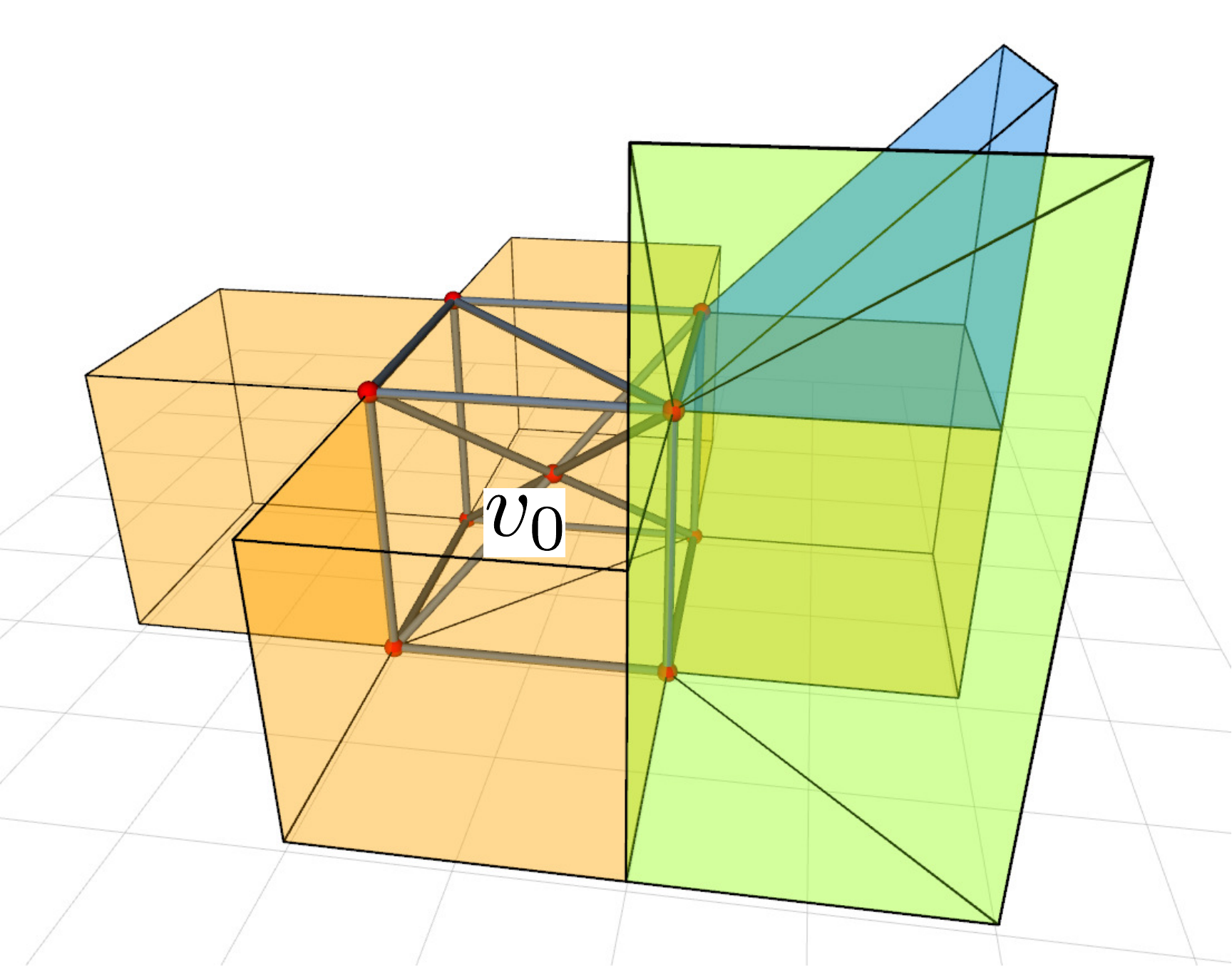}
  \caption{$v$ has to move from $\mathcal{C}_{\mathrm{free}}^v(q_i^v)$
    that is the yellow enclosed subspace to the green enclosed
    subspace, and then $\mathcal{C}_{\mathrm{free}}^v(q_g^v)$ that is
    the blue enclosed subspace.}
  \label{fig:topology-free-space-2}
\end{figure}

\begin{figure*}[b]
  \centering
  \subfloat[]{\includegraphics[width=0.16\textwidth]{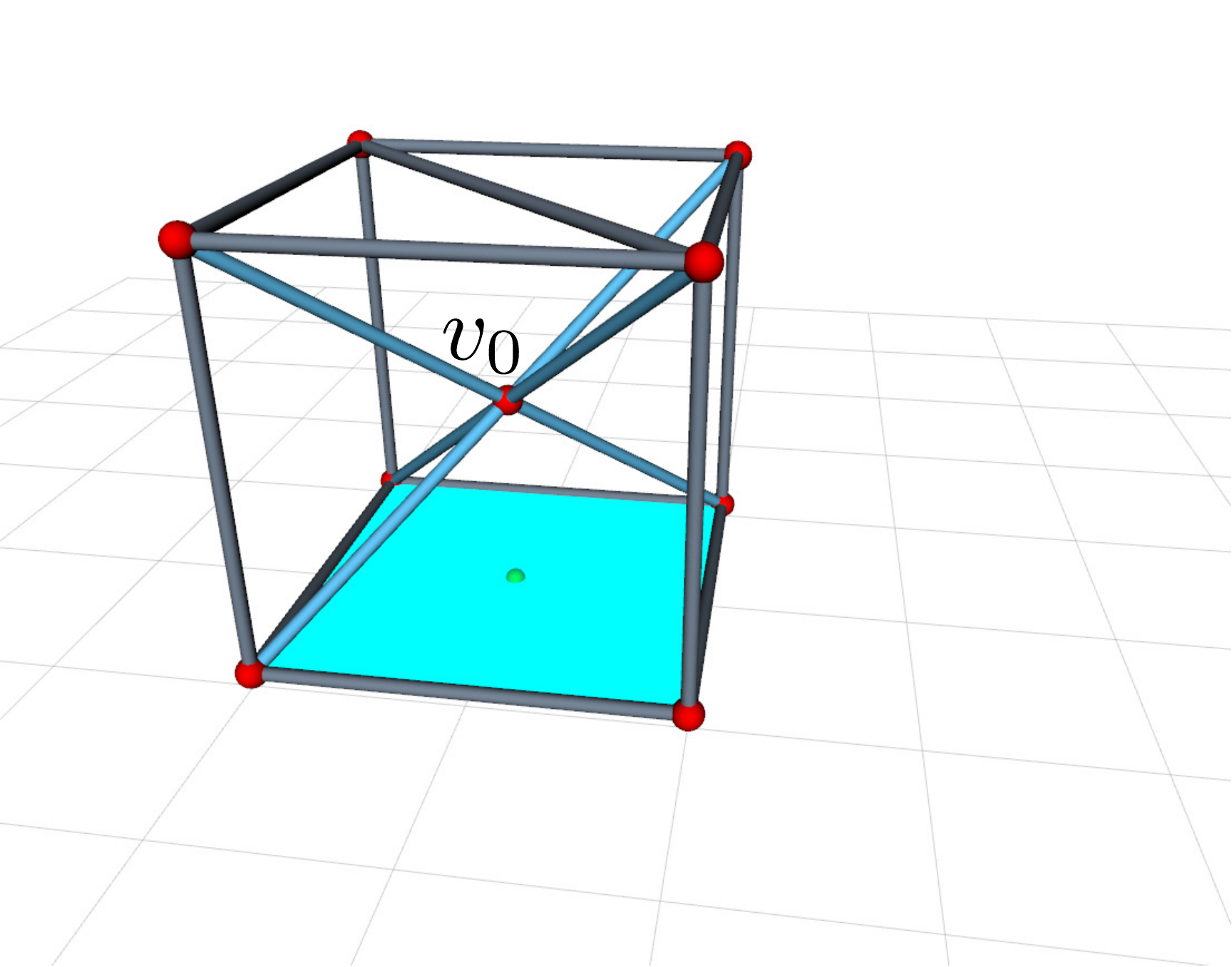}\label{fig:topo-multi-init-move}}
  \hfil
  \subfloat[]{\includegraphics[width=0.16\textwidth]{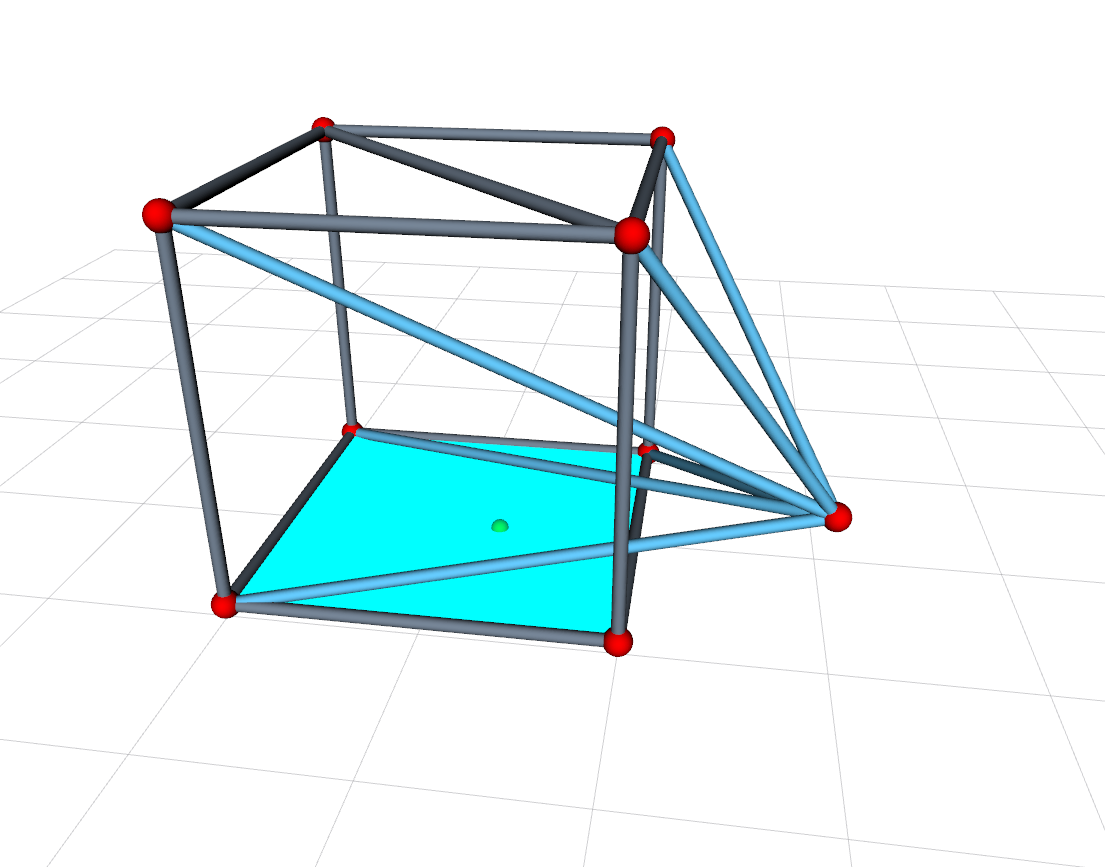}\label{fig:topo-multi-init-move-done}}
  \hfil
  \subfloat[]{\includegraphics[width=0.16\textwidth]{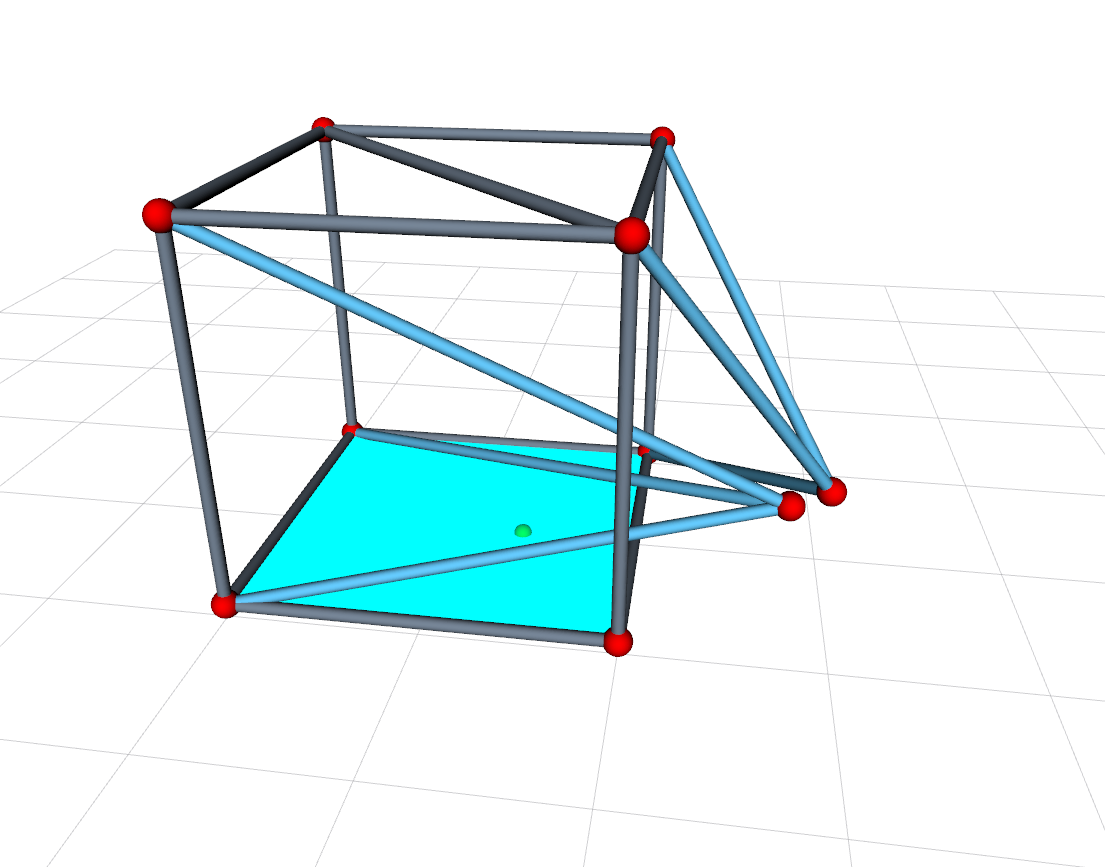}\label{fig:topo-multi-first-split}}
  \hfil
  \subfloat[]{\includegraphics[width=0.16\textwidth]{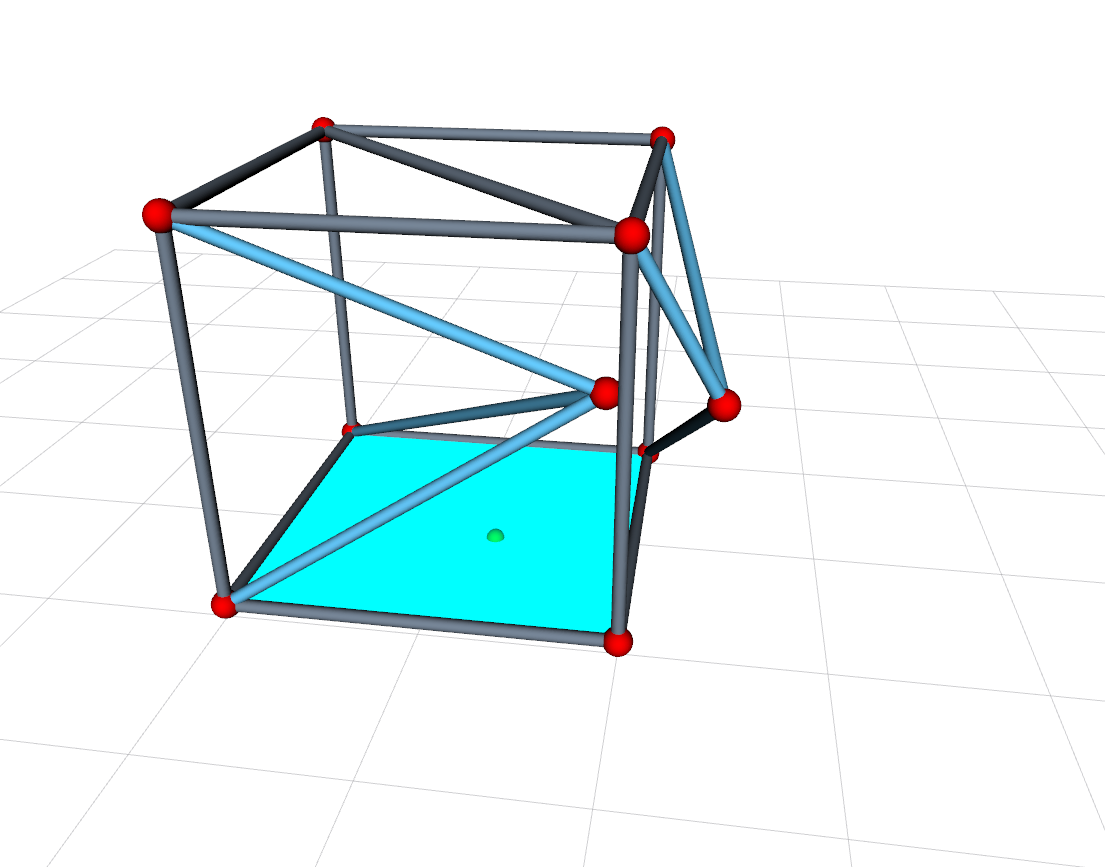}}
  \hfil
  \subfloat[]{\includegraphics[width=0.16\textwidth]{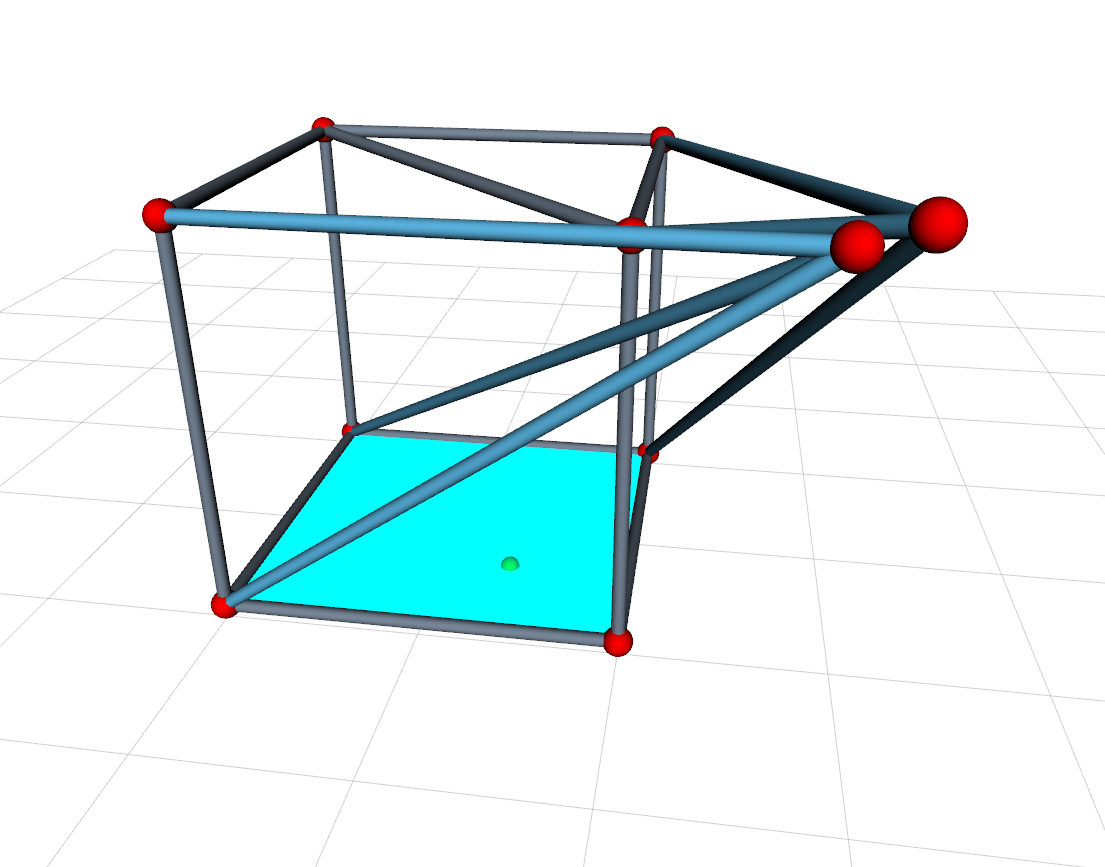}\label{fig:topo-multi-before-first-merge}}
  \hfil
  \subfloat[]{\includegraphics[width=0.16\textwidth]{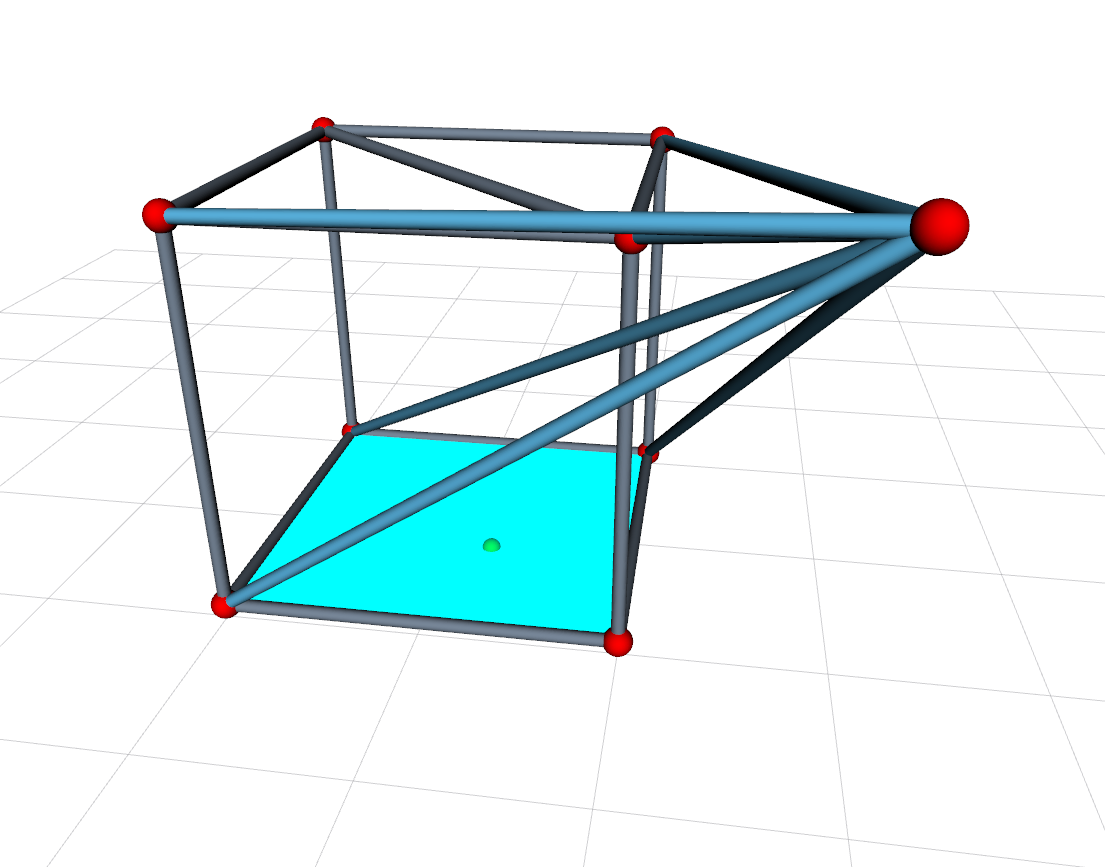}\label{fig:topo-multi-first-merge}}\\
  \hfil
  \subfloat[]{\includegraphics[width=0.16\textwidth]{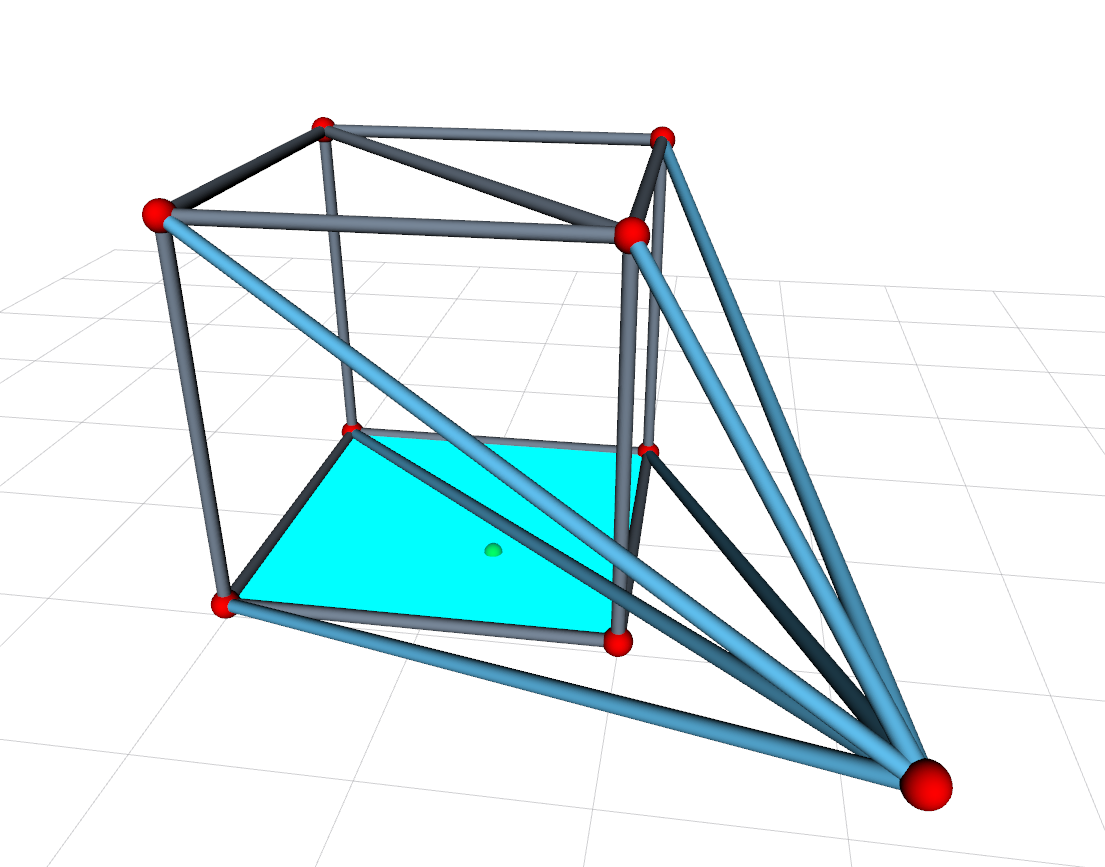}\label{fig:topo-multi-2nd-move}}
  \hfil
  \subfloat[]{\includegraphics[width=0.16\textwidth]{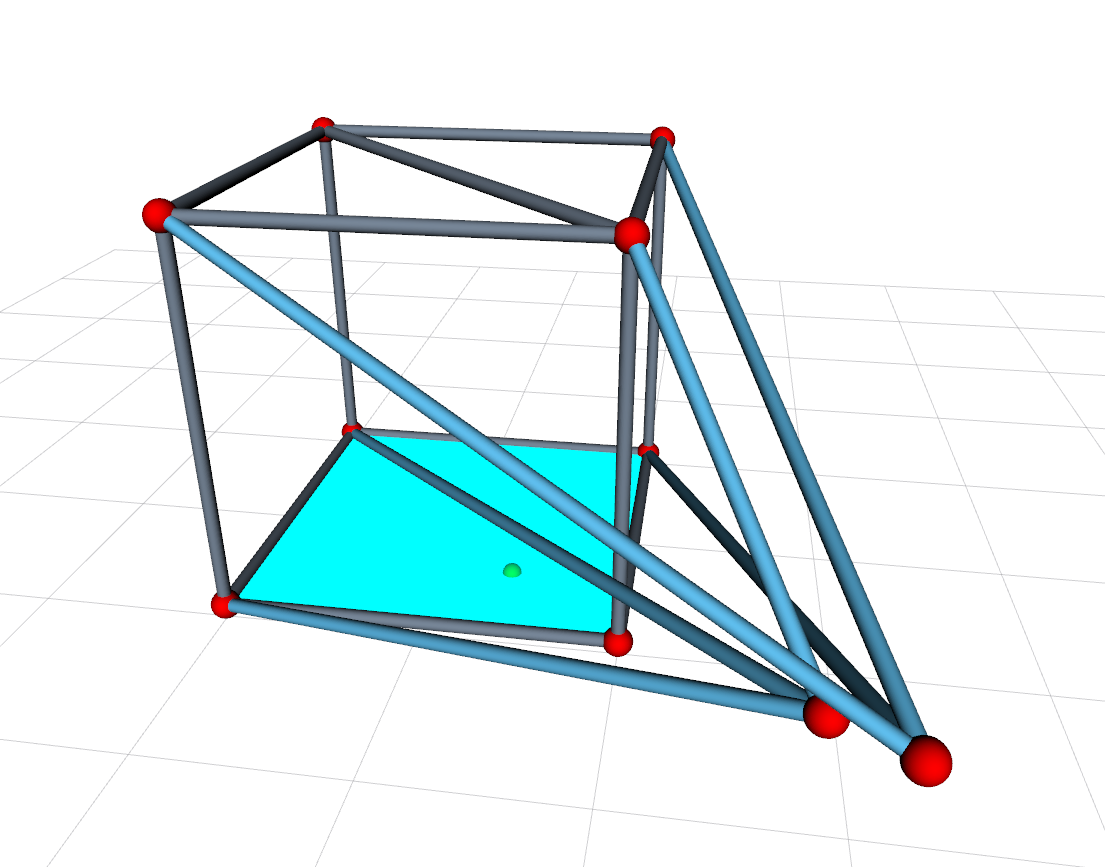}\label{fig:topo-multi-2nd-split}}
  \subfloat[]{\includegraphics[width=0.16\textwidth]{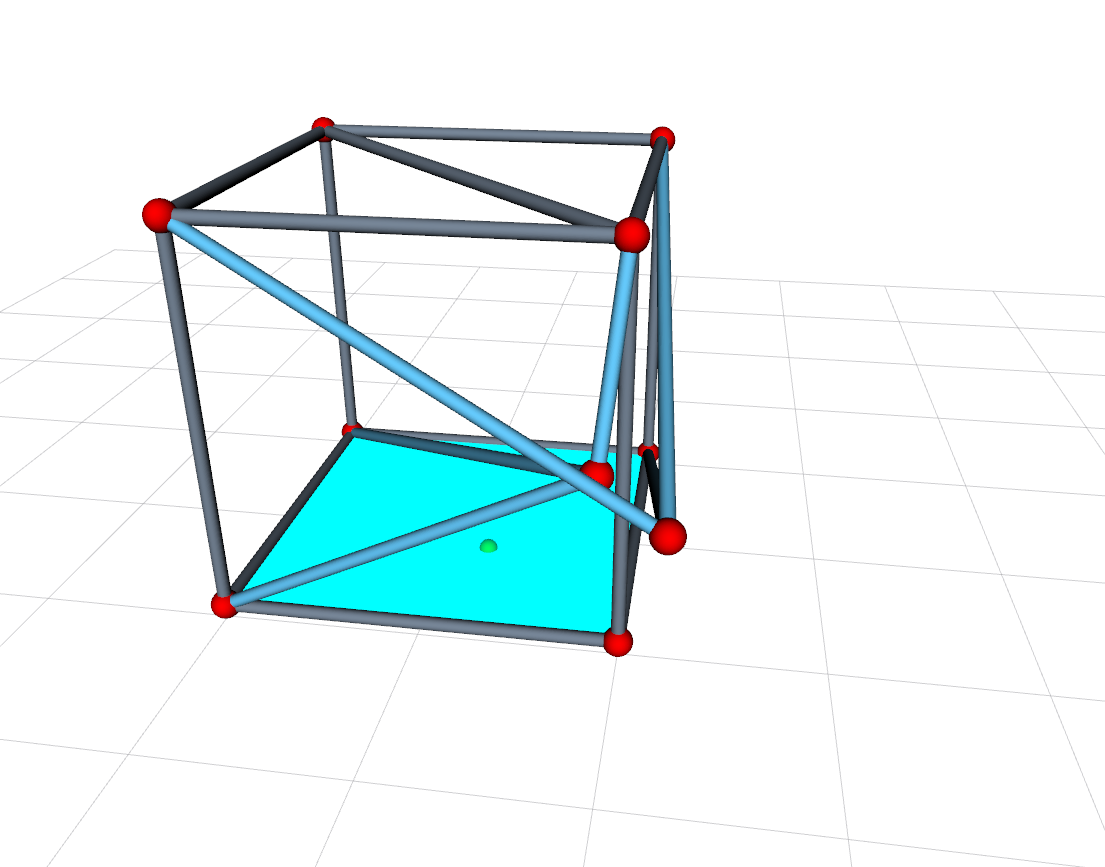}\label{fig:topo-multi-split-move}}
  \hfil
  \subfloat[]{\includegraphics[width=0.16\textwidth]{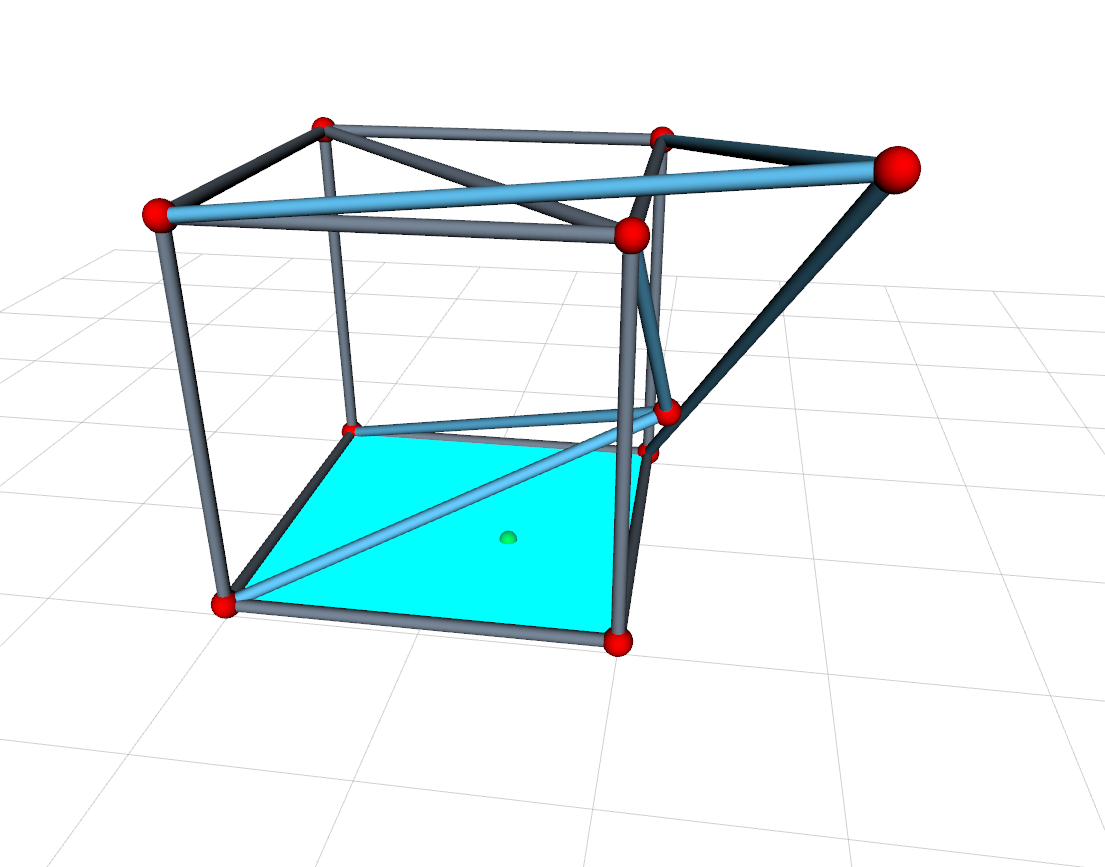}}
  \hfil
  \subfloat[]{\includegraphics[width=0.16\textwidth]{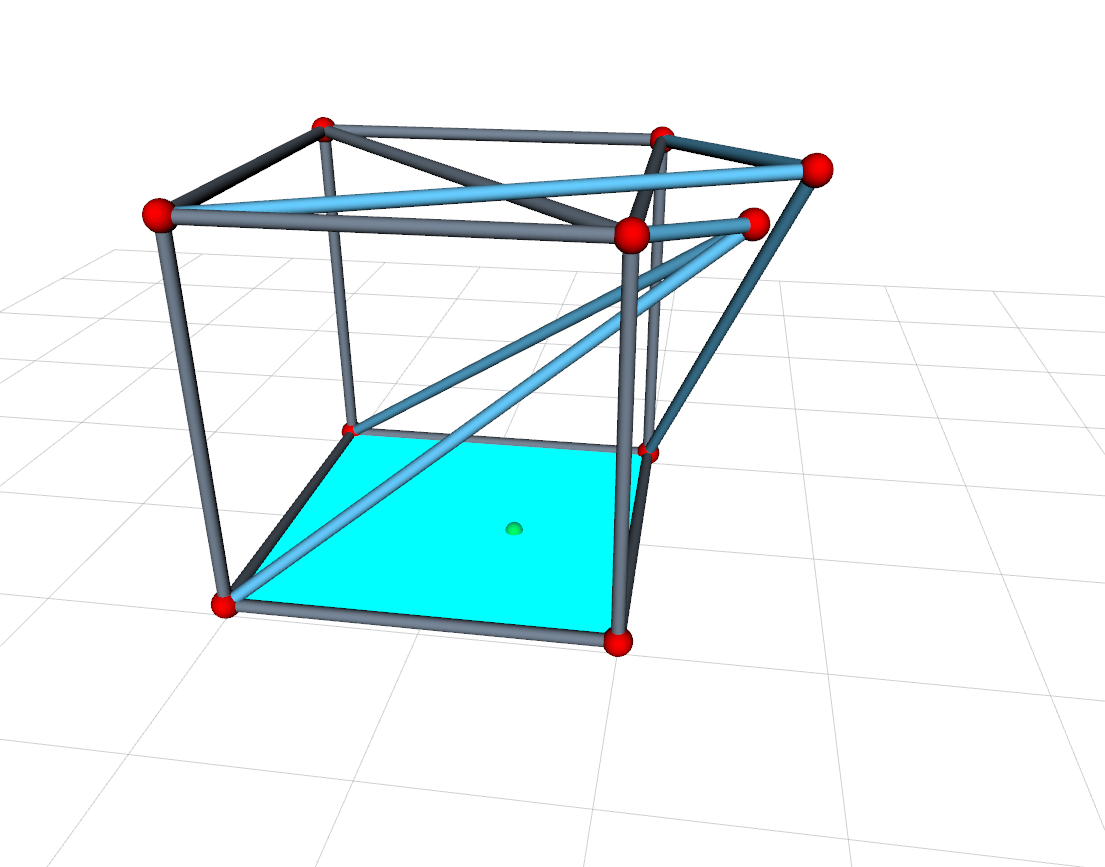}\label{fig:topo-multi-before-2nd-merge}}
  \hfil
  \subfloat[]{\includegraphics[width=0.16\textwidth]{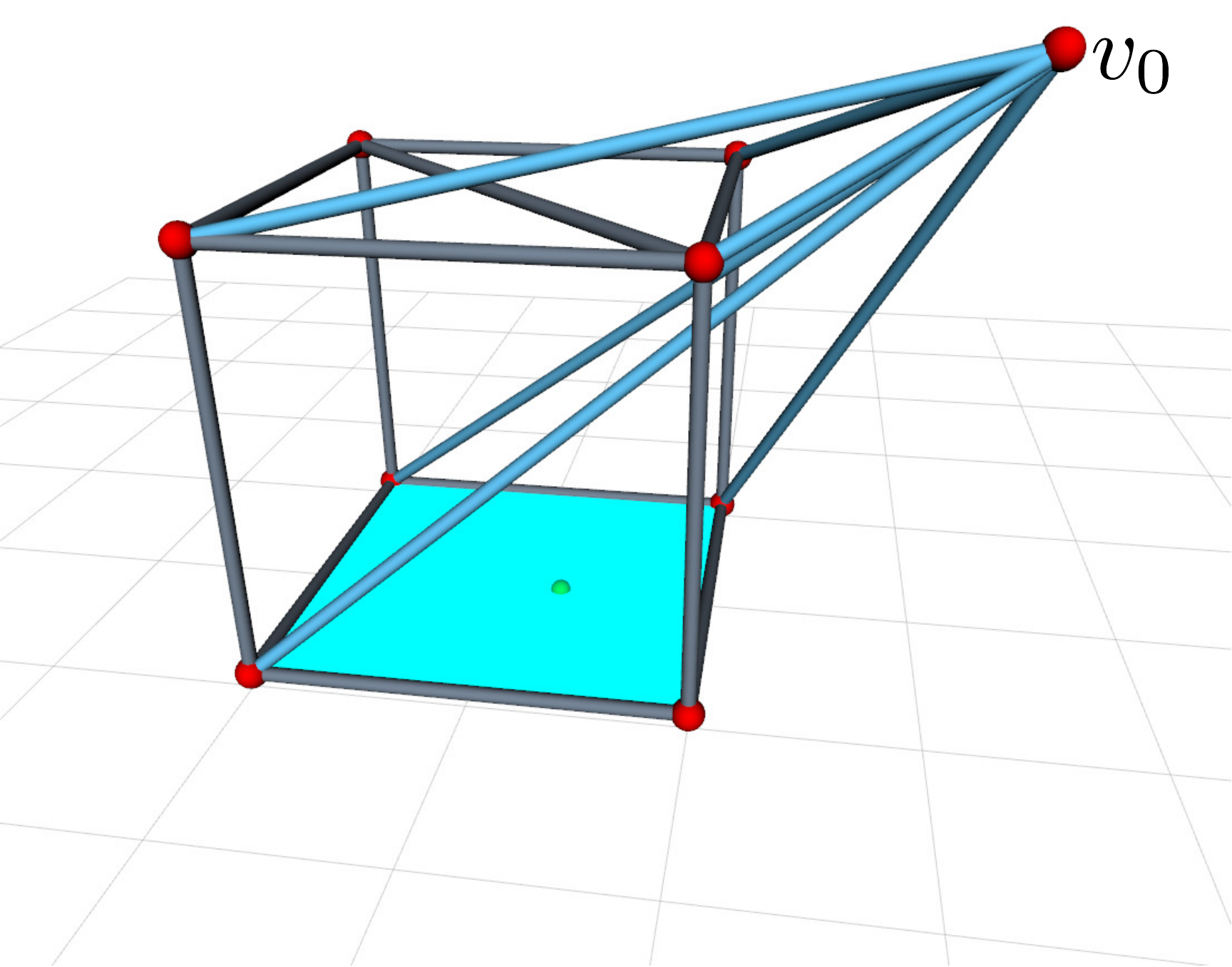}\label{fig:topo-multi-final}}
  \hfil
  \caption{The sequence to move $v_0$ from $q_i^{v_0}$ to $q_g^{v_0}$
    by traversing three enclosed subspaces in
    $\mathcal{C}_{\mathrm{free}}^v$ is shown. The support polygon is
    the aqua region ({\color[rgb]{0,1,1}$\blacksquare$}) and the green
    dot ({\color{green}{$\bullet$}}) is the center of mass represented
    on the ground. (a) --- (b) Move node $v_0$ to traverse the plane
    formed by node $v_3$, $v_4$, $v_7$, and $v_8$. (c) Split the node
    into a pair here so that both newly generated nodes can move
    around edge module $(v_3, v_7)$ while avoiding singular
    configuration. (d) --- (f) Two newly generated nodes are moved to
    a location inside the green enclosed subspace and merge. (g) ---
    (h) Move the merged node to a new location and split it in a
    different way to generate two new nodes. (i) --- (l) One node
    traverses the space inside the cubic truss to go to the blue
    enclosed subspace, and the other node moves upward. Then these two
    nodes merge at a location inside the blue enclosed subspace and
    then move to the target location.}
  \label{fig:topology-reconfig-process-2}
\end{figure*}

\subsubsection{Scenario 2}

Another motion task in which topology reconfiguration actions are
involved is shown in Fig.~\ref{fig:topology-reconfig-task-2} that is
to move $v_0$ from a position $q_i^{v_0}$ inside the cubic truss
(Fig.~\ref{fig:topology-reconfig-init-2}) to a new position
$q_g^{v_0}$ (Fig.~\ref{fig:topology-reconfig-goal-2}). The constraints
for this task are $\overline{L}_{\min}=\SI{0.5}{m}$,
$\overline{L}_{\max}=\SI{5.0}{m}$,
$\bar{\theta}_{\min}=\SI{0.15}{\radian}$, and $\bar{\mu}_{\min}=0.1$.

In this task, $q_i^{v_0}$ and $q_g^{v_0}$ are in two separated
enclosed subspaces, and one solution is to apply topology
reconfiguration actions twice to traverse three enclosed subspaces in
$\mathcal{C}_{\mathrm{free}}^v$ shown in
Fig.~\ref{fig:topology-free-space-2}. For this task, $d_{\min}=0.5$,
$N_{\max}$ is constrained to be greater than or equal to $3$, and
$K=8N_{\max}$.

\begin{figure}[t!]
  \centering
  \includegraphics[width=0.25\textwidth]{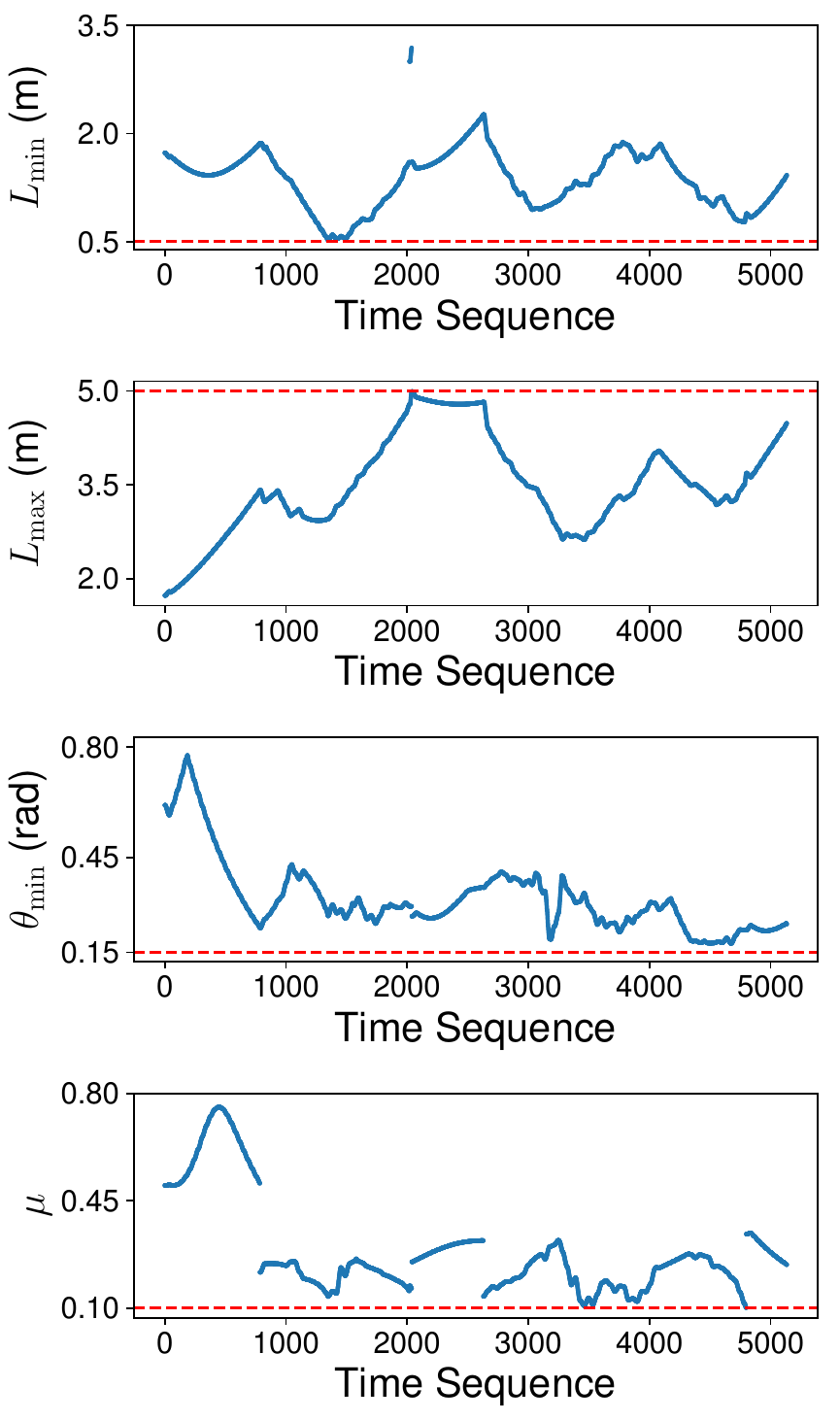}
  \caption{The minimum length ($L_{\min}$) and the maximum length
    ($L_{\max}$) of all moving edge modules, the minimum angle between
    every pair of edge modules ($\theta_{\min}$), and the motion
    manipulability ($\mu$) are measured throughout the topology
    reconfiguration process in
    Fig.~\ref{fig:topology-reconfig-process-2}.}
  \label{fig:topology-test2-data}
\end{figure}

\begin{figure}[t!]
  \centering
  \subfloat[]{\includegraphics[width=0.16\textwidth]{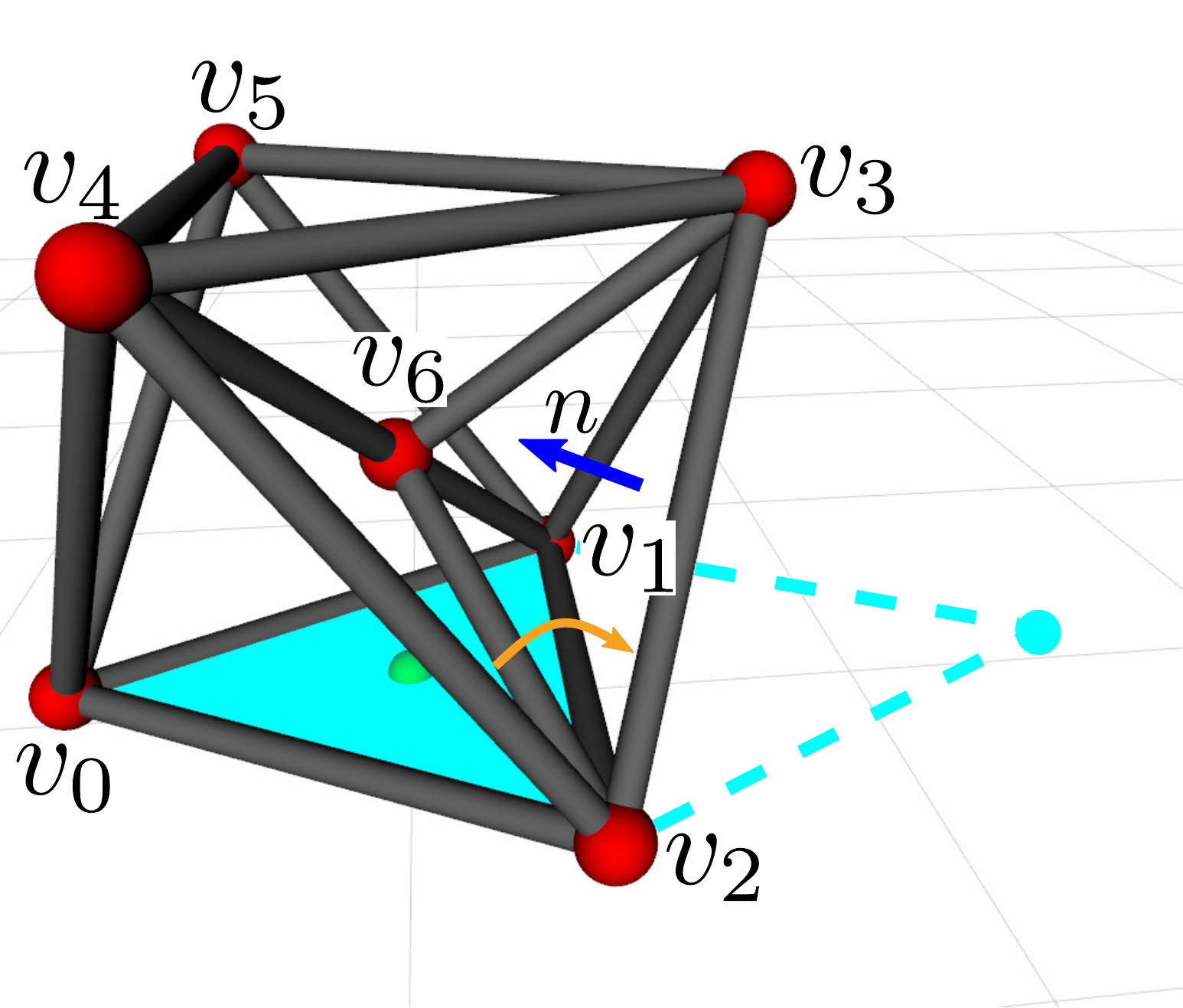}\label{fig:locomotion-test-init}}
  \subfloat[]{\includegraphics[width=0.16\textwidth]{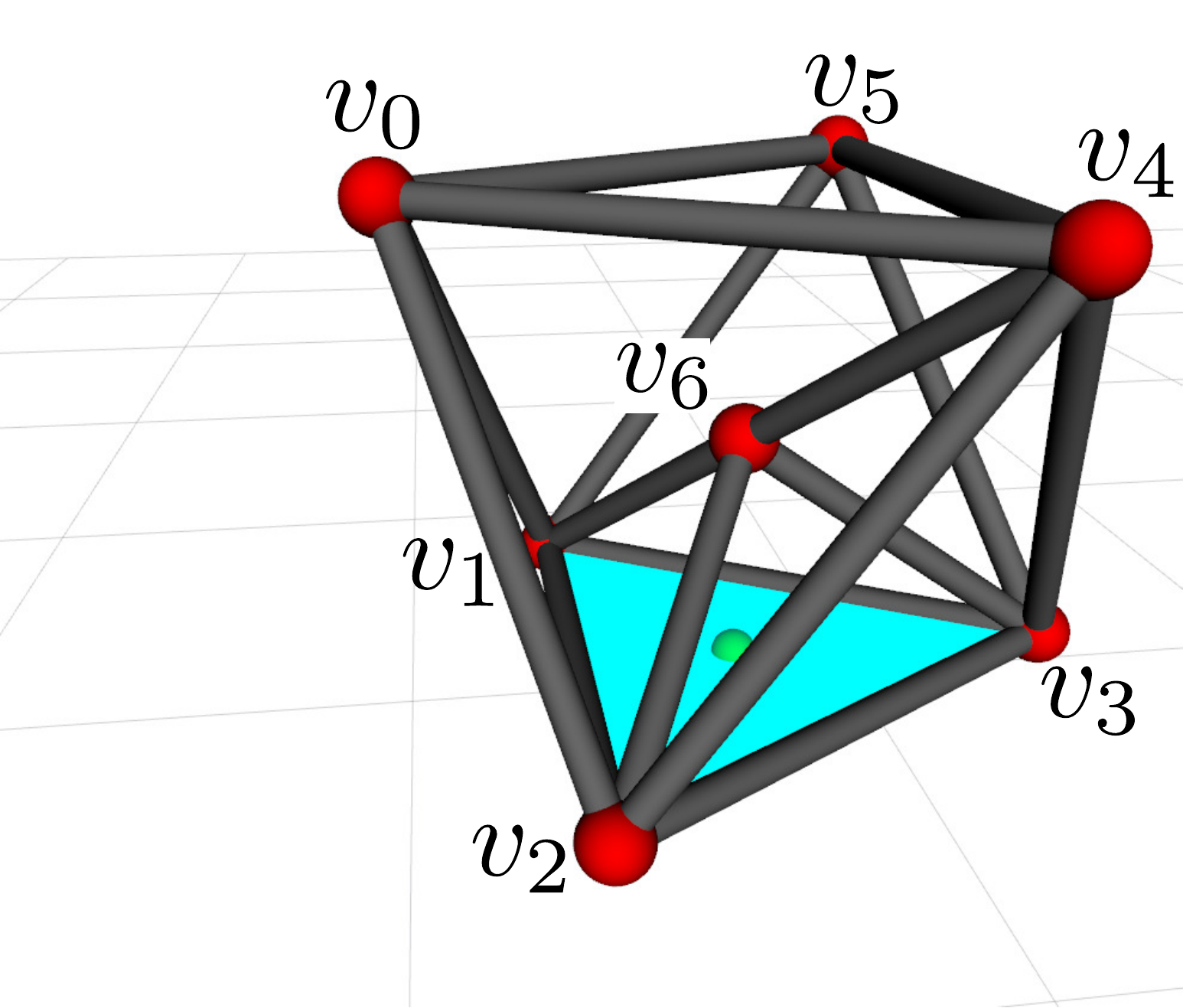}\label{fig:locomotion-test-goal}}
  \caption{The locomotion task is to roll the truss from (a) to (b).}
\end{figure}

\begin{figure*}[t!]
  \centering
  \subfloat[]{\includegraphics[width=0.16\textwidth]{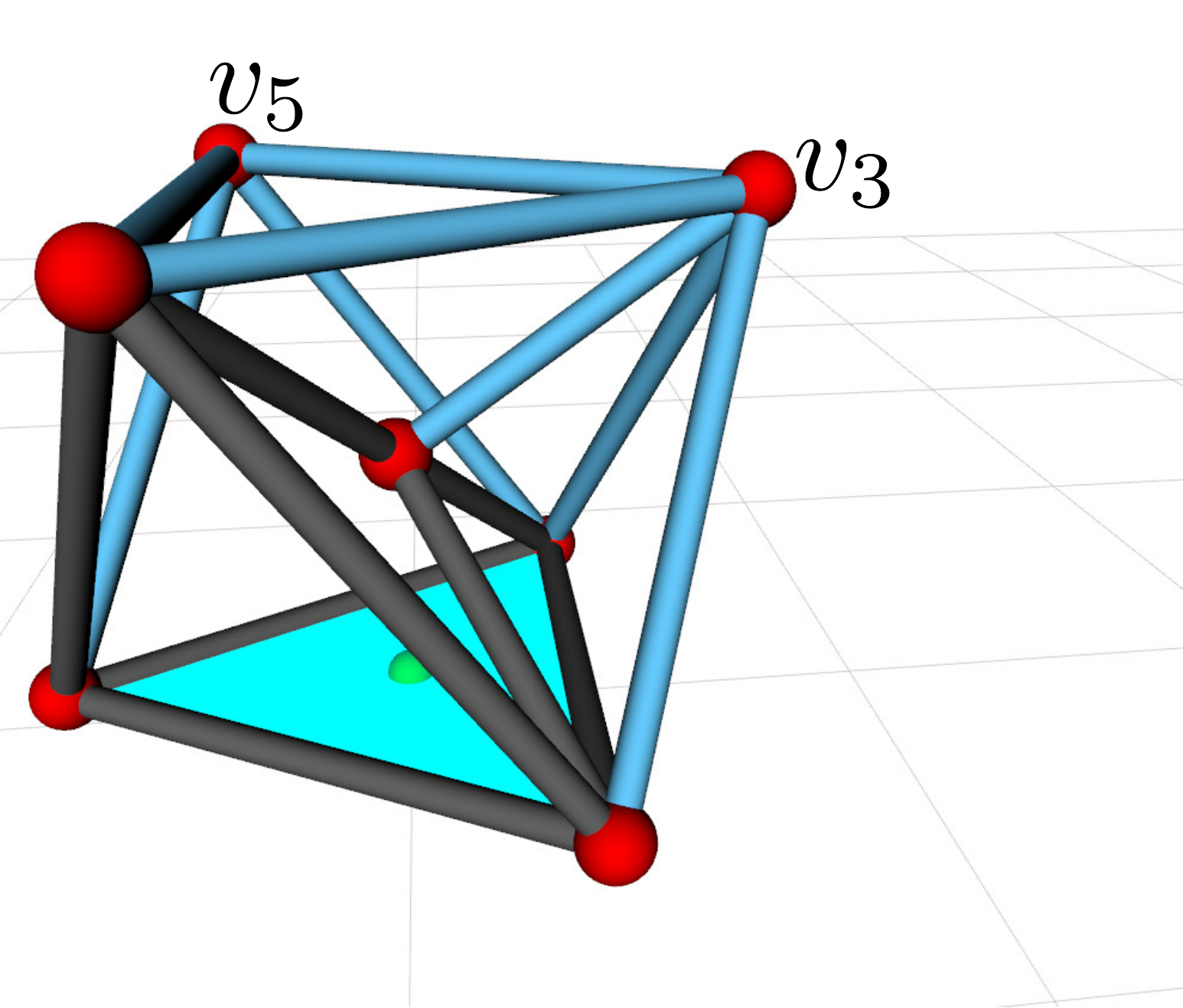}}
  \hfil
  \subfloat[]{\includegraphics[width=0.16\textwidth]{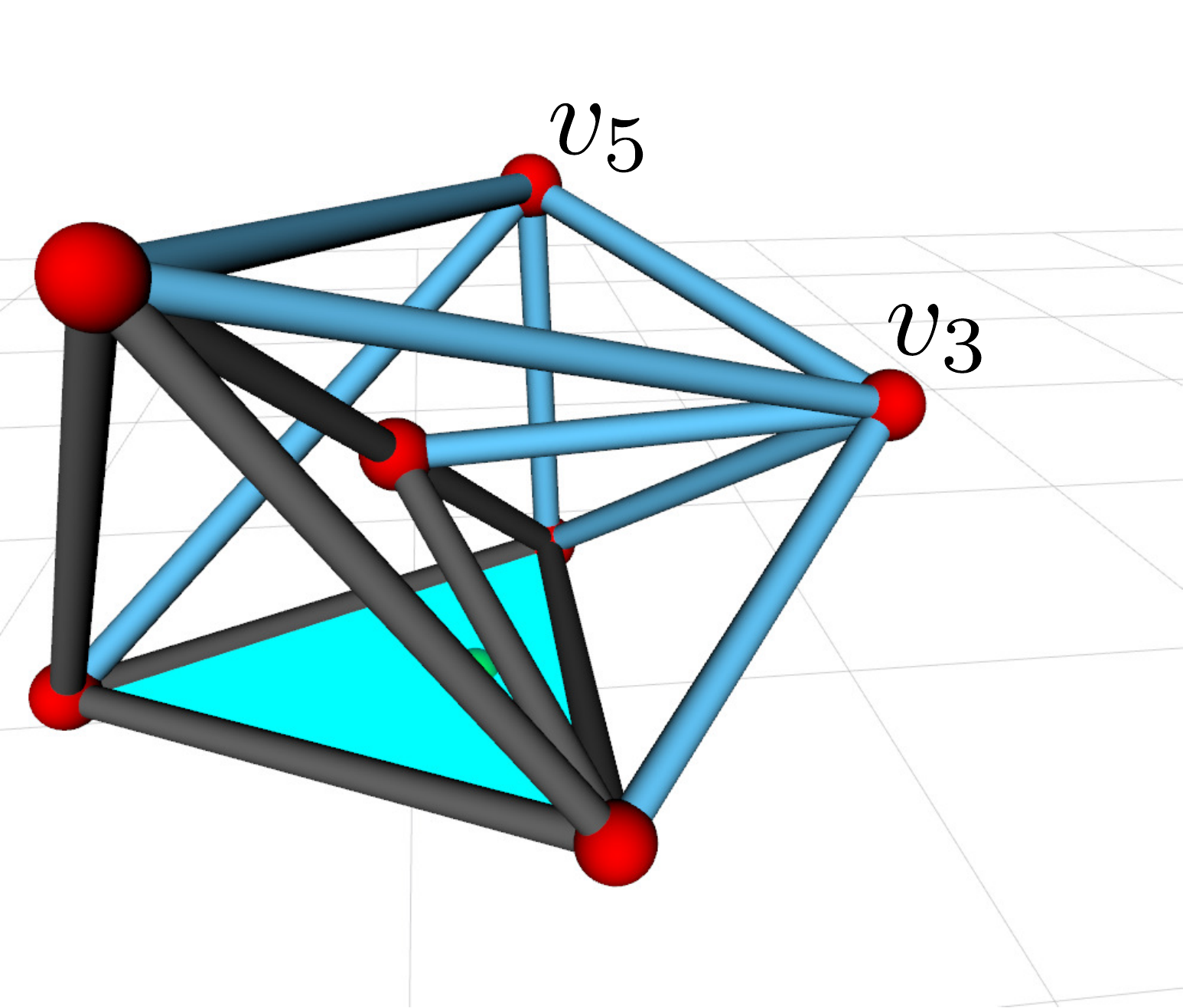}\label{fig:locomotion-test-tipping-start}}
  \hfil
  \subfloat[]{\includegraphics[width=0.16\textwidth]{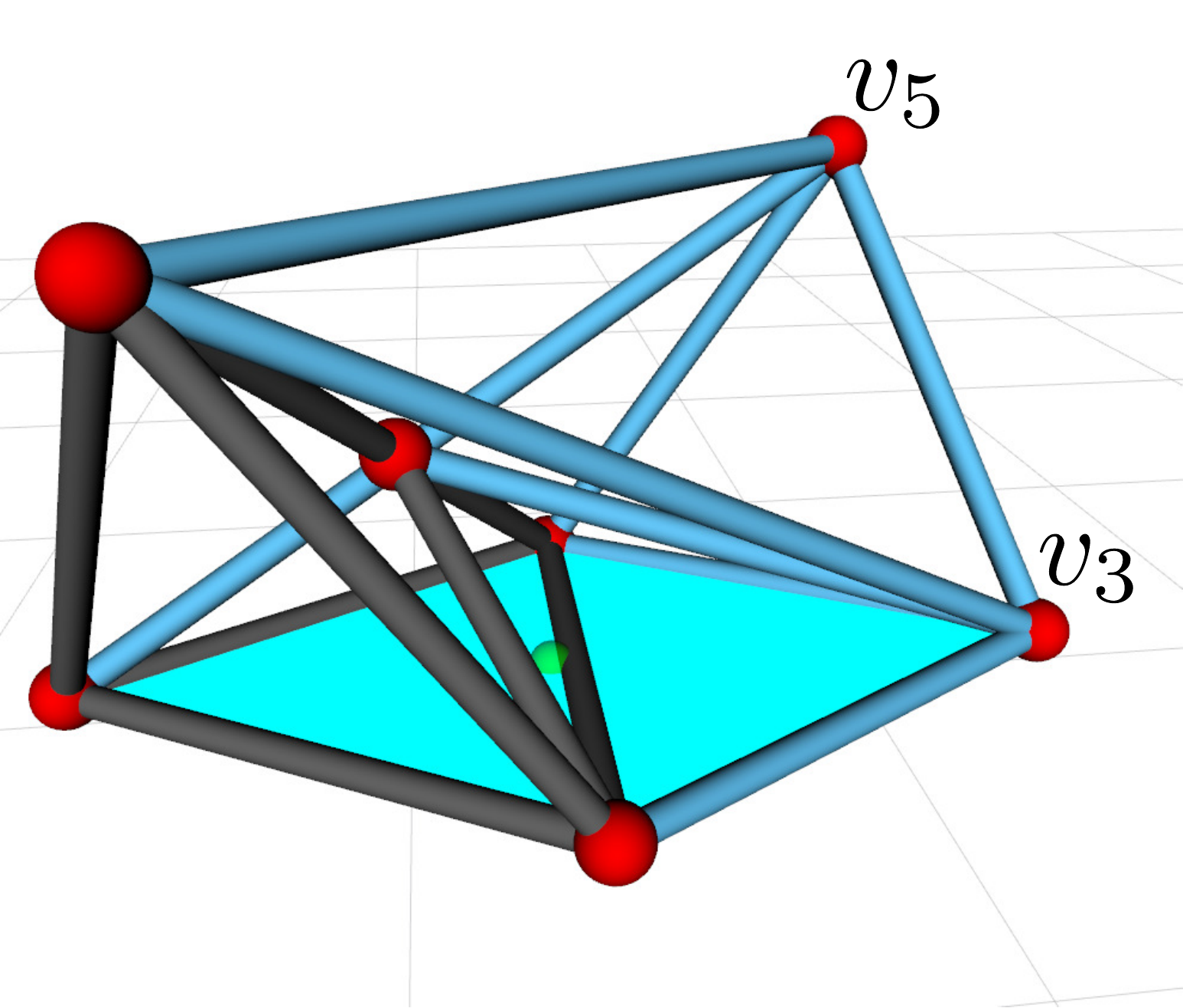}\label{fig:locomotion-test-tipping}}
  \hfil
  \subfloat[]{\includegraphics[width=0.16\textwidth]{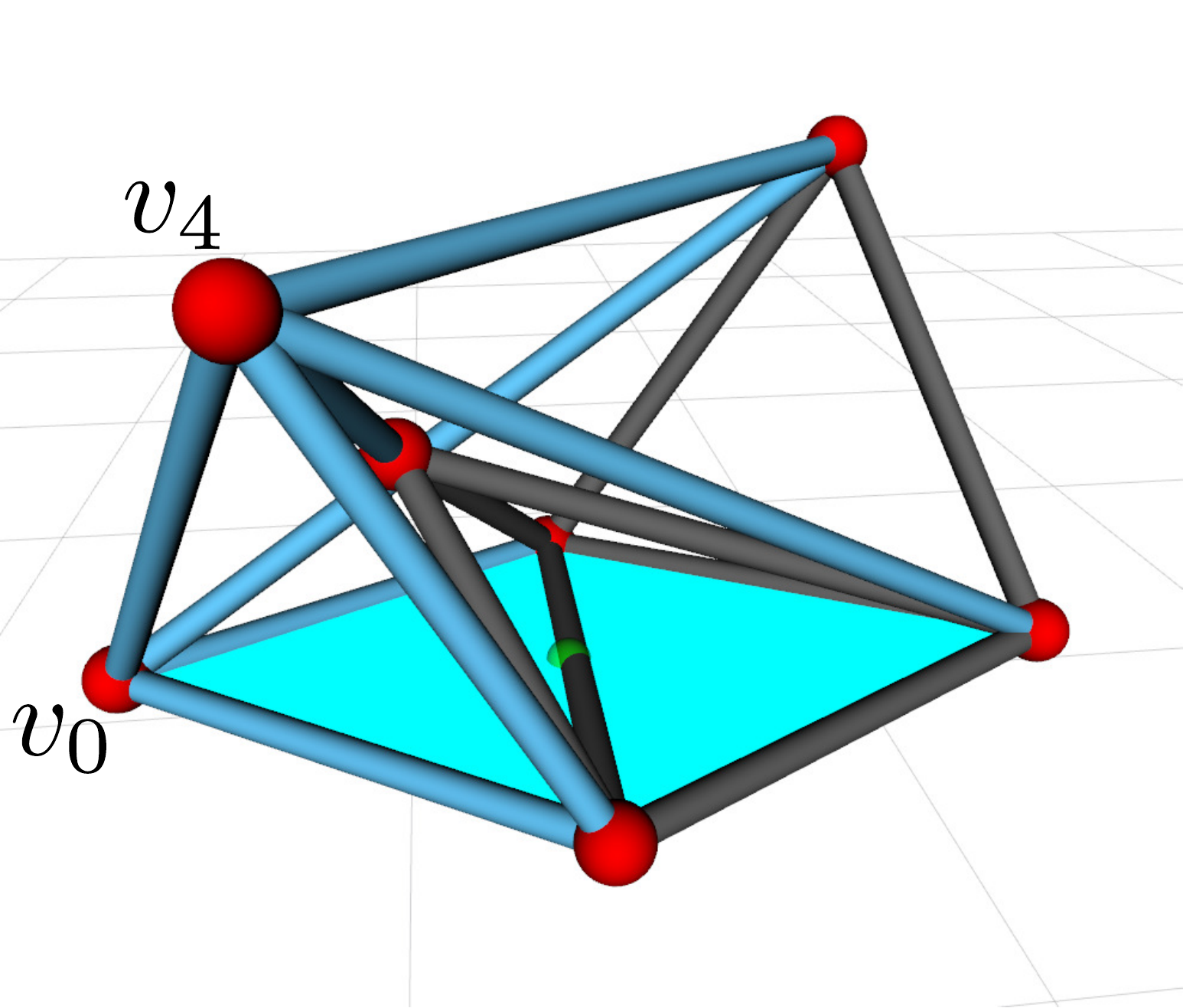}\label{fig:locomotion-test-move-com}}
  \hfil
  \subfloat[]{\includegraphics[width=0.16\textwidth]{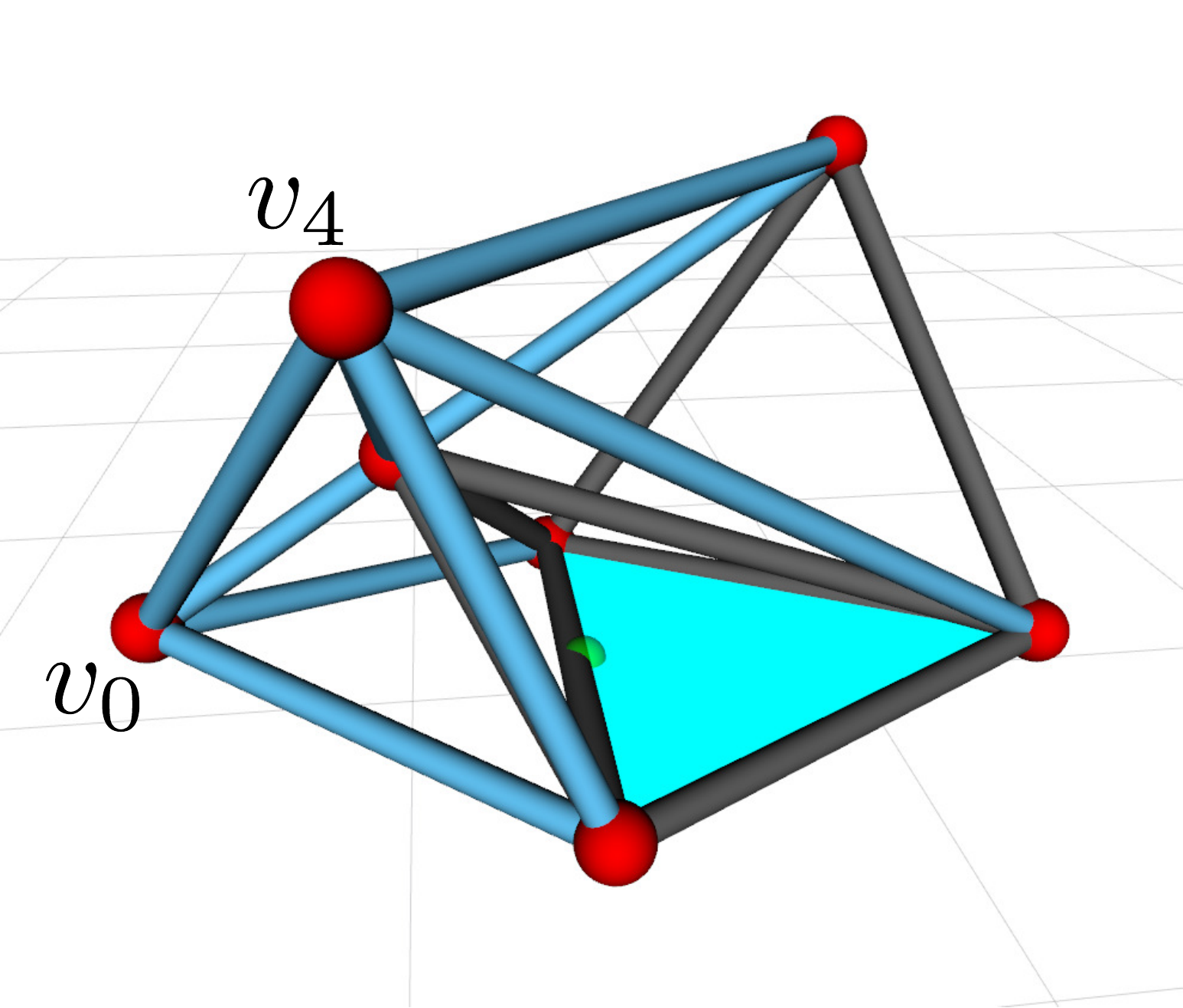}\label{fig:locomotion-test-lift}}\\
  \subfloat[]{\includegraphics[width=0.16\textwidth]{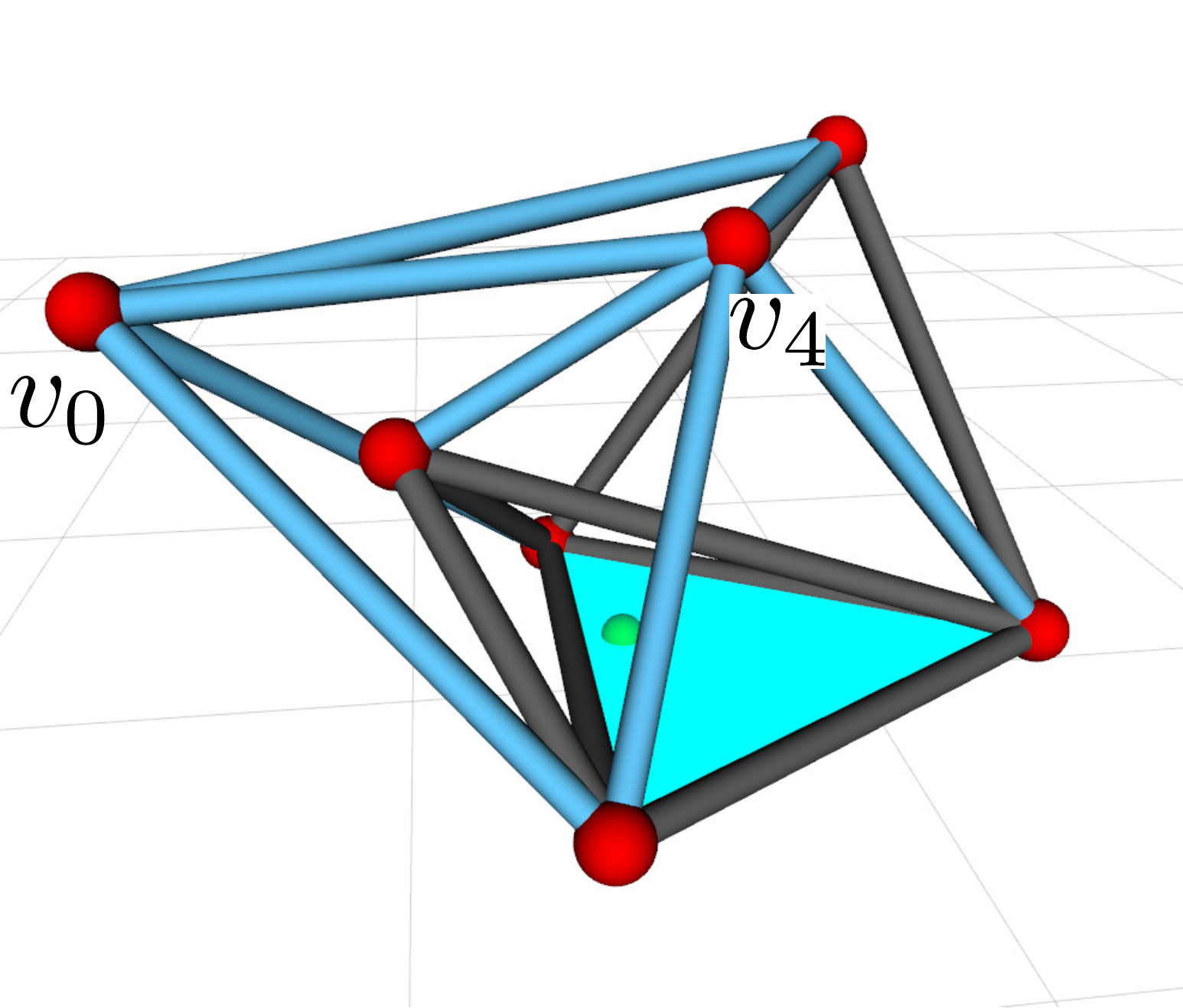}}
  \hfil
  \subfloat[]{\includegraphics[width=0.16\textwidth]{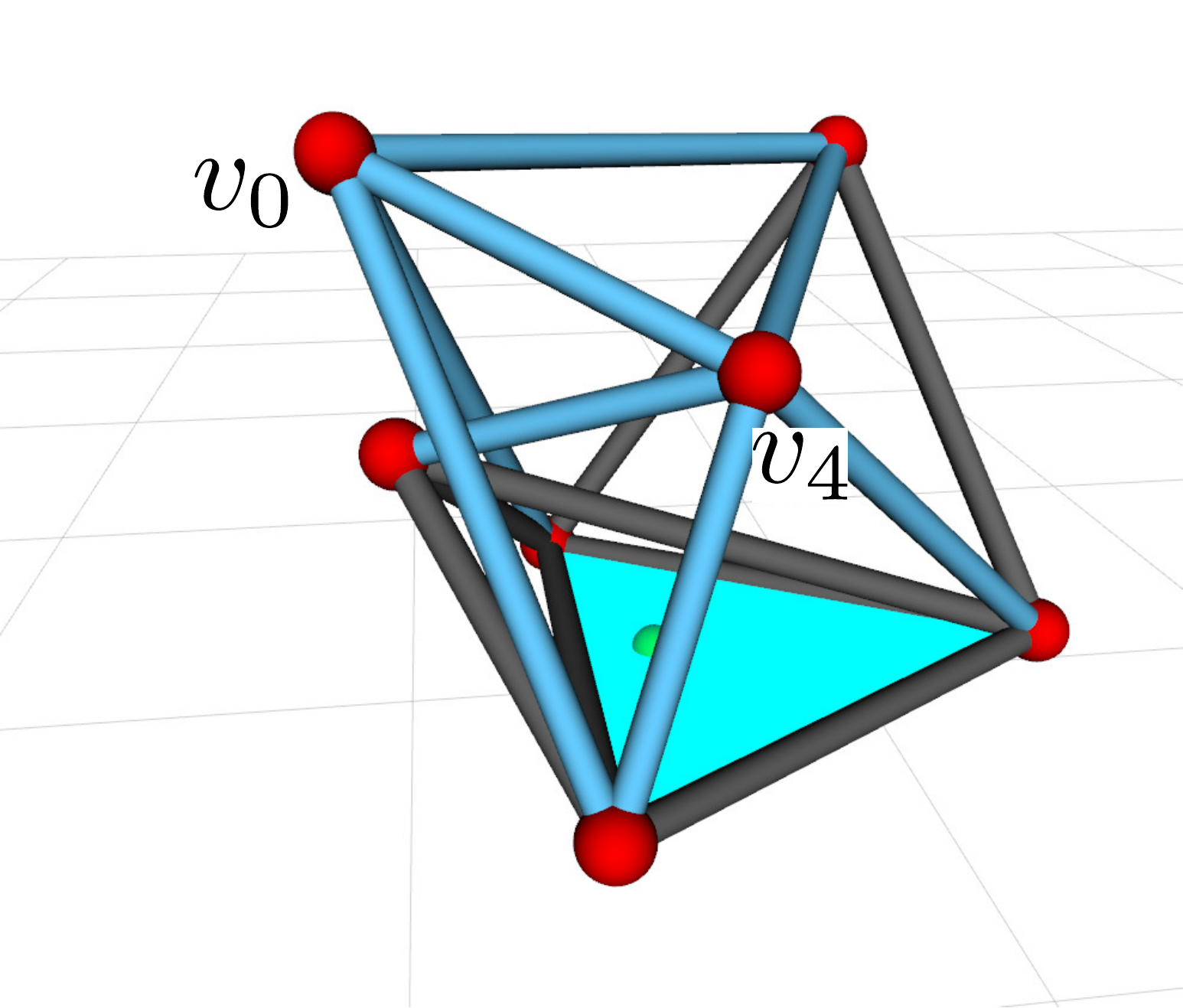}}
  \hfil
  \subfloat[]{\includegraphics[width=0.16\textwidth]{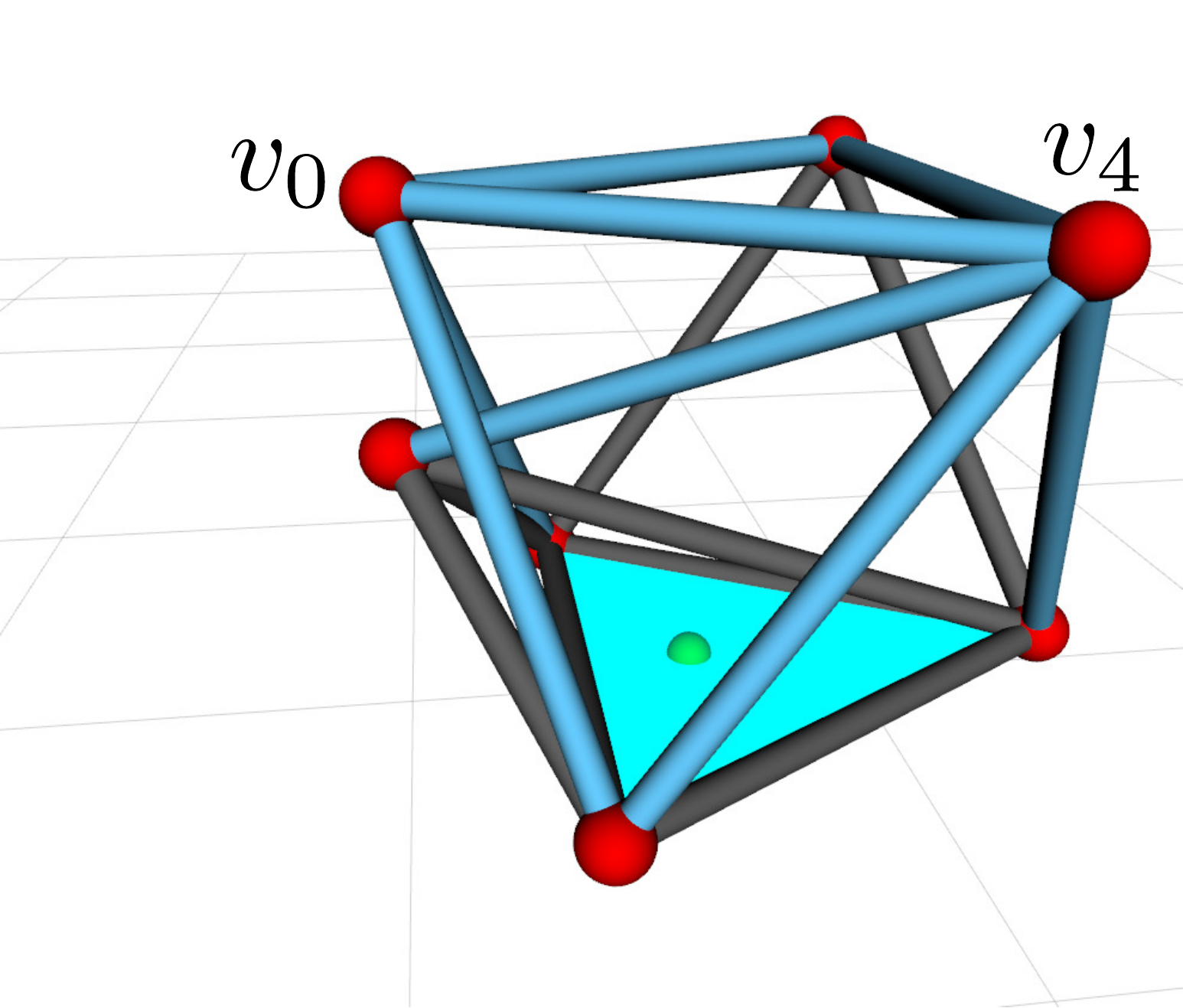}\label{fig:locomotion-test-lift-done}}
  \hfil
  \subfloat[]{\includegraphics[width=0.16\textwidth]{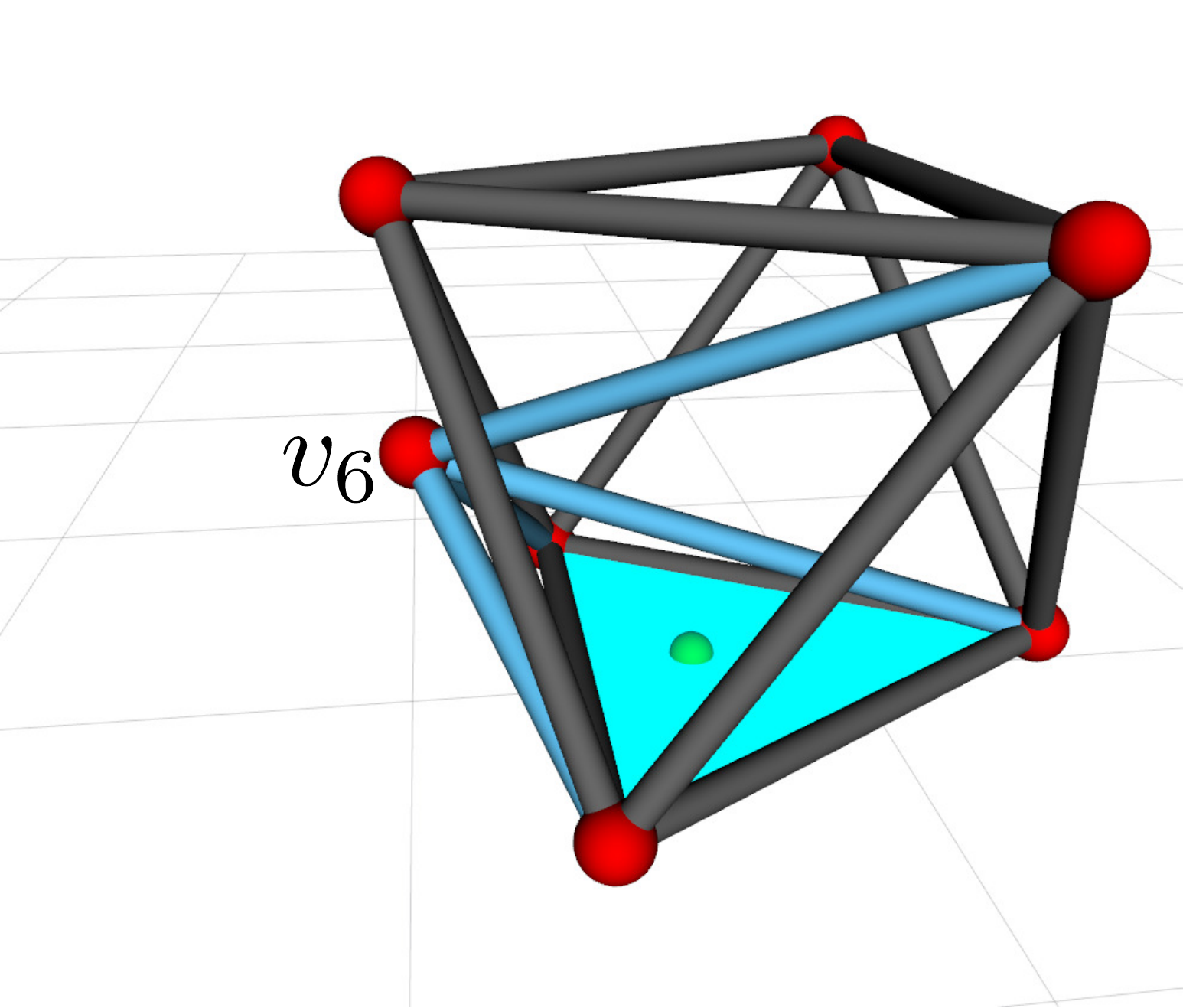}}
  \hfil
  \subfloat[]{\includegraphics[width=0.16\textwidth]{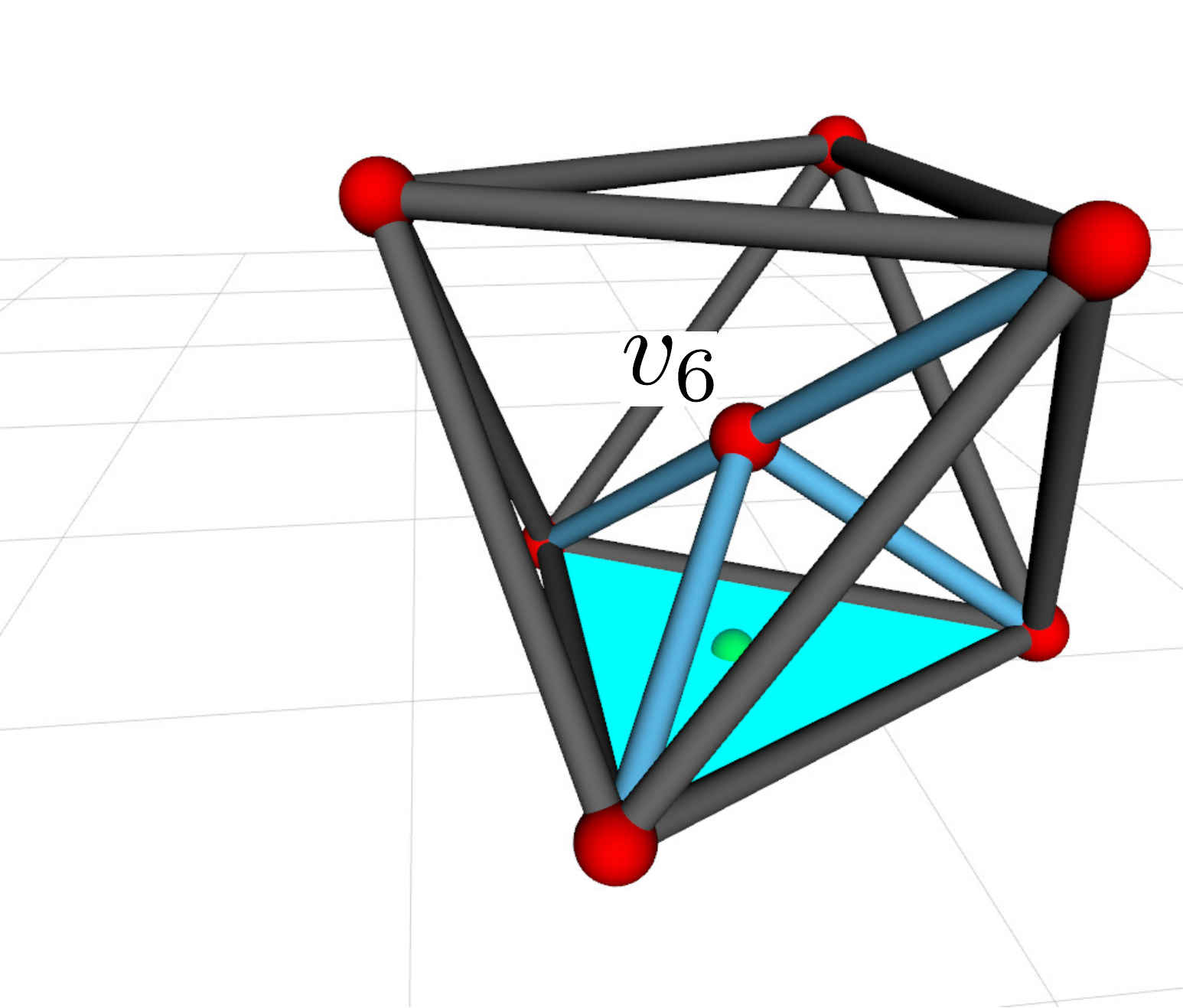}\label{fig:locomotion-test-final}}
  \caption{The planned motion for this locomotion task is shown. (a)
    --- (c) Move node $v_3$ and $v_5$ to first expand the support
    polygon that is the aqua region. (d) Move $v_0$ and $v_4$ to move
    the center of mass represented on the ground
    ({\color{green}{$\bullet$}}) toward the target support
    polygon. (e) --- (h) Once the center of mass represented on the
    ground is inside the target support polygon, lift $v_0$ and move
    both $v_0$ and $v_4$ to their target locations. (i) --- (j)
    Finally move $v_6$ to its target location to finish the locomotion
    process.}
  \label{fig:locomotion-test-process}
\end{figure*}

The detailed planning result is shown in
Fig.~\ref{fig:topology-reconfig-process-2}. We first move the node
$v_0$ to a location outside the cubic truss
(Fig.~\ref{fig:topo-multi-init-move} ---
Fig.~\ref{fig:topo-multi-init-move-done}), and then split it into
two. If the node is split inside the cubic truss, there is no way to
merge them inside the green enclosed subspace
(Fig.~\ref{fig:topo-multi-first-merge}) because one node has to
traverse a region formed by node $v_3$, $v_4$, $v_7$, and $v_8$
resulting in singular configurations. After this \texttt{Split}
action, do geometry reconfiguration planning to control them to the
green enclosed subspace (Fig.~\ref{fig:topo-multi-first-split} ---
Fig.~\ref{fig:topo-multi-before-first-merge}) and merge them
back. Then move the node to a different location inside the green
enclosed subspace (Fig.~\ref{fig:topo-multi-2nd-move}) and then split
the node in a different way (Fig.~\ref{fig:topo-multi-2nd-split}) in
order to navigate these two nodes to
$\mathcal{C}_{\mathrm{free}}^v(q_g^{v_0})$
(Fig.~\ref{fig:topo-multi-split-move} ---
Fig.~\ref{fig:topo-multi-before-2nd-merge}) and merge them
back. Eventually move the node to $q_g^{v_0}$
(Fig.~\ref{fig:topo-multi-final}). The minimum length ($L_{\min}$) and
the maximum length ($L_{\max}$) of all moving edge modules, the
minimum angle between every pair of edge modules ($\theta_{\min}$),
and the motion manipulability ($\mu$) are shown in
Fig.~\ref{fig:topology-test2-data}. Using the strict transition model,
the average planning time is \SI{319.83}{s} with a standard deviation
being \SI{21.31}{s} in 1000 trials, and the success rate is 74\%. In
these trials, the maximum planning time is \SI{392.47}{s} and the
minimum is \SI{172.05}{s}. The search space is larger and more samples
are generated in order to find the sequence of topology
reconfiguration actions which consumes more time. All the failures are
from the graph search phase, and more samples can solve this issue but
also make the search process more time consuming. With the same
parameters, when using the other transition model based on the group
free space, the average planning time is \SI{86.48}{s} with a standard
deviation being \SI{2.05}{s} in 1000 trials, and the success rate is
95.9\%. The maximum planning time is \SI{92.85}{s} and the minimum
planning time is \SI{79.98}{s}. The performance is improved a lot. All
the failures in these trials are from the path planning for a group of
node, namely even the transition model between two enclosed subspace
is valid, it is not guaranteed for the planner to find the
solution. Similar to the geometry reconfiguration test, it also takes
more time for the topology reconfiguration planning compared with our
previous conference paper. This is because the planner needs to
validate \texttt{Split} actions and generate enough samples to span
the space, and the transition model is also more complex.

\subsection{Locomotion}
\label{sec:locomotion-test}

The VTT used for the locomotion test is shown in
Fig.~\ref{fig:locomotion-test-init} that is an octahedron with three
additional edge modules internally. The constraints for this
locomotion task are $\overline{L}_{\min} = \SI{0.3}{m}$,
$\overline{L}_{\max}=\SI{2.3}{m}$,
$\bar{\theta}_{\min}=\SI{0.3}{\radian}$, and
$\bar{\mu}_{\min}=0.1$. The task is to roll this VTT to an adjacent
support polygon shown in Fig.~\ref{fig:locomotion-test-goal}, and the
locomotion command is $(v_1, v_2)$. Five nodes ($v_0$, $v_3$, $v_4$,
$v_5$, and $v_6$) are involved in this process and their goal
locations can be computed by Eq.~(\ref{eq:des-pos-locomotion}).

The detailed locomotion process is shown in
Fig.~\ref{fig:locomotion-test-process}. Five nodes are divided into
three groups and the whole process is also divided into three phases
automatically. In the first phase, $v_3$ and $v_5$ are moved to expand
the support polygon (Fig.~\ref{fig:locomotion-test-tipping-start} and
Fig.~\ref{fig:locomotion-test-tipping}). Then $v_0$ and $v_4$ are
moved gradually to move the center of mass toward the adjacent support
polygon shown in Fig.~\ref{fig:locomotion-test-move-com}. Once the
center of mass represented on the ground is inside the target support
polygon, $v_0$ can be lifted off the ground and both $v_0$ and $v_4$
can be moved to their target locations
(Fig.~\ref{fig:locomotion-test-lift}---Fig.~\ref{fig:locomotion-test-lift-done}). Finally,
move node $v_6$ to its target location shown in
Fig.~\ref{fig:locomotion-test-final} to finish the whole process.

\begin{figure}[t!]
  \centering
  \includegraphics[draft=false,width=0.25\textwidth]{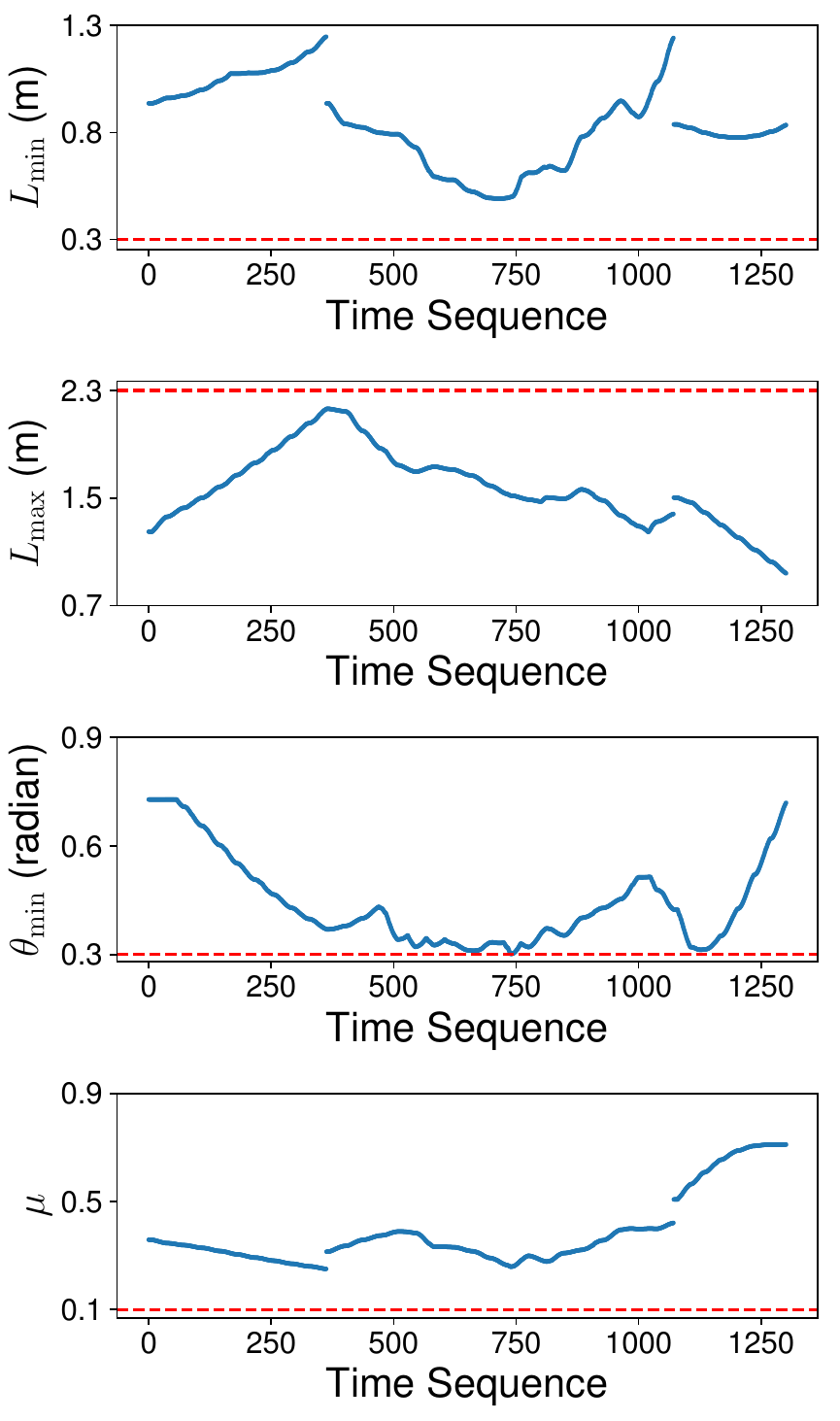}
  \caption{The minimum length ($L_{\min}$) and the maximum length
    ($L_{\max}$) of all moving edge modules, the minimum angle between
    every pair of edge modules ($\theta_{\min}$), and the motion
    manipulability ($\mu$) are measured throughout the locomotion
    process in Fig.~\ref{fig:locomotion-test-process}.}
  \label{fig:vtt-locomotion-data}
\end{figure}

The minimum length ($L_{\min}$) and the maximum length ($L_{\max}$) of
all moving edge modules, the minimum angle between every pair of edge
modules ($\theta_{\min}$), and the motion manipulability ($\mu$) are
shown in Fig.~\ref{fig:vtt-locomotion-data}. This rolling step is
tested with 1000 trials and the average planning time is \SI{8.30}{s}
with the standard deviation being \SI{5.94}{s} in 1000 trials, and the
success rate is 100\%. In these trials, the maximum planning time is
\SI{24.50}{s} and the minimum planning time is \SI{0.47}{s}. The
locomotion of this VTT is also tested
in~\cite{Park-vtt-locomotion-ral-2020} and the performance comparison
is shown in Table~\ref{tab:comp-park}. Our planner has better
performance in terms of both efficiency and robustness.

\begin{table}[t!]
  \centering
  \caption{Comparison with~\cite{Park-vtt-locomotion-ral-2020}.}
  \begin{tabular}{ccc}
    \toprule
    &Avg. Plan Time&Success Rate\\
    \midrule
    Optimization&\SI{18.84}{s}&38.3\% (820 trials)\\
    Sampling-Based (Ours)&\SI{8.30}{s}&100\% (1000 trials)\\
    \bottomrule
  \end{tabular}
  \label{tab:comp-park}
\end{table}

\begin{table}[t!]
  \centering
  \caption{Comparison with~\cite{Usevitch-lar-locomotion-tro-2020}.}
  \begin{tabular}{ccc}
    \toprule
    &Avg. Plan Time&Success Rate\\
    \midrule
    Optimization&\SI{166}{s}&Not provided\\
    Sampling-Based (Ours)&\SI{1.18}{s}&100\% (1000 trials)\\
    \bottomrule
  \end{tabular}
  \label{tab:comp-use}
\end{table}

An octahedron rolling gait is computed based on an optimization
approach in~\cite{Usevitch-lar-locomotion-tro-2020}. Impacts are not
considered in this approach and it takes \SI{166}{s} to compute the
whole process. The octahedron configuration is also tested with our
framework and the comparison is shown in Table~\ref{tab:comp-use}. The
solution (Fig.~\ref{fig:single-step-locomotion}) can be derived as
fast as \SI{1.18}{s} in average with a standard deviation being
\SI{1.03}{s} in 1000 trials, and the success rate is 100\%. In these
trials, the maximum planning time is \SI{3.44}{s} and the minimum is
\SI{0.10}{s}.

\section{Conclusion}
\label{sec:conclusion}

A motion planning framework for variable topology truss robots is
presented in this paper. Physical constraints such as joint limits,
truss member diameter, static stability during locomotion, and motion
manipulability are accounted for. An efficient algorithm to compute
the obstacle regions and the free space of a group is developed so
that sampling-based planners can be applied to solve the
self-collision-free motion of multiple nodes. A fast algorithm to
compute all enclosed subspaces in the free space of a node is
presented so that we can verify whether topology reconfiguration
actions are needed. A sample generation method is introduced to
efficiently provide valid topology reconfiguration actions over a wide
space. With our graph search algorithm, a sequence of topology
reconfiguration actions can then be computed with geometry
reconfiguration planning for a group of nodes, and the motion tasks
requiring topology reconfiguration can then be solved efficiently. A
non-impact rolling locomotion algorithm is presented to move an
arbitrary VTT easily by a simple command with our efficient geometry
reconfiguration planner. Our locomotion algorithm outperforms other
approaches in terms of efficiency and robustness. Future work includes
motion planning in more complex environments, such as locomotion in
complex terrains, and the exploration of heuristic functions for truss
robots to speed up the planning.

\ifCLASSOPTIONcaptionsoff
  \newpage
\fi

\addtolength{\textheight}{-13cm}

\bibliographystyle{IEEEtran}
\bibliography{IEEEabrv,reference.bib}

\end{document}